\documentclass[runningheads]{llncs}

\usepackage[T1]{fontenc}

\usepackage{graphicx}
\usepackage{amsmath}
\usepackage{amssymb}
\usepackage{booktabs}
\usepackage{multirow}
\usepackage{xspace}

\usepackage[table,dvipsnames]{xcolor}
\usepackage[breaklinks,colorlinks]{hyperref}

\usepackage[capitalize]{cleveref}
\crefname{section}{Sec.}{Secs.}
\Crefname{section}{Section}{Sections}
\Crefname{table}{Table}{Tables}
\crefname{table}{Tab.}{Tabs.}

\makeatletter
\DeclareRobustCommand\onedot{\futurelet\@let@token\@onedot}
\def\@onedot{\ifx\@let@token.\else.\null\fi\xspace}

\makeatother
\def\eg{\emph{e.g}\onedot} 
\def\ie{\emph{i.e}\onedot}

\def\etal{\emph{et al}\onedot}

\newcommand{\maketitlesupplementary}{%
  \begin{center}\Large\bfseries Supplementary Material\end{center}\bigskip%
}

\usepackage[shortlabels,inline]{enumitem}

\usepackage{algorithm}
\usepackage{algpseudocode}

\newcommand{\todo}[1]{} 
\newcommand{\TODO}[1]{} 
\usepackage{subcaption}
\usepackage{soul}
\usepackage{colortbl} 
\usepackage{changepage,threeparttable} 
\usepackage{tabularx}
\usepackage{tikz}
\usetikzlibrary{positioning, shapes.geometric, arrows.meta, shadows, backgrounds, fit, calc}

\usepackage{makecell}
\usepackage{longtable}
\usepackage{pdflscape}
\usepackage{comment}
\usepackage{adjustbox}
\usepackage{array}
\usepackage{pifont}
\usepackage[most]{tcolorbox}
\usepackage{setspace}
\usepackage{listings}
\usepackage{lipsum}
\usepackage{xparse}
\usepackage{csquotes}

\newcommand{\cmark}{\ding{51}}%
\newcommand{\xmark}{\ding{55}}%
\newcommand{\ourframework}{DocDjinn}

\NewDocumentCommand{\bert}{o}{%
    BERT%
    \IfNoValueT{#1}{~\cite{DCLT19}}%
}
\NewDocumentCommand{\lilt}{o}{%
    LiLT%
    \IfNoValueT{#1}{~\cite{WJD22}}%
}
\NewDocumentCommand{\layoutlm}{o}{%
    LayoutLMv3%
    \IfNoValueT{#1}{~\cite{HLCL22}}%
}
\NewDocumentCommand{\fasterrcnn}{o}{%
    Faster R-CNN%
    \IfNoValueT{#1}{~\cite{RHGS17}}%
}
\NewDocumentCommand{\cascadercnn}{o}{%
    Cascade R-CNN%
    \IfNoValueT{#1}{~\cite{CV21}}%
}

\NewDocumentCommand{\rvlcdip}{o}{%
    RVL-CDIP%
    \IfNoValueT{#1}{~\cite{HUD15}}%
}
\NewDocumentCommand{\tobacco}{o}{%
    Tobacco3482%
    \IfNoValueT{#1}{~\cite{KXD14}}%
}
\NewDocumentCommand{\doclaynetcls}{o}{%
    DocLayNet-CLS%
    \IfNoValueT{#1}{~\cite{PADN22}}%
}

\NewDocumentCommand{\cord}{o}{%
    CORD%
    \IfNoValueT{#1}{~\cite{PSLL19}}%
}
\NewDocumentCommand{\funsd}{o}{%
    FUNSD%
    \IfNoValueT{#1}{~\cite{JET19}}%
}
\NewDocumentCommand{\sroie}{o}{%
    SROIE%
    \IfNoValueT{#1}{~\cite{HCHB19}}%
}

\NewDocumentCommand{\docvqa}{o}{%
    DocVQA%
    \IfNoValueT{#1}{~\cite{MKJ21}}%
}
\NewDocumentCommand{\docvqahw}{o}{%
    DocVQA-HW%
    \IfNoValueT{#1}{}%
}
\NewDocumentCommand{\klc}{o}{%
    KLC%
    \IfNoValueT{#1}{~\cite{SGWL21}}%
}
\NewDocumentCommand{\wiki}{o}{%
    WTQ%
    \IfNoValueT{#1}{~\cite{PL15}}%
}
\NewDocumentCommand{\wikishort}{o}{%
    WTQ%
    \IfNoValueT{#1}{~\cite{PL15}}%
}

\NewDocumentCommand{\publaynet}{o}{%
    PubLayNet%
    \IfNoValueT{#1}{~\cite{ZTJ19}}%
}
\NewDocumentCommand{\doclaynetdla}{o}{%
    DocLayNet-DLA%
    \IfNoValueT{#1}{~\cite{PADN22}}%
}
\NewDocumentCommand{\icdarctdar}{o}{%
    ICDAR2019%
    \IfNoValueT{#1}{~\cite{GHDM19}}%
}
\NewDocumentCommand{\icdarctdarshort}{o}{%
    ICDAR19%
    \IfNoValueT{#1}{~\cite{GHDM19}}%
}
\NewDocumentCommand{\doclaynet}{o}{%
    DocLayNet%
    \IfNoValueT{#1}{~\cite{PADN22}}%
}

\newlength{\imgwidth}
\newlength{\imgheight}

\begin{document}

\title{\ourframework: Controllable Synthetic Document Generation with VLMs and Handwriting Diffusion}

\titlerunning{\ourframework: Controllable Synthetic Document Generation}

\author{
  Marcel Lamott$^{*}$\inst{1} \and
  Saifullah Saifullah$^{*}$\inst{2,3} \and
  Nauman Riaz$^{*}$\inst{2,3} \and
  Yves-Noel Weweler\inst{4} \and
  Tobias Alt-Veit\inst{4} \and
  Ahmad Sarmad Ali\inst{5} \and
  Muhammad Armaghan Shakir\inst{5} \and
  Adrian Kalwa\inst{1} \and
  Momina Moetesum\inst{5} \and
  Andreas Dengel\inst{2} \and
  Sheraz Ahmed\inst{2,3} \and
  Faisal Shafait\inst{5} \and
  Ulrich Schwanecke\inst{1} \and
  Adrian Ulges\inst{1}
}

\authorrunning{M. Lamott et al.}

\institute{
  RheinMain University of Applied Sciences, Wiesbaden, Germany \and
  German Research Center for Artificial Intelligence, Kaiserslautern, Germany \and
  DeepReader GmbH, Kaiserslautern, Germany \and
  Insiders Technologies GmbH, Kaiserslautern, Germany \and
  National University of Sciences and Technology (NUST), Islamabad, Pakistan
}

\maketitle

\renewcommand{\thefootnote}{*}
\footnotetext{Equal contribution.}
\renewcommand{\thefootnote}{\arabic{footnote}}

\begin{figure}[t]
    \centering
    \setlength{\imgwidth}{0.19\textwidth}
    \setlength{\imgheight}{0.25\textwidth}
    \resizebox{\textwidth}{!}{
    \begin{tabular}{ccccc}
        \includegraphics[width=\imgwidth,height=\imgheight]{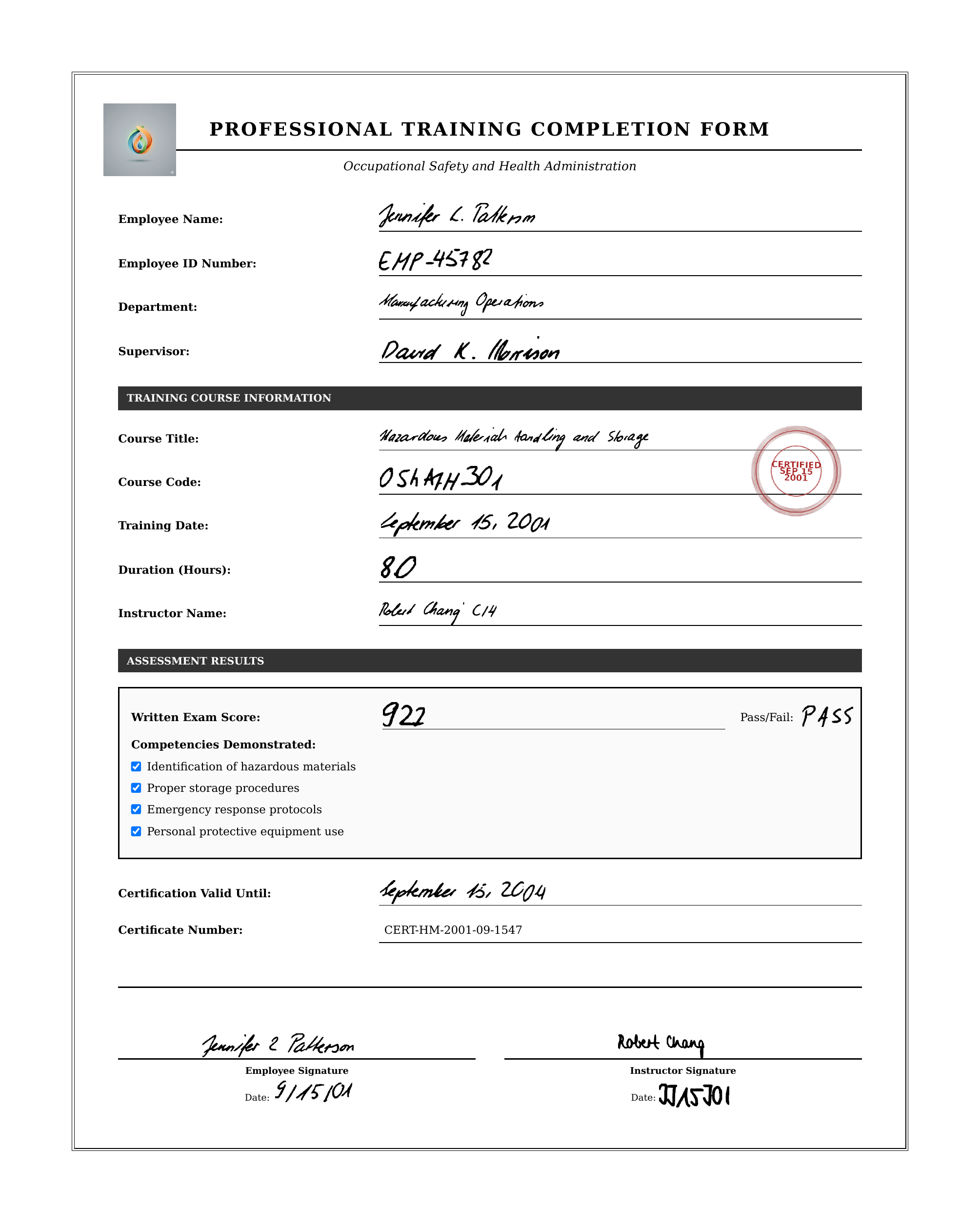} &
        \includegraphics[width=\imgwidth,height=\imgheight]{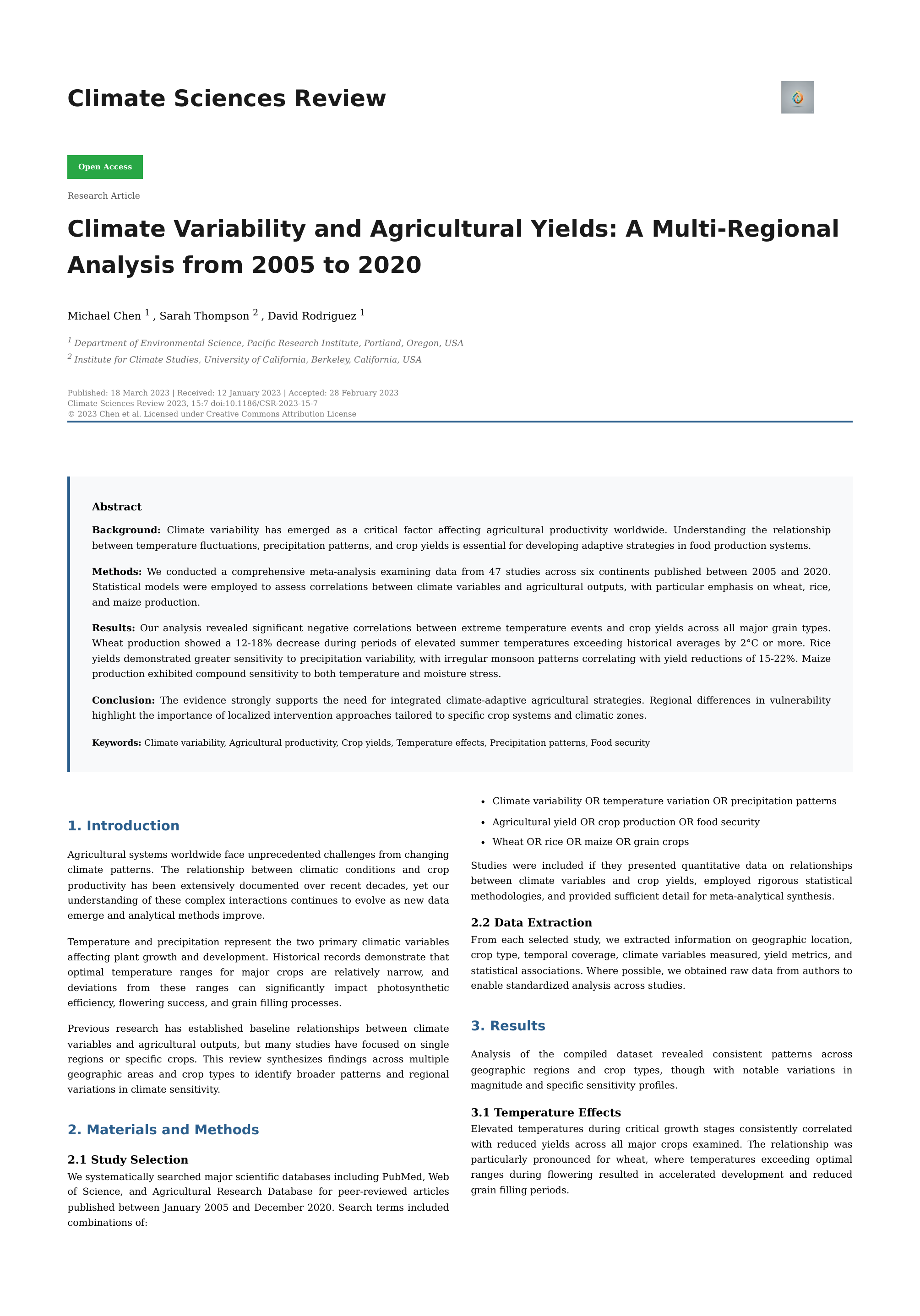} &
        \includegraphics[width=\imgwidth,height=\imgheight]{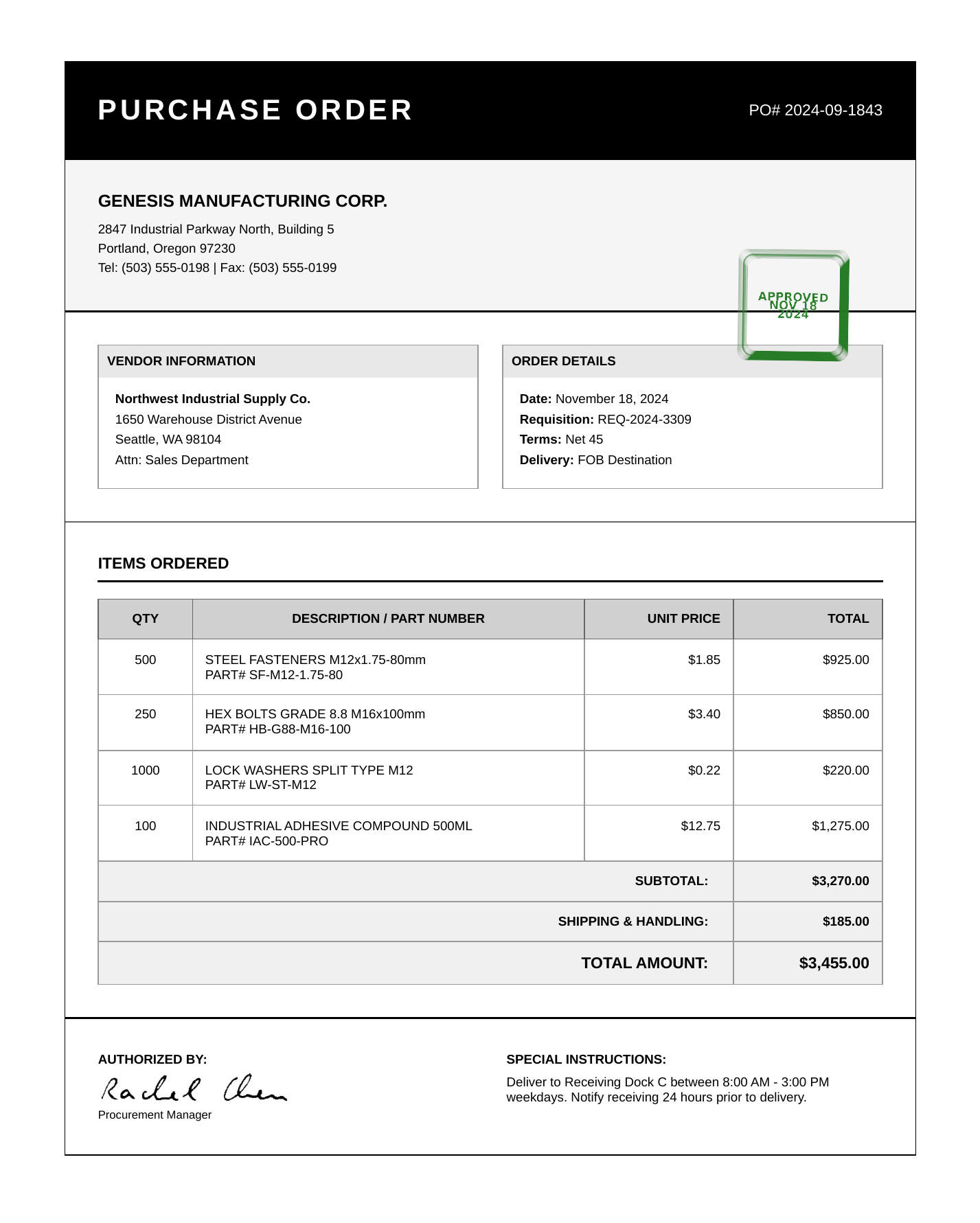} &
        \includegraphics[width=\imgwidth,height=\imgheight]{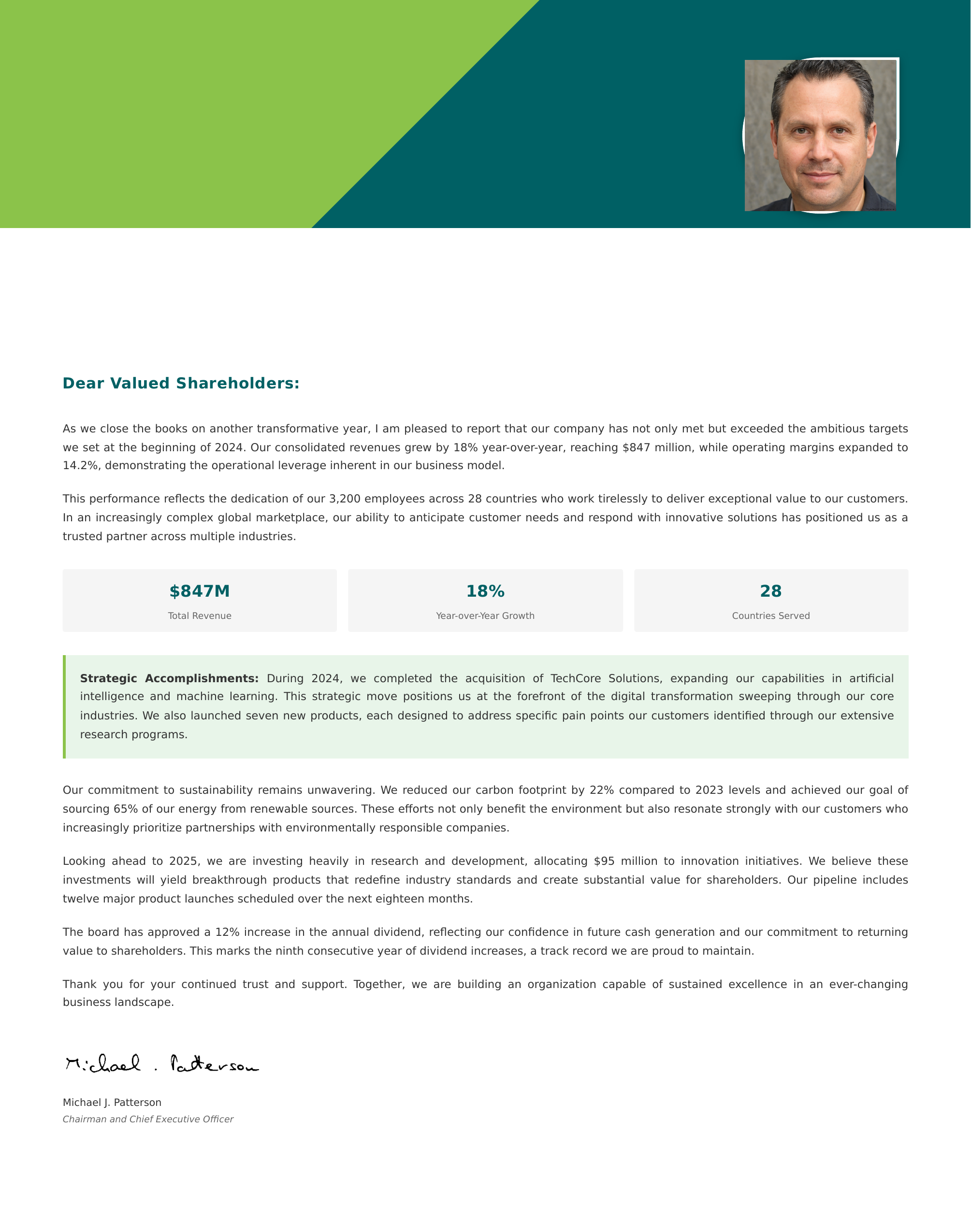} &
        \raisebox{\dimexpr0.49\imgheight}{%
            \begin{tabular}[t]{c}
                \includegraphics[width=\imgwidth,height=0.5\imgheight]{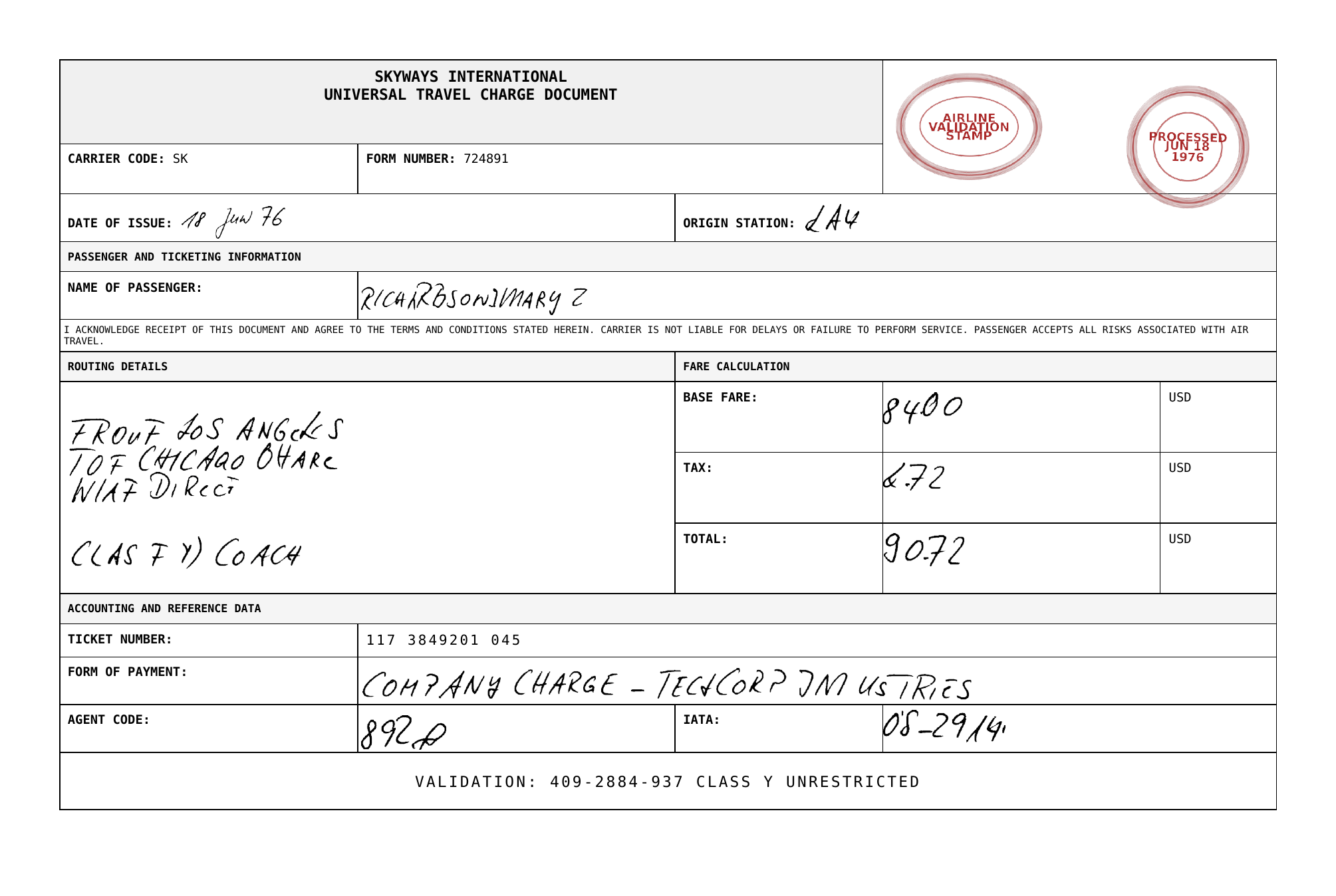} \\
                \includegraphics[width=\imgwidth,height=0.25\imgheight]{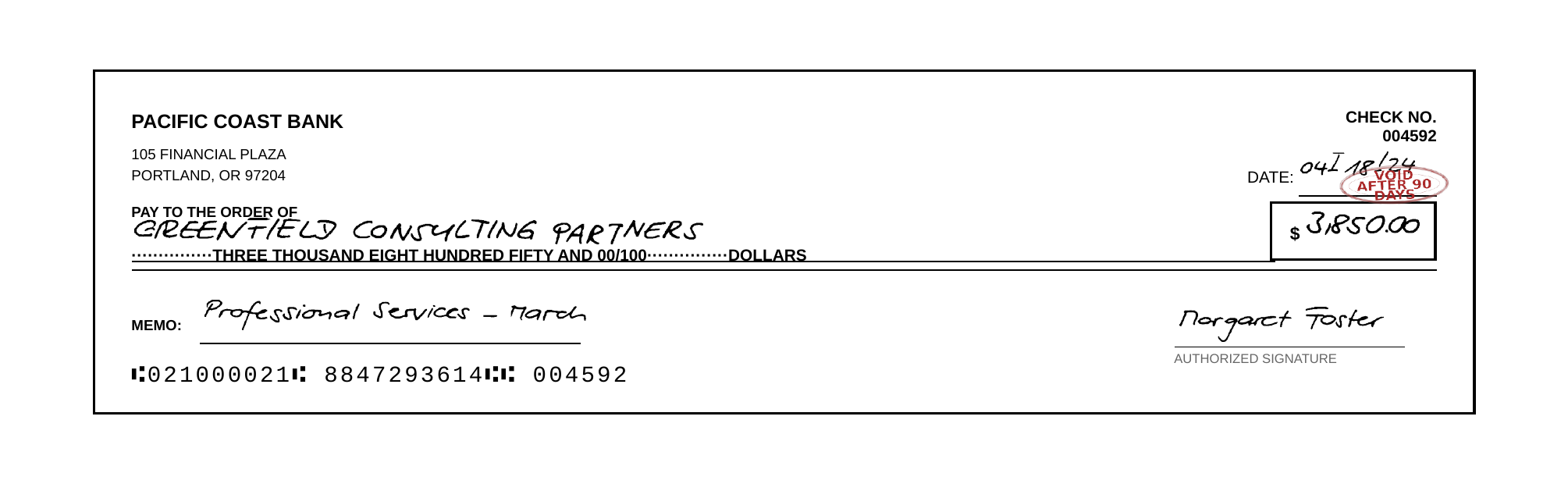}
            \end{tabular}%
        }
    \end{tabular}
    }
    \caption{Examples of synthetically generated documents across diverse domains and tasks. Our framework produces documents with realistic layouts, VLM-generated content, diffusion-based handwriting, and contextual visual elements.}
    \label{fig:selection}
\end{figure}

\let\origendabstract\endabstract
\renewcommand{\endabstract}{%
\keywords{Document synthesis \and Synthetic data generation \and Vision-Language Models \and Handwriting diffusion \and Document understanding \and Seed selection}
\origendabstract
}
\begin{abstract}
Effective document intelligence models rely on large amounts of annotated training data. 
However, procuring sufficient and high-quality data poses significant challenges due to the labor-intensive and costly nature of data acquisition. 
Additionally, leveraging language models to annotate real documents raises concerns about data privacy. 
Synthetic document generation has emerged as a promising, privacy-preserving alternative. 
We propose \mbox{\ourframework}, a novel framework for controllable synthetic document generation using Vision-Language Models (VLMs) that produces annotated documents from unlabeled seed samples.
Our approach generates visually plausible and semantically consistent synthetic documents that follow the distribution of an existing source dataset through clustering-based seed selection with parametrized sampling.
By enriching documents with realistic diffusion-based handwriting and contextual visual elements via semantic-visual decoupling, we generate diverse, high-quality annotated synthetic documents.
We evaluate across eleven benchmarks spanning key information extraction, question answering, document classification, and document layout analysis. 
To our knowledge, this is the first work demonstrating that VLMs can generate faithful annotated document datasets at scale from unlabeled seeds that can effectively enrich or approximate real, manually annotated data for diverse document understanding tasks. 
We show that with only 100 real training samples, our framework achieves on average $87\%$ of the performance of the full real-world dataset.
We publicly release our code and 140k+ synthetic document samples.
\end{abstract}

\renewcommand{\thefootnote}{\arabic{footnote}}

\section{Introduction}
\label{sec:intro}

Document intelligence systems employ deep learning and Vision-Language Models (VLMs) to transform documents into structured information via document layout analysis (DLA)~\cite{ZTJ19}, key information extraction (KIE)~\cite{HKJH22}, visual question answering (VQA)~\cite{MKJ21}, and classification (CLS). While VLMs are powerful, they remain prohibitively expensive for specialized, high-throughput applications. This has motivated smaller, task-specific models~\cite{Belck25SmallLM} that require substantial labeled training data. Despite recent datasets~\cite{VS21,FZSA25}, obtaining high-quality annotations for diverse document types remains costly and labor-intensive.

Synthetic data generation offers a promising solution. However, existing approaches either lack textual coherence~\cite{TUS23}, generate only task-specific content such as layout~\cite{HLCF23} or tables~\cite{HSAD24}, or produce documents without ground truth~(GT) annotations~\cite{BMT19,KHYN22,DLTL24,AH25}. DocGenie~\cite{HRVA25}, while capable of conditioning generation on seed documents, generates only visual content and text without task-specific labels required for supervised learning (\eg entity labels for KIE, bounding boxes for DLA, question-answer pairs for VQA). Consequently, these synthetic documents cannot directly train document understanding models.

We present \ourframework, addressing three key challenges: \textbf{(1) Multimodal realism} through VLM-generated content combined with diffusion-based handwriting synthesis and contextual visual element insertion, \textbf{(2) distribution alignment} via automatic clustering-based seed selection with parametrized sampling strategies that align synthetic data with source dataset distributions, and \textbf{(3) training suitability} by generating high-quality task-specific annotations alongside documents, enabling direct supervised learning across VQA, KIE, CLS, and DLA tasks.

Our evaluation demonstrates that synthetic-only training achieves \(70.8\%\) of real-data performance, while augmenting just \(100\) real samples with synthetic data reaches within \(9.45\) points of full real-data training. In low-resource scenarios with only \(100\) labeled samples, our framework achieves on average \(87\%\) of the performance of the full real-world dataset.

\noindent Our concrete contributions are as follows:
\begin{itemize}
    \item A scalable framework for synthetic document generation that produces automatic ground truth annotations from unlabeled seed documents across VQA, KIE, CLS, and DLA tasks.
    \item First integration of diffusion-generated handwriting into modern document synthesis with semantic-visual decoupling for stamps, barcodes, and logos.
    \item Clustering-based seed selection with parametrized sampling that preserves target distributions.
    \item Public release of eleven synthetic datasets (\(140\text{k}+\) samples) and \docvqahw, a handwriting-focused \docvqa~subset.
\end{itemize}

\begin{table*}[!t]
\centering
\caption{
Overview of recent synthetic document generation approaches compared to ours. 
\textbf{Target} specifies the generated modality, and \textbf{Source} the conditioning input. \textbf{Text}, \textbf{HW}, and \textbf{VE} indicate explicit support for readable text, readable handwritten text, and visual elements during generation. \textbf{U. GT.} represents the capability to automatically generate task annotations in an unsupervised manner. \textbf{Max Res.} denotes the maximum possible generation resolution, whereas \textbf{Dyn.\ Spec.} represents whether the generation process can be controlled via natural language. \textbf{ML} indicates multilingual generation ability, and \textbf{OS} indicates whether the framework is open source. Finally, \textbf{Editable} indicates whether generated documents are produced in an editable format (e.g., structured text) rather than as rasterized images.
}
\rowcolors{2}{gray!15}{white} 
\resizebox{\linewidth}{!}{
\begin{tabular}{lllllccccccccccc}
\textbf{Year} &\textbf{Name} &\textbf{Model} &\textbf{Target} &\textbf{Source} &\textbf{Text} &\textbf{HW} &\textbf{VE} &\textbf{U. GT.} &\textbf{Tasks} &\textbf{Max Res.} &\textbf{Dyn. Spec.} &\textbf{ML} &\textbf{OS} &\textbf{Editable} \\\midrule
2017 &DocCreator~\cite{JVMV17} &manual &full document &images &\cmark &\xmark &\xmark &\cmark &OCR / DLA &- &\xmark &\cmark &\cmark &\xmark \\
2019 &Bui et al.~\cite{BMT19} &GAN &document image &text &\cmark &\xmark &\xmark &\xmark &OCR &$512\times512$ &\xmark &\xmark &\xmark &\xmark \\
2021 &Genalog~\cite{GRMB21} &templates &document image &text &\cmark &\xmark &\xmark &\xmark &NER &Unspec. &\xmark &\xmark &\cmark &\xmark \\
2021 &DocSynth~\cite{BRLP21} &GAN &document image &layout &\xmark &\xmark &\cmark &\xmark &\xmark &$128\times128$ &\xmark &\xmark &\cmark &\cmark \\
2021 &Raman et al.~\cite{RSV22} &sampling &full document &layout &\cmark &\xmark &\cmark &\cmark &DLA &Unspec. &\xmark &\cmark &\xmark &\xmark \\
2022 &SynthDoG~\cite{KHYN22} &sampling &full document &- &\cmark &\xmark &\xmark &\cmark &DLA &$2560\times1920$ &\xmark &\cmark &\cmark &\xmark \\
2023 &Tanveer et al.~\cite{TUS23} &DPM &document image &layout &\xmark &\xmark &\cmark &\xmark &DLA &$256\times256$ &\xmark &\xmark &\xmark &\cmark \\
2023 &DocGen~\cite{AAMK23} &LLM &document text &text &\cmark &\xmark &\xmark &\xmark &IR &- &\cmark &\xmark &\cmark &\cmark \\
2023 &Fennir et al.~\cite{FLL23} &GAN &document image &layout &\xmark &\xmark &\cmark &\xmark &\xmark &$512\times512$ &\xmark &\xmark &\xmark &\cmark \\
2024 &Hamdani et al.~\cite{HSAD24} &DPM &table image &layout &\xmark &\xmark &\xmark &\xmark &TE &$512\times512$ &\xmark &\xmark &\xmark &\xmark \\
2024 &SynthDoc~\cite{DLTL24} &sampling &full document &text &\cmark &\xmark &\cmark &\cmark &KIE &$1280\times960$ &\xmark &\cmark &\xmark &\cmark \\
2024 &Hou et al.~\cite{HWQT24} &sampling &full table &text+layout &\cmark &\xmark &\xmark &\xmark &TE &- &\xmark &\cmark &\xmark &\cmark \\
2025 &Havas~\cite{AH25} &LLM &full document &template &\cmark &\xmark &\xmark &\cmark &KIE &- &\cmark &\cmark &\xmark &\cmark \\
2025 &DocGenie~\cite{HRVA25} &VLM &full document &images &\cmark &\xmark &\xmark &\xmark &\xmark &- &\cmark &\cmark &\xmark &\cmark \\
\midrule
\textbf{2025} &\textbf{DocDjinn (Ours)} &\textbf{VLM} &\textbf{full document} &\textbf{images} &\textbf{\cmark} &\textbf{\cmark} &\textbf{\cmark} &\textbf{\cmark} &\textbf{\makecell{\textbf{CLS} / \textbf{KIE} /\\\textbf{VQA} / \textbf{DLA}}} &- &\textbf{\cmark} &\textbf{\cmark} &\textbf{\cmark} &\textbf{\cmark} \\
\end{tabular}
}
\label{tab:related}
\end{table*}

\section{Related Work}
\label{sec:related}

{\tolerance=1500
Recent document intelligence models mainly rely on transformers~\cite{WJD22,HLCL22} and LLMs~\cite{DCLT19,BMRS20,RSRL20}. While specialized models exist for sub-tasks such as table analysis~\cite{NLLS22} or key information extraction~\cite{HKJH22}, recent models aim at multi-task full document understanding~\cite{WJD22,KHYN22,HLCL22,WJD22}.
Curating high-quality datasets from real-world documents is challenging if not infeasible, if these models need to be adapted to new domains or trained from scratch. 
Thus, researchers have turned towards synthetic training data~\cite{BTSL24}, fostering a plethora of synthetic document generation frameworks in recent years~\cite{HRVA25,BRLP21,KHYN22}. We list a selection thereof in \cref{tab:related}.
These works can be grouped mainly along two dimensions: the underlying model architecture, and the targeted modality of generated data.
In addition, we highlight the differences between approaches in terms of the input source, support for text (printed/handwritten) and visual elements, maximum possible generation resolution, controllability via natural language, multilingual support, and whether the documents can be easily edited post-generation.

Barring exceptions, four classes of models emerge: sampling-based strategies~\cite{DLTL24,KHYN22,HWQT24}, Generative Adversarial Networks (GANs)~\cite{BMT19,BRLP21,FLL23}, diffusion probabilistic models (DPMs)~\cite{SWMG15}, and Large Language Models (LLMs)~\cite{DCLT19,RSRL20}. 
Both GANs and DPMs excel at visual synthesis, enabling specialized applications in document image generation~\cite{TUS23,HSAD24}, document layout generation~\cite{HLCF23,LYHZ19}, and handwritten text generation~\cite{NRCS23,RSAD24,DZKG24}. 
However, they inherently lack the ability to generate coherent, contextually grounded text at large scale, as it requires discrete sequential modeling and linguistic reasoning capabilities beyond their visual architectures.

In contrast, the language generation capabilities of LLMs~\cite{DCLT19,RSRL20} represent a significant advancement, allowing for controllable synthetic generation of full documents. 
While LLMs lack the visual generation qualities of diffusion models to synthesize specialized visual content, such as realistic handwritten text with consistent style and natural variations, their recent extension to VLMs~\cite{RKHR21,WBTW24} enables them to reliably process visual data. 
As they have been trained on markup languages, among other data, they are capable of producing visual elements solely based on markup. This is sufficient for satisfactory synthetic document generation, as the majority of documents share a structured layout. 
Unsurprisingly, the state of the art in synthetic document generation employs LLMs and VLMs~\cite{AH25,HRVA25}.\par}


Out of all recent works, to the best of our knowledge, only six works fulfill the  requirements of full document generation~\cite{JVMV17,KHYN22,DLTL24,AH25,HRVA25,RSV22}. 
Among these, DocCreator~\cite{JVMV17}, SynthDoG~\cite{KHYN22}, SynthDoc~\cite{DLTL24}, and Raman et al.~\cite{RSV22} all rely on predefined templates or sampling from public corpora that limit their applications in specialized domains. 
Abarca \& Havas~\cite{AH25} generate full documents with LLMs but rely on manually crafted dataset-specific templates, limiting the generalization of their approach.
DocGenie~\cite{HRVA25} is the only approach that can leverage documents directly as a dynamic source without having to extract a specific modality from the source data or rely on handcrafted templates. However, since it does not produce task-specific ground-truth annotations, its applicability in downstream applications is limited.
In addition, DocGenie~\cite{HRVA25} randomly selects seed samples from the source distribution to generate new documents. 
This limits the scalability of the approach for large document corpora, which often exhibit substantial class imbalance and contain large clusters of highly similar documents within the data distribution.

We improve upon the limitations of DocGenie~\cite{HRVA25} by automating seed selection to obtain a representative set that captures the source data distribution and by enabling the VLM to generate task-specific ground-truth annotations. 
Furthermore, we combine the advantages of VLMs with those of DPMs by generating visual annotations with the VLM that serve as conditioning input for a DPM. The DPM then generates variable-style realistic handwritten text, which is injected into the documents.
Thus, we present a fully automatic and unsupervised pipeline to enrich real-world datasets with plausible synthetic samples that can be directly used in downstream tasks. 

\section{Our Framework: \ourframework}

\begin{figure*}[t]
    \setlength{\imgwidth}{1.00\textwidth}
    \setlength{\imgheight}{0.25\textwidth}
    \centering
    \includegraphics[width=\imgwidth,height=\imgheight]{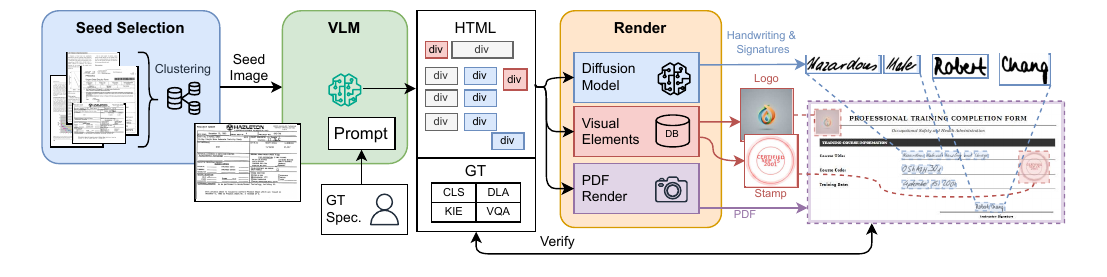}
    \vspace{-8mm}
    \caption{Overview of \ourframework~for synthetic document generation. After selecting representative seeds from a source dataset, a VLM generates an HTML representation of the document, along with multi-task ground truth information. This representation is enhanced with diffusion-generated handwriting and further visual elements. Finally, the ground truth is updated with bounding boxes and verified.
        \label{fig:framework}}
\end{figure*}

We propose \emph{\ourframework}, a VLM-based framework for generating synthetic documents with realistic content and structurally coherent layouts alongside task-specific ground-truth annotations.
Assuming a set of real unlabeled documents $\mathcal{D_\text{real}} = \{x_i\}_{i=1}^N$, we seek to substitute or complement it with a synthetic dataset $\mathcal{D}_\text{syn} = \{(x_i', y_i')\}_{i=1}^{N'}$ such that a model trained on the synthetic dataset, either alone or in combination with the real dataset, approximates the original model performance on a given task. Here,~$x'_i$ denotes a synthetically generated document sample, and~$y'_i$ its corresponding annotation.
\ourframework~operates in four stages, an overview of which is given in \cref{fig:framework}:
\textbf{(1)} intelligent seed sample selection,
\textbf{(2)} seed-guided VLM-based document and GT synthesis,
\textbf{(3)} visual realism enhancement via insertion of diffusion-based handwriting and visual elements, and
\textbf{(4)} bounding box extraction and GT verification.



\subsection{Intelligent Seed Sample Selection}\label{sec:seed_selection}
Unlike previous work~\cite{HRVA25}, where seed samples are randomly sampled from the source dataset, we introduce a clustering-based approach to select representative yet diverse seed samples. 
Seed-samples are document images supplied as few-shot examples to guide the VLM during the synthesis and condition it to produce similar documents.

\paragraph{Embeddings.}
To capture both the structural and semantic characteristics of each document, we first represent documents using embeddings derived from layout, image, and text modalities. For each document $d_i \in \mathcal{D_\text{real}}$, we compute embeddings using \layoutlm{} CLS tokens (\texttt{layoutlm}), CLIP~\cite{RKHR21} image features (\texttt{clip}), and Sentence Transformers~\cite{RG19} text representations (\texttt{sentence}). We propose a multimodal embedding (\texttt{combined}) by z-score normalizing and concatenating these three modalities, capturing layout structure, visual appearance, and textual semantics jointly. We additionally compare against pooled LayoutLMv3 embeddings (\texttt{pooled})~\cite{SM25}.


\paragraph{Clustering.}
We adopt the approach from~\cite{SM25} by combining HDBSCAN with minimum cluster size $\kappa$ and $k$-NN~\cite{FH51} with a fixed $k$. For each embedding type $\mathcal{E}$ and minimum cluster size $\kappa$, embeddings $\{\mathbf{e}_i\}_{i=1}^N$ are reduced to $d'$ dimensions via UMAP. HDBSCAN produces initial clusters with noise points, which are then reassigned using a $k$-NN classifier trained on non-noise embeddings, ensuring complete coverage.
We manually select the optimal clustering, assisted by a heuristic combining silhouette score and normalized entropy that empirically correlates well with clustering quality and favors high internal coherence and balanced cluster sizes (Appendix~\ref{app:clustering}).

\paragraph{Sampling.}
From $K$ clusters with sizes $\{n_c\}_{c=1}^K$, we sample clusters with probabilities $p(c) \propto n_c^\alpha$, where $\alpha$ controls cluster size bias. We compare two strategies for generations, in each of which $n$ seeds are supplied as few-shot examples: \textit{cross-cluster} ($\mathsf{CC}$) samples $n$ seeds independently (each according to $p(c)$), while \textit{intra-cluster} ($\mathsf{IC}$) first samples one cluster via $p(c)$, then draws all $n$ seeds from within that cluster. The resulting seed samples, consisting only of document images, are used to guide the VLM during document synthesis.

\subsection{VLM-Based Document and GT Synthesis}
\label{sec:vlm_synthesis}
Using the selected seed samples and a task-level prompt, we employ a VLM to synthesize HTML documents along with corresponding GT, generating $M$ documents per prompt call while supplying $2M$ seed images as guidance (compared to $10$ seed images used in~\cite{HRVA25}).
We distinguish two types of GT generation, corresponding to two prompt templates (see Appendix~\ref{app:prompt}): \textit{Macro} (document-level JSON annotations), where the VLM is instructed to produce GT for VQA and simple KIE tasks, and \textit{Micro} (element-level annotations with class labels), where labels are generated for layout- and structure-sensitive tasks, namely DLA and complex KIE.
Each dataset is further defined by three parameters in the prompt template: \textbf{document type}, a brief description; \textbf{GT type}, specifying the annotation task (QA pair creation, KIE class labeling, document classification, or region-level labeling); and \textbf{GT format}, defining the ground-truth structure, \ie, JSON or additional class groupings.
We extract element regions 
from the HTML via JavaScript and match them to the generated GT for tasks requiring spatial annotations (KIE, DLA). 
Additionally, we extract bounding boxes 
from a PDF rendering of the HTML for subsequent processing steps.

\subsection{Visual Realism Enhancement}
To improve realism, we add diffusion-based handwritten text and contextual visual elements such as figures and stamps to the documents, enhancing fidelity and bridging the domain gap to real documents. The VLM is prompted to produce HTML placeholders for such elements.

\paragraph{Region Identification.}
The VLM identifies regions requiring handwriting such as signatures or form fields, as well as visual elements (stamps, barcodes, logos, figures, and photos). For handwriting, it assigns author identifiers for multi-author generation. Each designated text element is rendered with a fixed font size, its region and word-level boxes
are extracted, and the placeholders are replaced by diffusion-generated handwriting while preserving layout and semantics.
Visual elements are typed and given textual content descriptions, such as \enquote{APPROVED 2024-03-15} for a stamp. Each visual element is rendered type-specific (details in Appendix~\ref{app:visual_elements}) and inserted into its corresponding region.

\paragraph{Diffusion-based Handwriting.}
To synthesize realistic handwritten text, we adopt a latent diffusion model~\cite{RBLE22,LZBC22,NRCS23,RSAD24} conditioned on both the target text and writer style. A pretrained Variational Autoencoder (VAE) 
is first used to encode the handwritten text images into latent variables \(z \in \mathbb{R}^{d_z}\). Then, a conditional UNet-based diffusion model is trained in the latent space using the standard DPM loss~\cite{HJA20}:
\begin{equation}
\mathcal{L}_{\text{DPM}}(\theta)
= \mathbb{E}_{t,\,z_0,\,\epsilon}\!\Big[
\big\|\,\epsilon -
\epsilon_\theta\!\big(z_t, t;\, c_{\text{text}}, c_{\text{style}}\big)
\big\|_2^2
\Big],
\end{equation}
where \(c_{\text{text}}\) denotes the text condition embedding obtained from a Transformer-based encoder model, and $c_{\text{style}}$ represents the writer-specific style embedding (writer class). Details of our diffusion training and inference parameters are provided in Appendix~\ref{app:implementation_handwriting}.

\begin{figure}[t]
  \centering


  \begin{subfigure}{1\linewidth}
    \centering
    \includegraphics[width=\linewidth]{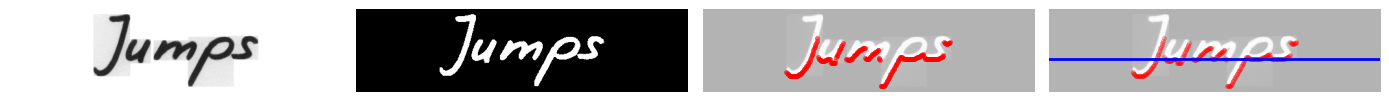}
  \end{subfigure}\hfill
  \begin{subfigure}{1\linewidth}
    \centering
    \includegraphics[width=\linewidth]{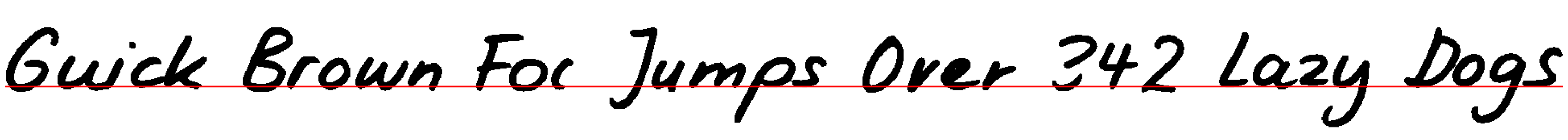}
  \end{subfigure}

  \vspace{-0.3em}
  \caption{
    Baseline alignment for sentence-level handwritten text. 
    \textbf{Top, left to right:} input image, word segmentation, lowest-ink pixel per column (in red), and computed baseline via percentile (in blue). 
    \textbf{Bottom:} example sentence-level handwritten text after baseline alignment (red).
  }
  \label{fig:baseline_composition}
  \vspace{-0.6em}
\end{figure}

\paragraph{Line-Segment Generation and Integration.}
Handwritten text lines are generated by concatenating style-conditioned word segments from our diffusion model with baseline alignment (\cref{fig:baseline_composition}), where the baseline is the median y-coordinate of lowest ink pixels. After horizontal concatenation, 
segments are refined with Gaussian blur, scaled to match their bounding box union, and positioned at their corresponding region location with random jitter (details in Appendix~\ref{app:implementation_handwriting}). A manual inspection of 200 generated handwritten sentences shows that baseline alignment is correct in 89\% of cases and remains acceptable in 84\% under stricter visual assessment.

\subsection{Bounding Box Extraction and GT Verification}
Following visual enhancement, we extract text bounding boxes via Optical Character Recognition (OCR) for documents with handwritten/visual elements or PDF rendering for typeset documents. VLM-generated GT is verified using task-specific constraints: For VQA, answers must appear in text, verified using averaged normalized Levenshtein distance (ANLS~\cite{BTMG19}). For CLS, class labels need to be valid. In DLA, labels must be valid and regions within extracted bounds. For KIE, key-values need to appear in text, and region annotations constrained to designated areas. Documents failing verification or rendering to multiple pages are excluded.

\section{Experiments}\label{sec:experiments}
We evaluate \ourframework{} along three axes: (\textbf{i}) its ability to generate visually and semantically faithful synthetic documents, (\textbf{ii}) the utility of these documents for downstream model training, and (\textbf{iii}) the impact of mixing synthetic and real-world data on model generalization. The experiments cover four major document understanding tasks: key information extraction (KIE), visual question answering (VQA), document classification (CLS), and document layout analysis (DLA).

\subsection{Experimental Setup}
\paragraph{Embeddings and Clustering.}
For embeddings, we use \layoutlm{} (\texttt{layout}), CLIP~\cite{RKHR21} (\texttt{clip}), and Sentence Transformers~\cite{RG19} (\texttt{sentence}).
For \texttt{pooled}~\cite{SM25} \layoutlm[nr]{} embeddings we use a kernel size of 4.
Embeddings are projected to $d'=100$ via UMAP~\cite{MHSG18} before two-stage clustering, where we evaluate different min cluster sizes $\kappa$ for HDBSCAN~\cite{CMS13} and set $k=5$ for $k$-NN~\cite{FH51}. 
We select the embedding and $\kappa$ for each dataset for downstream experiments according to~\cref{sec:seed_selection}, see Appendix~\ref{app:clustering} for details and dataset-specific clustering choices shown in \cref{tab:cluster_metrics}.

\paragraph{Document Synthesis.}
As VLM we use Claude Sonnet~4.5, as we deemed it the most capable VLM available for this task.
We generate $M=3$ documents per prompt call for VQA, KIE, and CLS tasks, and $M=2$ for DLA tasks due to their more complex annotation requirements. VLM-generated HTML undergoes post-processing with JavaScript-based dimension measurement to ensure single-page PDF rendering via Playwright with dynamically computed page sizes. As text similarity threshold ANLS$_{\tau}$ we use $75\%$.

\paragraph{Handwriting Synthesis.}
We employ a conditional latent diffusion model trained on IAM~\cite{MB02}, using a pretrained VAE to encode each canonical $128{\times}512$ word image into a $16{\times}64$ latent (8× downsampling) while preserving stroke scale. Conditioning is applied on both text and writer identity via a UNet denoiser and Transformer text encoder. Using Microsoft Document Intelligence OCR\footnotemark[1] and visual inspection, we retain the top nine writers (CER/WER: $0.092/0.249$ vs.\ $0.193/0.404$), enabling legible multi-writer synthesis across documents.
See Appendix~\ref{app:implementation_handwriting} for  training and architecture details.
\footnotetext[1]{Model version 2024-11-30.}

{\tolerance=1500
\paragraph{Datasets.}
We conduct experiments on eleven datasets spanning multiple document understanding tasks:
\textbf{VQA:} \docvqa{} and \wiki{};
\textbf{KIE:} \klc{}\footnotemark[2], \sroie{}, \cord{}, and \funsd{};
\textbf{CLS:} \tobacco{}, \rvlcdip{}, and \doclaynetcls{};
\textbf{DLA:} \publaynet{}, \icdarctdar{}, and \doclaynetdla{}.
\footnotetext[2]{\klc{} is modeled as VQA for downstream evaluation (matching its extractive format) but generated as KIE during synthesis (see \cref{app:syn_definitions}).}
To manage costs while maintaining sufficient data volume, we limit training sets to $4{,}000$ samples (except \docvqa[nr]{}, where we use the full train set to assess large-scale augmentation). This constraint reflects realistic resource limitations common in real-world applications. Details on dataset splits are given in Appendix~\ref{app:dataset_splits}.
We generate $1{,}000$--$10{,}000$ synthetic samples per dataset at sampling rates $\alpha \in \{0.5, 0.75, 1.0\}$, totaling over $140$k samples across all datasets (see \cref{tab:syn_datasets_details_final,tab:syn_datasets_details_ablation} in the appendix).\footnotemark[3]
Additionally, we introduce \docvqahw{}, a $103$-sample subset of \docvqa[nr]{} test split with handwritten content questions, to evaluate handwriting synthesis quality.\footnotemark[4]
\footnotetext[3]{For code release and data availability, see~Appendix~\ref{app:code_release}.}
\footnotetext[4]{For DocVQA-HW sample and question IDs, see Appendix~\ref{app:code_release}.}\par}


\paragraph{Models and Tasks.}
We benchmark a diverse set of models representing major architectures for document understanding. We consider document understanding models \bert{}, \lilt{}, and \layoutlm{} for CLS, KIE, and VQA tasks, and pure vision baselines \fasterrcnn{} and \cascadercnn{} for DLA.
Details to training hyperparameters are given in Appendix~\ref{app:training_hyperparameters}.

\paragraph{Metrics.}
Performance is measured using task-appropriate metrics: ANLS for \docvqa[nr]{}, WTQ for \wiki{}, exact-match accuracy for CLS, F1-score for KIE, and mean Average Precision (mAP) for DLA. For generation quality, we employ FID~\cite{HRU17} and Layout-FID~\cite{HRVA25} to assess distributional similarity between generated and real documents in pixel and learned feature spaces. We compute Layout-FID from \layoutlm[nr]{} CLS-token embeddings. For \publaynet[nr]{} and \icdarctdar[nr]{}, Layout-FID is computed from images only as text and bounding boxes are unavailable.

\subsection{Seed Selection Strategies}
We evaluate seed sampling strategies (\cref{sec:seed_selection}) by training \layoutlm[nr]{} (VQA, KIE, CLS) and \fasterrcnn[nr]{} (DLA) exclusively on synthetic data with cross-cluster ($\mathsf{CC}$) versus intra-cluster ($\mathsf{IC}$) sampling and varying $\alpha \in \{0.50, 0.75, 1.00\}$. \cref{tab:alpha_eval} reports means over five random seeds (std $<0.04$). Intra-cluster sampling consistently outperforms cross-cluster across all $\alpha$ values ($59.72\%$ vs. $56.91\%$ at $\alpha=1$), demonstrating that preserving structural coherence within document clusters is more critical than maximizing diversity across clusters. Within intra-cluster sampling, $\alpha=1$ achieves highest performance ($59.72\%$) and lowest Layout-FID (13.69), indicating that biasing generation toward dominant document patterns produces superior synthetic data. The improvement is particularly pronounced for classification tasks, where $\mathsf{IC}$ outperforms $\mathsf{CC}$ by $8.2$ points on \rvlcdip[nr]{} at $\alpha=1$. Based on these findings, we use intra-cluster sampling with $\alpha=1$ for all experiments. 

\begin{table}[!t]\centering

\rowcolors{2}{gray!15}{white} 
\resizebox{0.7\columnwidth}{!}{ 
\begin{tabular}{lrrrrrrrr}
\multirow{2}{*}{\textbf{Dataset}} &\multirow{2}{*}{\textbf{Smpl}} &\multicolumn{2}{c}{\textbf{{\boldmath$\alpha=0.50$}}} &\multicolumn{2}{c}{\textbf{{\boldmath$\alpha=0.75$}}} &\multicolumn{2}{c}{\textbf{{\boldmath$\alpha=1.00$}}} \\\cmidrule{3-8} \rowcolor{white}
& &\textbf{Score ($\uparrow$)} &\textbf{LFID ($\downarrow$)} &\textbf{Score ($\uparrow$)} &\textbf{LFID ($\downarrow$)} &\textbf{Score ($\uparrow$)} &\textbf{LFID ($\downarrow$)} \\\midrule
\docvqa[nr]{} &$\mathsf{CC}$ &62.46 &7.60 &61.51 &7.49 &\textbf{63.16 } &7.52 \\
\docvqa[nr]{} &$\mathsf{IC}$ &63.95 &6.88 &\textbf{64.27} &6.87 &63.64 &6.96 \\
\cord[nr]{} &$\mathsf{CC}$ &55.62 &37.31 &57.03 &36.95 &\textbf{57.74} &36.75 \\
\cord[nr]{} &$\mathsf{IC}$ &57.51 &36.92 &56.56 &37.01 &\textbf{58.40} &36.46 \\
\rvlcdip[nr]{} &$\mathsf{CC}$ &43.65 &11.15 &45.49 &12.45 &\textbf{45.74} &11.06 \\
\rvlcdip[nr]{} &$\mathsf{IC}$ &51.90 &8.62 &51.04 &9.88 &\textbf{53.94} &8.82 \\
\publaynet[nr]{} &$\mathsf{CC}$ &61.99 &3.53 &\textbf{63.09} &2.75 &61.00 &2.81 \\
\publaynet[nr]{} &$\mathsf{IC}$ &63.06 &2.63 &\textbf{63.41} &2.63 &62.90 &2.50 \\
\textbf{Average} &$\mathsf{CC}$ &55.93 &14.90 &56.78&14.91 &\textbf{56.91} &\textbf{14.54} \\
\textbf{Average} &$\mathsf{IC}$ &59.11 &13.76 &58.82 &14.10 &\textbf{59.72} &\textbf{13.69} \\
\end{tabular}
}
\caption{Layout-FID (LFID)~\cite{HRVA25} and performance comparison for $\mathsf{CC}$ and $\mathsf{IC}$ sampling across different $\alpha$ values on \layoutlm[nr]{} (VQA, KIE, CLS) and \fasterrcnn[nr]{} (DLA). Results averaged over 5 seeds (std $< 0.04$).}
\label{tab:alpha_eval}
\vspace{-1em}
\end{table}

\begin{table*}[!t]\centering

\rowcolors{2}{gray!15}{white} 
\resizebox{0.95\textwidth}{!}{ 
\begin{tabular}{lllrrrr|rr|rr|rr}
\multicolumn{4}{l}{}&\multicolumn{3}{c|}{$\mathbf{Full}$}&\multicolumn{2}{c|}{$\mathbf{Few_A}$} &\multicolumn{2}{c|}{$\mathbf{Few_B}$}&\multicolumn{2}{c}{$\mathbf{\Delta}$}\\
\cmidrule{5-13} 
\rowcolor{white}
\textbf{Model} &\textbf{Dataset} &\textbf{Task} &\textbf{Metric ($\uparrow$)} &$\mathbf{R}$ &$\mathbf{S}$ &$\mathbf{R+S}$ &$\mathbf{R}$ &$\mathbf{R+S}$ &$\mathbf{R}$ &$\mathbf{R+S}$ &$\mathbf{Full{-}Few_B}$ & $\mathbf{R{-}S}$ \\\midrule
\bert[nr]{} &\docvqa[nr]{} &VQA &ANLS &57.97 &52.92 &61.33 &46.63 &55.45 &25.61 &54.04 &3.93 &5.05 \\
\lilt[nr]{} &\docvqa[nr]{} &VQA &ANLS &70.50 &64.34 &72.53 &58.99 &66.95 &38.59 &64.11 &6.39 &6.16 \\
\layoutlm[nr]{} &\docvqa[nr]{} &VQA &ANLS &71.45 &66.03 &\textbf{73.26} &62.91 &\textbf{68.04} &43.61 &\textbf{65.82} &5.63 &5.42 \\
\bert[nr]{} &\docvqahw[nr]{} &VQA &ANLS &48.24 &42.26 &51.19 &40.94 &44.07 &26.47 &42.72 &5.51 &5.97 \\
\lilt[nr]{} &\docvqahw[nr]{} &VQA &ANLS &58.94 &50.26 &58.24 &50.21 &50.95 &35.88 &50.78 &8.16 &8.67 \\
\layoutlm[nr]{} &\docvqahw[nr]{} &VQA &ANLS &\textbf{59.41} &51.25 &58.36 &52.31 &\textbf{53.31} &41.20 &\textbf{51.24} &8.17 &8.16 \\
\bert[nr]{} &\wiki[nr]{} &VQA &WTQ &16.51 &9.24 &19.39 &15.68 &18.50 &6.65 &13.84 &2.68 &7.27 \\
\lilt[nr]{} &\wiki[nr]{} &VQA &WTQ &26.71 &14.79 &\textbf{30.90} &24.39 &\textbf{29.50} &10.91 &\textbf{22.46} &4.25 &11.92 \\
\layoutlm[nr]{} &\wiki[nr]{} &VQA &WTQ &25.64 &12.76 &29.47 &24.20 &28.66 &7.86 &21.61 &4.04 &12.89 \\
\multicolumn{4}{r}{\textbf{Average VQA}} &48.37 &40.43 &50.52 &41.81 &46.16 &26.31 &42.96 &\textbf{5.42} &\textbf{7.95} \\\midrule
\bert[nr]{} &\cord[nr]{} &KIE &F1 &93.78 &49.06 &93.92 &90.27 &90.97 &84.16 &85.80 &7.99 &44.72 \\
\lilt[nr]{} &\cord[nr]{} &KIE &F1 &94.61 &55.76 &94.88 &93.04 &93.17 &88.28 &88.90 &5.71 &38.85 \\
\layoutlm[nr]{} &\cord[nr]{} &KIE &F1 &95.92 &58.84 &\textbf{96.56} &94.64 &\textbf{95.13} &90.39 &\textbf{92.05} &3.87 &37.08 \\
\bert[nr]{} &\funsd[nr]{} &KIE &F1 &56.33 &40.84 &59.19 &-&-&54.18 &56.46 &-0.13 &15.49 \\
\lilt[nr]{} &\funsd[nr]{} &KIE &F1 &74.03 &49.13 &74.90 &-&-&71.91 &72.49 &1.54 &24.90 \\
\layoutlm[nr]{} &\funsd[nr]{} &KIE &F1 &\textbf{88.56} &49.10 &87.74 &-&- &\textbf{87.46} &85.69 &2.87 &39.46 \\
\bert[nr]{} &\klc[nr]{} &KIE &F1 &45.22 &41.90 &45.00 &43.98 &44.46 &33.66 &43.11 &2.11 &3.32 \\
\lilt[nr]{} &\klc[nr]{} &KIE &F1 &46.08 &43.66 &\textbf{46.25} &45.07 &45.32 &37.44 &\textbf{44.08} &2.00 &2.41 \\
\layoutlm[nr]{} &\klc[nr]{} &KIE &F1 &46.11 &43.24 &46.11 &\textbf{45.43} &45.22 &37.84 &43.64 &2.47 &2.86 \\
\bert[nr]{} &\sroie[nr]{} &KIE &F1 &88.12 &60.94 &88.97 &83.78 &85.32 &74.90 &79.35 &8.77 &27.18 \\
\lilt[nr]{} &\sroie[nr]{} &KIE &F1 &94.03 &70.79 &93.49 &91.61 &91.29 &83.21 &87.62 &6.42 &23.24 \\
\layoutlm[nr]{} &\sroie[nr]{} &KIE &F1 &94.17 &72.32 &\textbf{94.60} &91.05 &\textbf{93.12} &83.82 &\textbf{90.49} &3.68 &21.85 \\
\multicolumn{4}{r}{\textbf{Average KIE}} &76.41 &52.97 &76.80 &75.43 &76.00 &68.94 &72.47 &\textbf{3.94} &\textbf{23.45} \\\midrule
\bert[nr]{} &\doclaynetcls[nr]{} &CLS &Acc &95.59 &81.12 &94.78 &93.98 &93.37 &72.03 &83.09 &12.50 &14.47 \\
\lilt[nr]{} &\doclaynetcls[nr]{} &CLS &Acc &96.35 &83.97 &94.31 &95.32 &93.47 &80.24 &84.79 &11.56 &12.39 \\
\layoutlm[nr]{} &\doclaynetcls[nr]{} &CLS &Acc &\textbf{97.33} &82.89 &97.27 &\textbf{96.33} &96.31 &66.94 &\textbf{89.90} &7.43 &14.45 \\
\bert[nr]{} &\rvlcdip[nr]{} &CLS &Acc &76.87 &44.40 &75.23 &70.06 &67.65 &33.48 &55.42 &21.45 &32.46 \\
\lilt[nr]{} &\rvlcdip[nr]{} &CLS &Acc &78.78 &48.72 &77.68 &72.41 &69.43 &41.16 &57.99 &20.79 &30.06 \\
\layoutlm[nr]{} &\rvlcdip[nr]{} &CLS &Acc &\textbf{86.26} &53.84 &85.64 &\textbf{80.84} &80.29 &19.68 &\textbf{64.93} &21.33 &32.42 \\
\bert[nr]{} &\tobacco[nr]{} &CLS &Acc &86.05 &59.62 &84.57 &81.48 &79.86 &36.91 &63.52 &22.53 &26.43 \\
\lilt[nr]{} &\tobacco[nr]{} &CLS &Acc &88.05 &61.76 &84.19 &86.57 &79.19 &50.67 &68.48 &19.57 &26.29 \\
\layoutlm[nr]{} &\tobacco[nr]{} &CLS &Acc &92.43 &61.14 &\textbf{93.38} &\textbf{92.14} &91.62 &37.90 &\textbf{78.86} &\textbf{13.57} &\textbf{31.28} \\
\multicolumn{4}{r}{\textbf{Average CLS}} &88.63 &64.16 &87.45 &85.46 &83.47 &48.78 &71.89 &16.75 &24.47 \\\midrule
\cascadercnn[nr]{} &\doclaynetdla[nr]{} &DLA &AP &49.74 &10.39 &\textbf{50.20} &36.96 &36.16 &13.76 &\textbf{19.55} &30.19 &39.35 \\
\fasterrcnn[nr]{} &\doclaynetdla[nr]{} &DLA &AP &50.03 &6.60 &48.47 &\textbf{37.33} &35.42 &7.89 &17.84 &32.19 &43.43 \\
\cascadercnn[nr]{} &\icdarctdar[nr]{} &DLA &AP &87.69 &64.06 &\textbf{91.13} &84.08 &\textbf{88.09} &67.48 &\textbf{84.65} &3.05 &23.64 \\
\fasterrcnn[nr]{} &\icdarctdar[nr]{} &DLA &AP &85.52 &62.10 &88.56 &82.64 &85.04 &71.37 &83.26 &2.26 &23.42 \\
\cascadercnn[nr]{} &\publaynet[nr]{} &DLA &AP &\textbf{90.84} &62.25 &90.97 &\textbf{88.95} &87.92 &\textbf{77.97} &78.06 &12.78 &28.59 \\
\fasterrcnn[nr]{} &\publaynet[nr]{} &DLA &AP &85.72 &58.94 &85.70 &82.93 &82.36 &71.85 &72.84 &12.88 &26.78 \\
\multicolumn{4}{r}{\textbf{Average DLA}} &74.92 &44.06 &75.84 &68.81 &69.17 &51.72 &59.37 &\textbf{15.56} &\textbf{30.87} \\\midrule
\multicolumn{4}{r}{\textbf{Average all}} &72.21 &51.15 &72.73 &67.79 &68.66 &50.37 &62.76 &\textbf{9.45} &\textbf{21.06} \\

\end{tabular}
}
\caption{Performance across datasets and tasks with three training settings: (1)~$\mathrm{Full}$: models trained on all real (R), all synthetic (S), or both (R+S); (2)~$\mathrm{Few}_A$: $300$--$1000$ real samples with/without synthetic augmentation; (3)~$\mathrm{Few}_B$: 100 real samples with/without synthetic augmentation. Gap columns ($\Delta$) report: $\mathrm{R_{Full}} - (\mathrm{R{+}S})_{\mathrm{Few}_B}$ (how close 100 augmented samples approaches full training) and $\mathrm{R_{Full}} - \mathrm{S_{Full}}$ (real vs synthetic quality). Bold indicates best per setting (std $< 0.02$). Synthetic augmentation achieves +12.38 average improvement in $\mathrm{Few}_B$ scenarios.}


\vspace{-1em}

\label{tab:main_results}
\end{table*}

\subsection{Downstream Task Performance}
We evaluate models on VQA, KIE, CLS, and DLA tasks under three data regimes: full-shot ($\mathrm{Full}$), few-shot with $300-1000$ real samples ($\mathrm{Few}_A$), and few-shot with $100$ real samples ($\mathrm{Few}_B$). Results in \cref{tab:main_results} report means over three random seeds (std $<0.02$).

Synthetic-only training demonstrates substantial quality. On \docvqa{}, real data outperforms pure synthetic by $5.05$ points, but augmenting just $100$ real samples reduces this gap to $3.93$ points. On \klc{}, the gap is even smaller: real data exceeds pure synthetic by only $2.41$ points, narrowing to $2.00$ points with $100$ real samples added. Performance gaps are larger on specialized datasets like \cord{} ($37$--$45$ points) and \doclaynetdla{} ($39$--$43$ points), reflecting domain-specific challenges in replicating real-world capture artifacts and complex annotations.

Combining real and synthetic data consistently matches or exceeds real-only performance, with improvements on \docvqa[nr]{}, \wiki[nr]{}, and \icdarctdar[nr]{}, averaging $+0.51$ points. Notably, vision-based \layoutlm[nr]{} degrades on \docvqahw[nr]{} when adding synth data while text-only \bert[nr]{} improves, revealing that synthetic handwriting lacks authentic visual characteristics despite recognizable content, though the difficulty of this handwriting-focused subset makes isolated quality assessment challenging. 

Synthetic augmentation proves most valuable in data-scarce settings. In Setting $\mathrm{Few}_B$ with only $100$ real samples, adding synthetic data yields $+12.38$ average improvement, bringing performance within $9.45$ points of full real-data training. This demonstrates substantial annotation cost reduction: augmenting minimal labeled data with synthetic samples achieves $87\%$ of full-dataset performance.

\subsection{Visual Quality Comparison}

\cref{tab:layoutfid} shows our framework achieves strong visual fidelity across diverse document types, with Layout-FID~\cite{HRVA25} scores below $10$ for most datasets: \wiki[nr]{} ($3.13$), \doclaynet[nr]{} ($6.35$--$6.45$), \docvqa[nr]{} ($6.96$), and \klc[nr]{} ($7.98$). Performance on \cord[nr]{} is noticeably worse ($36.46$ Layout-FID, $139.52$ FID). However, this is expected as \cord[nr]{} consists of camera-captured receipt images that contain real-world artifacts such as blur, lighting variation, and complex real-world backgrounds.

While direct comparison is challenging due to the different nature of setting across multiple document synthesis frameworks (as in ~\cref{tab:related}), we also compare the FID scores achieved by our approach with multiple existing works~\cite{HRVA25,BRLP21,TUS23,FLL23}.
Against DocGenie~\cite{HRVA25}, which we extend, our approach achieves lower FID on \cord[nr]{} ($139.52$ vs. $155.34$) and \sroie[nr]{} ($63.50$ vs. $109.31$), though DocGenie reports superior Layout-FID on these datasets ($31.30$ vs. $36.46$ on \cord[nr]{}; $3.52$ vs. $17.18$ on \sroie[nr]{})\footnotemark[5]\footnotetext[5]{DocGenie~\cite{HRVA25} is closed-source, preventing verification of their exact Layout-FID computation. FID-based metrics can be sensitive to sample size and implementation details. We compute Layout-FID using LayoutLMv3 CLS token embeddings, which may differ from DocGenie's unspecified approach.}.
Note that our $\mathsf{CC}$ sampling with $\alpha=1$ replicates DocGenie's~\cite{HRVA25} seed-guided generation strategy; however, DocGenie uses $10$ seeds, while we use $6$ seeds for these datasets with prompting optimized for GT generation.
Other methods~\cite{BRLP21,TUS23,FLL23} which are mostly diffusion-based achieve better FID scores on \publaynet[nr]{} ($33.75$ at $128\times128$, $15.02$ at $256\times256$ vs. our $35.28$ at full resolution) and \doclaynetdla[nr]{} ($20.58$ at $256\times256$ vs. $37.80$) but this is expected since all these approaches synthesize new documents by training on the training distribution of the same dataset. 
Furthermore, these approaches typically synthesize at lower resolutions ($128\times128$ to $256\times256$) and require ground truth layout annotations as input, fundamentally differing from our annotation-free approach. 
Our framework trades some visual fidelity for complete annotated dataset synthesis from unlabeled documents, enabling direct supervised learning across multiple tasks.
For additional qualitative visual results of our framework, refer to Appendix~\ref{app:synthdata}.



\begin{table}[!t]\centering
\rowcolors{2}{gray!15}{white} 
\resizebox{0.5\linewidth}{!}{

\begin{tabular}{lllrrr}\toprule
\textbf{Dataset} &\textbf{Method} &\textbf{Task} &\textbf{FID ($\downarrow$)} &\textbf{LayoutFID ($\downarrow$)} \\\midrule
\cellcolor{white}\docvqa[nr]{} &Ours &VQA &41.36 &6.96 \\
\cellcolor{white}\wiki[nr]{} &Ours &VQA &52.43 &3.13 \\
\midrule
&DocGenie~\cite{HRVA25} &KIE &155.34 &31.30 \\
\multirow{-2}{*}{\cellcolor{white}\cord[nr]{}}&Ours &KIE &139.52 &36.46 \\
\cellcolor{white}\funsd[nr]{} &Ours &KIE &44.57 &9.60 \\
\cellcolor{white}\klc[nr]{} &Ours &KIE &26.98 &7.98 \\
&DocGenie~\cite{HRVA25} &KIE &109.31 &3.52 \\
\multirow{-2}{*}{\cellcolor{white}\sroie[nr]{}}&Ours &KIE &63.50 &17.18 \\
\midrule
\cellcolor{white}\rvlcdip[nr]{} &Ours &CLS &86.59 &8.82 \\
\cellcolor{white}\tobacco[nr]{} &Ours &CLS &61.86 &14.43 \\
\cellcolor{white}\doclaynetcls[nr]{} &Ours &CLS &36.62 &6.45 \\
\midrule

\cellcolor{white}&DocSynth~\cite{BRLP21} &DLA & 33.75 @ $128\times128$ &- \\
\cellcolor{white}&Tanveer\etal~\cite{TUS23} &DLA &15.02 @ $256\times256$ &- \\
\cellcolor{white}&Fenrir\etal~\cite{FLL23} &DLA &248 @ $256 \times256$ &- \\
\multirow{-4}{*}{\cellcolor{white}\publaynet[nr]{}}\cellcolor{white}&Ours &DLA &35.28 &2.50 \\
\cellcolor{white}\icdarctdar[nr]{} &Ours &DLA &43.52 &7.19 \\
&Tanveer\etal~\cite{TUS23} &DLA &20.58 @ $256\times256$ &- \\
\multirow{-2}{*}{\cellcolor{white}\doclaynetdla[nr]{}}&Ours &DLA &37.80 &6.35 \\
\bottomrule
\end{tabular}
}
\caption{FID~\cite{HRU17} and Layout-FID~\cite{HRVA25} scores comparing our method to prior work. Lower scores indicate better distributional similarity to real documents. Resolution annotations indicate synthesis resolution before upscaling for evaluation. 
}
\label{tab:layoutfid}

\vspace{-1em}
\end{table}


\subsection{Analysis of Failure-Cases}\label{sec:failure_cases}
While synthetic data improves few-shot performance and maintains competitive full-shot results (\cref{tab:main_results}), qualitative analysis reveals systematic failure modes. For KIE, despite reasonable spatial distributions (Appendix~\ref{app:synthdata_gt_kie}), synth-only achieves $49$--$72$ F1 vs. $88$--$95$ real. 
Real samples are camera captures with scanning artifacts and distortions absent in pristine synthetic documents; \cord[nr]{} additionally applies artificial selective blur to non-KIE regions. Unlike DocGenie~\cite{HRVA25}, which applies synthetic degradation post-generation, our focus on multi-task GT generation produces clean documents, creating a visual domain gap evidenced by \layoutlm[nr]{} degrading on \funsd[nr]{} ($88.56$$\rightarrow$$87.74$ F1) while text-only models improve. For CLS, severe class imbalance (\cref{fig:cls_distributions}~in Appendix) yields synth-only accuracy of $44$--$61\%$ vs. $76$--$92\%$ real, with VLMs generating memos ($\sim45\%$) while neglecting specialized classes ($<2\%$). For DLA, synth-only achieves $6$--$10$ AP vs. $49$--$50$ real; however, qualitative analysis confirms reasonable predictions, indicating low scores stem from annotation inconsistencies rather than synthesis failure. 
Overall, GT analysis (Appendix~\ref{app:synthdata_gt}) validates high semantic and spatial annotation quality - question embeddings align closely between real and synthetic samples, and spatial entity distributions are well-preserved - though class imbalance remains a limitation for classification tasks. Manual inspection reveals $\sim$3\% of documents across all tasks exhibit rendering failures or anomalous layouts.

\subsection{Discussion}
{\tolerance=1200 Our experiments demonstrate that VLM-based synthesis generates high-quality document distributions suitable for model training. Synthetic-only training achieves $70.8\%$ of real-data performance on average ($51.15\%$ vs. $72.21\%$), closely approximating real data on several datasets with gaps as small as $2.41$ points (\klc{}) and $5.05$ points (\docvqa{}). Combining real and synthetic data consistently improves results: $+0.51$ in full-shot and $+0.83$ in few-shot with $300-1000$ samples. Most notably, augmenting only $100$ real samples with synthetic data yields $+12.38$ improvement, achieving $87\%$ of full real-data performance and demonstrating substantial annotation cost reduction. Seed selection analysis (\cref{tab:alpha_eval}) confirms that intra-cluster sampling with $\alpha=1$ ($59.72\%$ vs. $56.91\%$) preserves structural coherence more effectively than cross-cluster approaches.


To assess reproducibility with open-weight models, we additionally evaluate Gemma 3 27B (Appendix~\ref{app:opensource_vlms}). While it achieves comparable visual quality when successful (e.g., similar FID/Layout-FID), pipeline success rates are significantly lower — particularly for GT generation — reflecting instruction-following limitations in current open-source VLMs rather than framework constraints. As open-weight VLMs continue to improve, we expect this gap to narrow, making our framework fully reproducible without proprietary models.

Though successful, challenges persist:
pristine synthetic documents lack real-world degradations that vision encoders utilize, class imbalance emerges in classification tasks, and annotation taxonomy differences affect DLA scores despite qualitatively reasonable predictions. Integrating document degradation techniques (as in DocGenie~\cite{HRVA25}) with our GT generation framework could address visual domain gaps, while constrained sampling strategies could improve class balance. Overall, our framework offers a scalable, privacy-preserving approach that substantially reduces annotation costs while maintaining competitive performance across document understanding tasks.\par}

\section{Conclusion}
We present a scalable framework for synthetic document generation that addresses labeled data scarcity in document understanding through VLM-based content generation, automatic ground truth annotation from unlabeled seed documents, and intelligent clustering-based seed selection. Our approach produces visually realistic documents with task-specific annotations across VQA, KIE, CLS, and DLA tasks. We release $140K$+ synthetic samples across eleven datasets and \docvqahw[nr]{} to support other researchers.

Comprehensive evaluation demonstrates substantial annotation cost reduction: 100 real samples augmented with synthetic data achieves $87\%$ of full real-data performance. Synthetic-only training reaches competitive performance compared to real data, while Layout-FID scores predominantly below $10$ validate strong visual fidelity across diverse document types. Our clustering-based seed selection with intra-cluster sampling effectively preserves structural coherence and target distributions.

Future work should integrate existing degradation techniques (as in DocGenie~\cite{HRVA25}) with our multi-task GT generation framework to bridge the visual domain gap, implement constrained sampling strategies to address class imbalance, and explore content-aware generation for all visual element types. We are convinced that our framework helps accelerate data-efficient document understanding research and enables practitioners to train competitive models with minimal annotation costs.

\begin{credits}
\subsubsection{\ackname}
This work was partially funded by the German Federal Ministry of Education and Research (BMBF).

\subsubsection{\discintname}
The authors have no competing interests to declare that are relevant to the content of this article.
\end{credits}

\bibliographystyle{splncs04}
\bibliography{biblography}

\newpage
\clearpage
\setcounter{page}{1}
\maketitlesupplementary

\appendix

\section{Claude vs. Gemma 3 27B Comparison}\label{app:opensource_vlms}
To improve reproducibility we've done formal evaluation with Gemma 3 27B (\cref{tab:gemma_layoutfid}), which shows significantly lower pipeline success but comparable visual quality when successful---reflecting instruction-following limitations in current open-source VLMs, not framework constraints.

\begin{table}[htb]\centering
\rowcolors{5}{white}{gray!15}
\resizebox{\linewidth}{!}{
\begin{tabular}{ll l rrr rr r}\toprule
& & & \multicolumn{3}{c}{\textbf{Pipeline Success (\%)}} & \multicolumn{3}{c}{\textbf{Quality}} \\
\cmidrule(lr){4-6} \cmidrule(lr){7-9}
\textbf{Dataset} &\textbf{Task} &\textbf{VLM} & \textbf{GT} & \textbf{SP} & \textbf{Vis} & \textbf{FID $\downarrow$} &\textbf{LFID $\downarrow$} & \textbf{N} \\
\midrule
 &  & Claude & \textbf{92.6} & \textbf{92.2} & \textbf{95.9} & \textbf{45.9} & \textbf{7.0} & 6864 \\
\multirow{-2}{*}{\cellcolor{white} DocVQA} & \multirow{-2}{*}{\cellcolor{white} VQA} & Gemma & 69.48 & 0.2 & 73.47 & 73.9 & 9.4 & 6864 \\
\midrule
 &  & Claude & \textbf{95.5} & \textbf{95.5} & \textbf{93.5} & \textbf{51.2} & \textbf{9.6} & 140 \\
\multirow{-2}{*}{\cellcolor{white} FUNSD} & \multirow{-2}{*}{\cellcolor{white} KIE} & Gemma & 71.9 & 0.0 & 71.1 & 83.2 & 11.2 & 134 \\
\midrule
 & & Claude & \textbf{91.6} & \textbf{97.1} & \textbf{98.6} & 93.5 & \textbf{10.6} & 850 \\
\multirow{-2}{*}{\cellcolor{white} RVL-CDIP} & \multirow{-2}{*}{\cellcolor{white} CLS} & Gemma & 8.5 & 0.7 & 94.6 & \textbf{93.4} & 14.3 & 850 \\
\midrule
 & & Claude & \textbf{89.0} & \textbf{91.2} & \textbf{97.8} & \textbf{37.5} & \textbf{3.0} & 1235 \\
\multirow{-2}{*}{\cellcolor{white} PubLayNet} & \multirow{-2}{*}{\cellcolor{white} DLA} & Gemma & 12.2 & 0.3 & 94.2 & 81.2 & 9.8 & 1235 \\
\bottomrule
\end{tabular}
}
\caption{Claude vs. Gemma 3 27B comparison. \textbf{Pipeline Success}: GT = valid annotations, SP = single-page renders, Vis = valid visual element and handwriting definitions (documents without handwriting/visual elements are counted as valid). \textbf{Quality}: FID/LFID (Layout-FID) computed on N samples (lower = better).}
\label{tab:gemma_layoutfid}
\end{table}

\section{Dataset Splits}\label{app:dataset_splits}
\begin{table}[htb] 
\centering
\rowcolors{2}{gray!15}{white} 

\resizebox{0.5\linewidth}{!}{ 
\begin{threeparttable}
\begin{tabular}{lrrrrr}
\textbf{Dataset} &\textbf{Train} &\textbf{Validation} &\textbf{Test} &\textbf{Train (Synth)} \\\midrule
\docvqa{} &10194 &1286 &1287 &8082 \\
\docvqahw{} &N/A &N/A &103 &N/A \\
\wiki{} &1350 &337 &421 &1479 \\
\sroie{} &626 &N/A &347 &1008 \\
\funsd{} &149 &N/A &50 &259 \\
\cord{} &800 &100 &100 &1182 \\
\klc{} &3641 &953 &1309 &3441 \\
\tobacco{} &2782 &N/A &700 &4092 \\
\rvlcdip{} &4000* &4000 &39998 &3819 \\
\doclaynetcls{} &4000* &1000* &4999 &3978 \\
\doclaynetdla{} &4000* &1000* &4999 &3732 \\
\publaynet{} &4000* &11245 & 11405 &3835 \\
\icdarctdar{} &600 &N/A &240 &1515 \\
\end{tabular}

\begin{tablenotes}
\item[*] For these datasets, we use a subset of the original training splits.
\end{tablenotes}
\end{threeparttable}
}
\caption{
Train/validation/test splits and synthetic training sets for all datasets used in our experiments. For datasets with no validation set, we use 5\% of the training set as the validation set.
}\label{tab:train-eval-splits}
\vspace{-1em}
\end{table}
Details on our dataset splits are given in \cref{tab:train-eval-splits}. As discussed in \cref{sec:experiments}, we limit training sets to $4{,}000$ samples (except \docvqa[nr]{}, where we use the full train set to assess large-scale augmentation)  to manage costs while maintaining sufficient data volume. This constraint reflects realistic resource limitations common in real-world applications.

\section{Embeddings and Clustering}\label{app:clustering}
\begin{sloppypar}
To create the embeddings, we use the following checkpoints:
\texttt{micro\-soft/\-layout\-lmv3-base} for \texttt{layout} and \texttt{pooled},
\texttt{openai/\-clip-vit-base-patch32} for \texttt{clip}, and
\texttt{all-mpnet-base-v2} for \texttt{sentence}.
\end{sloppypar}
\begin{table}[!b]\centering
\label{tab:clustering_rankings}
\begin{tabular}{lrrr}\toprule
\textbf{Embedding} &\textbf{$\kappa$} &\textbf{Rank Score $\uparrow$} \\\midrule
combined &10 &74 \\
combined &5 &72 \\
sentence &5 &65 \\
clip &5 &56 \\
clip &10 &55 \\
sentence &10 &51 \\
pooled &10 &50 \\
pooled &5 &48 \\
layout &10 &35 \\
layout &5 &27 \\
\bottomrule
\end{tabular}
\caption{Clustering configurations ranked using cumulative position scores over our base datasets.}
\end{table}
\begin{table*}[t]\centering\label{tab:cluster_metrics}
\resizebox{0.8\textwidth}{!}{ 
\rowcolors{2}{gray!15}{white} 
\begin{tabular}{lrrrrrrr}
\textbf{Dataset} &\textbf{Embedding} &\textbf{$\kappa$} &\textbf{Num Clusters} &\textbf{Silhouette Score $\uparrow$} &\textbf{Norm. Entropy $\uparrow$} &\textbf{Final Score $\uparrow$} \\\midrule
\sroie[nr]{} &combined &10 &14 &0.64 &0.94 &0.79 \\
\icdarctdar[nr]{} &clip &5 &9 &0.64 &0.82 &0.73 \\
\wiki[nr]{} &combined &5 &50 &0.41 &0.95 &0.68 \\
\cord[nr]{} &combined &10 &22 &0.39 &0.96 &0.68 \\
\tobacco[nr]{} &combined &10 &31 &0.42 &0.93 &0.67 \\
\doclaynet[nr]{} &combined &10 &48 &0.51 &0.82 &0.66 \\
\rvlcdip[nr]{} &combined &10 &49 &0.38 &0.92 &0.65 \\
\funsd[nr]{} &combined &10 &4 &0.36 &0.92 &0.64 \\
\klc[nr]{} &combined &10 &41 &0.35 &0.86 &0.61 \\
\publaynet[nr]{} &clip &5 &106 &0.30 &0.89 &0.60 \\
\docvqa[nr]{} &sentence &5 &408 &0.41 &0.96 &0.69 \\
\docvqa[nr]{} &combined &5 &362 &0.37 &0.96 &0.67 \\
\docvqa[nr]{} &sentence &10 &192 &0.39 &0.92 &0.66 \\
\docvqa[nr]{} &combined &10 &187 &0.37 &0.94 &0.65 \\
\docvqa[nr]{} &pooled &5 &370 &0.35 &0.95 &0.65 \\
\docvqa[nr]{} &pooled &10 &181 &0.35 &0.94 &0.65 \\
\docvqa[nr]{} &clip &10 &123 &0.33 &0.91 &0.62 \\
\docvqa[nr]{} &clip &5 &259 &0.29 &0.91 &0.60 \\
\docvqa[nr]{} &layout &5 &282 &0.25 &0.95 &0.60 \\
\docvqa[nr]{} &layout &10 &128 &0.23 &0.93 &0.58 \\
\end{tabular}
}
\caption{Metrics of the selected clusterings for all datasets (top) and listing of all clusterings for \docvqa[nr]{} (bottom), where we used \texttt{combined} embeddings and $\kappa=10$ for our experiments.}
\end{table*}
We select the optimal clustering $(\mathcal{E}^*, \kappa^*)$ by maximizing a heuristic quality score:
\begin{equation}
    (\mathcal{E}^*, \kappa^*) = \arg\max_{\mathcal{E}, \kappa} \left[ S(C_{\mathcal{E},\kappa}) + H(C_{\mathcal{E},\kappa}) \right]
\end{equation}
where $S(C)$ is the silhouette score~\cite{Rou87} measuring cluster compactness and $H(C) = -\sum_{c=1}^K p_c \log p_c$ is normalized entropy measuring cluster balance (where $p_c$ denotes a cluster's proportion of the samples).
This heuristic prioritizes clusterings with both high internal coherence and balanced cluster sizes, which aligns with our manual inspection showing that such configurations produce semantically meaningful, interpretable document groupings suitable for seed selection.

\begin{figure*}[p] 
    \centering
    \begin{subfigure}[b]{0.3\textwidth}
        \includegraphics[width=\textwidth]{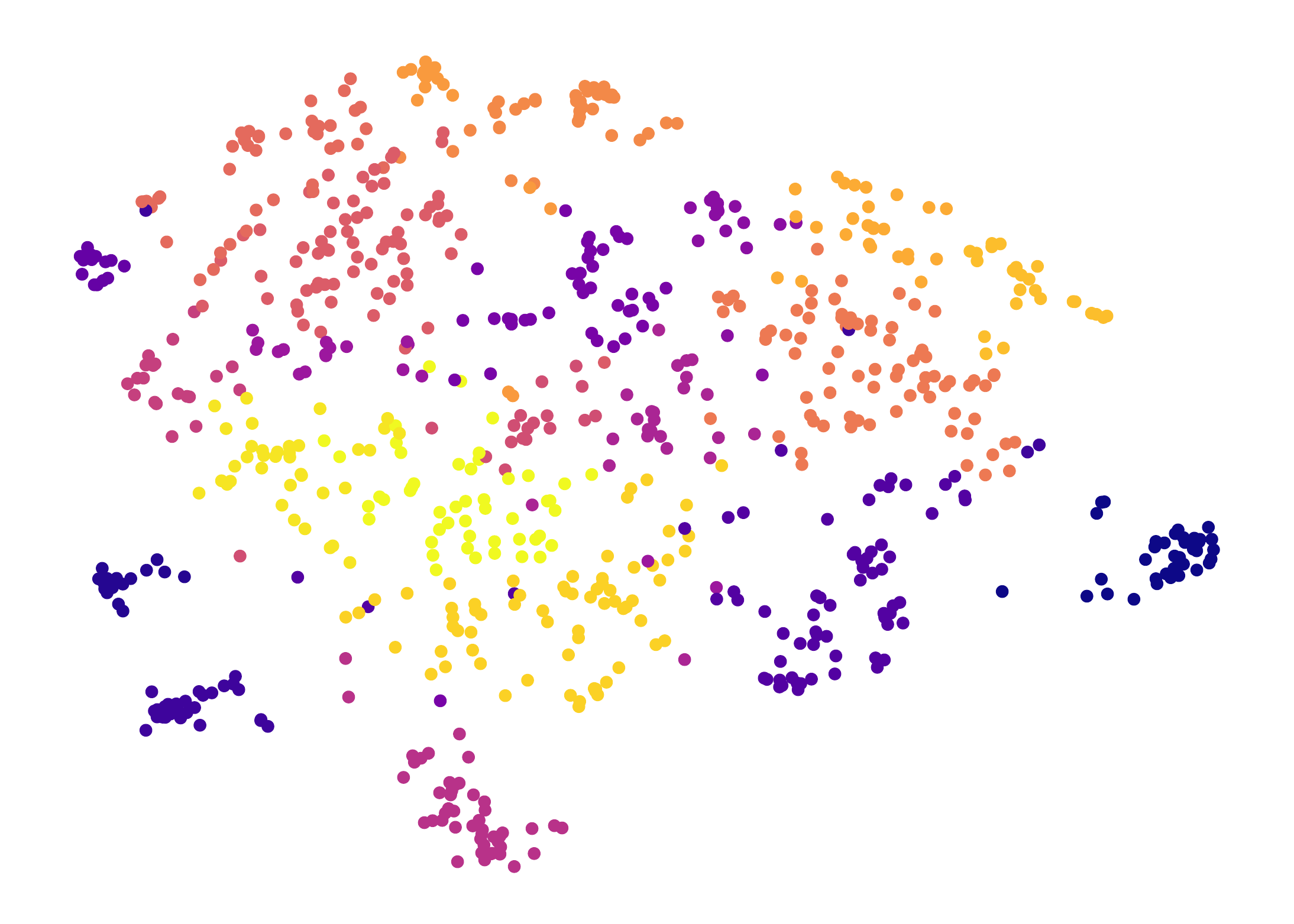}
        \caption{\cord{}, \texttt{combined}, $\kappa=10$, 22 Clusters}
    \end{subfigure} \hfill
    \begin{subfigure}[b]{0.3\textwidth}
        \includegraphics[width=\textwidth]{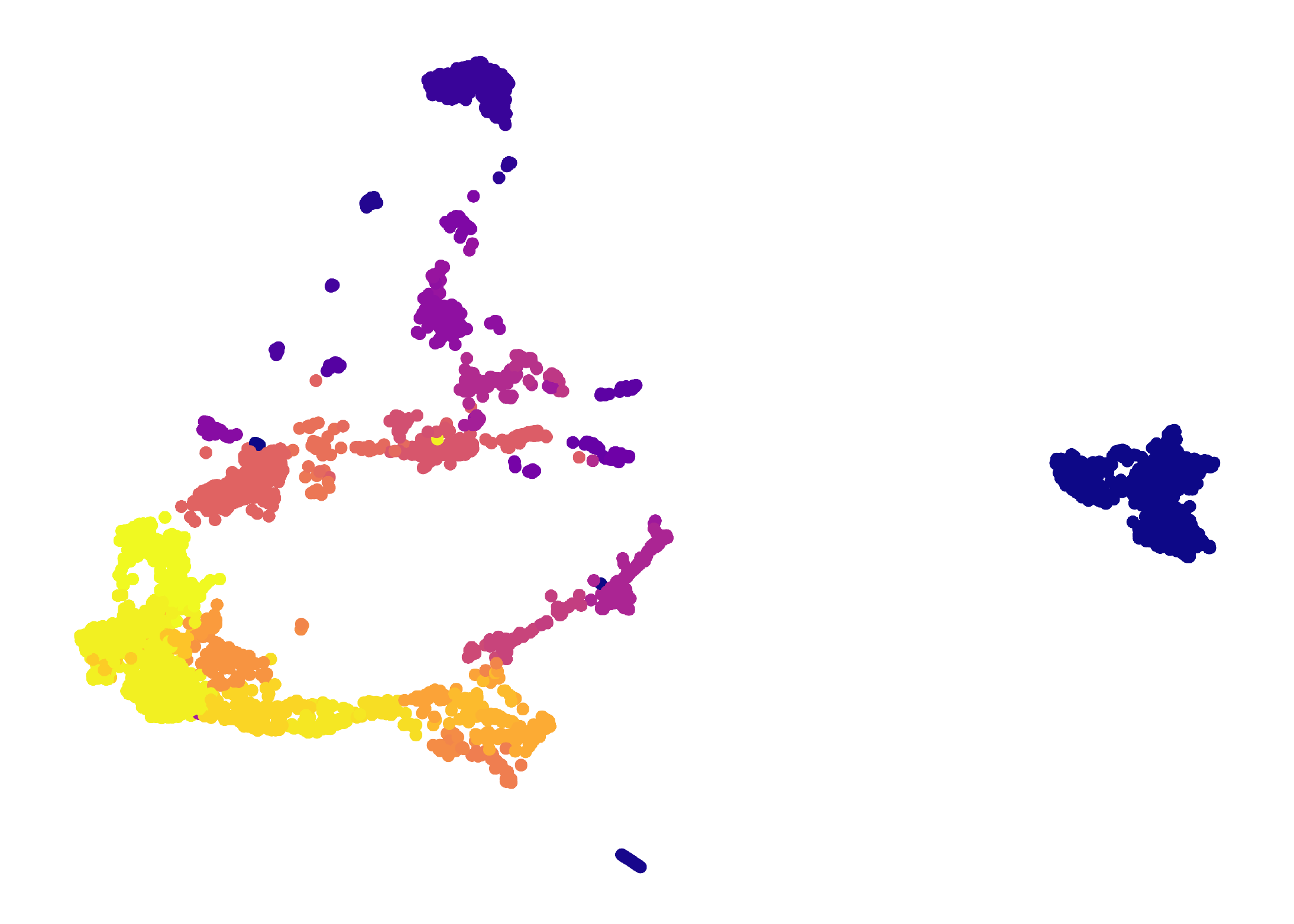}
        \caption{\doclaynet{}, \texttt{combined}, $\kappa=10$, 48 Clusters}
    \end{subfigure} \hfill
    \begin{subfigure}[b]{0.3\textwidth}
        \includegraphics[width=\textwidth]{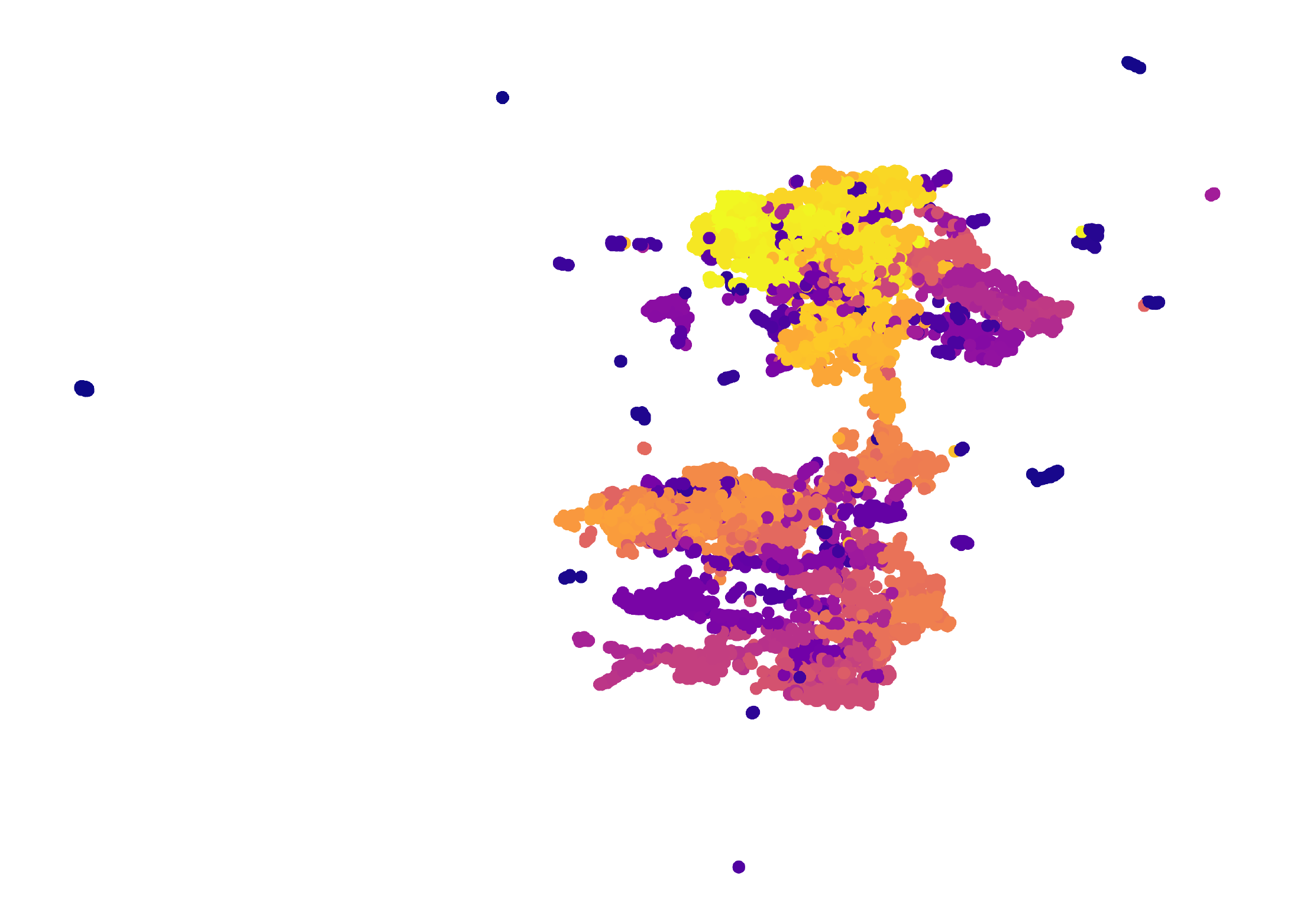}
        \caption{\docvqa{}, \texttt{combined}, $\kappa=10$, 187 Clusters}
    \end{subfigure}
    \vspace{2mm}
    \begin{subfigure}[b]{0.3\textwidth}
        \includegraphics[width=\textwidth]{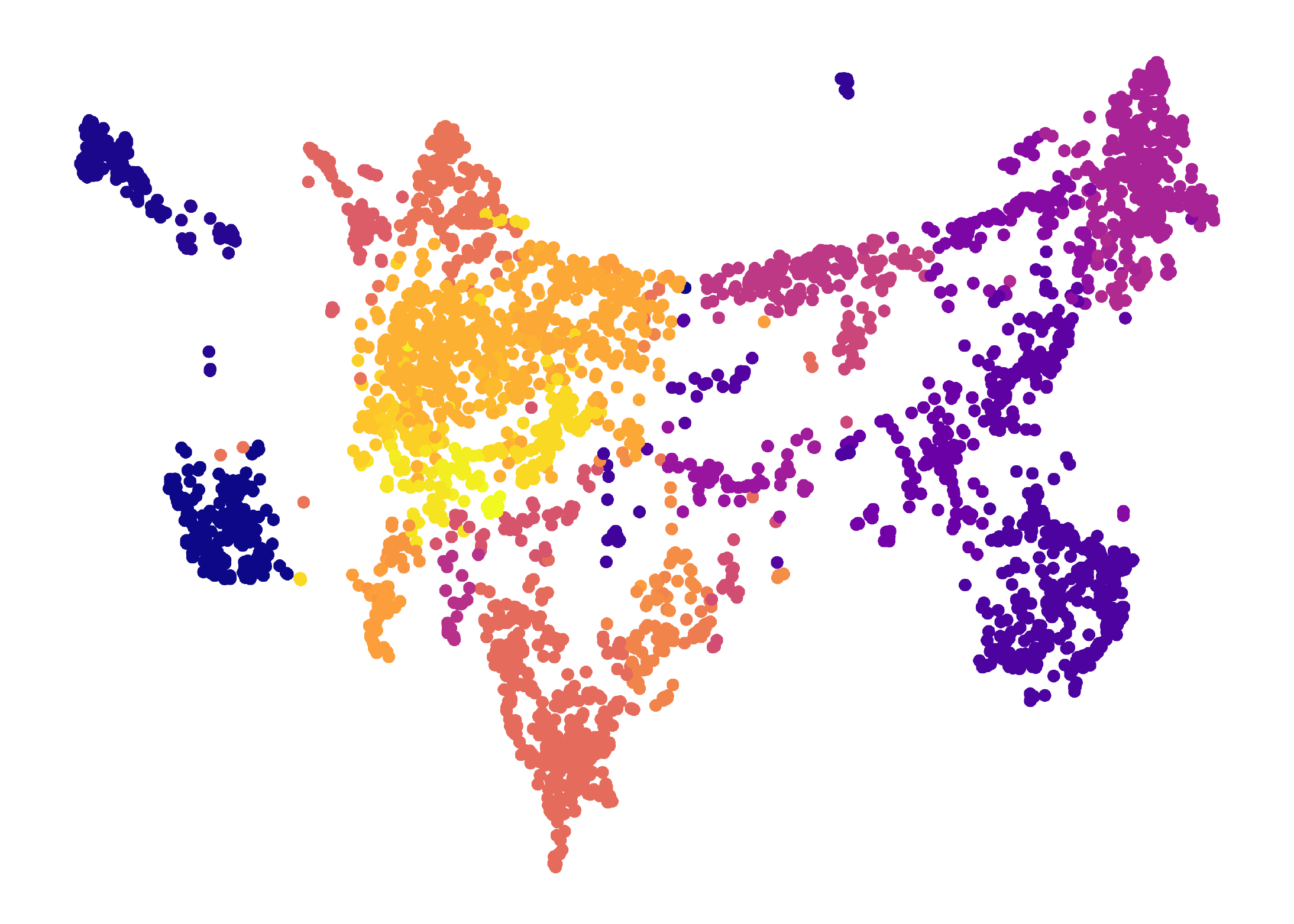}
        \caption{\klc{}, \texttt{combined}, $\kappa=10$, 41 Clusters}
    \end{subfigure} \hfill
    \begin{subfigure}[b]{0.3\textwidth}
        \includegraphics[width=\textwidth]{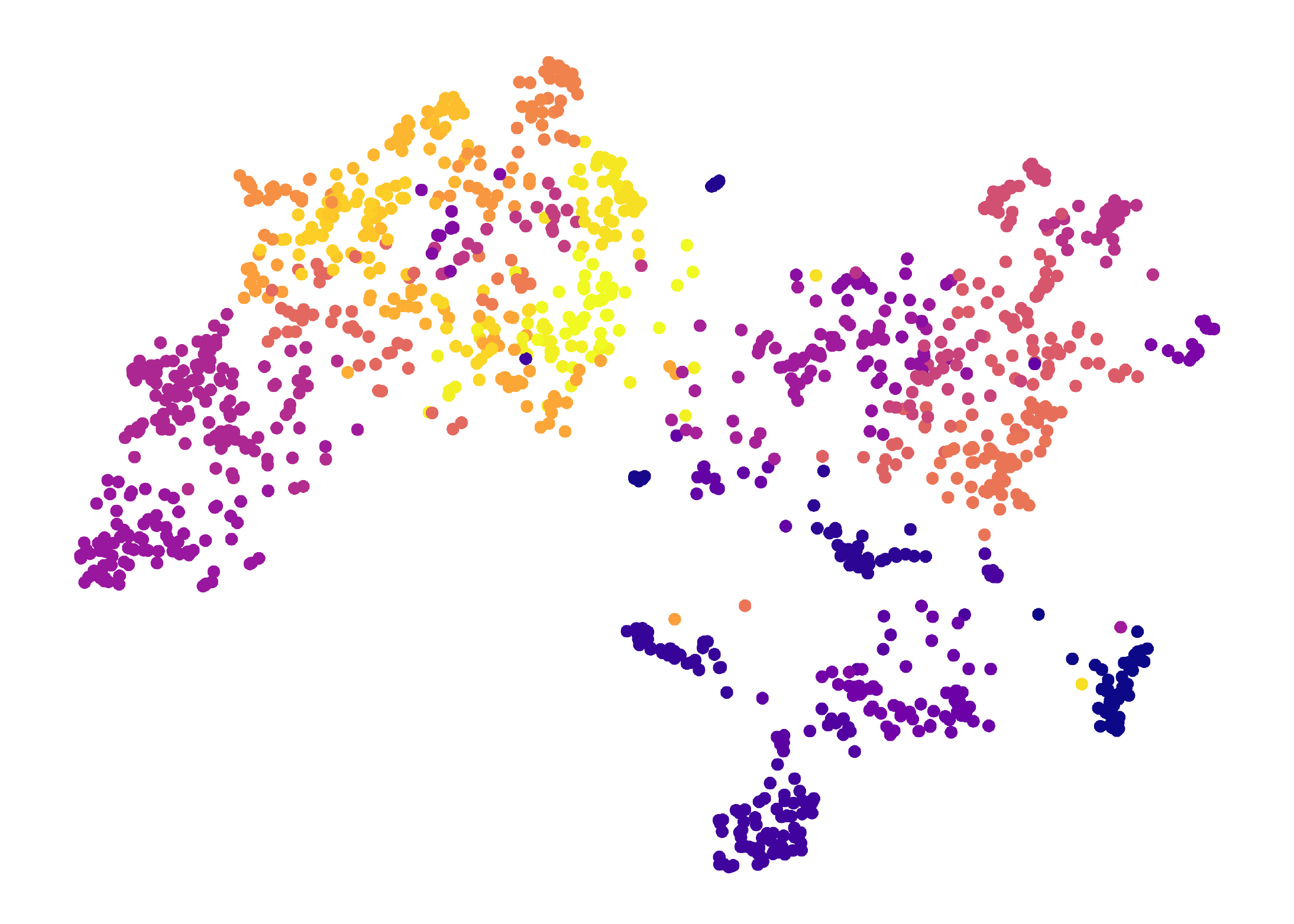}
        \caption{\wiki{}, \texttt{combined}, $\kappa=5$, 50 Clusters}
    \end{subfigure} \hfill
    \begin{subfigure}[b]{0.3\textwidth}
        \includegraphics[width=\textwidth]{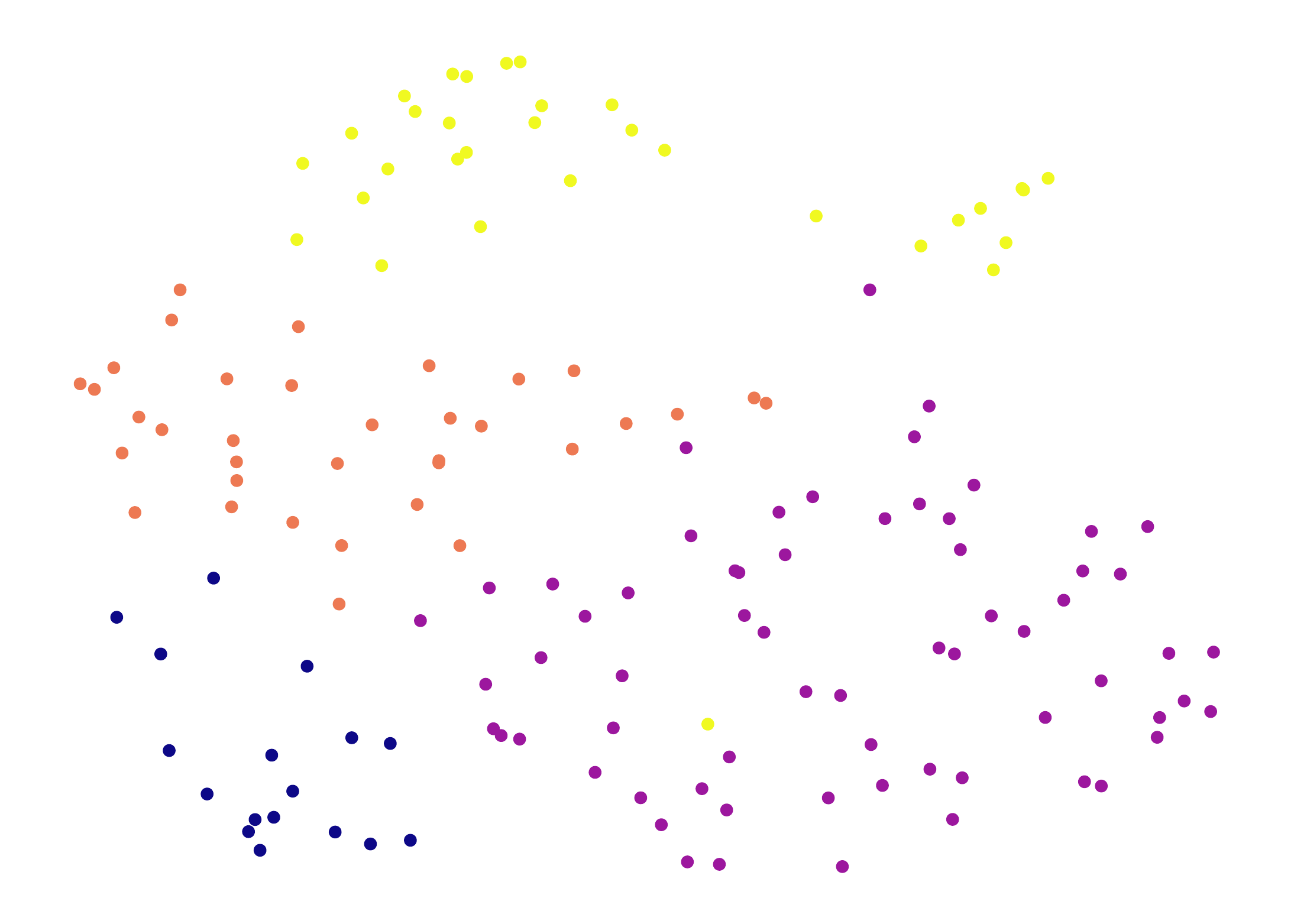}
        \caption{\funsd{}, \texttt{combined}, $\kappa=10$, 4 Clusters}
    \end{subfigure}
    \vspace{2mm}
    \begin{subfigure}[b]{0.3\textwidth}
        \includegraphics[width=\textwidth]{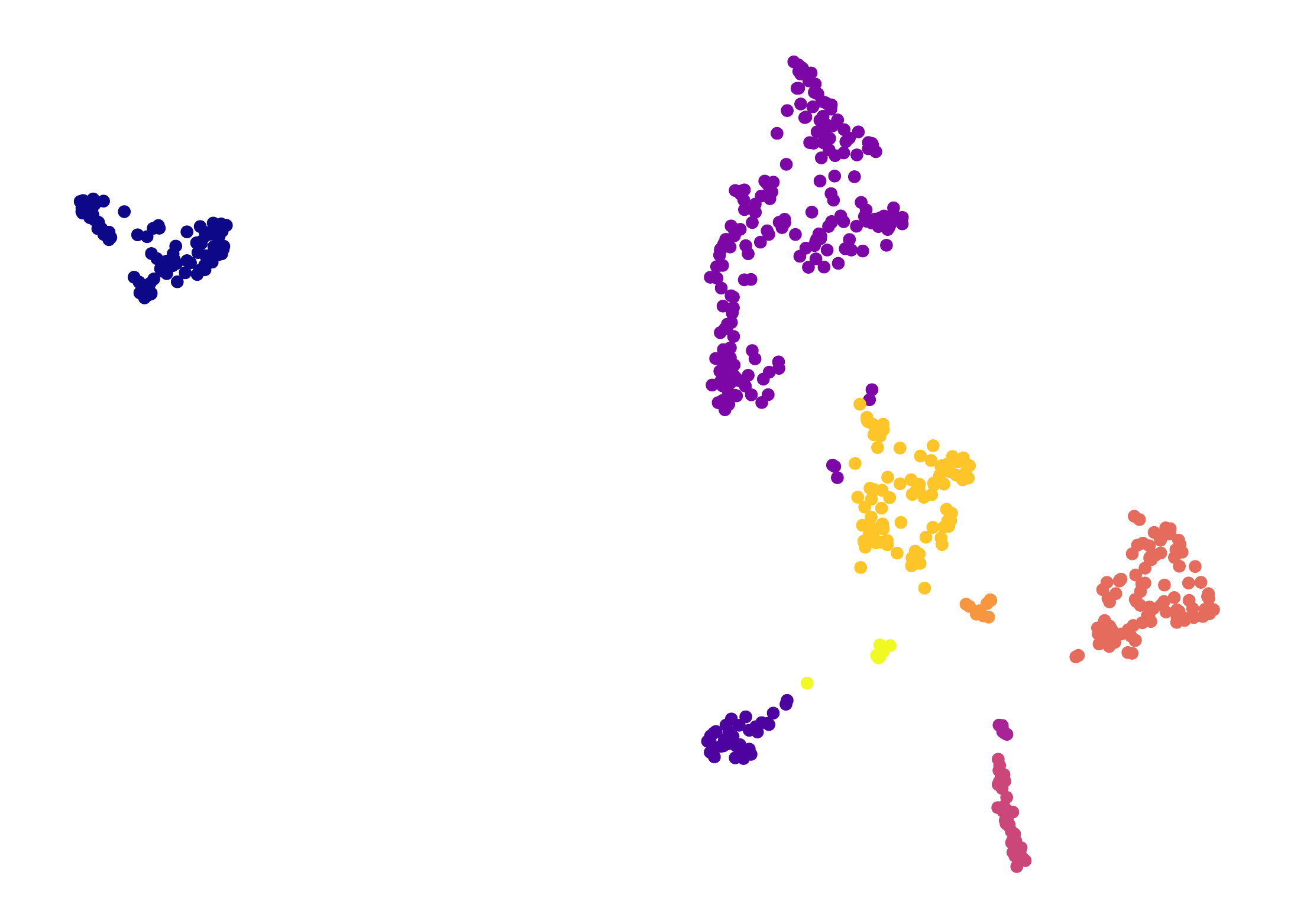}
        \caption{\icdarctdar{}, \texttt{clip}, $\kappa=5$, 9 Clusters}
    \end{subfigure} \hfill
    \begin{subfigure}[b]{0.3\textwidth}
        \includegraphics[width=\textwidth]{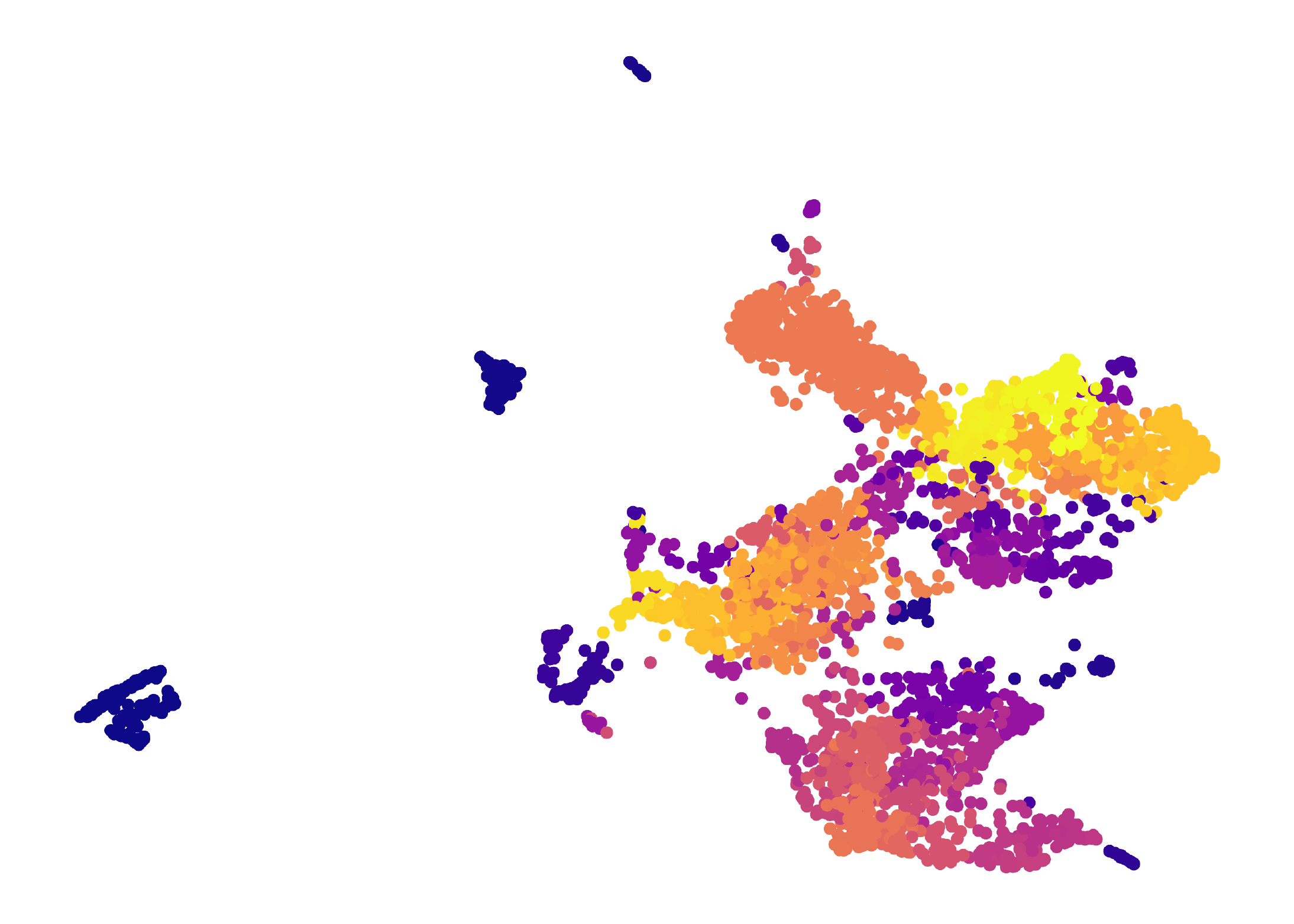}
        \caption{\publaynet, \texttt{clip}, $\kappa=5$, 106 Clusters}
    \end{subfigure} \hfill
    \begin{subfigure}[b]{0.3\textwidth}
        \includegraphics[width=\textwidth]{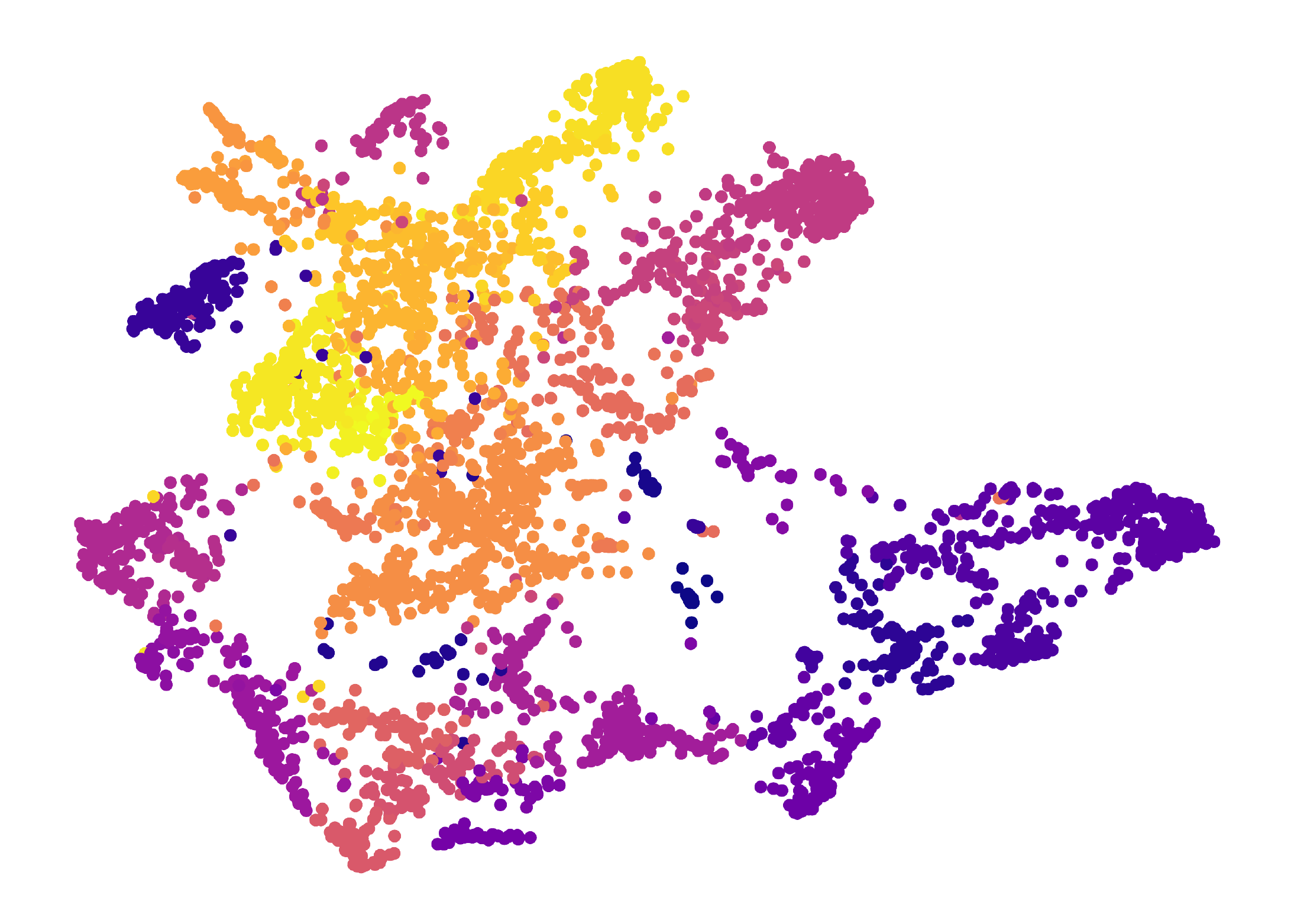}
        \caption{\rvlcdip{}, \texttt{combined}, $\kappa=10$, 49 Clusters}
    \end{subfigure}
    \vspace{2mm}
    \hfill
    \begin{subfigure}[b]{0.3\textwidth}
        \includegraphics[width=\textwidth]{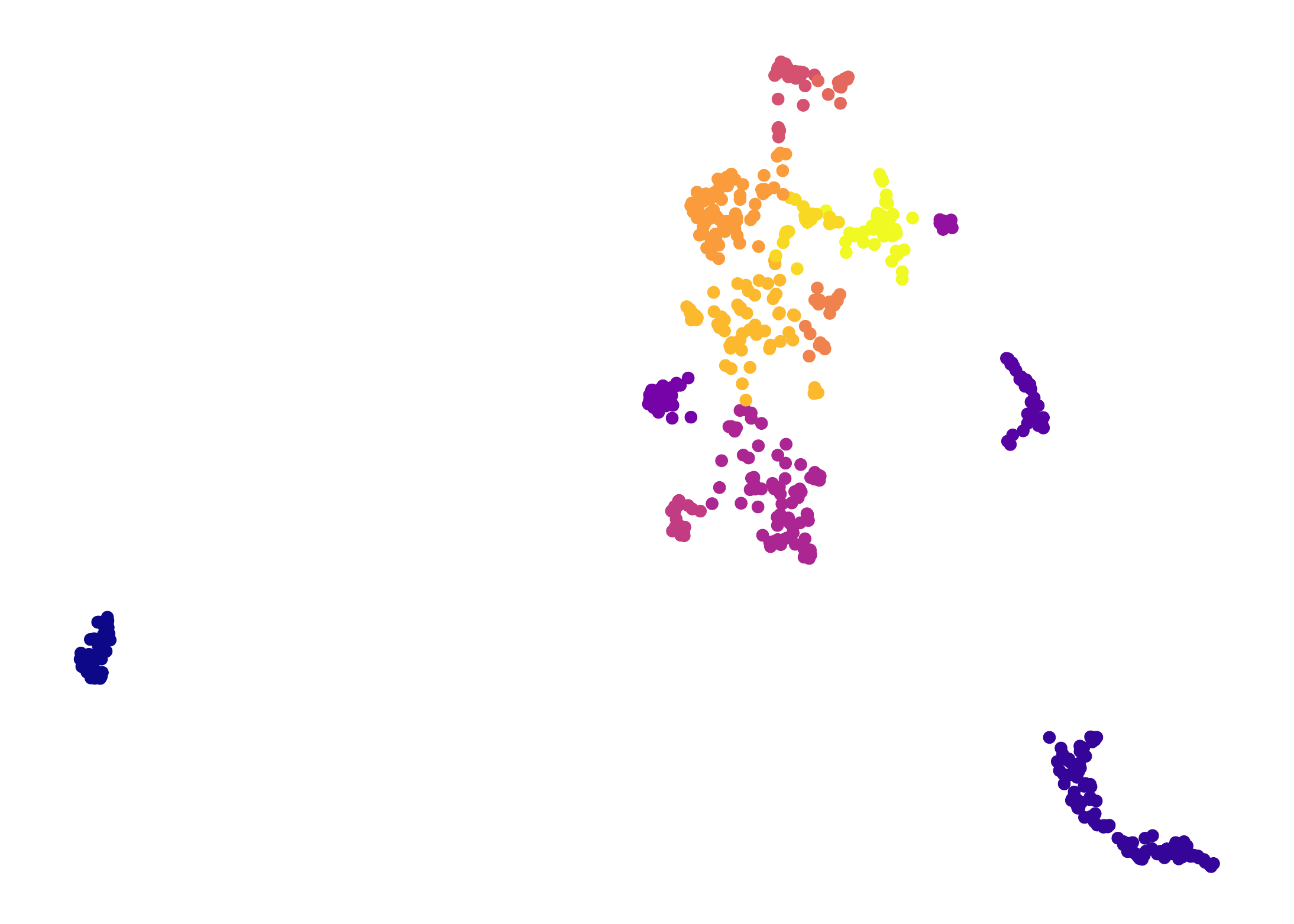}
        \caption{\sroie{}, \texttt{combined}, $\kappa=10$, 14 Clusters}
    \end{subfigure} \hfill
    \begin{subfigure}[b]{0.3\textwidth}
        \includegraphics[width=\textwidth]{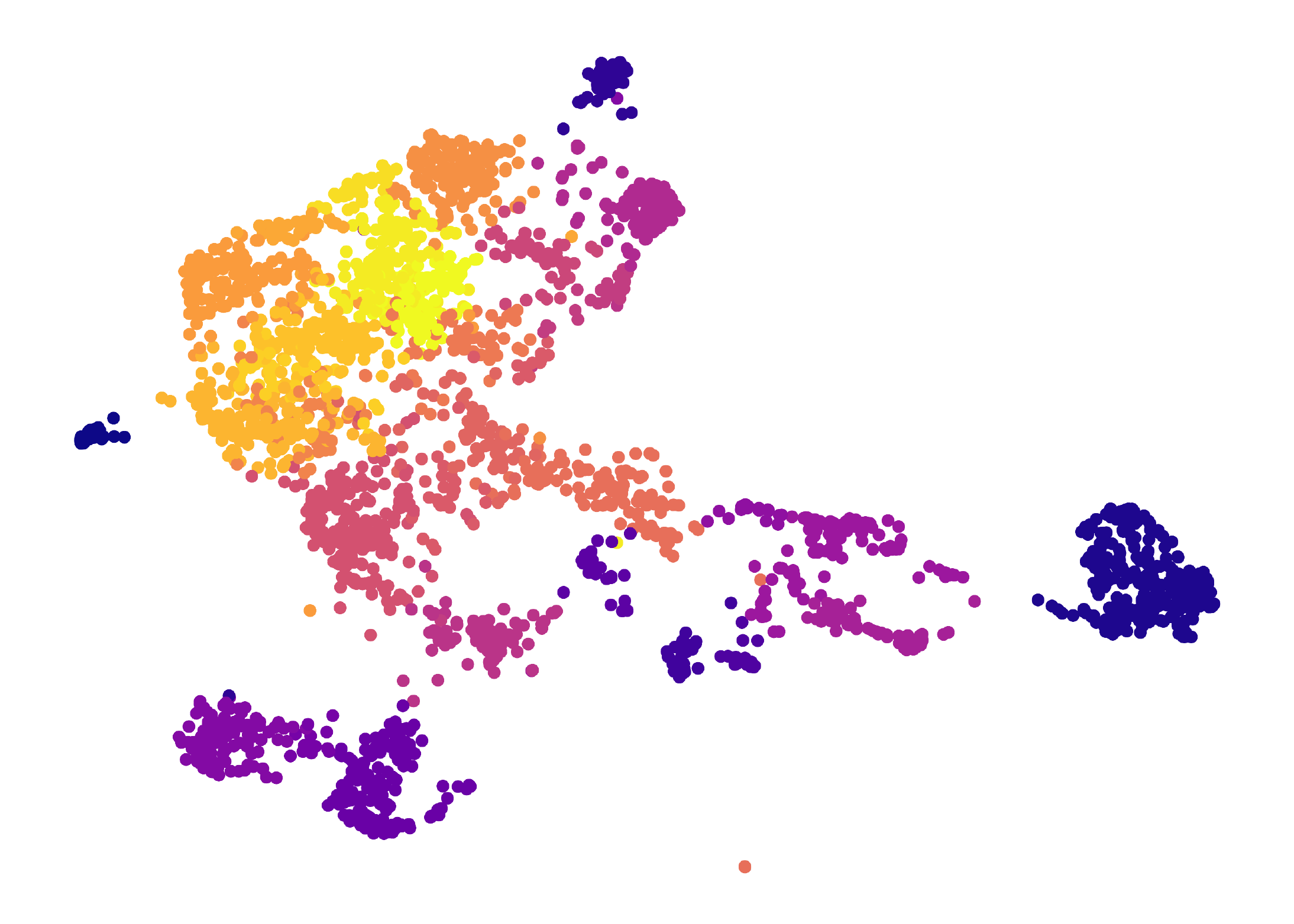}
        \caption{\tobacco{}, \texttt{combined}, $\kappa=10$, 31 Clusters}
    \end{subfigure}
    \hfill\null 
    \caption{Overview of our used clusters for all datasets. Each clustering lists embedding type, HDBSCAN~\cite{CMS13} minimum cluster size $\kappa$ and number of resulting clusters.}
    \label{fig:clusters_used}
\end{figure*}

\begin{figure*}[p] 
    \centering

    \begin{subfigure}[b]{0.45\textwidth}
        \centering
        \includegraphics[height=0.17\textheight]{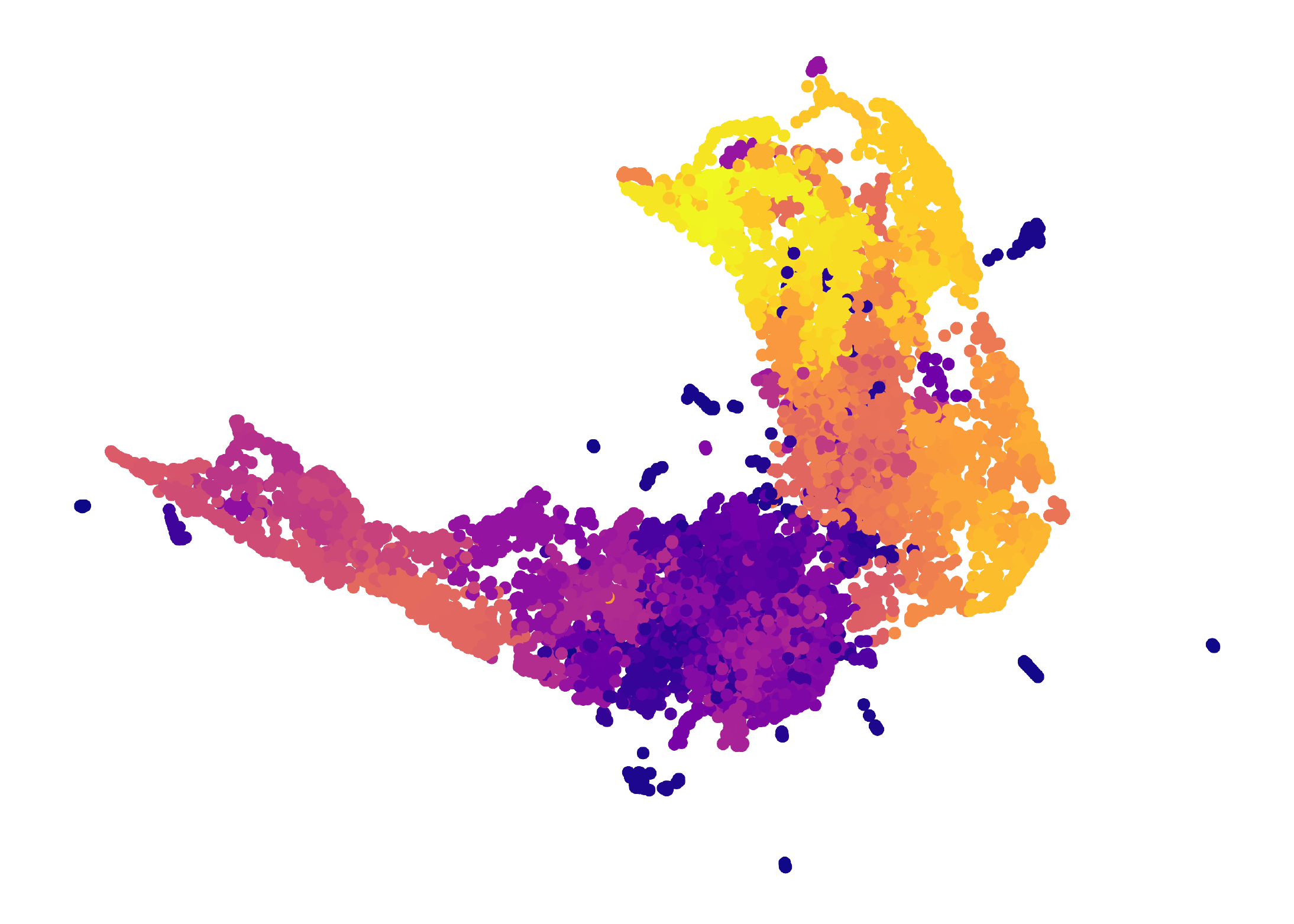}
        \caption{\texttt{layoutlm}, $\kappa=5$, 282 Clusters}
    \end{subfigure} \hfill
    \begin{subfigure}[b]{0.45\textwidth}
        \centering
        \includegraphics[height=0.17\textheight]{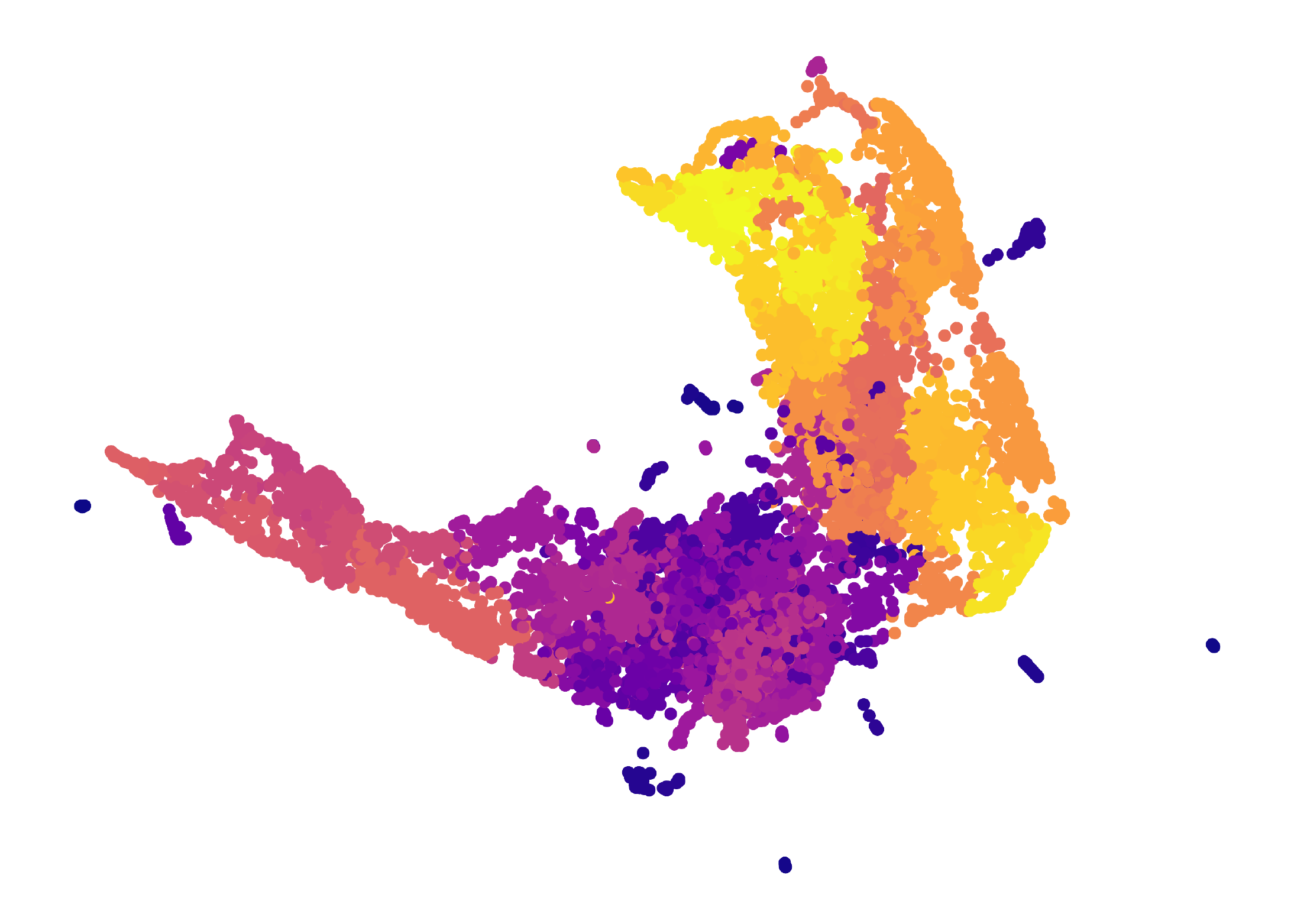}
        \caption{\texttt{layoutlm}, $\kappa=10$, 128 Clusters}
    \end{subfigure}

    \begin{subfigure}[b]{0.45\textwidth}
        \centering
        \includegraphics[height=0.17\textheight]{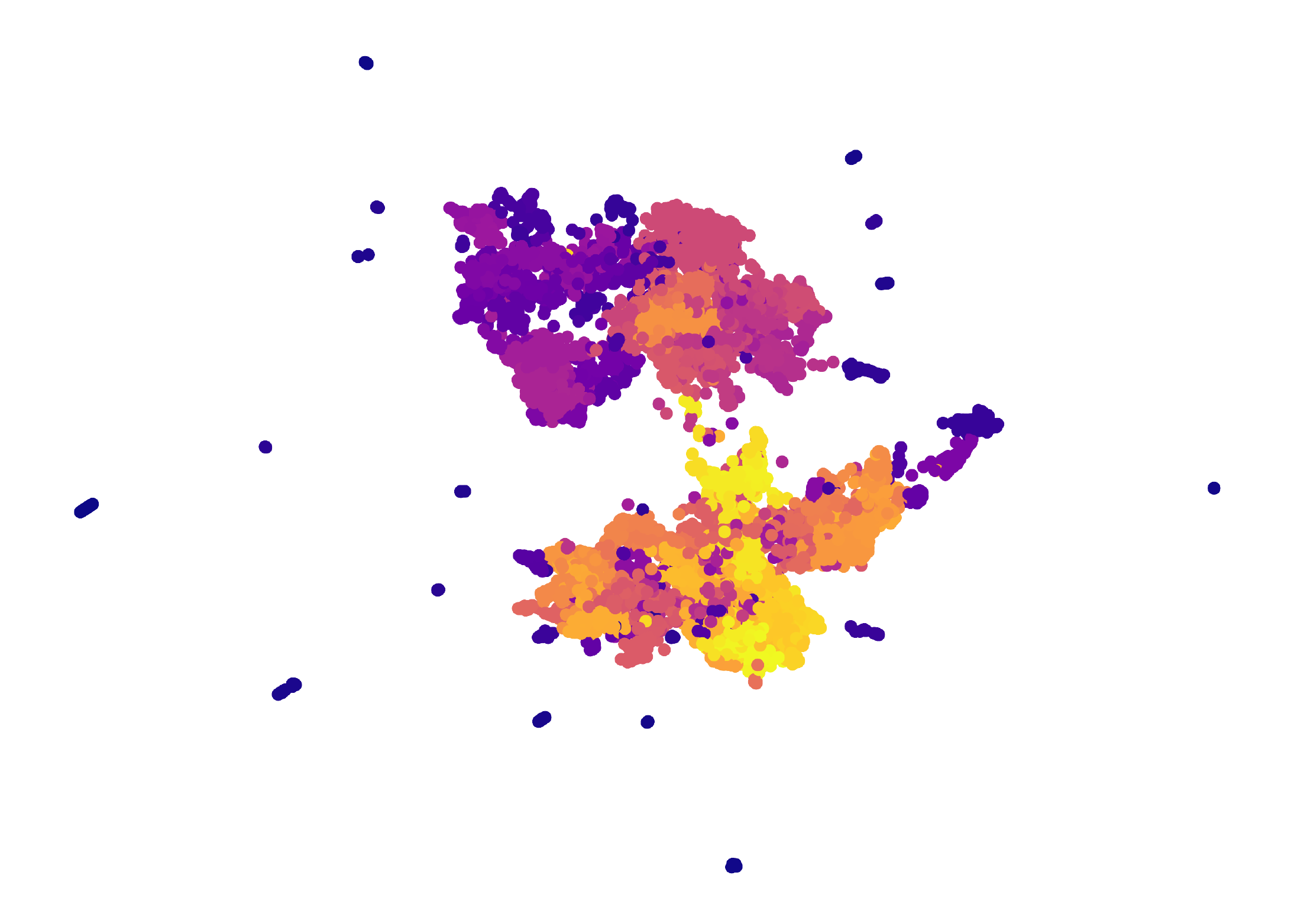}
        \caption{\texttt{clip}, $\kappa=5$, 259 Clusters}
    \end{subfigure} \hfill
    \begin{subfigure}[b]{0.45\textwidth}
        \centering
        \includegraphics[height=0.17\textheight]{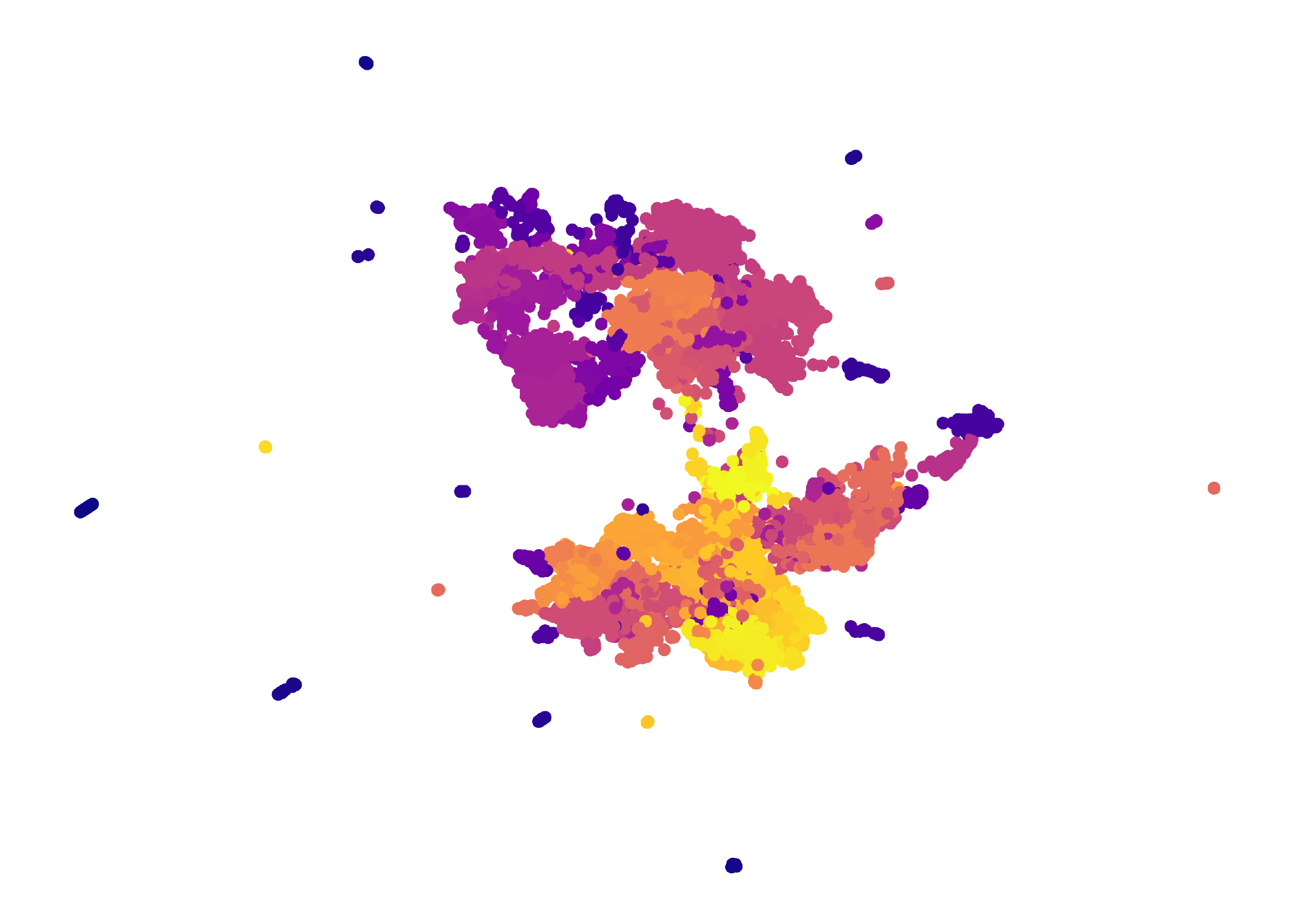}
        \caption{\texttt{clip}, $\kappa=10$, 123 Clusters}
    \end{subfigure}

    \begin{subfigure}[b]{0.45\textwidth}
        \centering
        \includegraphics[height=0.17\textheight]{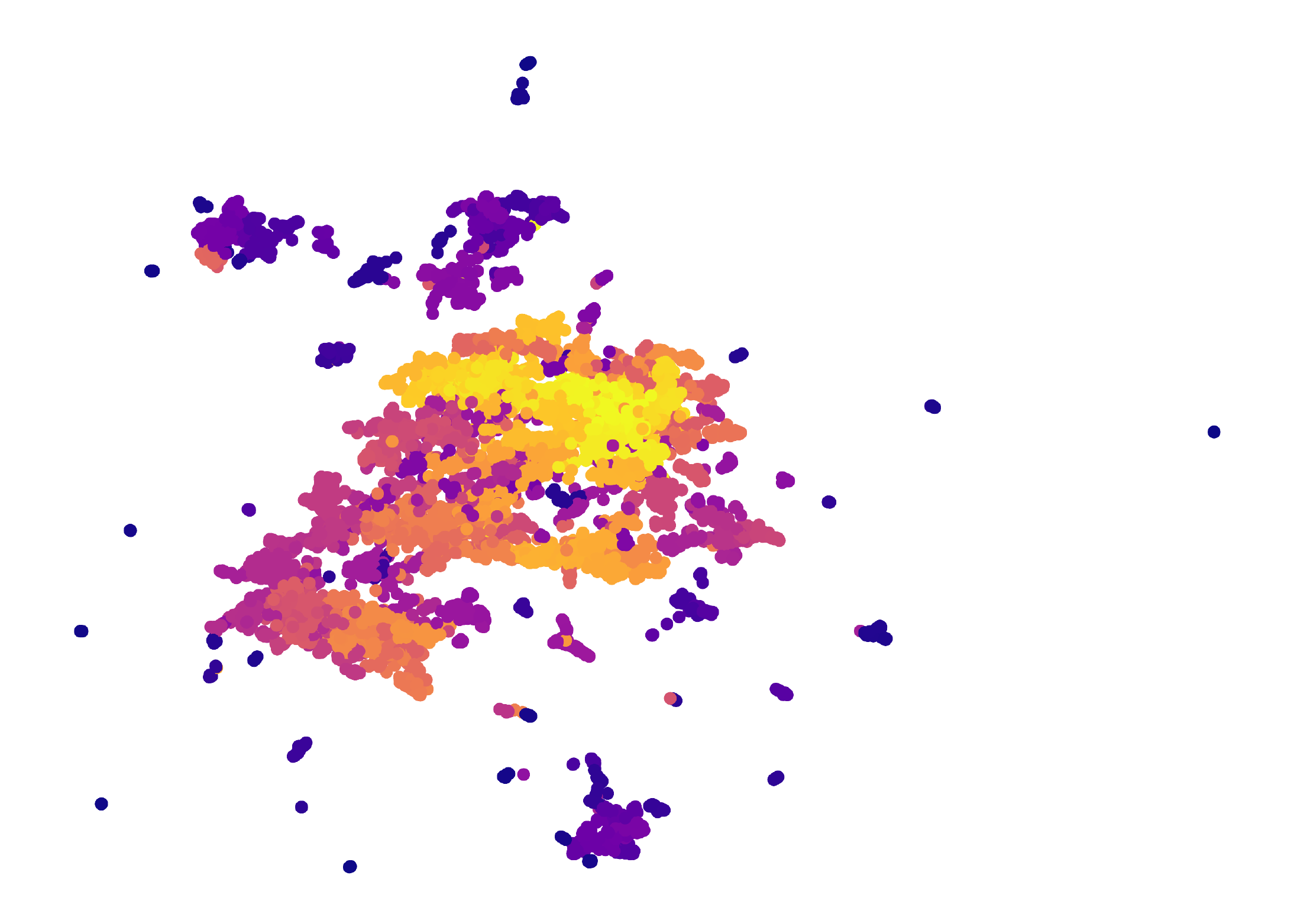}
        \caption{\texttt{sentence}, $\kappa=5$, 408 Clusters}
    \end{subfigure} \hfill
    \begin{subfigure}[b]{0.45\textwidth}
        \centering
        \includegraphics[height=0.17\textheight]{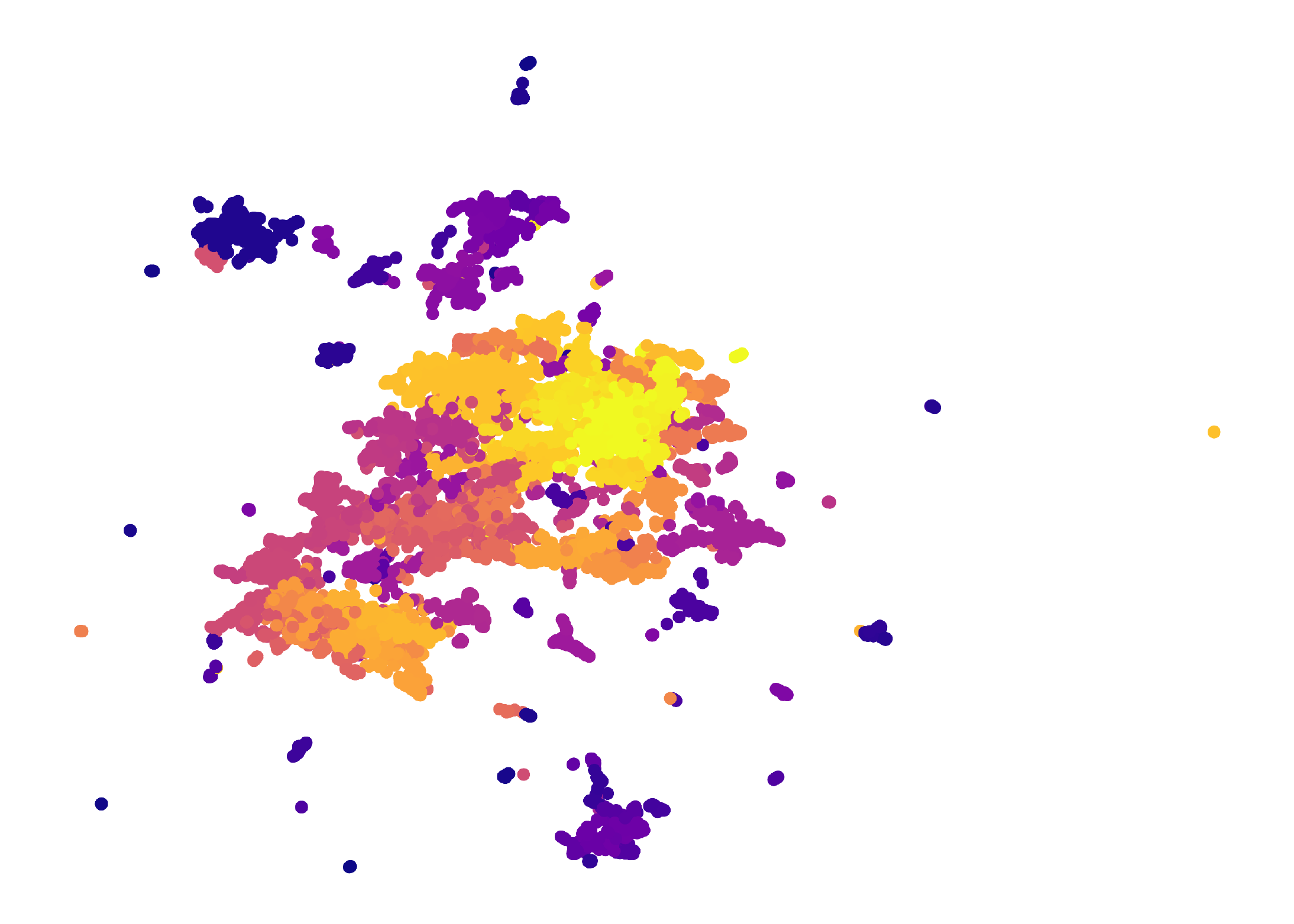}
        \caption{\texttt{sentence}, $\kappa=10$, 192 Clusters}
    \end{subfigure}

    \begin{subfigure}[b]{0.45\textwidth}
        \centering
        \includegraphics[height=0.17\textheight]{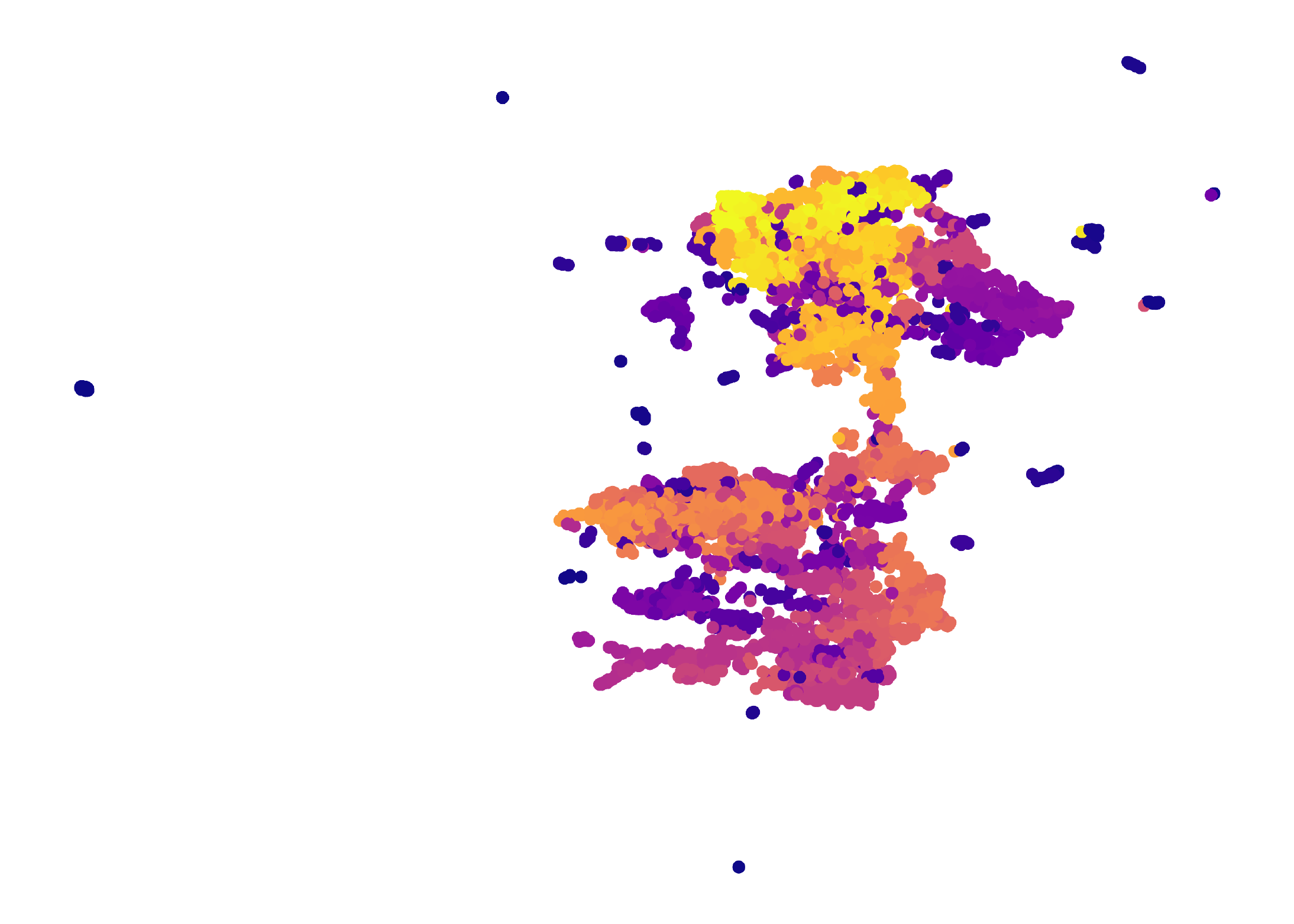}
        \caption{\texttt{combined}, $\kappa=5$, 362 Clusters}
    \end{subfigure} \hfill
    \begin{subfigure}[b]{0.45\textwidth}
        \centering
        \includegraphics[height=0.17\textheight]{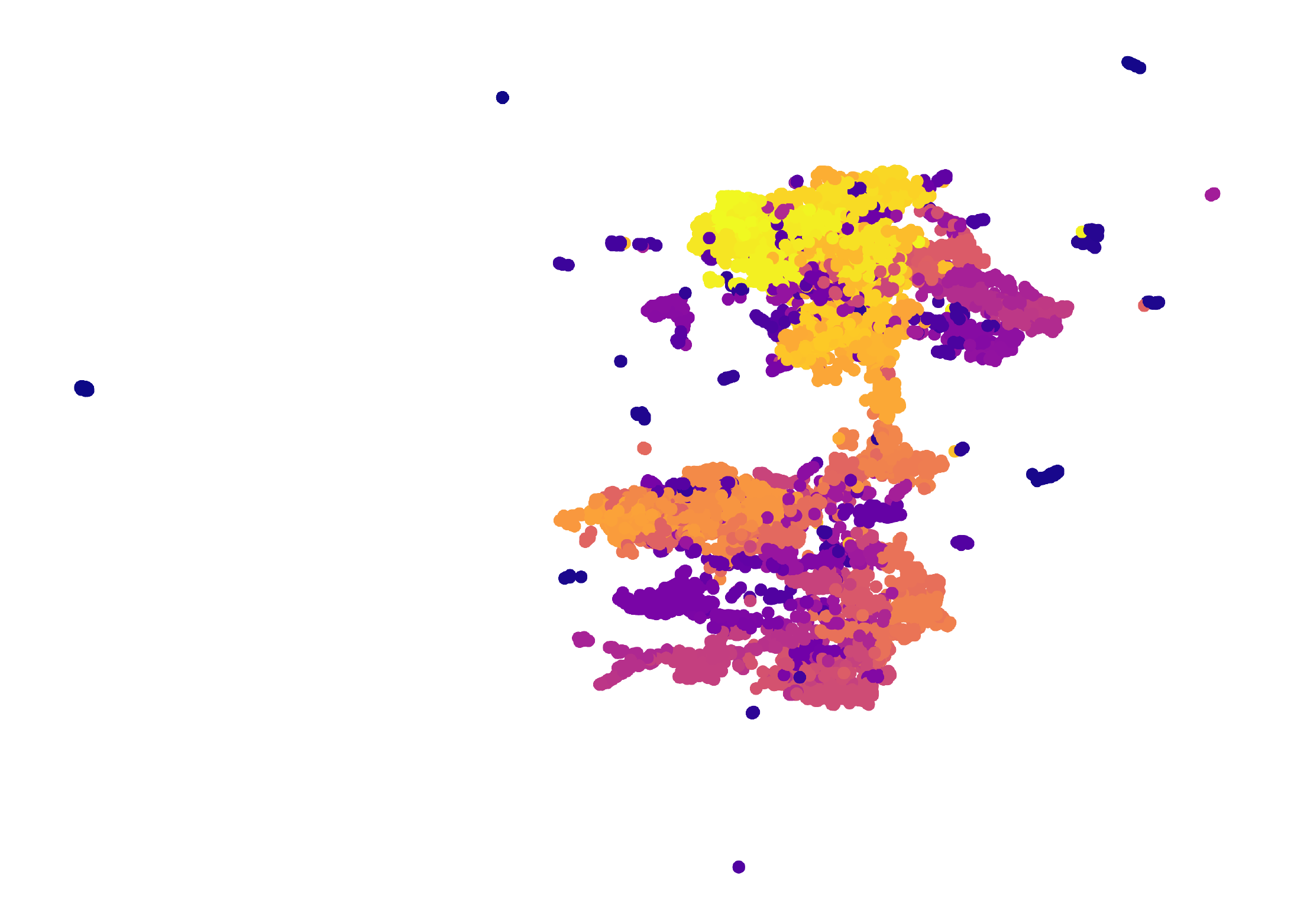}
        \caption{\texttt{combined}, $\kappa=10$, 187 Clusters}
    \end{subfigure}

    \begin{subfigure}[b]{0.45\textwidth}
        \centering
        \includegraphics[height=0.17\textheight]{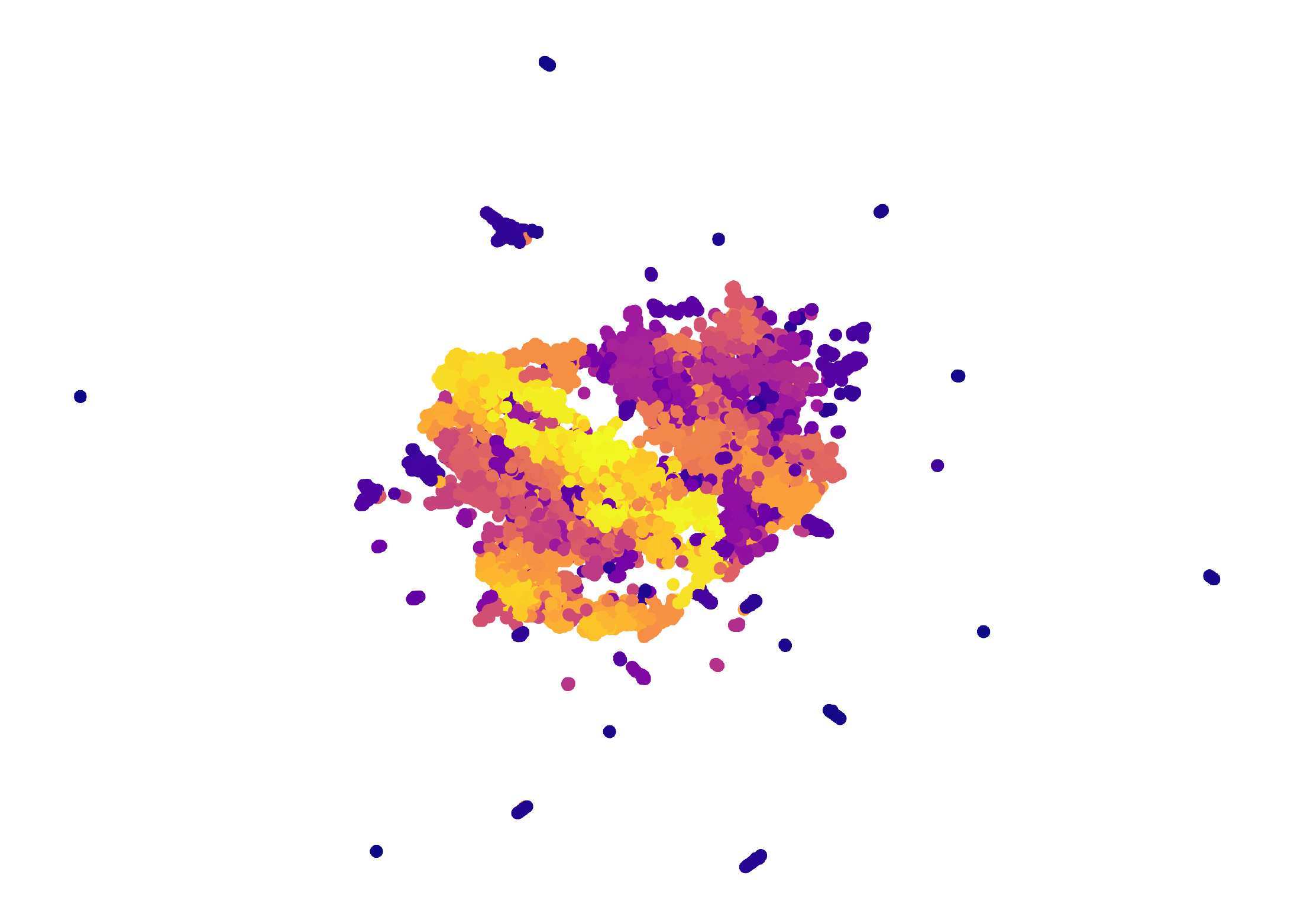}
        \caption{\texttt{pooled}, $\kappa=5$, 370 Clusters}
    \end{subfigure} \hfill
    \begin{subfigure}[b]{0.45\textwidth}
        \centering
        \includegraphics[height=0.17\textheight]{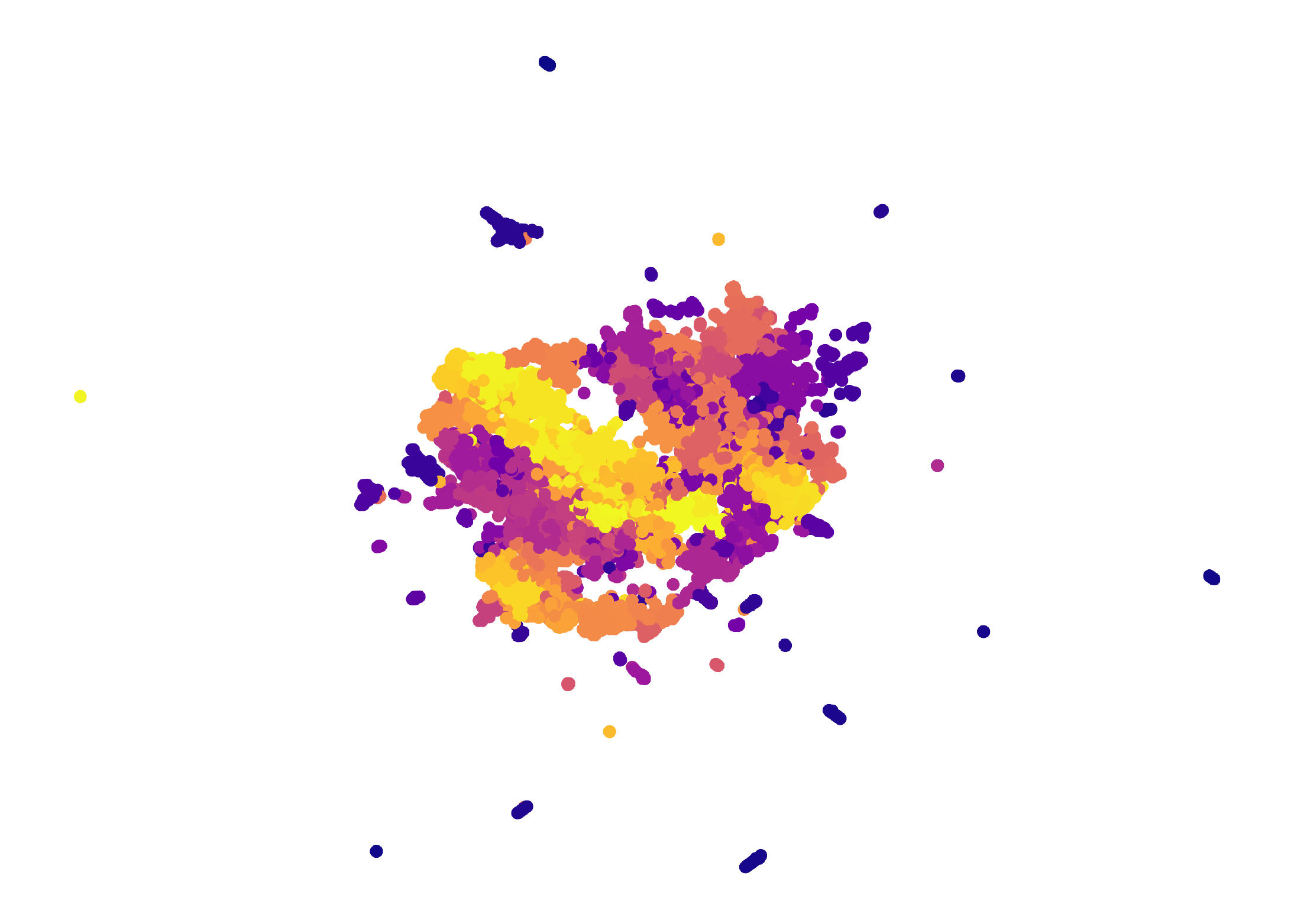}
        \caption{\texttt{pooled}, $\kappa=10$, 181 Clusters}
    \end{subfigure}

    \caption{Clustering results across different embeddings and HDBSCAN~\cite{CMS13} minimum cluster sizes $\kappa$ for \docvqa{}.}
    \label{fig:clusters_all}
\end{figure*}

Configurations $(\mathcal{E}, \kappa)$ are ranked using cumulative position scores: on each dataset, 
the top $N$ configurations receive points from $N$ down to 1 based on their 
composite metric ranking. Final rankings aggregate these scores across all 
datasets as $R(\mathcal{E}, \kappa) = \sum_{d} r_d(\mathcal{E}, \kappa)$ and are listed in \cref{tab:clustering_rankings}.
Based on these rankings and manual inspection we select a clustering configuration $(\mathcal{E}^*, \kappa^*)$ for each dataset.
Metrics for the selected configurations and metrics for all configurations on \docvqa{} are shown in \cref{tab:cluster_metrics}, with the corresponding clusters visualized in \cref{fig:clusters_used,fig:clusters_all}.



\section{Implementation Details for Handwriting Synthesis}
\label{app:implementation_handwriting}
\noindent\textbf{Dataset Preparation.}
All experiments were conducted using the IAM handwriting dataset. Each word image was center-padded to a fixed spatial size of $128\times512$ pixels without resizing to ensure consistent scale across all samples. This dimension covers over $95\%$ of IAM words and aligns with the $8\times$ spatial reduction of the VAE encoder, yielding latent tensors of size $[4,16,64]$. Each image was encoded to the latent space using the pretrained \texttt{stabilityai/\allowbreak sd\hbox{-}vae\hbox{-}ft\hbox{-}mse} autoencoder with a scaling factor of $0.18215$. The LMDB dataset stored per-sample latent, grayscale image, writer ID, and text transcription.

\noindent\textbf{Model Architecture.}
The baseline model is a conditional latent diffusion model trained on VAE-encoded handwriting latents. The denoising network is a conditional UNet with cross-attention layers and residual blocks, conditioned jointly on text and writer identity. The text conditioning network is a transformer encoder with hidden dimension $d=512$, $L=6$ layers, $H=8$ attention heads, feedforward width $d_{ff}=2048$, and dropout rate $0.1$. The UNet operates on $4\times16\times64$ latent inputs and includes class embeddings for writer conditioning. The diffusion scheduler follows the DDPM formulation with 1000 timesteps and a linear $\beta$ schedule from $\beta_{start}=1\times10^{-4}$ to $\beta_{end}=0.02$.

\noindent\textbf{Training Hyperparameters.}
The model was trained using AdamW optimizer with learning rate $1\times10^{-4}$, $\beta_1=0.9$, $\beta_2=0.999$, and weight decay $0.01$. Gradient clipping was set to $1.0$. A cosine learning rate schedule was employed across $200$ epochs. Mixed-precision training used \texttt{fp16} with automatic gradient scaling. EMA of model weights was applied with decay $0.9999$ and power $1.0$. The batch size per GPU was $16$, with gradient accumulation for an effective batch size of $64$. Random seed was fixed to $42$. No image augmentations were applied to preserve text legibility. 

\noindent\textbf{Inference and Generation.}
At inference time, handwriting was generated from text tokens and corresponding writer embeddings. Generation used $30$ diffusion steps with a DPMSolver++ multistep scheduler (order 3) and a temperature of $0.5$. The VAE decoder scaled latents by $1/0.18215$ before decoding to image space. To ensure consistent scale and aspect ratio, all generations were performed directly at the $128\times512$ resolution without any resizing. For variable-length text, words longer than six characters were internally divided into balanced subsegments before generation, as the IAM corpus has an average word length of approximately six characters. Each subsegment was decoded separately and horizontally concatenated after generation.

\noindent\textbf{Scaling and Alignment.}
Two issues were explicitly addressed. \emph{(1) Scaling:} generation scale was fixed to the canonical $128\times512$ resolution to prevent variation in stroke thickness and character proportion. \emph{(2) Alignment:} baseline alignment was used for horizontal stitching. The baseline position of each segment was estimated from the bottom $50$th percentile of the ink mask, and subsegments were vertically aligned by matching these baselines before compositing. Refer to algorithm\ref{alg:baseline} for baseline calculation. 

\begin{algorithm}[t]
\caption{Percentile Baseline Estimation}
\label{alg:baseline}
\begin{algorithmic}[1]
\Require RGBA segment image $\tilde{I}$; opacity threshold $\tau$; fixed percentile $p=50$
\Ensure Robust baseline $b^\star$
\State Extract alpha channel $A$; let $\mathcal{C}$ be columns with any pixel $A > \tau$
\For{each column $j \in \mathcal{C}$}
  \State $b_j \leftarrow$ lowest row index with $A(r,j) > \tau$
\EndFor
\State $b^\star \leftarrow \operatorname{percentile}\big(\{b_j\}_{j \in \mathcal{C}},\, p\big)$
\State \Return $b^\star$
\end{algorithmic}
\end{algorithm}

\noindent\textbf{Post-Processing.}
To remove discretization artifacts and simulate pen spread, a Gaussian blur was applied with radius $r\sim\mathcal{U}(0.35,0.85)$. An anti-aliasing pass was optionally performed using a downscale–upscale factor of $0.75$. Additional postprocessing parameters included contrast multiplier $1.02$, ink gamma $0.98$, additive Gaussian noise $\sigma=0.35$ (pixel intensity units), and unsharp mask parameters $(r,p,t)=(0.5,30,2)$. The blurred outputs were composited with the alpha channel preserved to maintain soft ink boundaries. 

\noindent\textbf{Summary.}
The overall pipeline consists of: dataset padding and latent encoding $\rightarrow$ conditional diffusion training $\rightarrow$ text-conditioned inference with sub-word segmentation $\rightarrow$ baseline alignment $\rightarrow$ Gaussian and anti-aliasing refinement. This design ensures uniform spatial scale, stable conditioning, and visually realistic handwriting suitable for integration into synthetic printed documents.

\begin{figure*}[t]
  \centering
  \begin{subfigure}{0.95\linewidth}
    \centering
    \includegraphics[width=\linewidth]{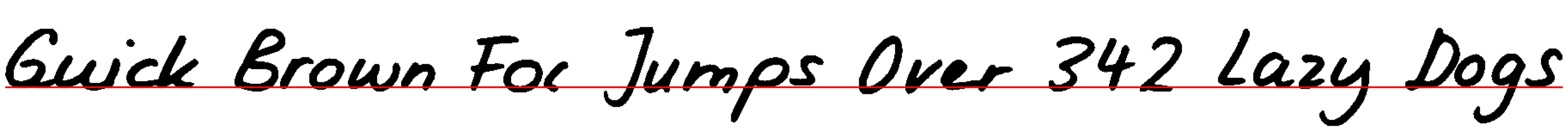}
    \caption{Sentence-level synthesis with a common baseline (red) produced by our diffusion model.}
    \label{fig:hw-sentence}
  \end{subfigure}
  \vspace{0.5em}

  \begin{subfigure}{0.95\linewidth}
    \centering
    \includegraphics[width=\linewidth]{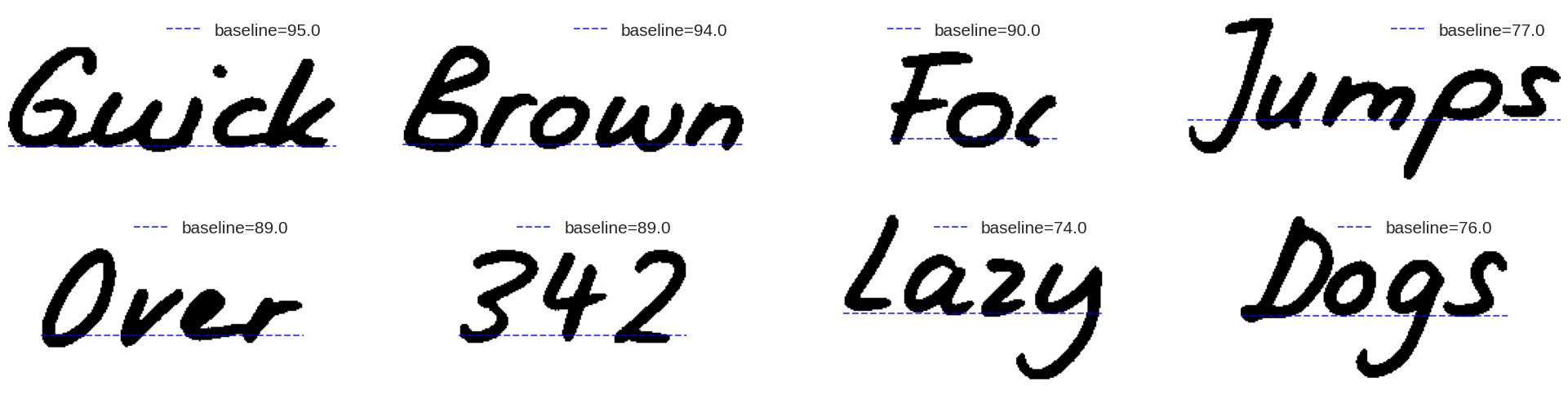}
    \caption{Word-level segments with estimated baselines (blue dashed lines) used to align the final sentence.}
    \label{fig:hw-words}
  \end{subfigure}

  \caption{\textbf{Diffusion-based handwriting generation and baseline alignment.}
  (a) The model synthesizes the full sentence “Quick Brown Fox Jumps Over 342 Lazy Dogs” with a coherent global baseline.
  (b) For each word segment, we estimate a robust baseline which is then used to compose the globally aligned line.}
  \label{fig:hw-baseline-alignment}
\end{figure*}

\begin{figure*}[t]
  \centering

  \includegraphics[width=0.90\linewidth]{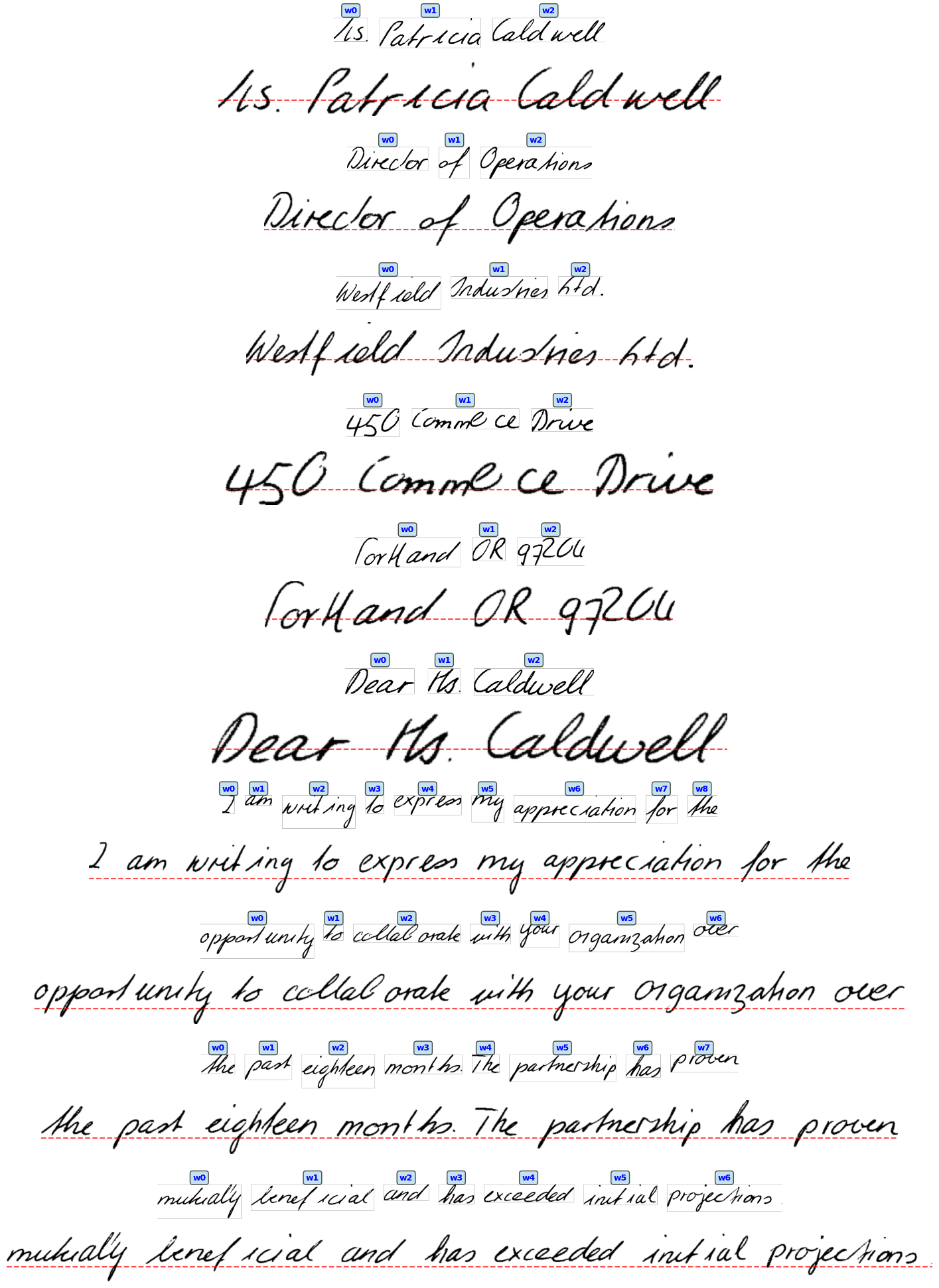}

  \caption{\textbf{Baseline-aligned handwriting synthesis.}
  The figure shows multiple handwriting segments generated by our latent diffusion model. Individually synthesized words, each with an automatically estimated baseline and composition of these word segments into globally baseline-aligned sentences, demonstrating consistent geometric structure across words and lines.}
  \label{fig:hw-baseline-examples}
\end{figure*}

\section{Implementation Details for Visual Elements}\label{app:visual_elements}
\paragraph{Visual Element Rendering.}
Each visual element type requires specialized rendering: \texttt{stamp} elements use custom text-based generators; \texttt{barcode} elements encode numeric content (or random values if non-numeric) using the python-barcode library; \texttt{logo}, \texttt{figure}, and \texttt{photo} elements sample from image banks generated with Gemini 2.5 Flash~\cite{CBSP25}, with \texttt{photo} additionally incorporating synthetic faces from StyleGAN2~\cite{KLAH20} via ThisPersonDoesNotExist.com.

To generate the image banks, we prompt Gemini 2.5 Flash with the following instructions for each element type:
\begin{itemize}[leftmargin=*, itemsep=2pt]
    \item \texttt{figure}: \textit{``Create an arbitrary scientific figure without any visible text and any additional requests.''}
    \item \texttt{logo}: \textit{``Create an arbitrary, abstract logo without any visible text and any additional requests.''}
    \item \texttt{photo}: \textit{``Create an arbitrary photo without any visible text and any additional requests.''}
\end{itemize}

\begin{figure*}[htbp]
    \newlength{\vefigwidth}
    \setlength{\vefigwidth}{0.19\textwidth}
    \centering
    \begin{subfigure}{\vefigwidth}
        \includegraphics[width=\linewidth]{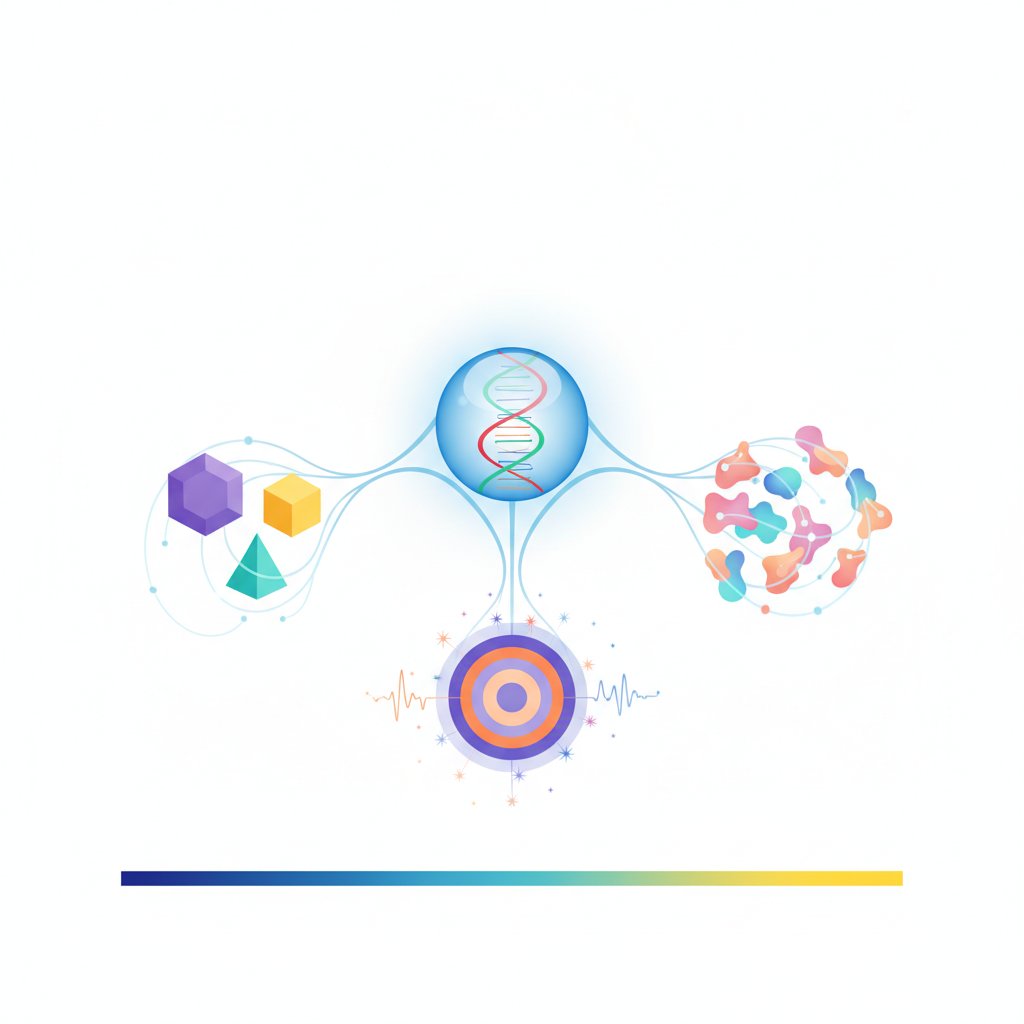}
    \end{subfigure}%
    \hfill
    \begin{subfigure}{\vefigwidth}
        \includegraphics[width=\linewidth]{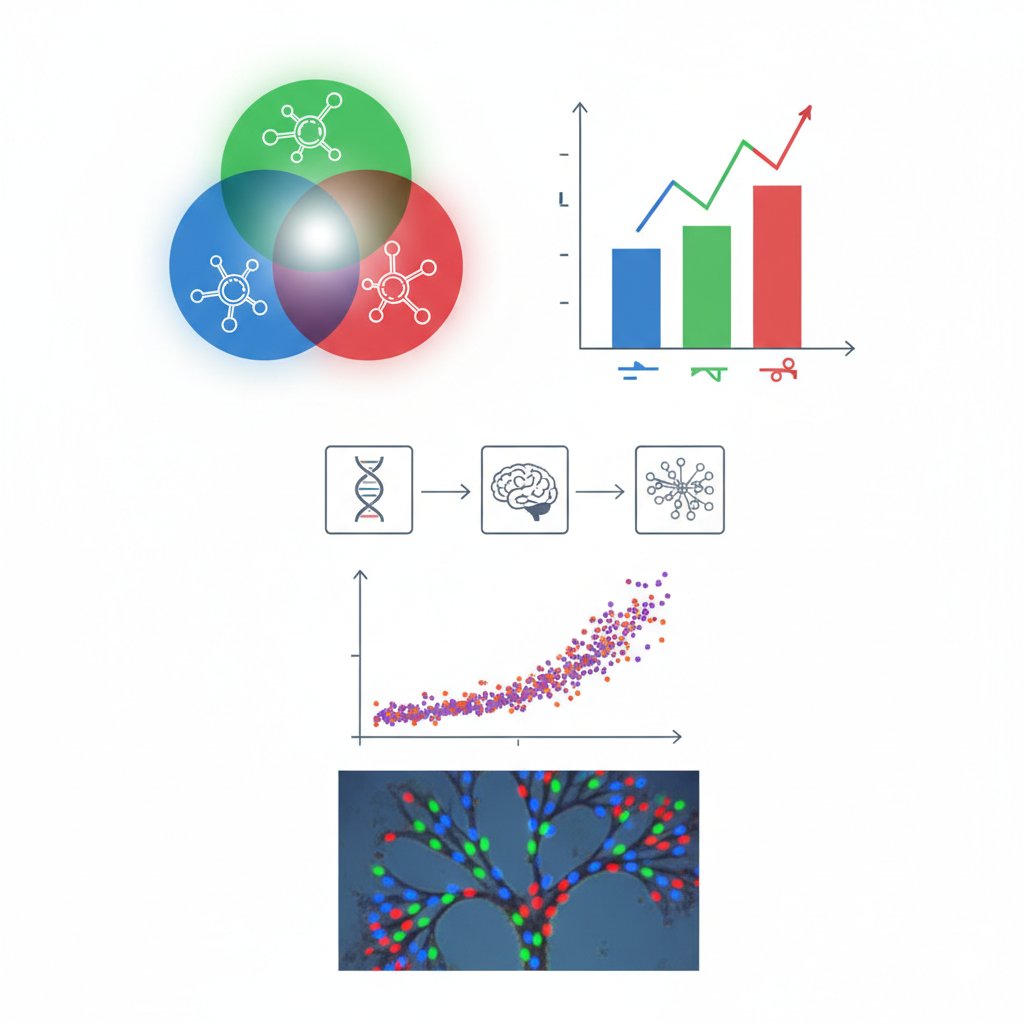}
    \end{subfigure}%
    \hfill
    \begin{subfigure}{\vefigwidth}
        \includegraphics[width=\linewidth]{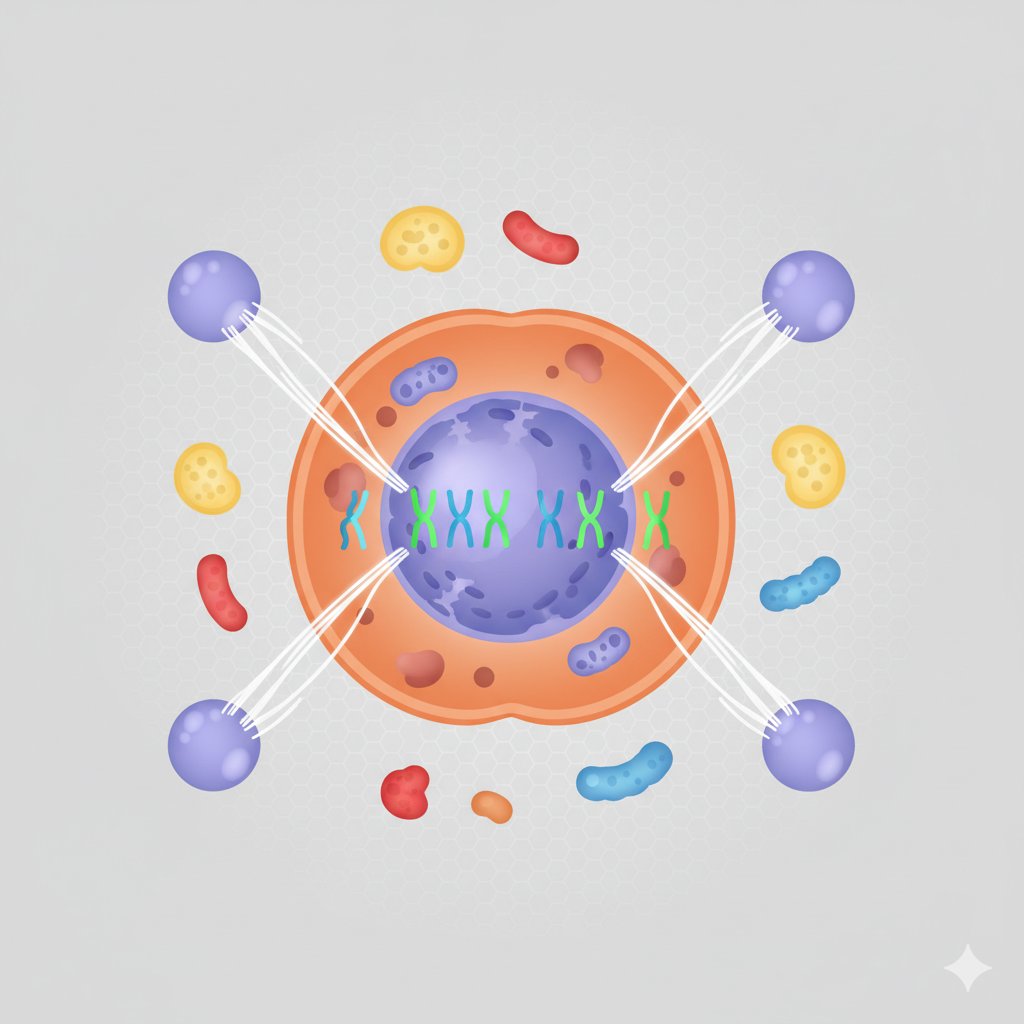}
    \end{subfigure}%
    \hfill
    \begin{subfigure}{\vefigwidth}
        \includegraphics[width=\linewidth]{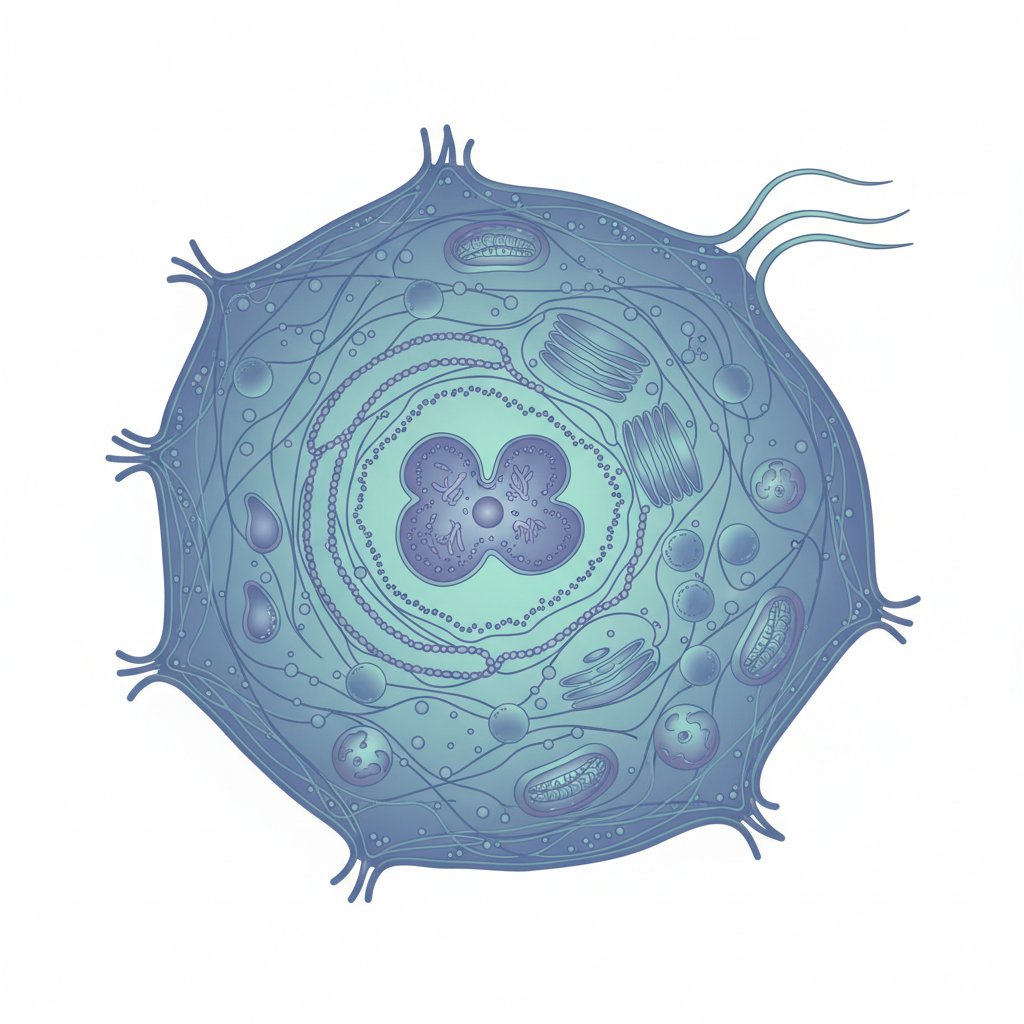}
    \end{subfigure}%
    \hfill
    \begin{subfigure}{\vefigwidth}
        \includegraphics[width=\linewidth]{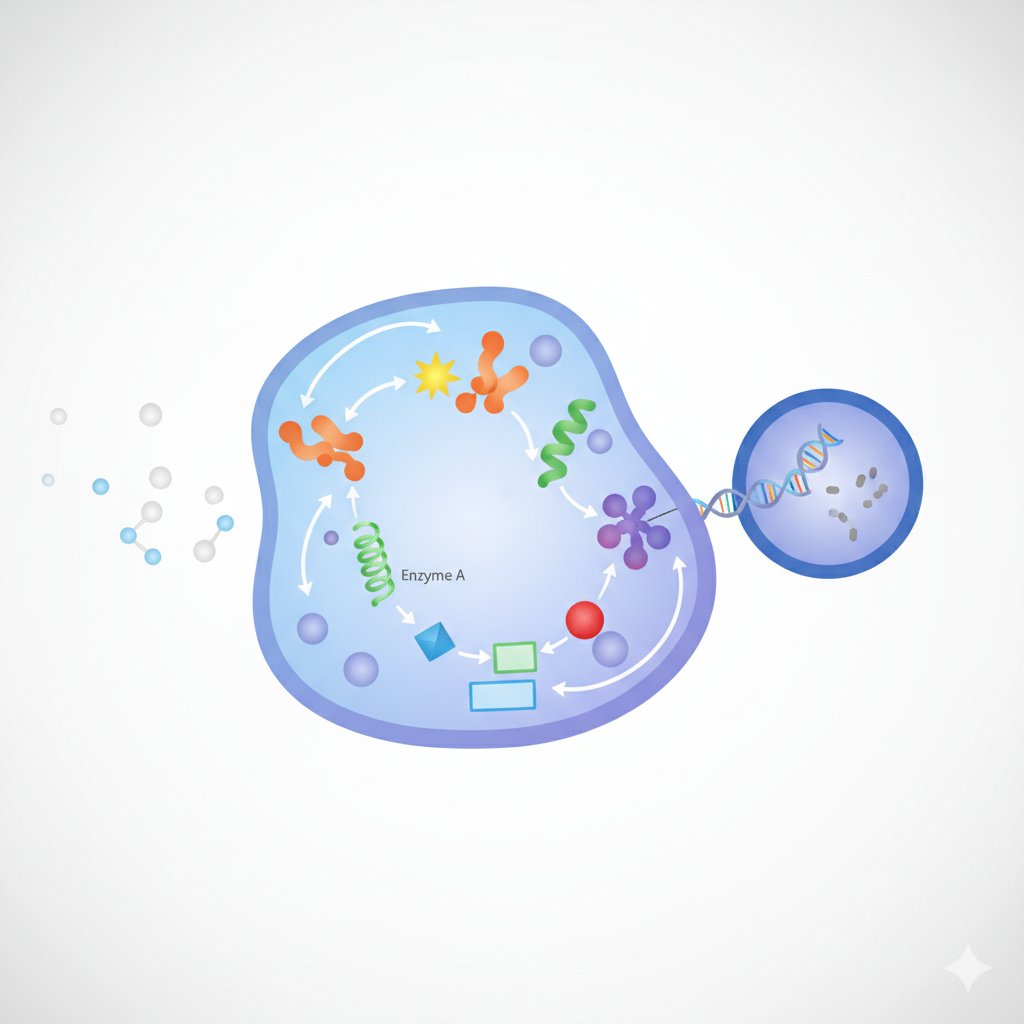}
    \end{subfigure}%
    \caption{Image bank for visual elements of type \texttt{figure}, generated with Gemini 2.5 Flash~\cite{CBSP25}.}\label{app:fig_ve_figure}
\end{figure*}

\begin{figure*}[htbp]
    \setlength{\vefigwidth}{0.15\textwidth}
    \centering
    \begin{subfigure}{\vefigwidth}
        \includegraphics[width=\linewidth]{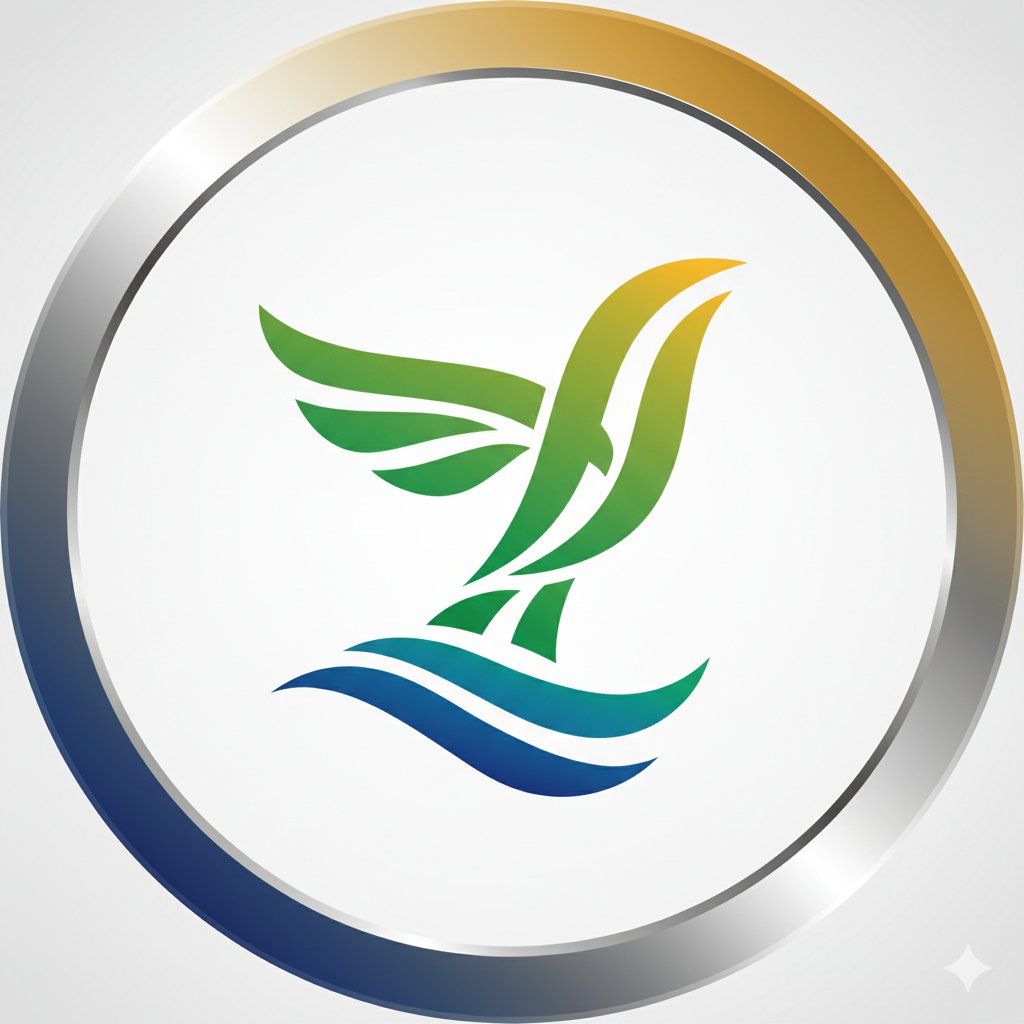}
    \end{subfigure}%
    \hfill
    \begin{subfigure}{\vefigwidth}
        \includegraphics[width=\linewidth]{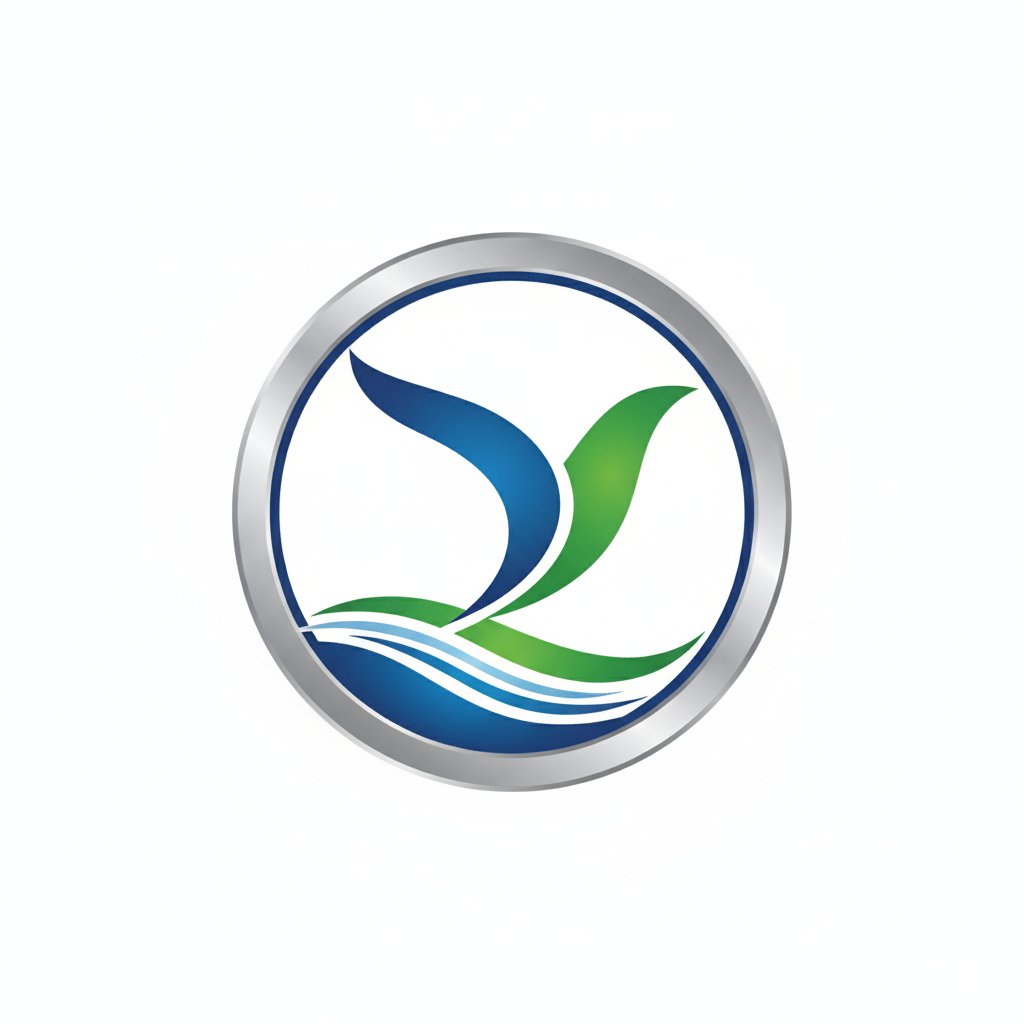}
    \end{subfigure}%
    \hfill
    \begin{subfigure}{\vefigwidth}
        \includegraphics[width=\linewidth]{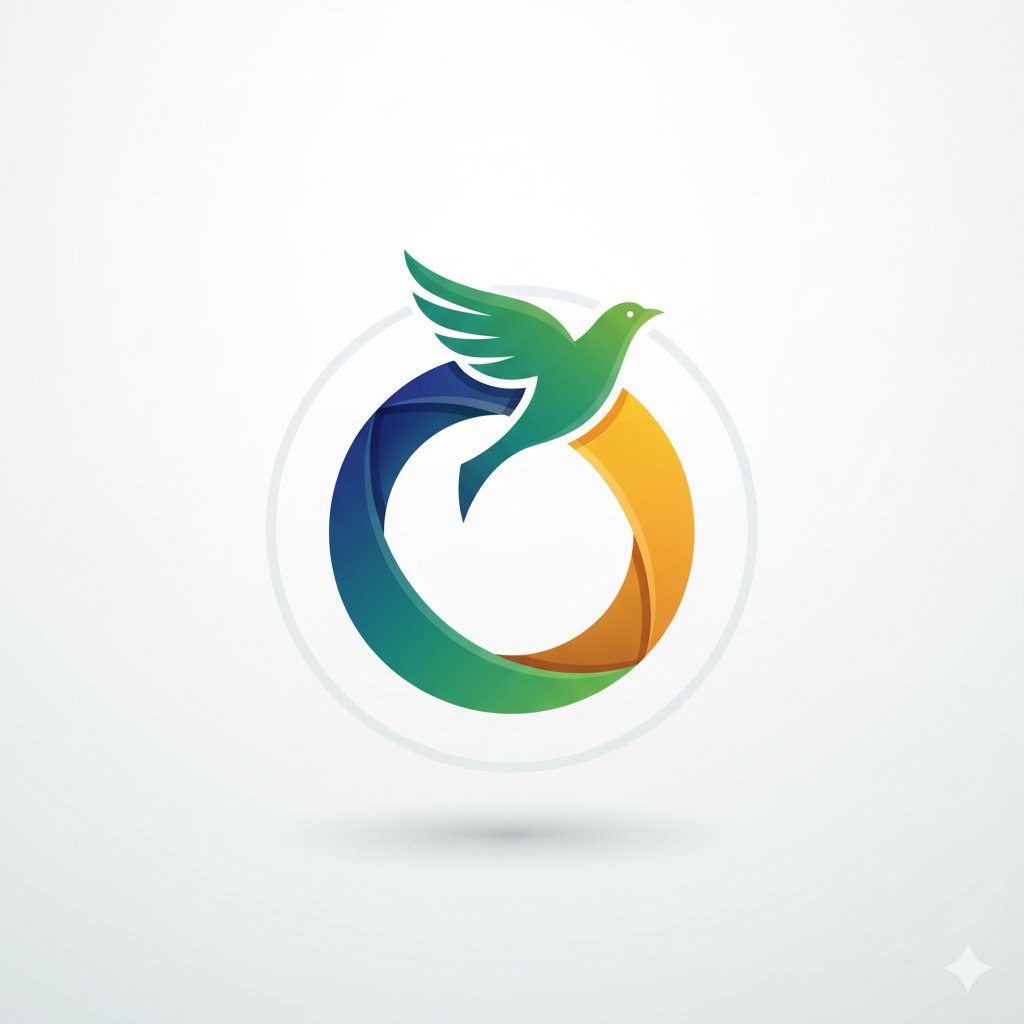}
    \end{subfigure}%
    \hfill
    \begin{subfigure}{\vefigwidth}
        \includegraphics[width=\linewidth]{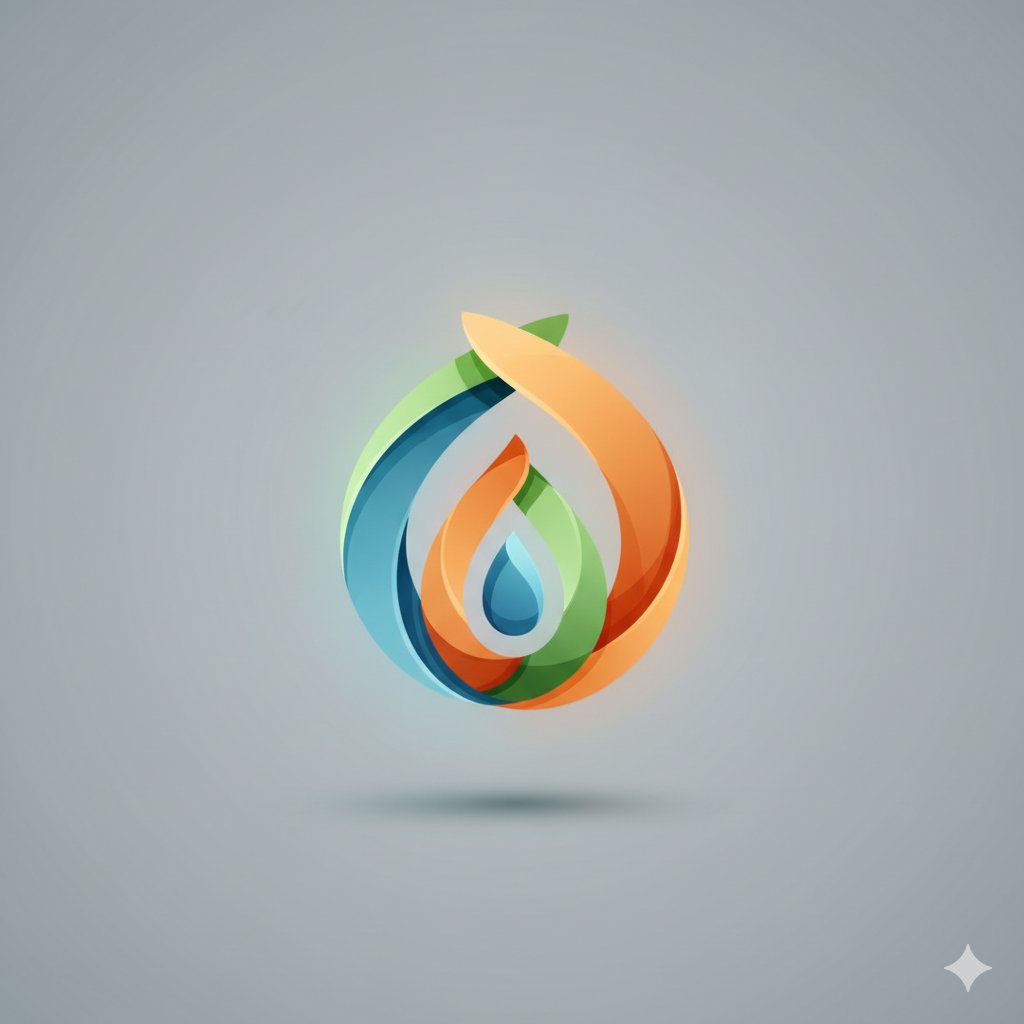}
    \end{subfigure}%
    \hfill
    \begin{subfigure}{\vefigwidth}
        \includegraphics[width=\linewidth]{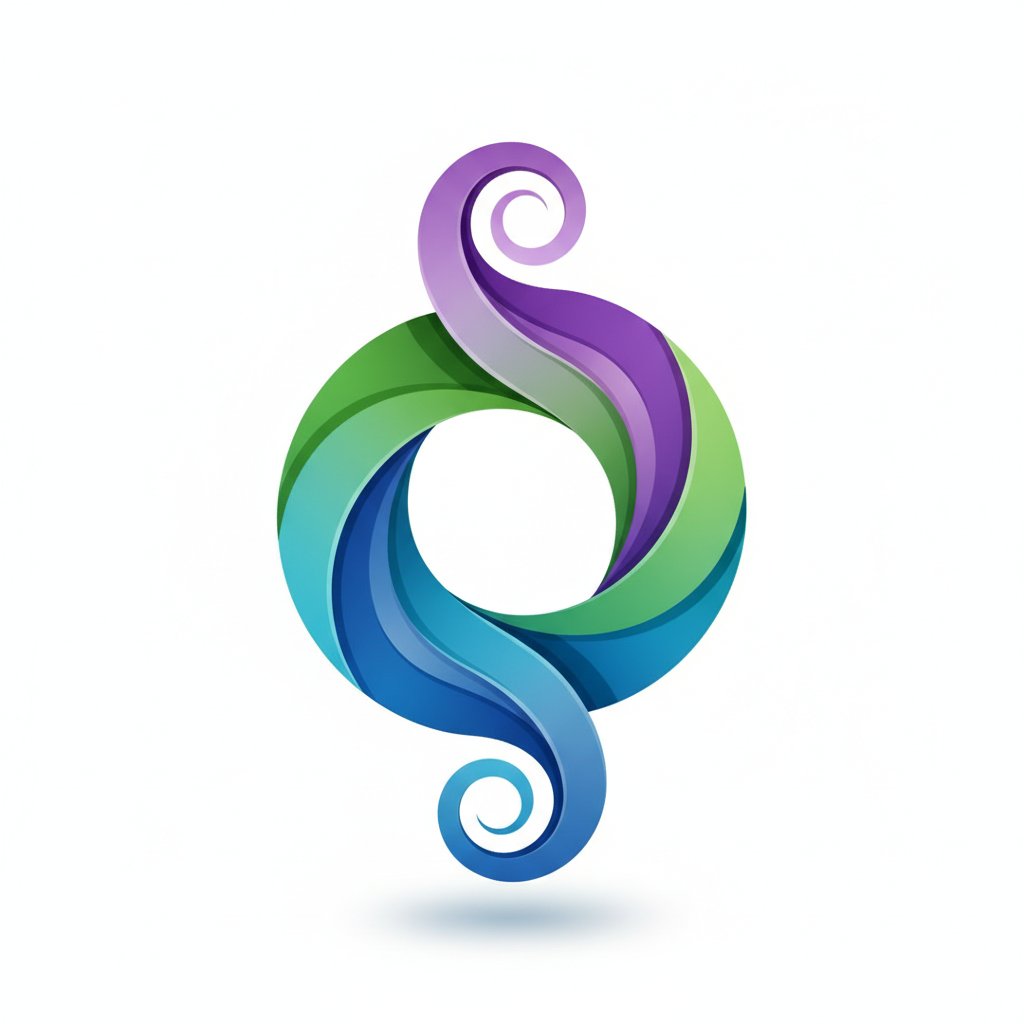}
    \end{subfigure}%
        \hfill
    \begin{subfigure}{\vefigwidth}
        \includegraphics[width=\linewidth]{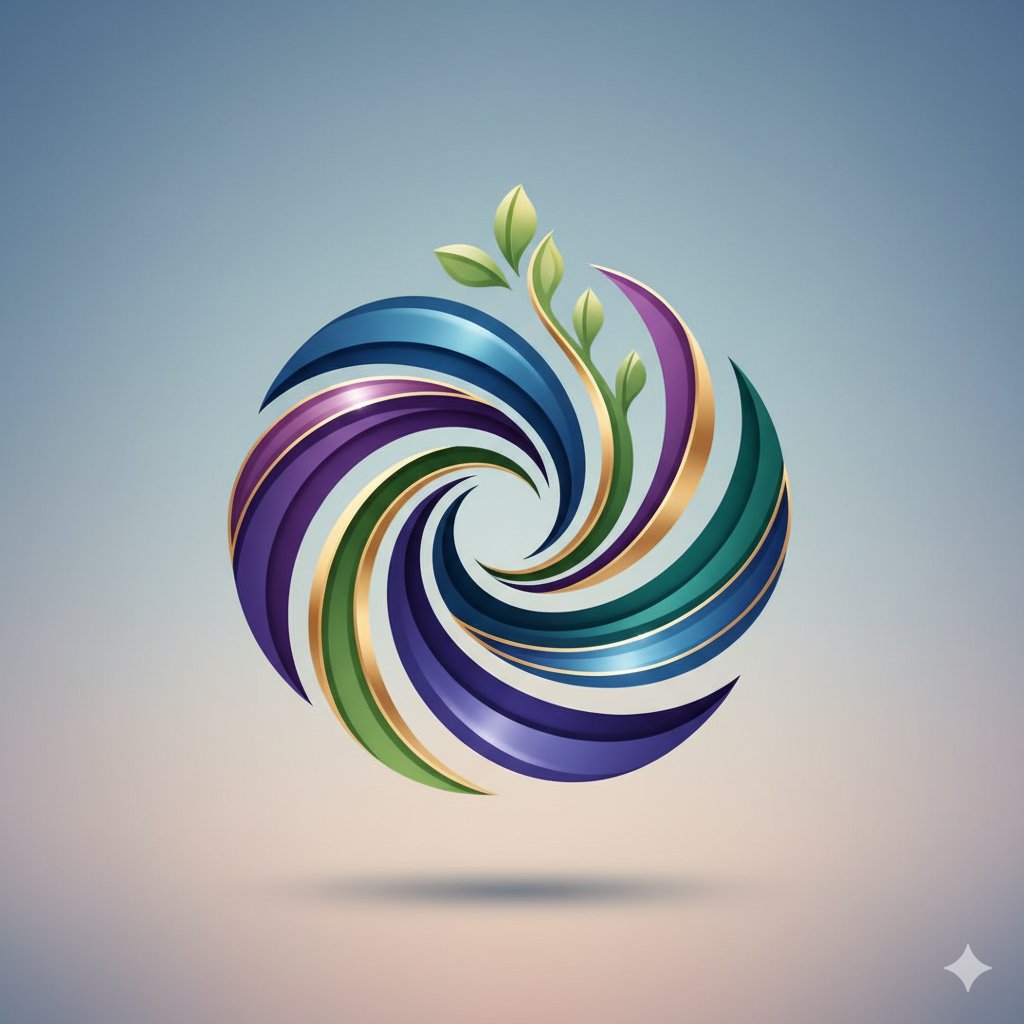}
    \end{subfigure}%
    \caption{Image bank for visual elements of type \texttt{logo}, generated with Gemini 2.5 Flash~\cite{CBSP25}.}\label{app:fig_ve_logo}
\end{figure*}

\begin{figure*}[htbp]
    \setlength{\vefigwidth}{0.24\textwidth}
    \centering
    \begin{subfigure}{\vefigwidth}
        \includegraphics[width=\linewidth]{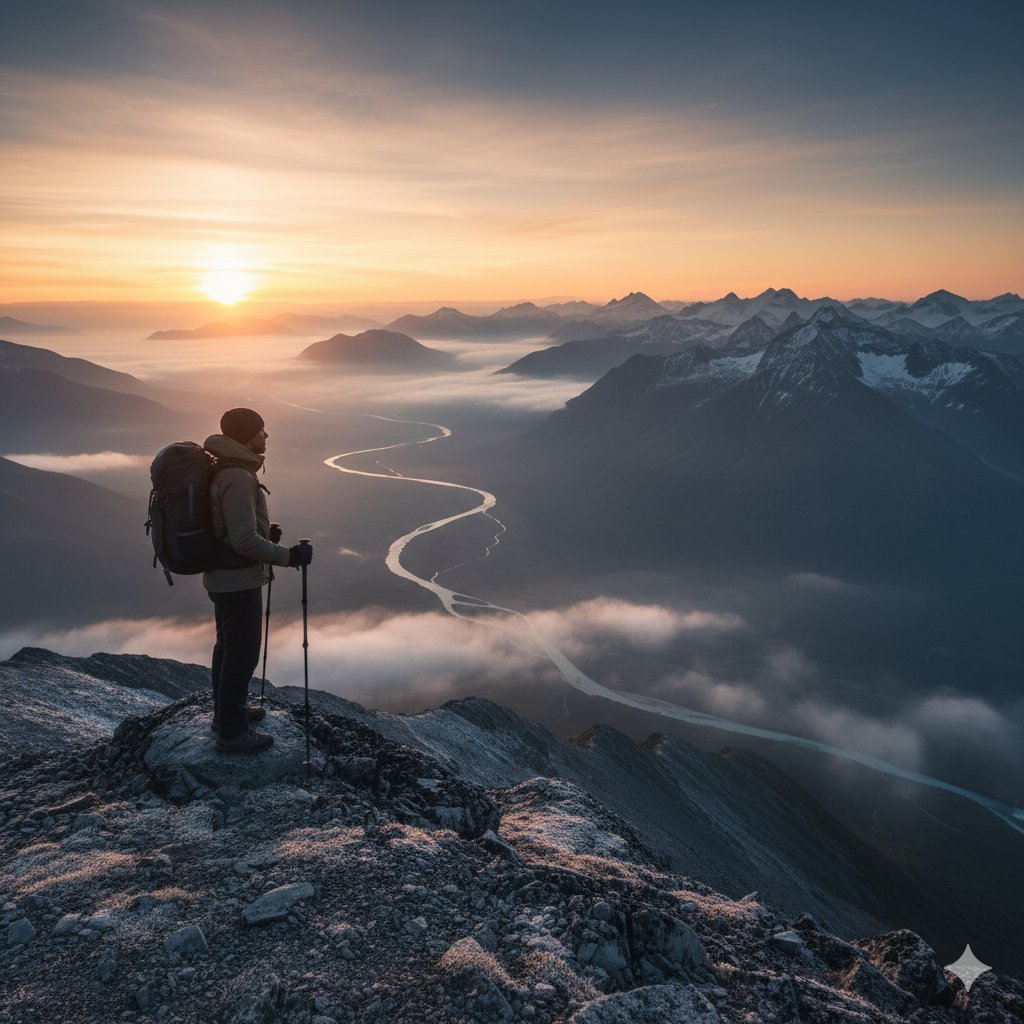}
    \end{subfigure}%
    \hfill
    \begin{subfigure}{\vefigwidth}
        \includegraphics[width=\linewidth]{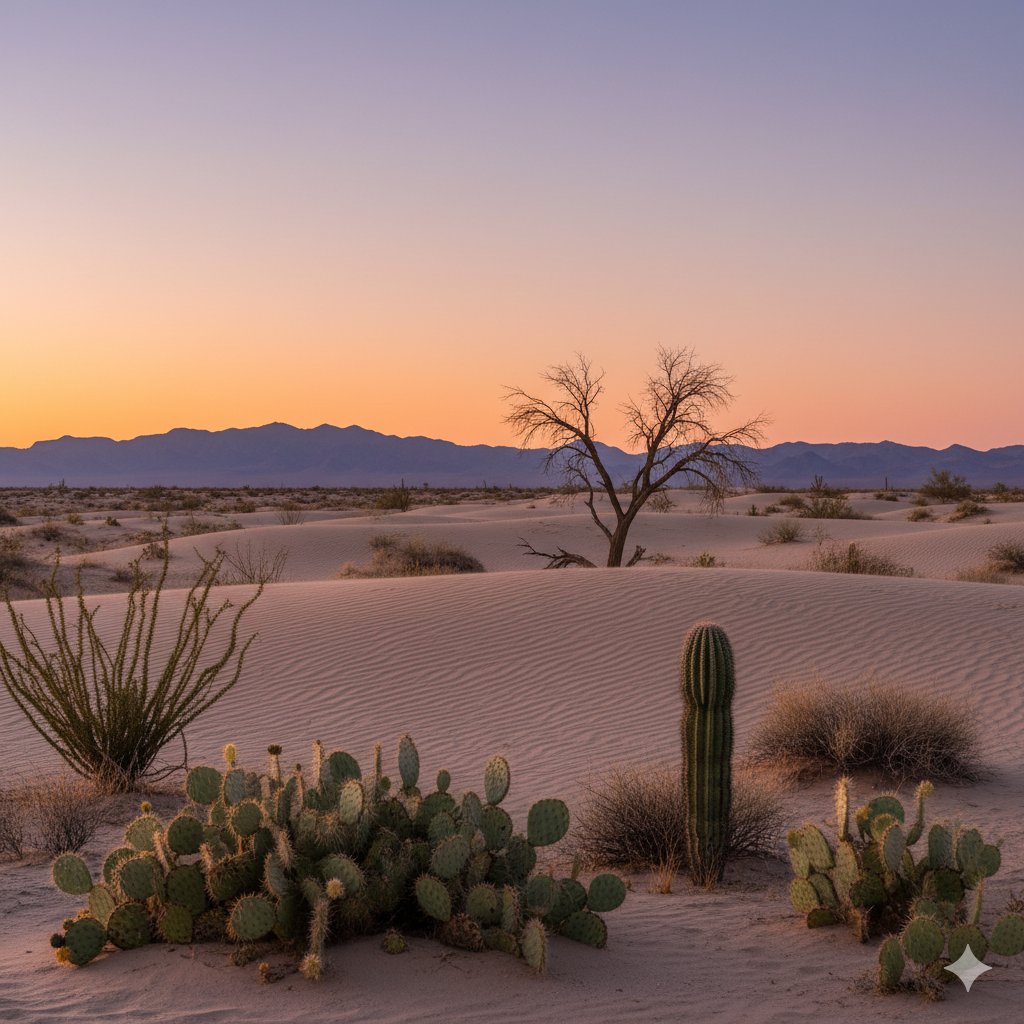}
    \end{subfigure}%
    \hfill
    \begin{subfigure}{\vefigwidth}
        \includegraphics[width=\linewidth]{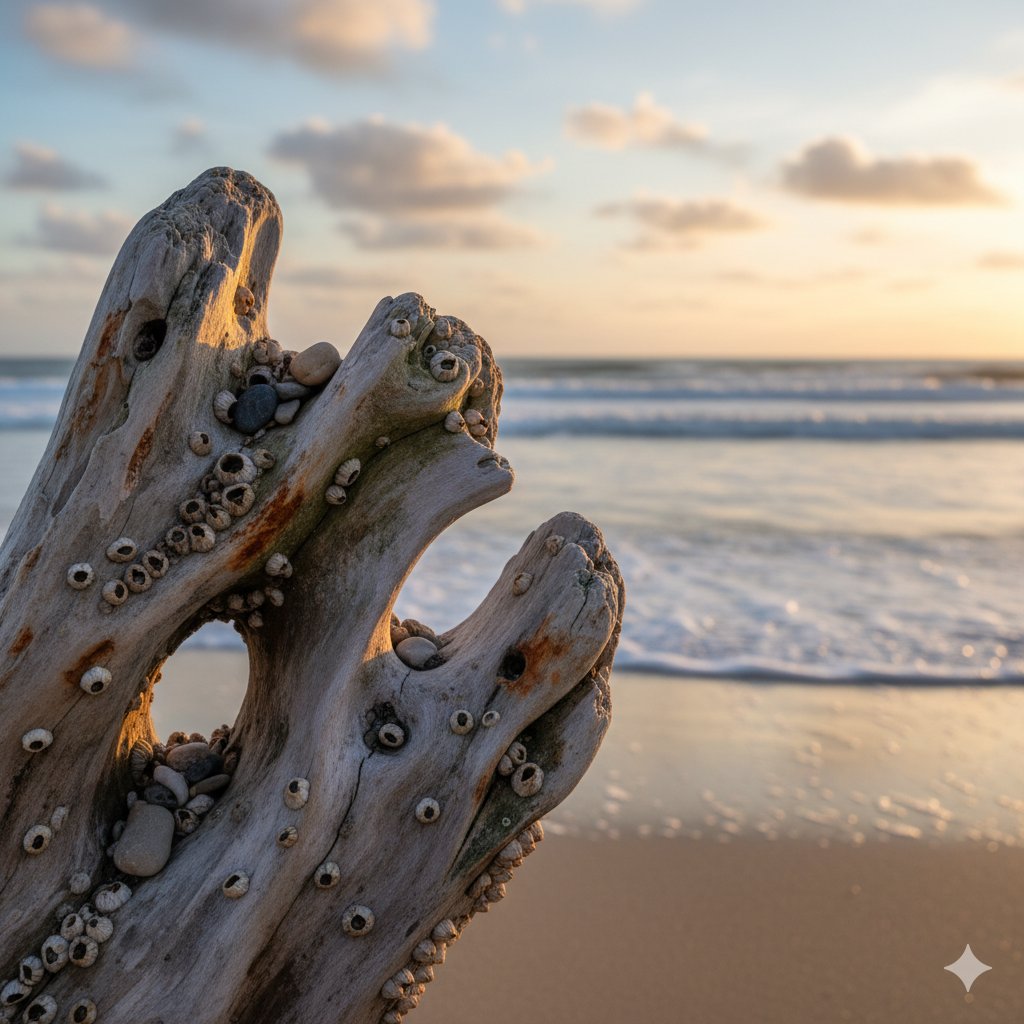}
    \end{subfigure}%
    \hfill
    \begin{subfigure}{\vefigwidth}
        \includegraphics[width=\linewidth]{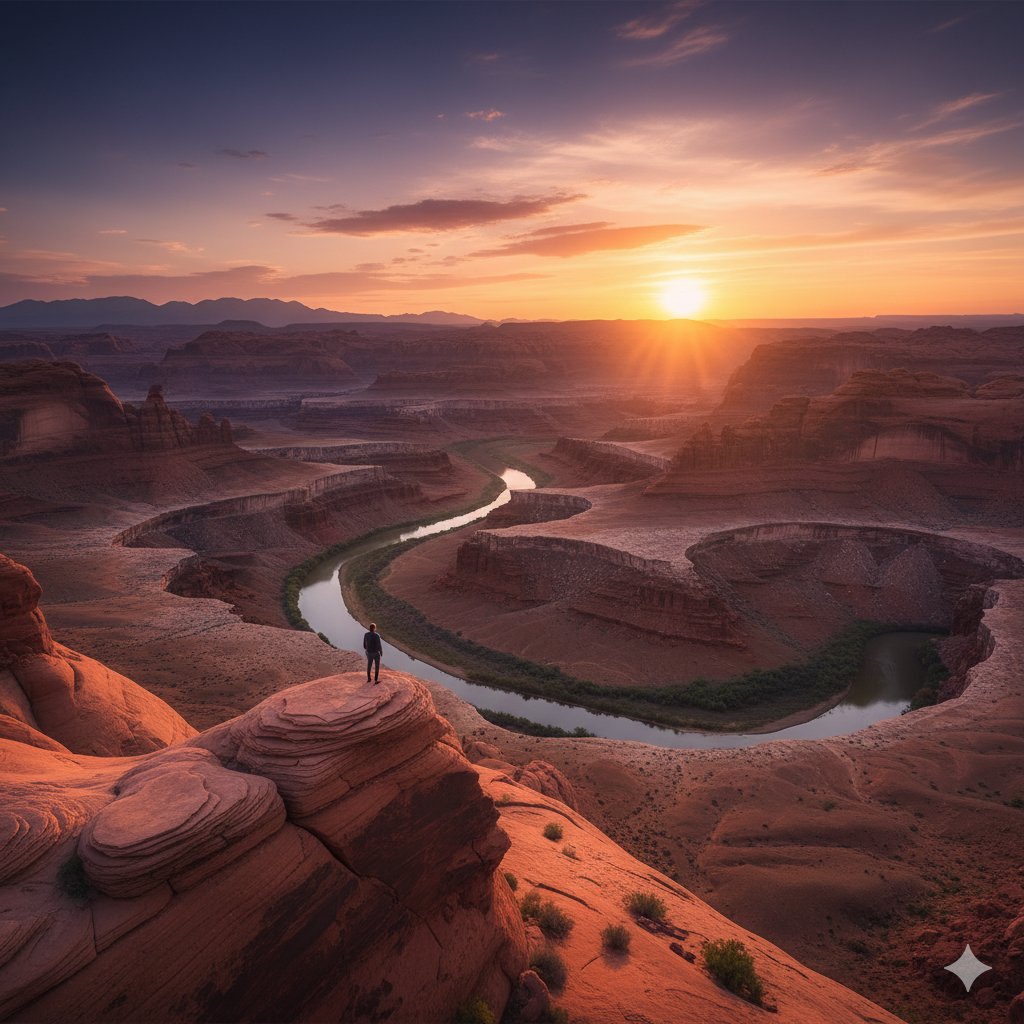}
    \end{subfigure}

    \setlength{\vefigwidth}{0.19\textwidth}
    \begin{subfigure}{\vefigwidth}
        \includegraphics[width=\linewidth]{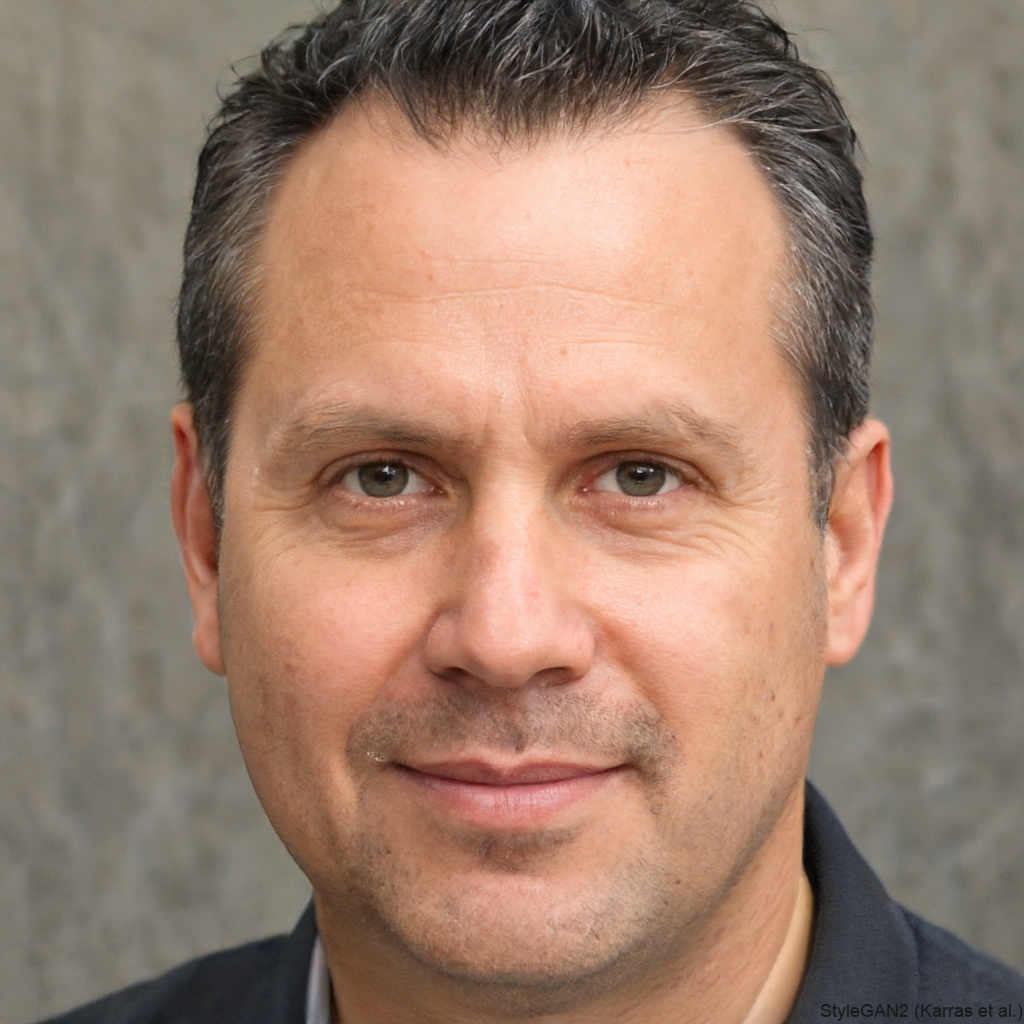}
    \end{subfigure}%
    \hfill
    \begin{subfigure}{\vefigwidth}
        \includegraphics[width=\linewidth]{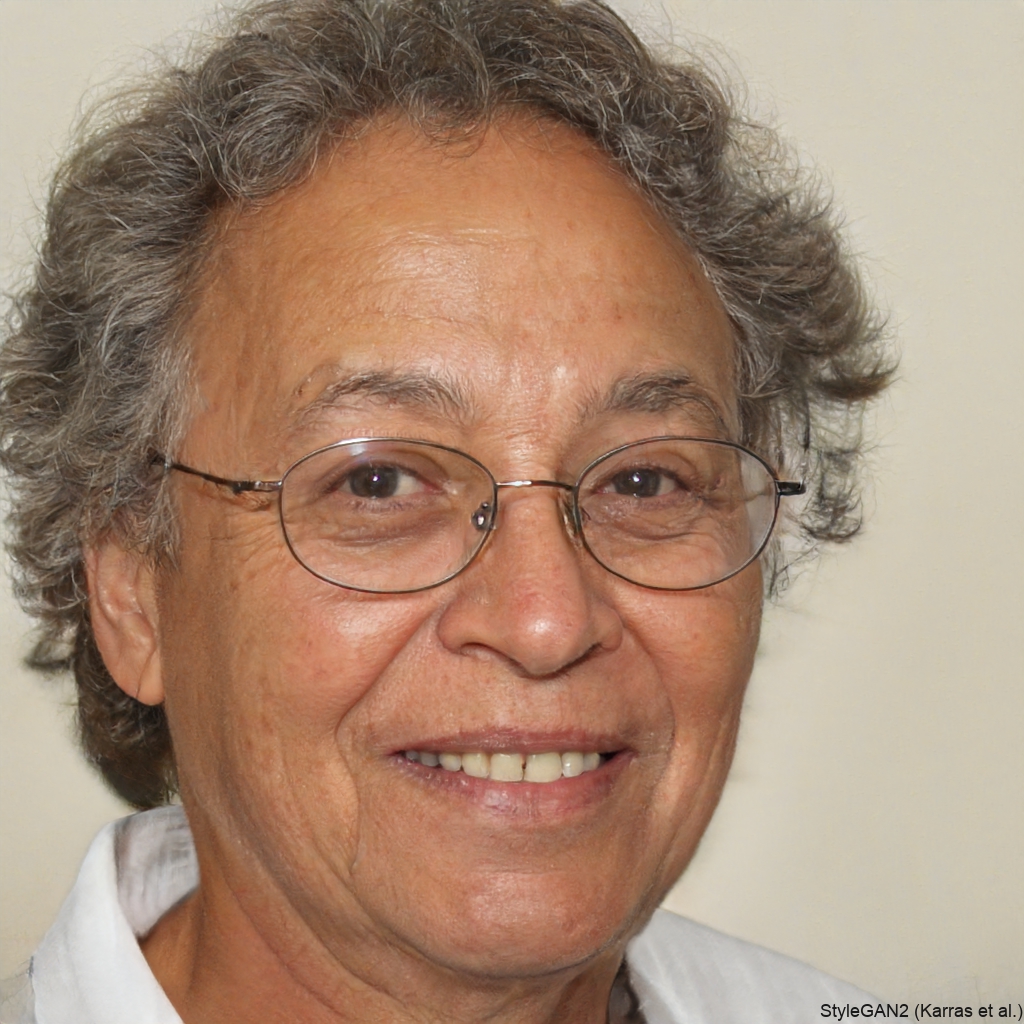}
    \end{subfigure}%
    \hfill
    \begin{subfigure}{\vefigwidth}
        \includegraphics[width=\linewidth]{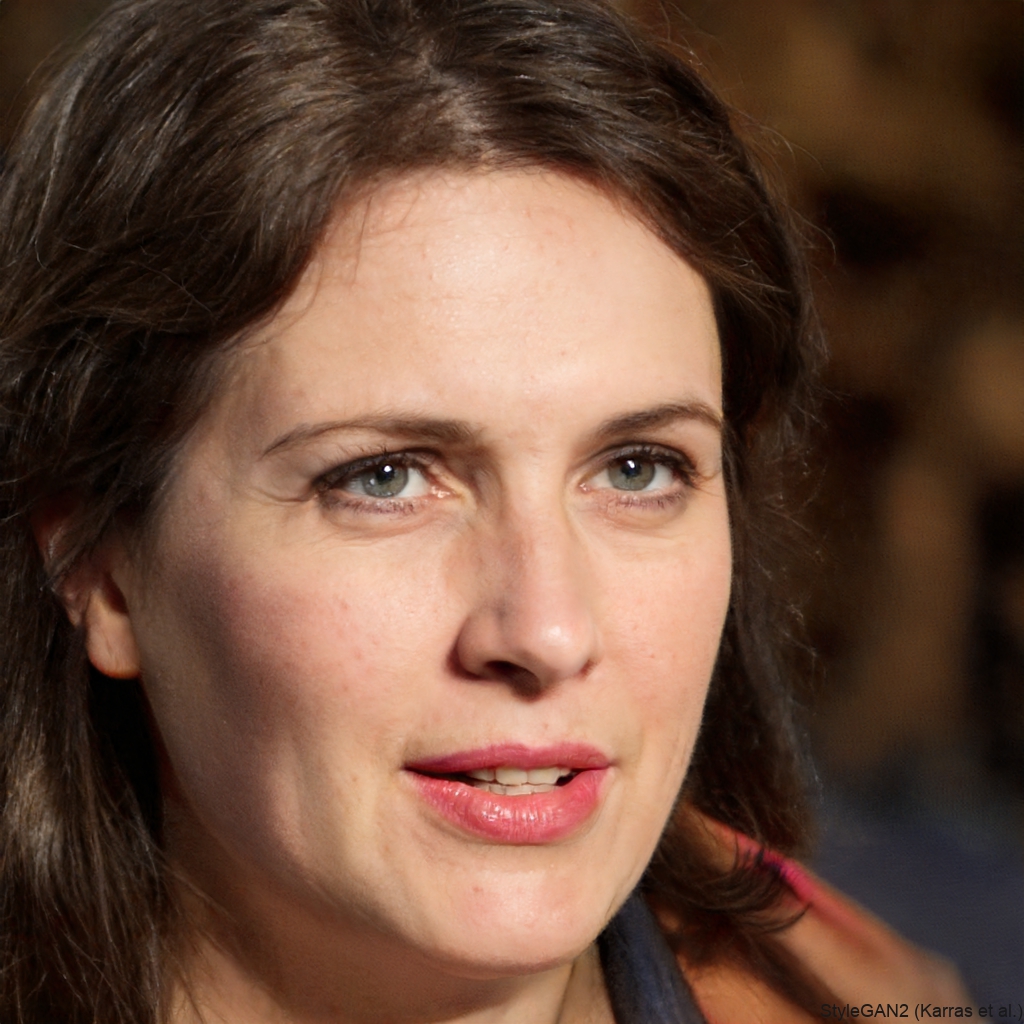}
    \end{subfigure}%
    \hfill
    \begin{subfigure}{\vefigwidth}
        \includegraphics[width=\linewidth]{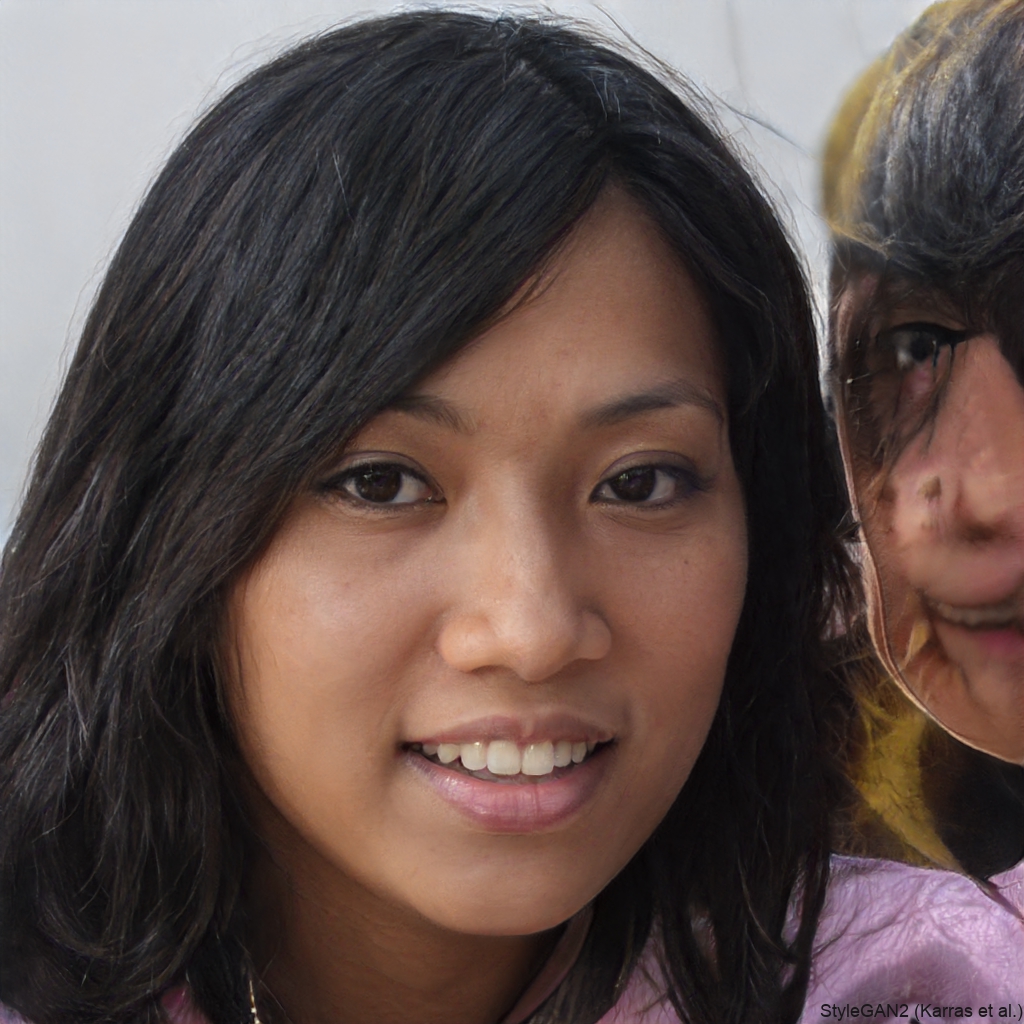}
    \end{subfigure}%
    \hfill
    \begin{subfigure}{\vefigwidth}
        \includegraphics[width=\linewidth]{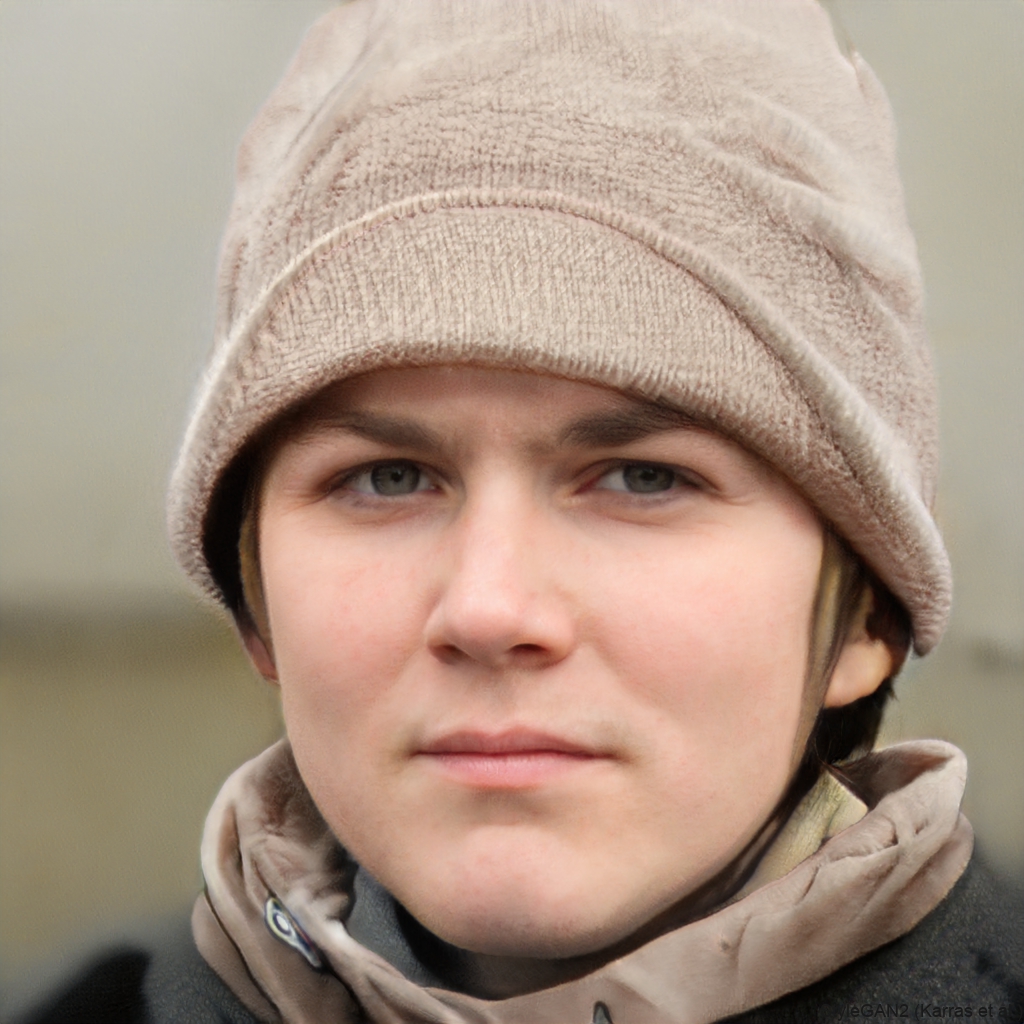}
    \end{subfigure}%
    \caption{Image bank for visual elements of type \texttt{photo}, generated with Gemini 2.5 Flash~\cite{CBSP25} (top). Synthetic faces (bottom) generated with StyleGAN2~\cite{KLAH20} via ThisPersonDoesNotExist.com.}\label{app:fig_ve_photo}
\end{figure*}

The resulting images for \texttt{figure}, \texttt{logo} and \texttt{photo} are shown in \cref{app:fig_ve_figure,app:fig_ve_logo,app:fig_ve_photo}, respectively.

\paragraph{Type Mapping.}
\begin{sloppypar}
As a post-processing step, we map certain VLM-predicted types to canonical categories:
\texttt{chart}, \texttt{diagram}, \texttt{plot}, \texttt{graph}, \texttt{illus\-tra\-tion}, and \texttt{info\-graphic} $\rightarrow$~\texttt{figure};
\texttt{image} $\rightarrow$~\texttt{photo};
\texttt{seal} $\rightarrow$~\texttt{stamp}. While such mislabelings are rare, this mapping helps retain more synthesized documents. For DLA, we augment ground truth annotations with \textit{Figure}/\textit{Picture} regions where needed, ensuring consistency between layout structure and annotations.
\end{sloppypar}

\section{Environmental Impact and Energy Estimates}\label{app:environmental_impact}
We provide conservative estimates of the energy consumption and carbon footprint of this work.

\textbf{Model Training:} Training the DocDjinn models consumed approximately 2,507 GPU hours, corresponding to $\sim$752~kWh of energy. Using a carbon intensity of 0.385~kg~CO$_2$/kWh (typical for European grids), this results in approximately 290~kg~CO$_2$.

\textbf{VLM Inference:} Direct measurements for Claude Sonnet 4.5 are not yet available. Based on benchmarking of similar frontier models\footnotemark[1], we conservatively estimate that our 533M-token workload consumed $\sim$113~kWh. Using a carbon intensity of 0.287~kg~CO$_2$/kWh (reported for Claude infrastructure\footnotemark[1]), this corresponds to approximately 33~kg~CO$_2$.

\textbf{Total Estimated Footprint:} The combined estimated carbon footprint is $\sim$323~kg~CO$_2$. These estimates are conservative and intended as upper bounds. Carbon offsets have been purchased through Climeworks to compensate for these emissions.

\footnotetext[1]{Nidhal Jegham, Marwan F. Abdelatti, Lassad Elmoubarki, Abdeltawab M. Hendawi, "How Hungry is AI? Benchmarking Energy, Water, and Carbon Footprint of LLM Inference," \emph{CoRR}, vol. abs/2505.09598, 2025.}

\section{Prompt Templates}\label{app:prompt}
We employ two prompt templates that differ in how the VLM generates ground truth (GT), corresponding to the annotation granularity required by each task family:

\noindent\textbf{(1) Macro Template (Document-Level GT):}
This template instructs the VLM to generate document-level GT for tasks such as VQA and simple KIE. The GT is embedded as a JSON object within \texttt{<script></script>} tags in the HTML, separate for each synthetic document. For instance, VQA tasks generate JSON in the form: \texttt{\{"Q1": "A1", "Q2": "A2", ...\}}, where keys are question texts and values are corresponding answers. The macro prompt template is provided below:
\tcbset{
  metricbox/.style={
    enhanced,
    breakable,
    sharp corners,
    colback=gray!5,
    colframe=gray!50,
    boxrule=0.5pt,
    fonttitle=\bfseries,
    coltitle=black,
    title={#1},
    attach boxed title to top left={yshift=-2mm, xshift=2mm},
    boxed title style={
      colback=gray!20,
      boxrule=0pt,
      sharp corners,
    },
  }
}

\begin{tcolorbox}[metricbox={Macro: Document level JSON annotations}, fontupper=\small\ttfamily,label={macro}]
\begin{verbatim}
You are an AI creating authentic HTML 
representations of documents based on 
seed images. Analyze the seed images for 
structural and semantic content and 
generate authentic variations. 
The generated documents will be printed.

## Requirements
1. **Authenticity**: Reflect stylistic 
elements from seed images without 
copying text/layouts verbatim
2. **Format**: Single-page documents 
with dimensions appropriate to the 
document type
3. **Language**: {language}
4. **Static Only**: No animations, 
transitions, or dynamic effects

## Technical
- Wrap each document in 
`<HTML>...</HTML>` 
tags, numbered sequentially
- Static CSS only for single-page 
layout
- Generate only minified CSS, HTML, 
JS.

## Content Guidelines
**DO**: Adapt cultural elements, vary 
layouts/colors/typography, use static 
styling
**DON'T**: Copy text/code blocks, reuse 
identical sections, include dynamic 
effects

## Handwritten Fields (if document 
type requires)
- Mark with class 'handwritten' 
and use regular text
- Apply no special styles to 
'handwritten', except generously 
increased size, in line with realistic
handwriting
- Assign author ID via class 
('author1', 'author2', etc.) to 
distinguish different people
- If the handwriting represents a 
signature mark it additionally with 
class 'signature'

## Visual Placeholders (if document 
type requires)
- Insert `<div data-placeholder="type" 
style="...">` for non-text elements at 
appropriate positions
- Valid types are: stamp, logo, figure, 
barcode, photo
- Add data-content attribute with 
actual content description
- For stamps, use `position:absolute;
z-index:10;` and specify 'top' 
and 'right'
- Always provide appropiate dimensions
- Example: `<div data-placeholder=
"stamp" data-content="APPROVED 
2024-03-15" 
style="position:absolute;top:50mm;
right:20mm;width:35mm;height:35mm;
z-index:10;"></div>`
- Example: `<div data-placeholder="logo" 
data-content="ACME Corp Logo" 
style="width:150mm;height:100mm;">
</div>`

## Output Format
Generate minified HTML like this:
```
1. <HTML><!DOCTYPE html><html ... 
document 1 ... </html></HTML>
2. <HTML><!DOCTYPE html><html ... 
document 2 ... </html></HTML>
...
```
## Ground Truth
Generate ground truth as JSON in 
`<script type="application/json" 
id="GT">...</script>` tag.
Ground truth specification: {gt_type}
Ground truth must follow the 
format: {gt_format}

## Quality Checklist
- [ ] Authentic variations without 
verbatim copying from seed images
- [ ] Static styling only 
(no animations or dynamic effects)
- [ ] Single-page format with 
minified HTML/CSS
- [ ] Content in {language}
- [ ] GT JSON present, correctly 
formatted and semantically coherent
- [ ] Visual elements are semantically 
coherent

Generate {num_solutions} distinct 
{doc_type} documents based on 
{num_seed_images} seed images.
\end{verbatim}
\end{tcolorbox}

\noindent\textbf{(2) Micro Template (Element-Level GT):}
This template instructs the VLM to generate element-level GT for tasks requiring fine-grained spatial annotations, such as DLA and complex KIE. The VLM assigns each applicable HTML element a class label from a predefined set \{gt\_type\} to uniquely identify its semantic role. For example, all elements containing figures, images, or visuals are assigned the class \texttt{"LE-FIGURE"}. The micro prompt template is provided below:
\tcbset{
  metricbox/.style={
    enhanced,
    breakable,
    sharp corners,
    colback=gray!5,
    colframe=gray!50,
    boxrule=0.5pt,
    fonttitle=\bfseries,
    coltitle=black,
    title={#1},
    attach boxed title to top left={yshift=-2mm, xshift=2mm},
    boxed title style={
      colback=gray!20,
      boxrule=0pt,
      sharp corners,
    },
  }
}

\begin{tcolorbox}[metricbox={Micro: Element level annotations}, fontupper=\small\ttfamily,label={micro}]
\begin{verbatim}
You are an AI creating authentic HTML 
representations of documents based on 
seed images. Analyze the seed images for 
structural and semantic content and 
generate authentic variations. 
The generated documents will be printed.

## Requirements
1. **Authenticity**: Reflect stylistic 
elements from seed images without 
copying text/layouts verbatim
2. **Format**: Single-page documents 
with dimensions appropriate to the 
document type
3. **Language**: {language}
4. **Static Only**: No animations, 
transitions, or dynamic effects

## Technical
- Wrap each document in 
`<HTML>...</HTML>` 
tags, numbered sequentially
- Static CSS only for single-page 
layout
- Generate only minified CSS, HTML, 
JS.

## Content Guidelines
**DO**: Adapt cultural elements, vary 
layouts/colors/typography, use static 
styling
**DON'T**: Copy text/code blocks, reuse 
identical sections, include dynamic 
effects

## Handwritten Fields (if document 
type requires)
- Mark with class 'handwritten' 
and use regular text
- Apply no special styles to 
'handwritten', except generously 
increased size, in line with realistic
handwriting
- Assign author ID via class 
('author1', 'author2', etc.) to 
distinguish different people
- If the handwriting represents a 
signature mark it additionally with 
class 'signature'

## Visual Placeholders (if document 
type requires)
- Insert `<div data-placeholder="type" 
style="...">` for non-text elements at 
appropriate positions
- Valid types are: stamp, logo, figure, 
barcode, photo
- Add data-content attribute with 
actual content description
- For stamps, use `position:absolute;
z-index:10;` and specify 'top' 
and 'right'
- Always provide appropiate dimensions
- Example: `<div data-placeholder=
"stamp" data-content="APPROVED 
2024-03-15" 
style="position:absolute;top:50mm;
right:20mm;width:35mm;height:35mm;
z-index:10;"></div>`
- Example: `<div data-placeholder="logo" 
data-content="ACME Corp Logo" 
style="width:150mm;height:100mm;">
</div>`

## Output Format
Generate minified HTML like this:
```
1. <HTML><!DOCTYPE html><html ... 
document 1 ... </html></HTML>
2. <HTML><!DOCTYPE html><html ... 
document 2 ... </html></HTML>
...
```
## Ground Truth
Generate ground truth by assigning each 
applicable element in HTML a class from 
the list below to uniquely identify 
its label:
{gt_type}
{gt_format}


## Quality Checklist
- [ ] Authentic variations without 
verbatim copying from seed images
- [ ] Static styling only 
(no animations or dynamic effects)
- [ ] Single-page format with 
minified HTML/CSS
- [ ] Content in {language}
- [ ] GT labels via class annotations 
are present and assigned to 
correct elements
- [ ] Visual elements are semantically 
coherent

Generate {num_solutions} distinct 
{doc_type} documents based on 
{num_seed_images} seed images.
\end{verbatim}
\end{tcolorbox}

\noindent Both templates are instantiated with five parameters: language, document type (\texttt{\{doc\_type\}}), GT type (\texttt{\{gt\_type\}}), GT format (\texttt{\{gt\_format\}}), and the number of documents to generate (\texttt{\{num\_solutions\}}), as described in Section~\ref{sec:vlm_synthesis}.


\section{Synthetic Dataset Definitions}\label{app:syn_definitions}
Below we list the configurations for all synthetic datasets generated in this work. Each definition instantiates the prompt templates from Section~\ref{app:prompt}, specifying the Task Type (VQA, DLA, KIE, CLS), Prompt Type (JSON or Annotation), and prompt parameters in YAML format.

\tcbset{
  datasetbox/.style={
    enhanced,
    breakable,
    colback=gray!5,
    colframe=gray!50,
    boxrule=0.5pt,
    fonttitle=\bfseries,
    coltitle=black,
    title={#1},
    attach boxed title to top left={yshift=-2mm, xshift=2mm},
    boxed title style={
      colback=gray!20,
      boxrule=0pt,
      sharp corners,
    },
  }
}
\lstdefinelanguage{yaml}{ 
  keywords={num_solutions,doc_type,gt_type,gt_format},
  keywordstyle=\bfseries,
  basicstyle=\ttfamily\small,
  sensitive=true,
}
\newcounter{romancount}
\newcommand{\syndef}[1]{%
  \stepcounter{romancount}
  Dataset Definition \Roman{romancount}: #1%
}
\begin{tcolorbox}[datasetbox={\syndef{DocVQA}}, fontupper=\small\ttfamily,label={docvqa}]
\medskip
\textbf{Task Type:} VQA \\
\textbf{Prompt Type:} JSON 
\tcblower 
\begin{lstlisting}[language=yaml]
num_solutions: 3
doc_type: "business and administrative"

gt_type: "Multiple questions about each document, with their answers taken **verbatim** from the document."
gt_format: '{"<Text of question 1>": "<Answer to question 1>", "<Text of question 2>": "<Answer to question 2>", ...}'
\end{lstlisting}
\end{tcolorbox}

\begin{tcolorbox}[datasetbox={\syndef{WTQ}}, fontupper=\small\ttfamily,label={wikitables}]
\medskip
\textbf{Task Type:} VQA \\
\textbf{Prompt Type:} JSON 
\tcblower 
\begin{lstlisting}[language=yaml]
num_solutions: 3
doc_type: "semi-structures table"

gt_type: |
    Multiple complex question-answer pairs in everyday language that can be answered from the associated table, with their answers taken **verbatim** from the document.
    Common Question Types:
    * Lookup: Finding specific cell values ("What is the capital of France?")
    * Aggregation: Counting, summing, averaging ("How many players scored over 20 points?")
    * Comparison: Finding max/min ("Which country has the largest population?")
    * Reasoning: Requiring multiple steps ("What team did the highest scorer play for?")
gt_format: '{"<Text of question 1>": "<Answer to question 1>", "<Text of question 2>": "<Answer to question 2>", ...}'



\end{lstlisting}
\end{tcolorbox}

\begin{tcolorbox}[datasetbox={\syndef{CORD}}, fontupper=\small\ttfamily,label={cord}]
\medskip
\textbf{Task Type:} KIE \\
\textbf{Prompt Type:} annotation 
\tcblower 
\begin{lstlisting}[language=yaml]
num_solutions: 3
doc_type: "receipt"

gt_type: |
    (if applicable, provide as plaintext values from the document)
      // Menu items (multiple menu items are allowed)
      * "MENU_NM": The menu item name.
      * "MENU_NUM": The menu item number or identifier.
      * "MENU_UNITPRICE": The price per unit of the menu item.
      * "MENU_CNT": The quantity or count of the menu item.
      * "MENU_DISCOUNTPRICE": The discount amount applied to the menu item.
      * "MENU_PRICE": The final price of the menu item.
      * "MENU_ITEMSUBTOTAL": The subtotal for this menu item line.
      * "MENU_VATYN": The VAT indicator (yes/no) for the menu item.
      * "MENU_ETC": Other miscellaneous menu item information.
      * "MENU_SUB_NM": The name of a sub-item or modifier.
      * "MENU_SUB_UNITPRICE": The price per unit of the sub-item.
      * "MENU_SUB_CNT": The quantity of the sub-item.
      * "MENU_SUB_PRICE": The price of the sub-item.
      * "MENU_SUB_ETC": Other sub-item information.
      // Menu items that were canceled
      * "VOID_MENU_NM": The name of a cancelled or voided item.
      * "VOID_MENU_PRICE": The price of the cancelled item.
      // Generic receipt data
      * "SUB_TOTAL_SUBTOTAL_PRICE": The subtotal before additional charges.
      * "SUB_TOTAL_DISCOUNT_PRICE": The total discount amount.
      * "SUB_TOTAL_SERVICE_PRICE": The service charge or fee.
      * "SUB_TOTAL_OTHERSVC_PRICE": Other service charges.
      * "SUB_TOTAL_TAX_PRICE": The tax amount.
      * "SUB_TOTAL_ETC": Other subtotal information.
      * "TOTAL_TOTAL_PRICE": The final total amount on the receipt.
      * "TOTAL_TOTAL_ETC": Other total-related information.
      * "TOTAL_CASHPRICE": The amount paid in cash.
      * "TOTAL_CHANGEPRICE": The change given back to the customer.
      * "TOTAL_CREDITCARDPRICE": The amount paid by credit card.
      * "TOTAL_EMONEYPRICE": The amount paid by electronic money or digital payment.
      * "TOTAL_MENUTYPE_CNT": The count of different menu item types.
      * "TOTAL_MENUQTY_CNT": The total quantity of all items ordered.

gt_format: | Group individual menu items in groups using the menu item enumerator class MENU_<idx> and a sub-field class from the list above (e.g. "MENU_1 MENU_NM", "MENU_1 MENU_CNT", "MENU_2 MENU_NM", ...). For void/canceled menu items use the class "VOID_MENU" instead of the enumeration. For generic receipt data use the class "GENERIC".

\end{lstlisting}
\end{tcolorbox}

\begin{tcolorbox}[datasetbox={\syndef{FUNSD}}, fontupper=\small\ttfamily,label={funsd}]
\medskip
\textbf{Task Type:} KIE \\
\textbf{Prompt Type:} annotation 
\tcblower 
\begin{lstlisting}[language=yaml]
num_solutions: 3
doc_type: "form"

gt_type: |
    keys and their values structured as QA pairs
      * "HEADER": The header of the question answer pair.
      * "QUESTION": The question i.e. a key.
      * "ANSWER": The answer i.e a value.
gt_format: |
    Group individual annotations in groups using the enumerator class PAIR_<idx> and a annotation class from the list above (e.g. "PAIR_1 QUESTION", "PAIR_1 ANSWER", "PAIR_2 HEADER", ...). Ensure to annotate exact using spans, i.e. "QUESTION" element should not contain "ANSWER".

\end{lstlisting}
\end{tcolorbox}

\begin{tcolorbox}[datasetbox={\syndef{Kleister Charity}}, fontupper=\small\ttfamily,label={kleister}]
\medskip
\textbf{Task Type:} KIE \\
\textbf{Prompt Type:} JSON 
\tcblower 
\begin{lstlisting}[language=yaml]
num_solutions: 3
doc_type: "UK charity annual financial report"

gt_type: |
    keys and their values (if applicable, provide as plaintext values from the document):
      * "address__post_town": Post town of the address of the charitable organization.
      * "address__postcode": Postcode of the address of the charitable organization.
      * "address__street_line": Street line of the address of the charitable organization.
      * "charity_name": The name of the charitable organization.
      * "charity_number": The registered number of the charitable organization.
      * "income_annually_in_british_
      pounds": The annual income in British Pounds of the charitable organization.
      * "report_date": The reporting date of the annual document of the charitable organization.
      * "spending_annually_in_british_
      pounds": The annual spending in British Pounds of the charitable organization.
gt_format: '{"address__post_town": "<value>", "spending_annually_in_british_pounds": "<value>", ...}'

\end{lstlisting}
\end{tcolorbox}

\begin{tcolorbox}[datasetbox={\syndef{SROIE}}, fontupper=\small\ttfamily,label={sroie}]
\medskip
\textbf{Task Type:} KIE \\
\textbf{Prompt Type:} JSON 
\tcblower 
\begin{lstlisting}[language=yaml]
num_solutions: 3
doc_type: "receipt"

gt_type: |
    keys and their values
      * "COMPANY": The company name.
      * "DATE": The date on the receipt.
      * "ADDRESS": The address of the company.
      * "TOTAL": The total amount.
gt_format: '{"COMPANY": "<value>", "DATE": "<value>", "ADDRESS": "<value>", "TOTAL": "<value>"}'



\end{lstlisting}
\end{tcolorbox}

\begin{tcolorbox}[datasetbox={\syndef{DocLayNet (CLS)}}, fontupper=\small\ttfamily,label={doclaynet_cls}]
\textbf{Task Type:} Classification \\
\textbf{Prompt Type:} JSON 
\tcblower 
\begin{lstlisting}[language=yaml]
num_solutions: 3
doc_type: "single A4 pages out of diverse business and technical"

gt_type: |
    document class label
      * financial_reports
      * scientific_articles
      * laws_and_regulations
      * government_tenders
      * manuals
      * patents
gt_format: 'JSON object {"label": "<class label>"}'

\end{lstlisting}
\end{tcolorbox}

\begin{tcolorbox}[datasetbox={\syndef{RVL-CDIP}}, fontupper=\small\ttfamily,label={rvlcdip}]
\medskip
\textbf{Task Type:} CLS \\
\textbf{Prompt Type:} JSON 
\tcblower 
\begin{lstlisting}[language=yaml]
num_solutions: 3
doc_type: "business correspondence and corporate"

gt_type: |
    document class label
      * letter
      * form
      * email
      * handwritten
      * advertisement
      * scientific report
      * scientific publication
      * specification
      * file folder
      * news article
      * budget
      * invoice
      * presentation
      * questionnaire
      * resume
      * memo
gt_format: 'JSON object {"label": "<class label>"}'



\end{lstlisting}
\end{tcolorbox}

\begin{tcolorbox}[datasetbox={\syndef{Tobacco3482}}, fontupper=\small\ttfamily,label={tobacco3482}]
\medskip
\textbf{Task Type:} CLS \\
\textbf{Prompt Type:} JSON 
\tcblower 
\begin{lstlisting}[language=yaml]
num_solutions: 3
doc_type: "legal and corporate"

gt_type: |
    document class labels:
      * ADVERTISEMENT: Advertisement
      * EMAIL: Email
      * FORM: Form
      * LETTER: Letter
      * MEMO: Memo
      * NEWS_ARTICLE: News article
      * NOTE: Note/handwritten note
      * REPORT: Report
      * RESUME: Resume/CV
      * SCIENTIFIC: Scientific publication
gt_format: 'JSON object {"label": "<class label>"}'



\end{lstlisting}
\end{tcolorbox}

\begin{tcolorbox}[datasetbox={\syndef{Doclaynet (DLA)}}, fontupper=\small\ttfamily,label={doclaynet_dla}]
\medskip
\textbf{Task Type:} DLA \\
\textbf{Prompt Type:} annotation 
\tcblower 
\begin{lstlisting}[language=yaml]
num_solutions: 2
doc_type: "single A4 pages out of diverse business and technical"

gt_type: |
      * "LE-CAPTION": Text that accompanies and explains figures, tables, or other visual elements, typically appearing above or below the referenced element.
      * "LE-FOOTNOTE": Supplementary notes or citations placed at the bottom of a page, providing additional context or references to the main text, distinct from footers.
      * "LE-FORMULA": Mathematical equations, chemical formulas, or symbolic expressions, whether displayed inline or as standalone elements.
      * "LE-LIST-ITEM": Individual items within enumerated, bulleted, or definition lists, with each list item annotated separately rather than as a unified list structure.
      * "LE-PAGE-FOOTER": Recurring content at the bottom of pages such as page numbers, copyright notices, document identifiers, or footer text.
      * "LE-PAGE-HEADER": Recurring content at the top of pages including running headers, document titles, chapter names.
      * "LE-PICTURE": Photographs, diagrams, charts, graphs, illustrations, and other visual content excluding tables.
      * "LE-SECTION-HEADER": Section and subsection headings.
      * "LE-TABLE": Complete table structure including grid content, inline captions, and column/row headers as a unified element.
      * "LE-TEXT": Contains regular body text including paragraphs, abstracts, definitions, descriptions, and other primary textual content.
      * "LE-TITLE": The main document title appearing prominently at the beginning of the document, distinct from section headers.
gt_format:
\end{lstlisting}
\end{tcolorbox}

\begin{tcolorbox}[datasetbox={\syndef{ICDAR2019 cTDaR}}, fontupper=\small\ttfamily,label={icda_2019}]
\medskip
\textbf{Task Type:} DLA \\
\textbf{Prompt Type:} annotation 
\tcblower 
\begin{lstlisting}[language=yaml]
num_solutions: 2
doc_type: "single A4 pages out of diverse modern digital-born and historical archival scanned"

gt_type: |
    * "LE-TABLE": Any tabular structure containing data organized in rows and columns. Include the complete table region from border to border.
gt_format:

\end{lstlisting}
\end{tcolorbox}

\begin{tcolorbox}[datasetbox={\syndef{Publaynet}}, fontupper=\small\ttfamily,label={publaynet}]
\medskip
\textbf{Task Type:} DLA \\
\textbf{Prompt Type:} annotation 
\tcblower 
\begin{lstlisting}[language=yaml]
num_solutions: 2
doc_type: "single A4 pages out of one and two column scientific article"

gt_type: |
    * "LE-TEXT": Contains regular body text including paragraphs, abstracts, authors, affiliations, keywords, footnotes, footer, references, and captions for figures and tables.
    * "LE-TITLE": Comprises all document titles and headings, article titles as well as standalone section or subsection headings that appear on their own line rather than inline with text.
    * "LE-TABLE": Denotes the main body content of tables, excluding captions and labels.
    * "LE-FIGURE": Indicates the main visual content of figures and illustrations, with multi-panel figures annotated as complete units rather than individual sub-figures.
    * "LE-LIST": Represents enumerated or bulleted list structures, with nested lists annotated as single unified objects.

gt_format:

\end{lstlisting}
\end{tcolorbox}

\section{Synthetic Dataset Samples}\label{app:synthdata}
This section provides visual examples from all synthetic datasets specified in Section~\ref{app:syn_definitions}. For each dataset, we display representative synthetic documents generated by the VLM-based synthesis pipeline in \cref{app:sds_ex_docvqa,app:sds_ex_wtq,app:sds_ex_cord,app:sds_ex_funsd,app:sds_ex_klc,app:sds_ex_sroie,app:sds_ex_doclaynetcls,app:sds_ex_rvlcdip,app:sds_ex_tobacco3482,app:sds_ex_doclaynetdla,app:sds_ex_icdar2019,app:sds_ex_publaynet}. The examples demonstrate the variety in layout, typography, and content while maintaining task-specific authenticity. 

\newlength{\sdsexamplewidth}
\setlength{\sdsexamplewidth}{0.19\textwidth}

\newlength{\sdsexampleheight}
\setlength{\sdsexampleheight}{0.45\textheight}
\begin{figure*}[htbp]
    \centering
    \begin{subfigure}{\sdsexamplewidth}
        \includegraphics[width=\linewidth,height=\sdsexampleheight,keepaspectratio]{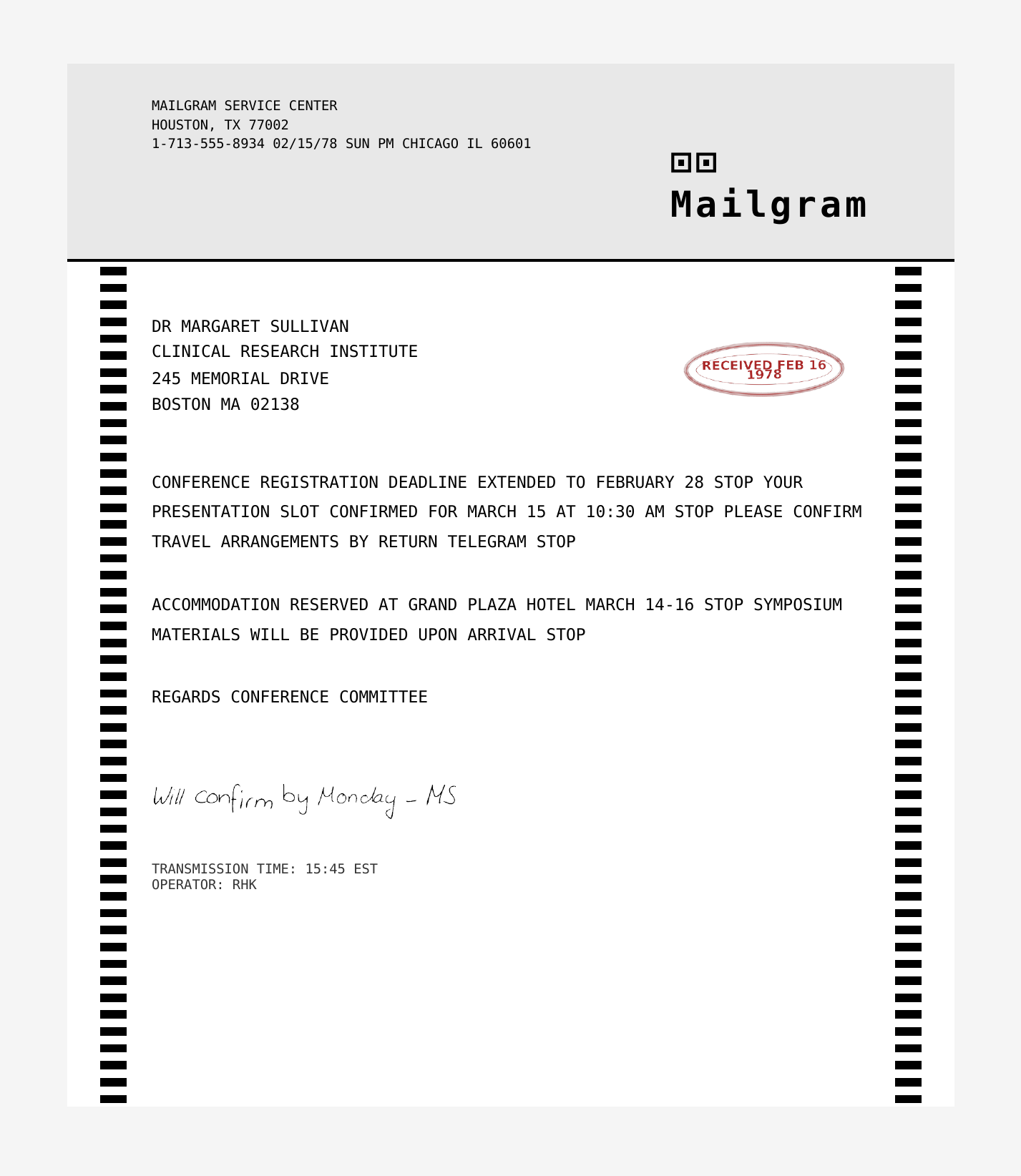}
    \end{subfigure}%
    \hfill
    \begin{subfigure}{\sdsexamplewidth}
        \includegraphics[width=\linewidth,height=\sdsexampleheight,keepaspectratio]{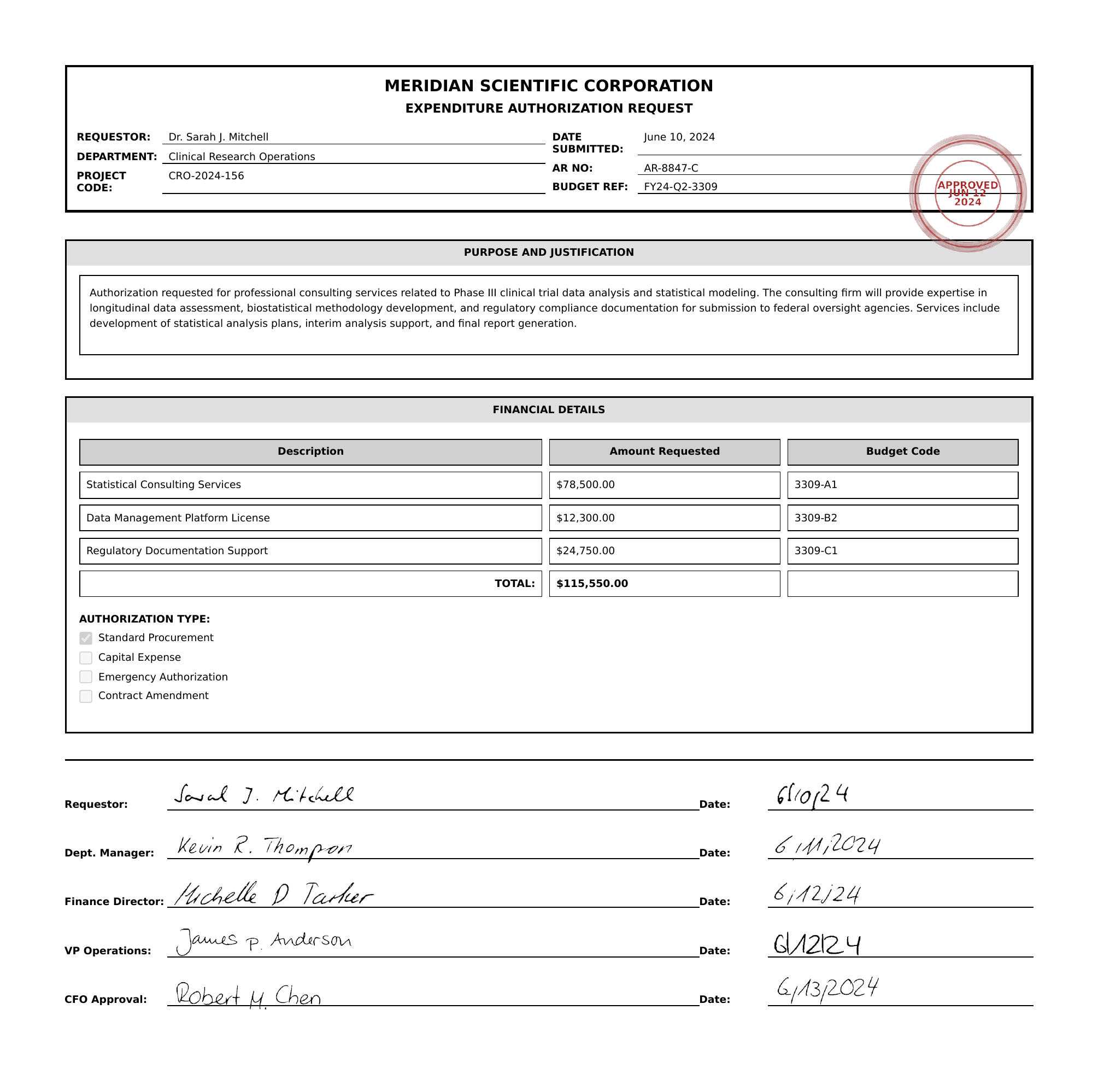}
    \end{subfigure}%
    \hfill
    \begin{subfigure}{\sdsexamplewidth}
        \includegraphics[width=\linewidth,height=\sdsexampleheight,keepaspectratio]{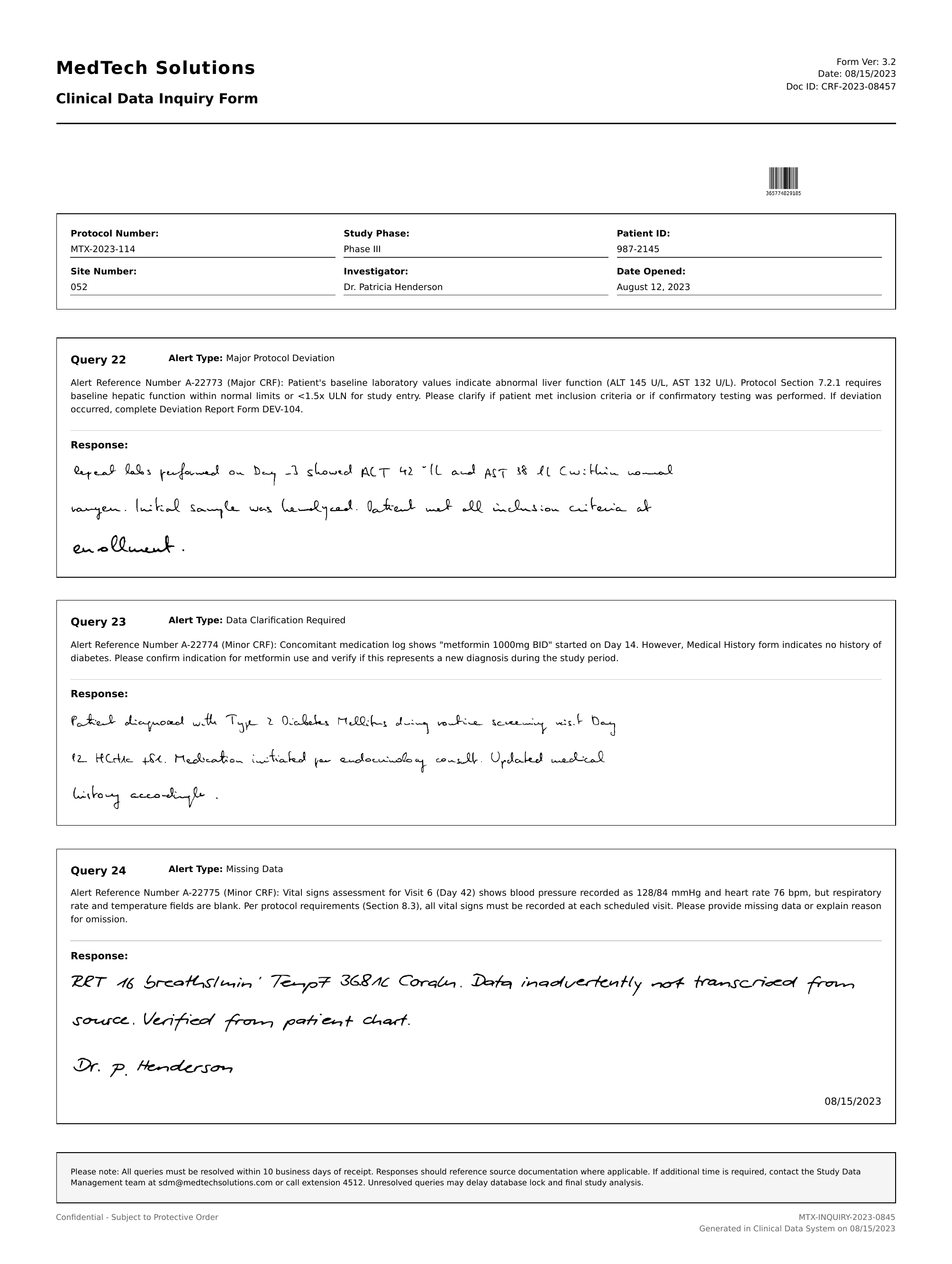}
    \end{subfigure}%
    \hfill
    \begin{subfigure}{\sdsexamplewidth}
        \includegraphics[width=\linewidth,height=\sdsexampleheight,keepaspectratio]{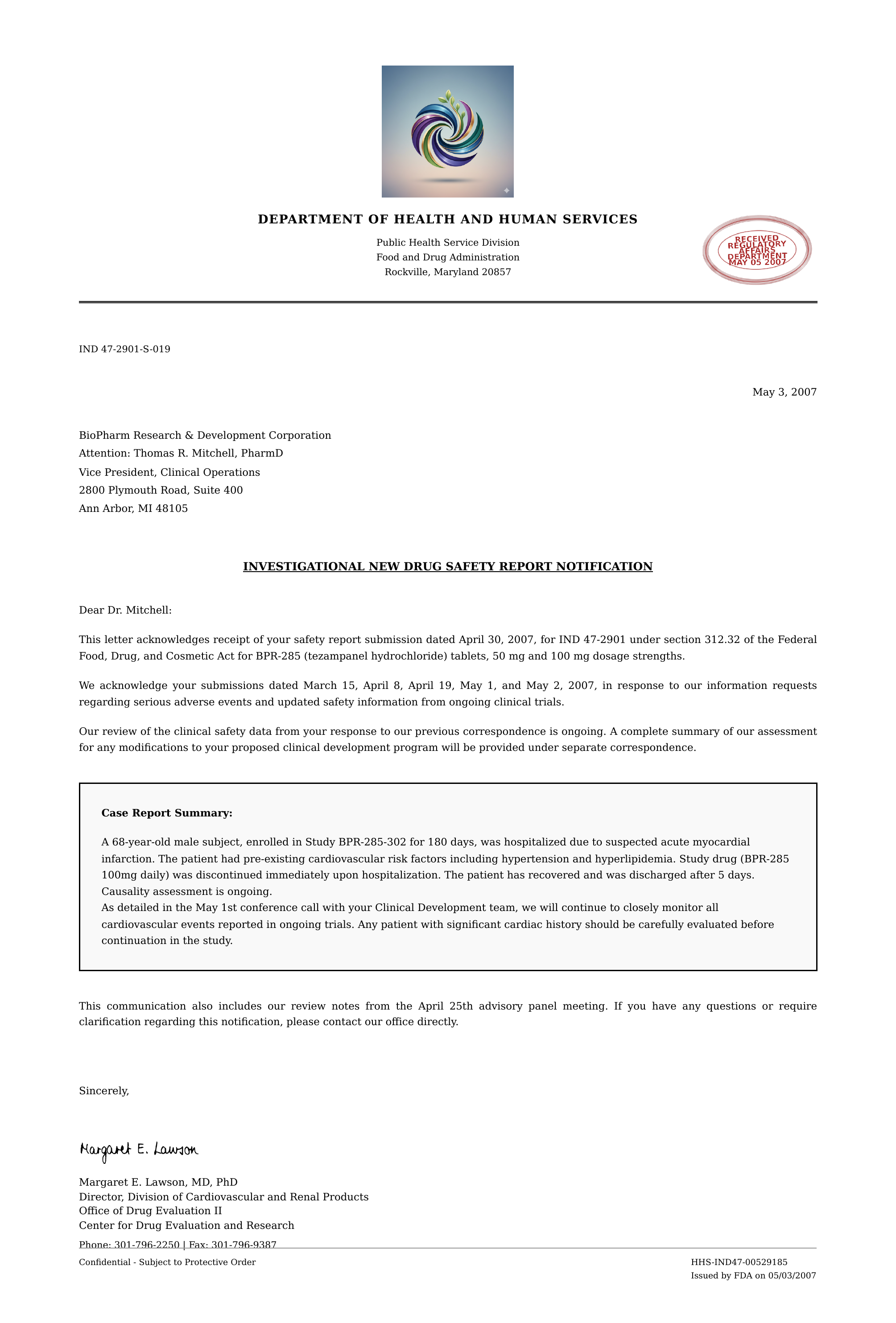}
    \end{subfigure}%
    \hfill
    \begin{subfigure}{\sdsexamplewidth}
        \includegraphics[width=\linewidth,height=\sdsexampleheight,keepaspectratio]{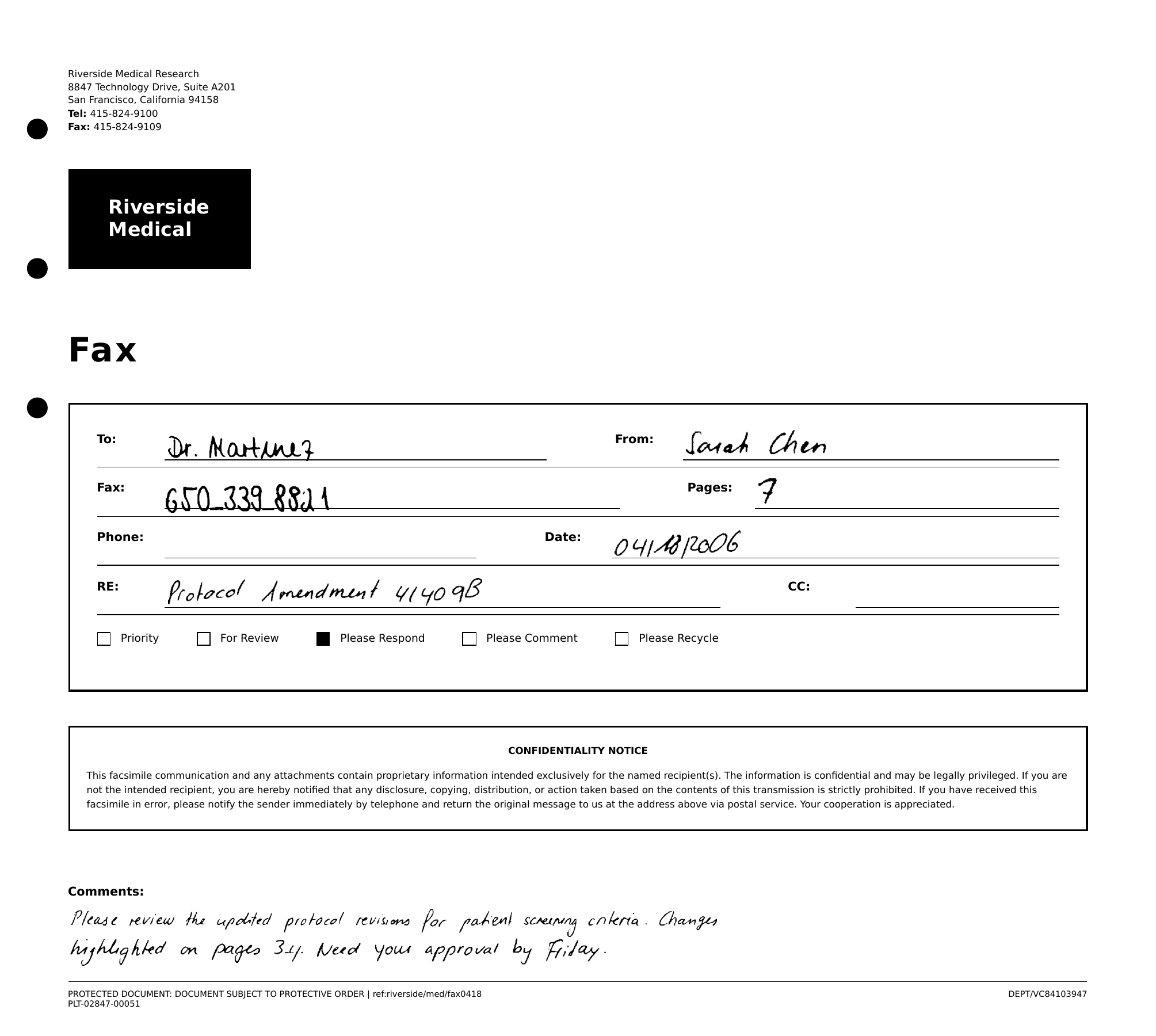}
    \end{subfigure}%
    \caption{\docvqa{} synthetic dataset samples for VQA task.}\label{app:sds_ex_docvqa}
\end{figure*}
 \begin{figure*}[htbp]
    \centering
    \begin{subfigure}{\sdsexamplewidth}
        \includegraphics[width=\linewidth,height=\sdsexampleheight,keepaspectratio]{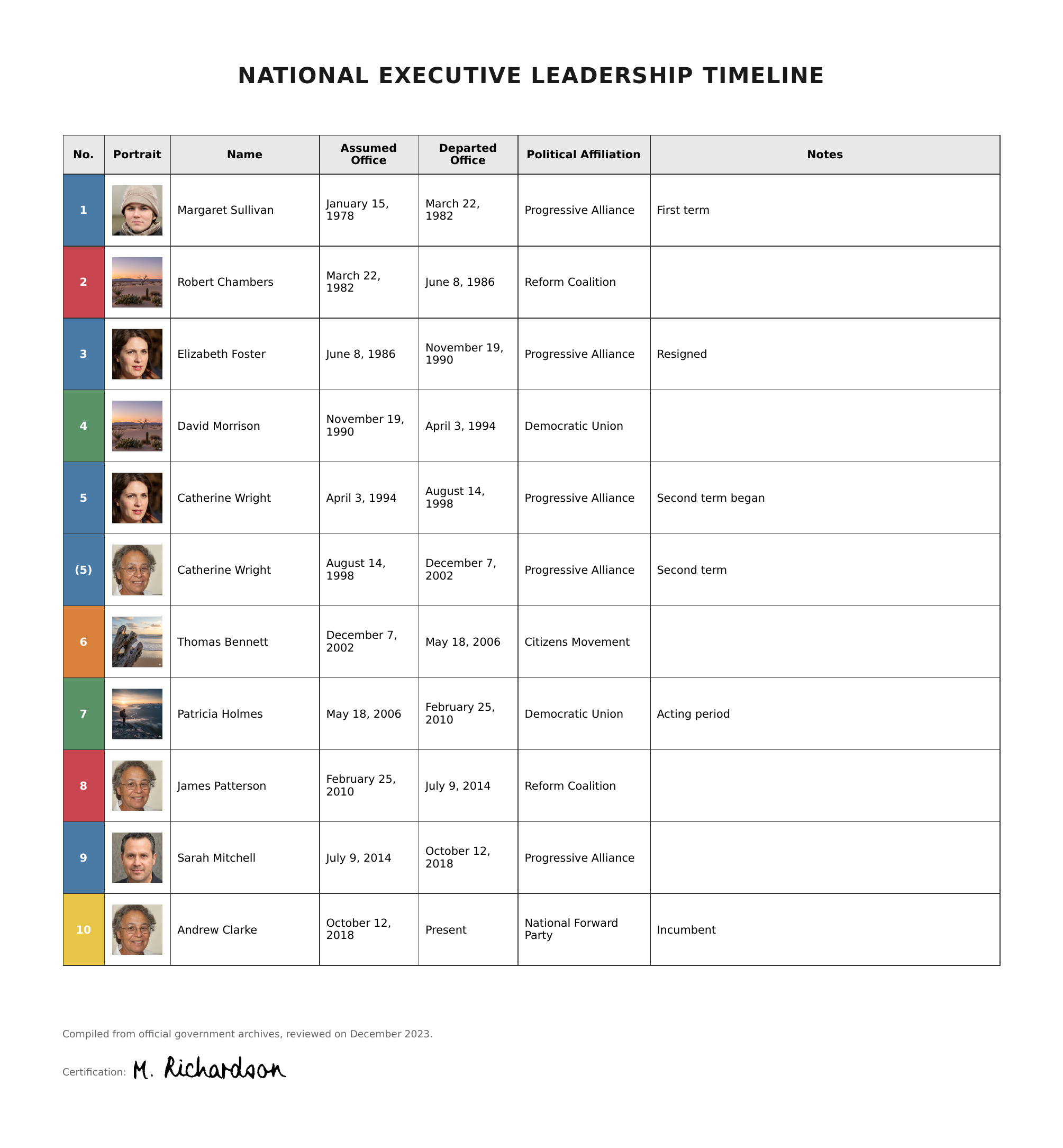}
    \end{subfigure}%
    \hfill
    \begin{subfigure}{\sdsexamplewidth}
        \includegraphics[width=\linewidth,height=\sdsexampleheight,keepaspectratio]{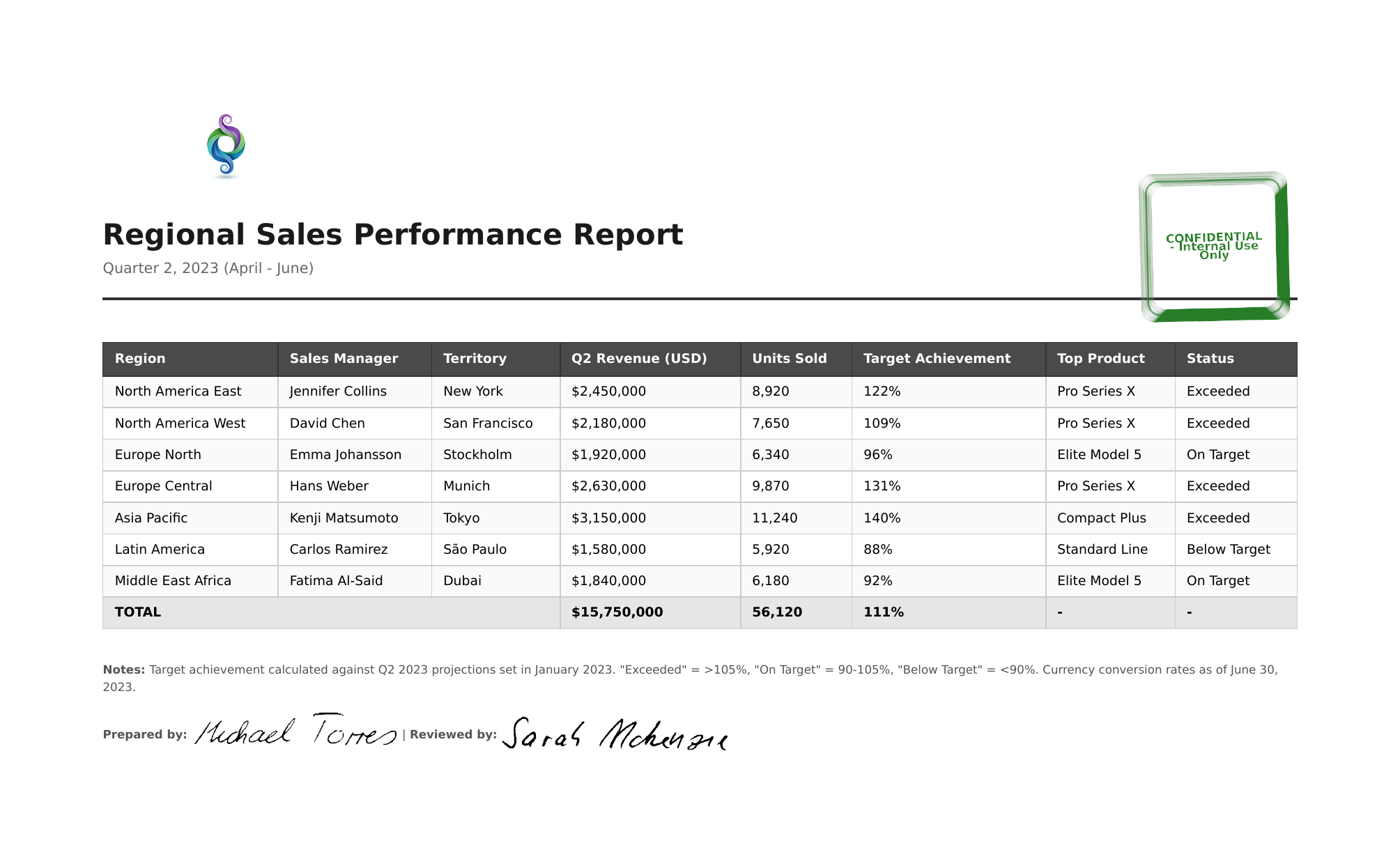}
    \end{subfigure}%
    \hfill
    \begin{subfigure}{\sdsexamplewidth}
        \includegraphics[width=\linewidth,height=\sdsexampleheight,keepaspectratio]{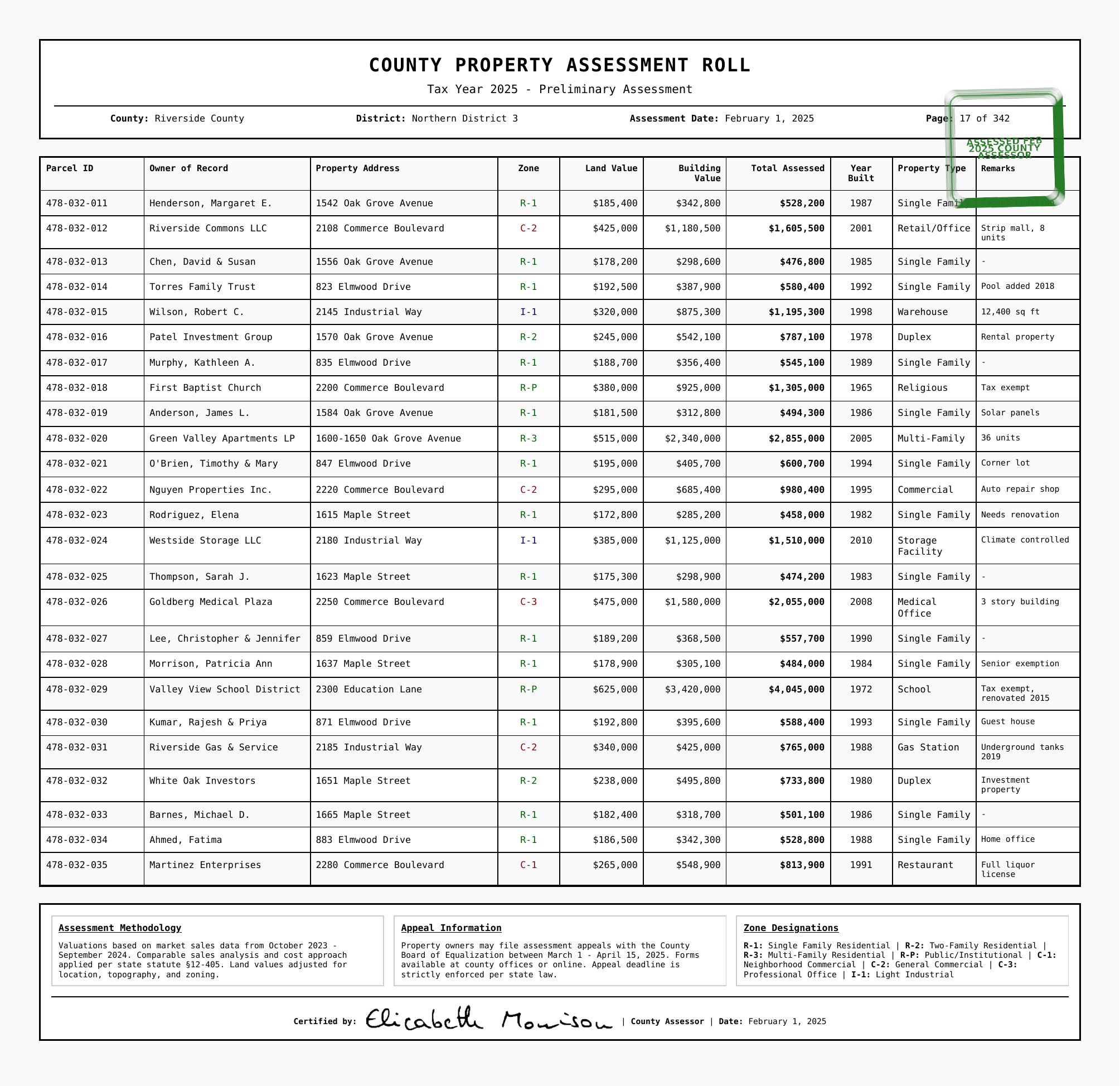}
    \end{subfigure}%
    \hfill
    \begin{subfigure}{\sdsexamplewidth}
        \includegraphics[width=\linewidth,height=\sdsexampleheight,keepaspectratio]{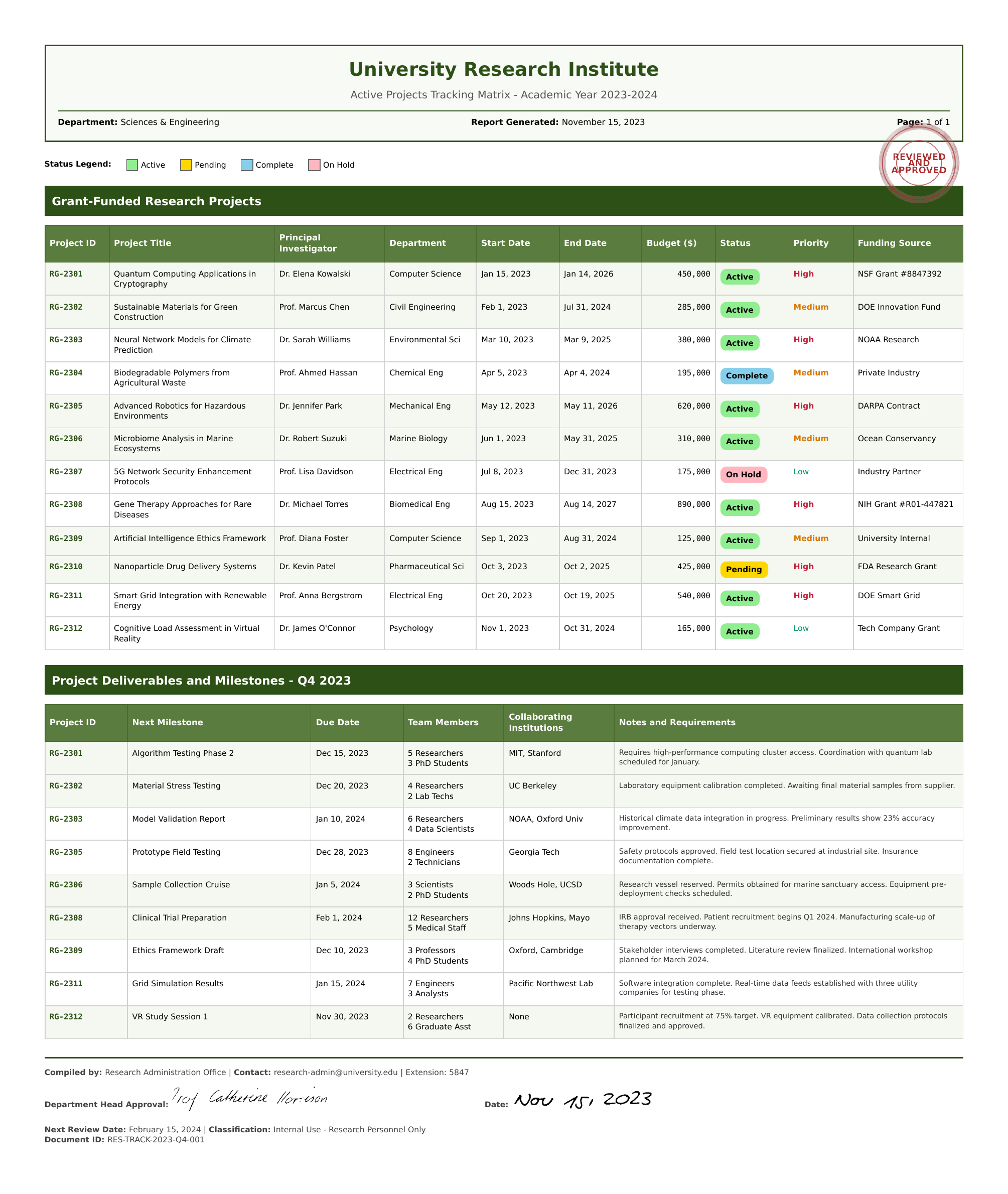}
    \end{subfigure}%
    \hfill
    \begin{subfigure}{\sdsexamplewidth}
        \includegraphics[width=\linewidth,height=\sdsexampleheight,keepaspectratio]{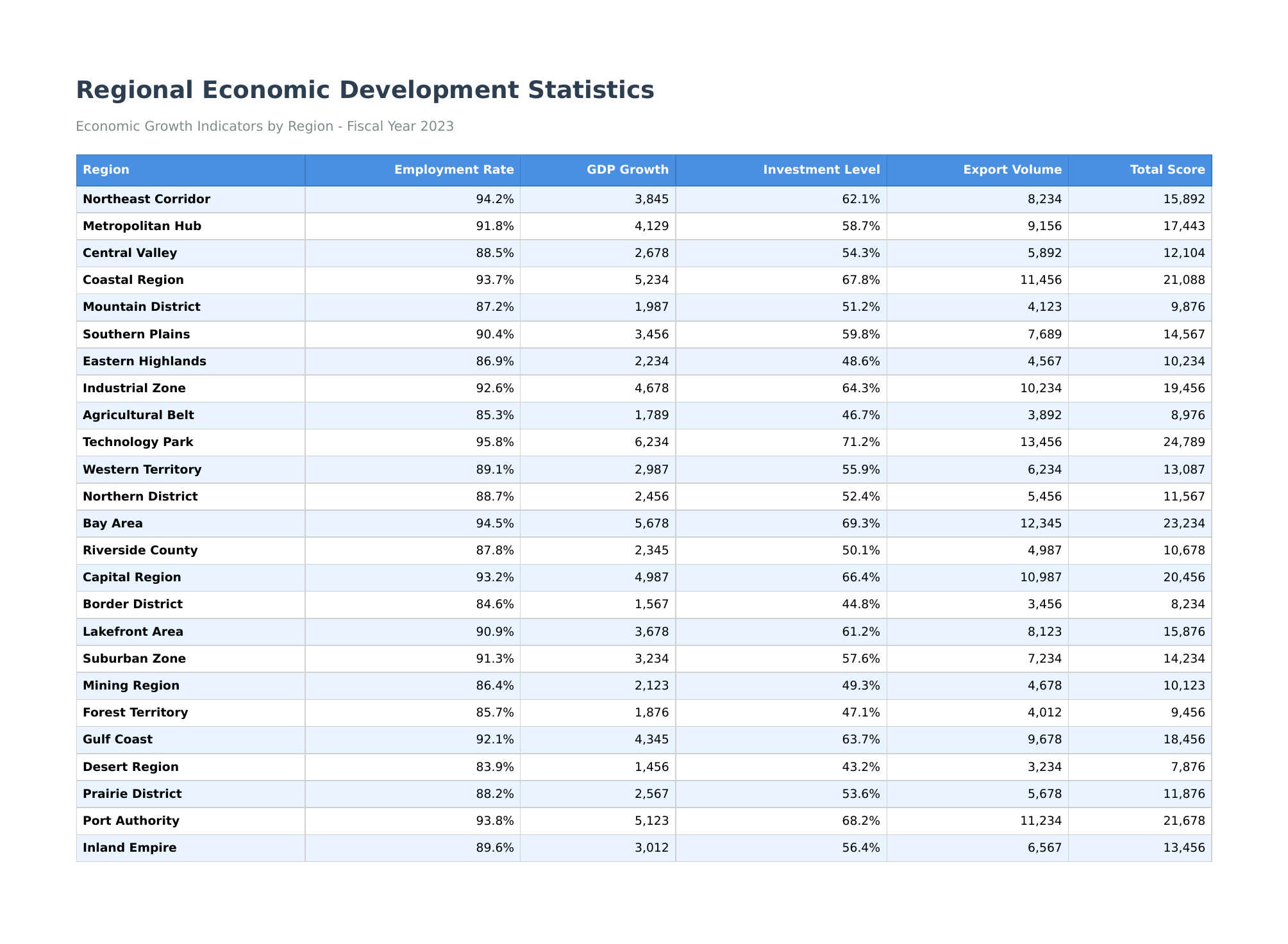}
    \end{subfigure}%
    \caption{\wiki{} synthetic dataset samples for VQA task.}\label{app:sds_ex_wtq}
\end{figure*}
\begin{figure*}[htbp]
    \centering
    \begin{subfigure}{\sdsexamplewidth}
        \includegraphics[width=\linewidth,height=\sdsexampleheight,keepaspectratio]{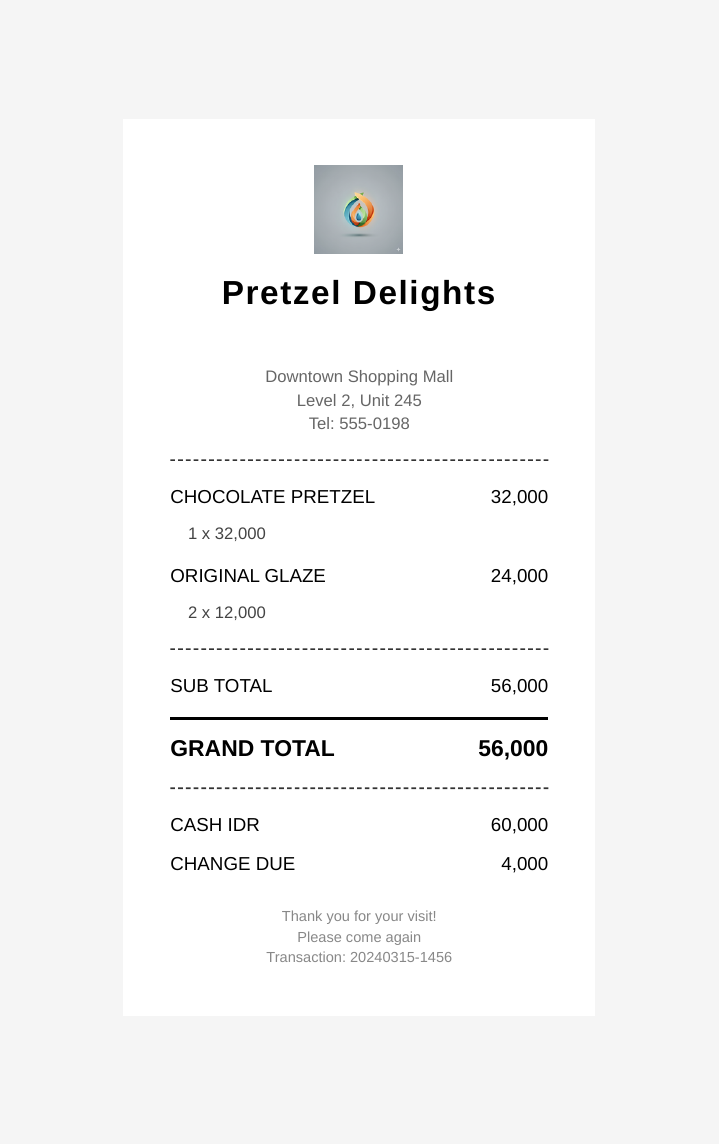}
    \end{subfigure}%
    \hfill
    \begin{subfigure}{\sdsexamplewidth}
        \includegraphics[width=\linewidth,height=\sdsexampleheight,keepaspectratio]{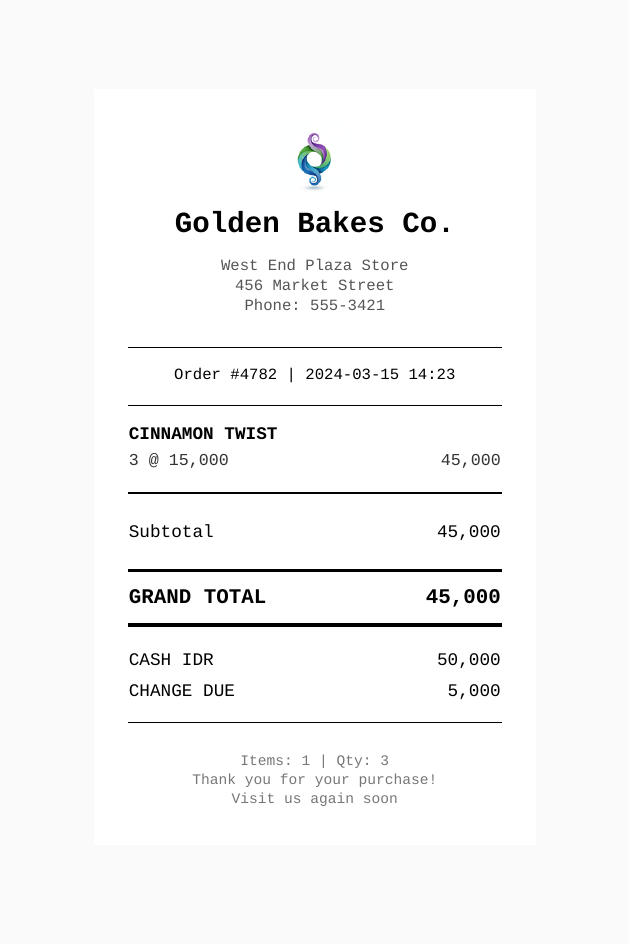}
    \end{subfigure}%
    \hfill
    \begin{subfigure}{\sdsexamplewidth}
        \includegraphics[width=\linewidth,height=\sdsexampleheight,keepaspectratio]{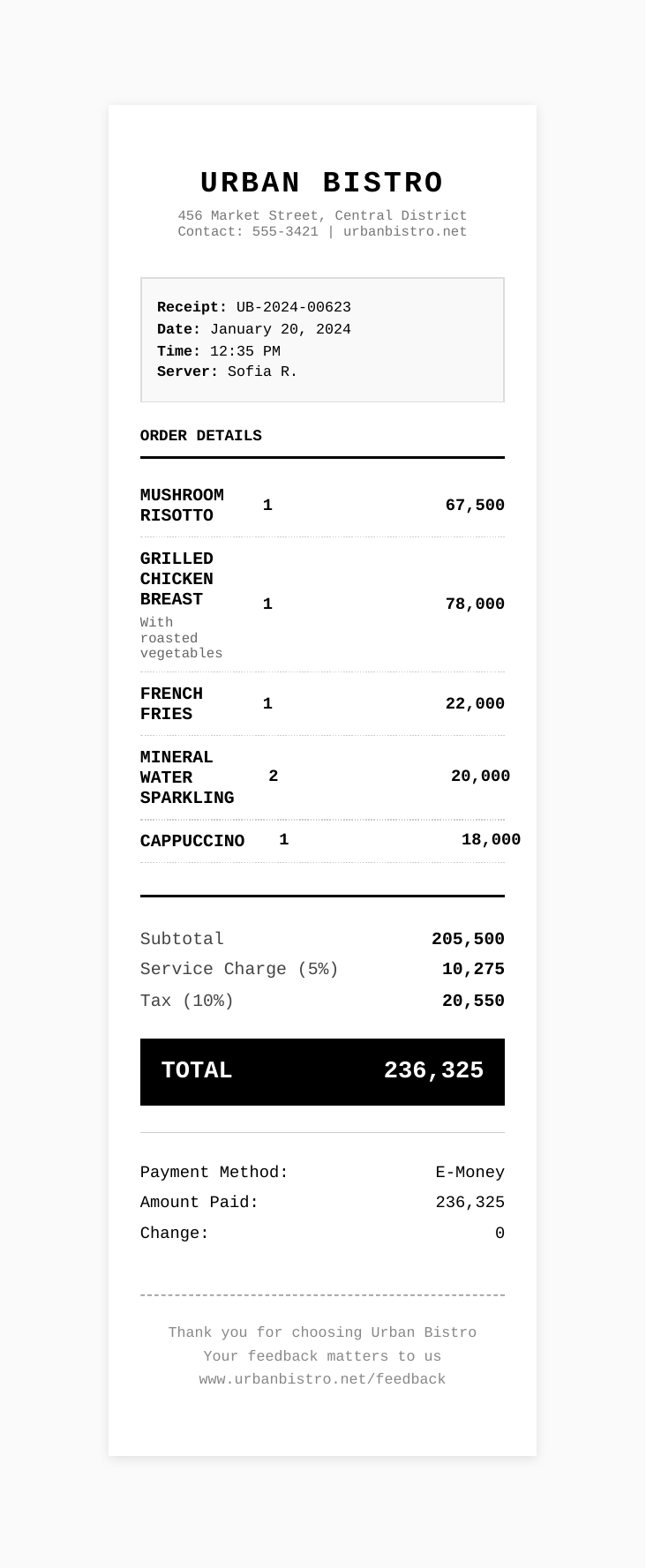}
    \end{subfigure}%
    \hfill
    \begin{subfigure}{\sdsexamplewidth}
        \includegraphics[width=\linewidth,height=\sdsexampleheight,keepaspectratio]{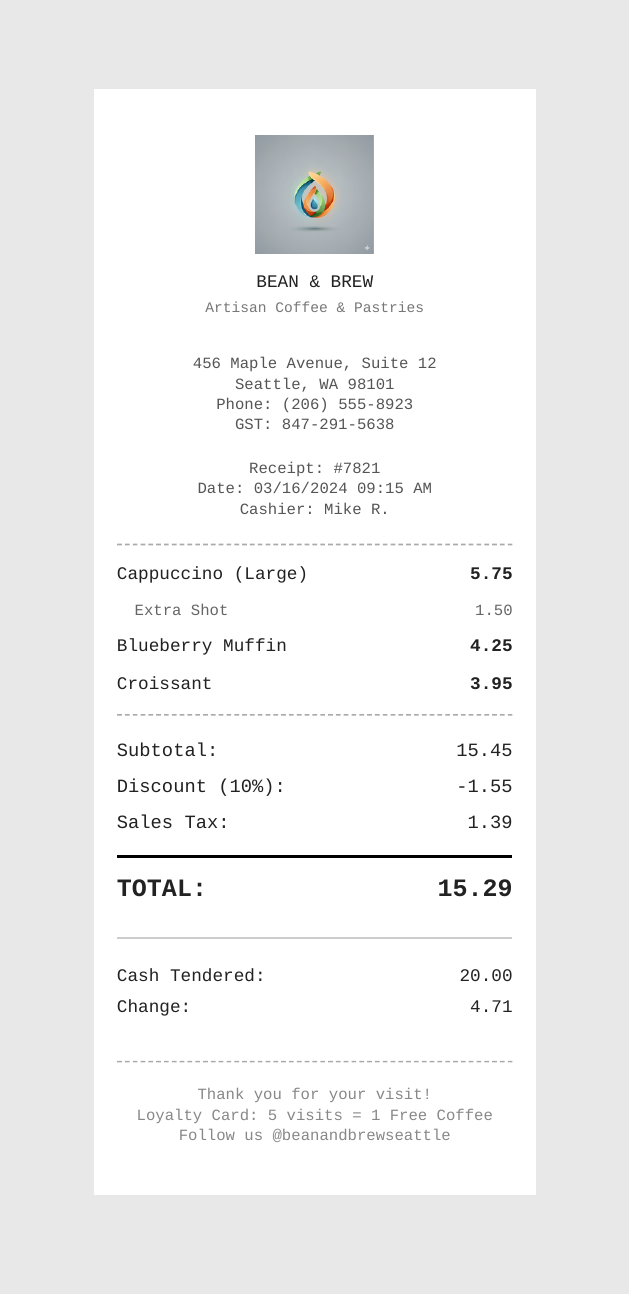}
    \end{subfigure}%
    \hfill
    \begin{subfigure}{\sdsexamplewidth}
        \includegraphics[width=\linewidth,height=\sdsexampleheight,keepaspectratio]{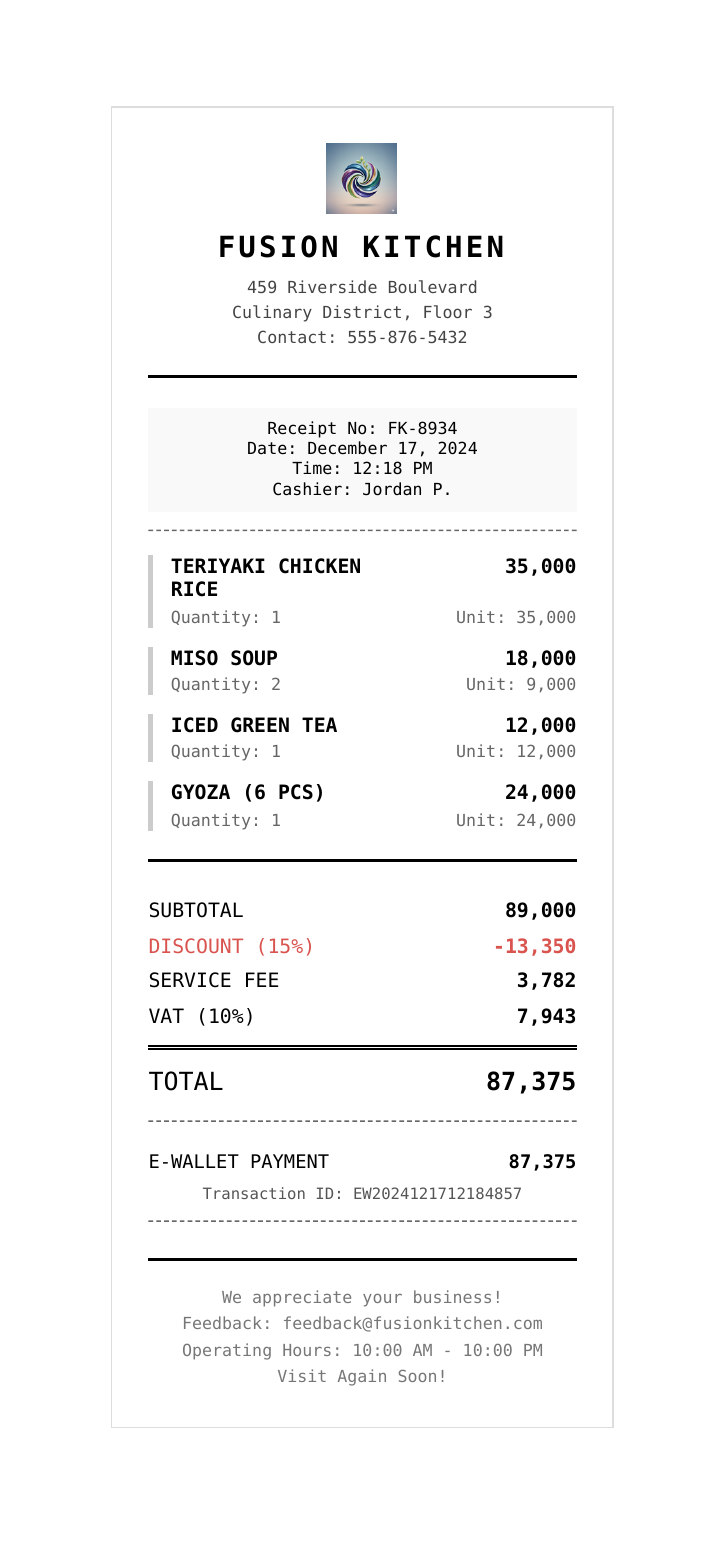}
    \end{subfigure}%
    \caption{\cord{} synthetic dataset samples for KIE task.}\label{app:sds_ex_cord}
\end{figure*}
\begin{figure*}[htbp]
    \centering
    \begin{subfigure}{\sdsexamplewidth}
        \includegraphics[width=\linewidth,height=\sdsexampleheight,keepaspectratio]{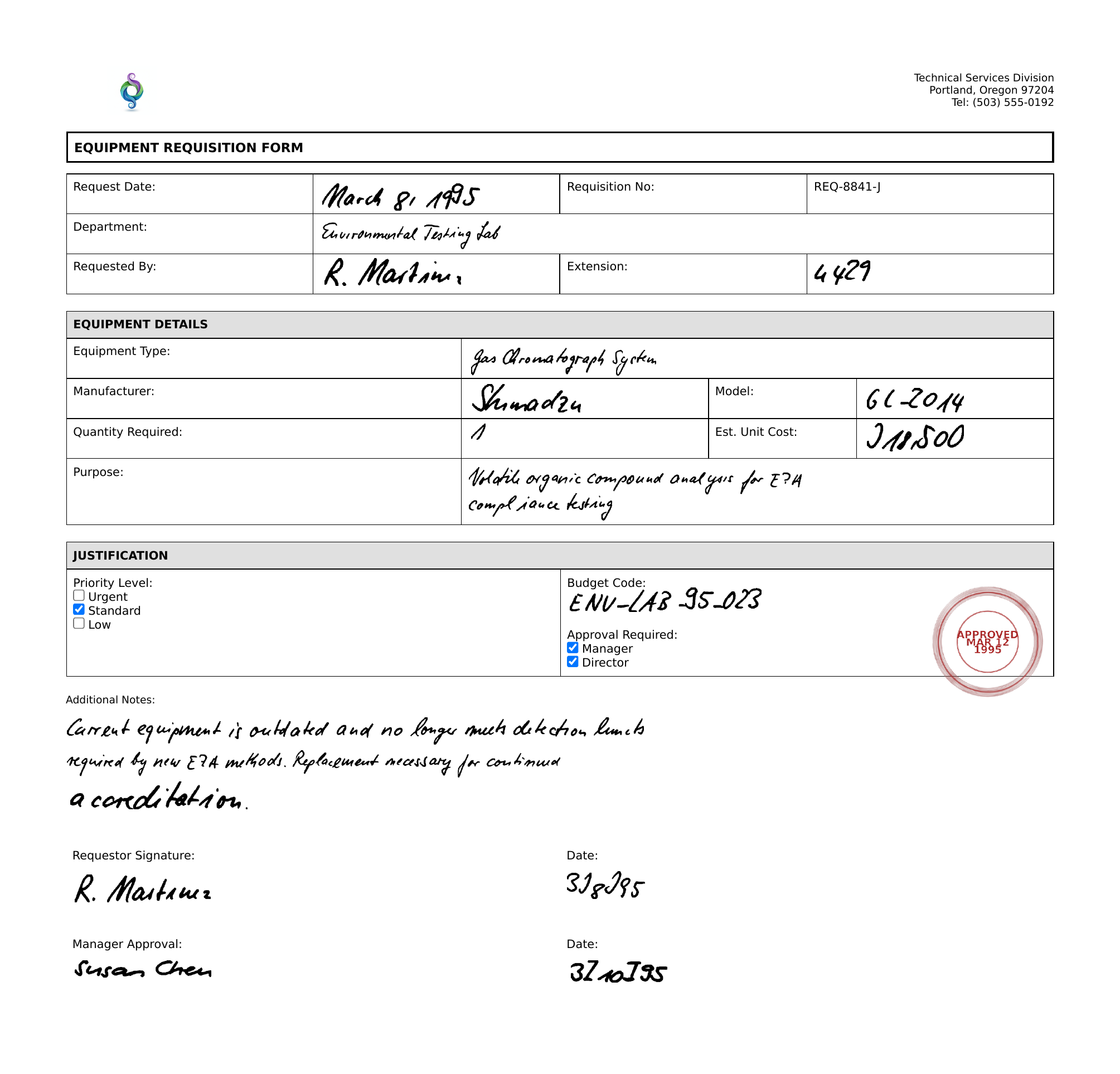}
    \end{subfigure}%
    \hfill
    \begin{subfigure}{\sdsexamplewidth}
        \includegraphics[width=\linewidth,height=\sdsexampleheight,keepaspectratio]{images/synthetic_dataset_images/FUNSD/sample_2.pdf}
    \end{subfigure}%
    \hfill
    \begin{subfigure}{\sdsexamplewidth}
        \includegraphics[width=\linewidth,height=\sdsexampleheight,keepaspectratio]{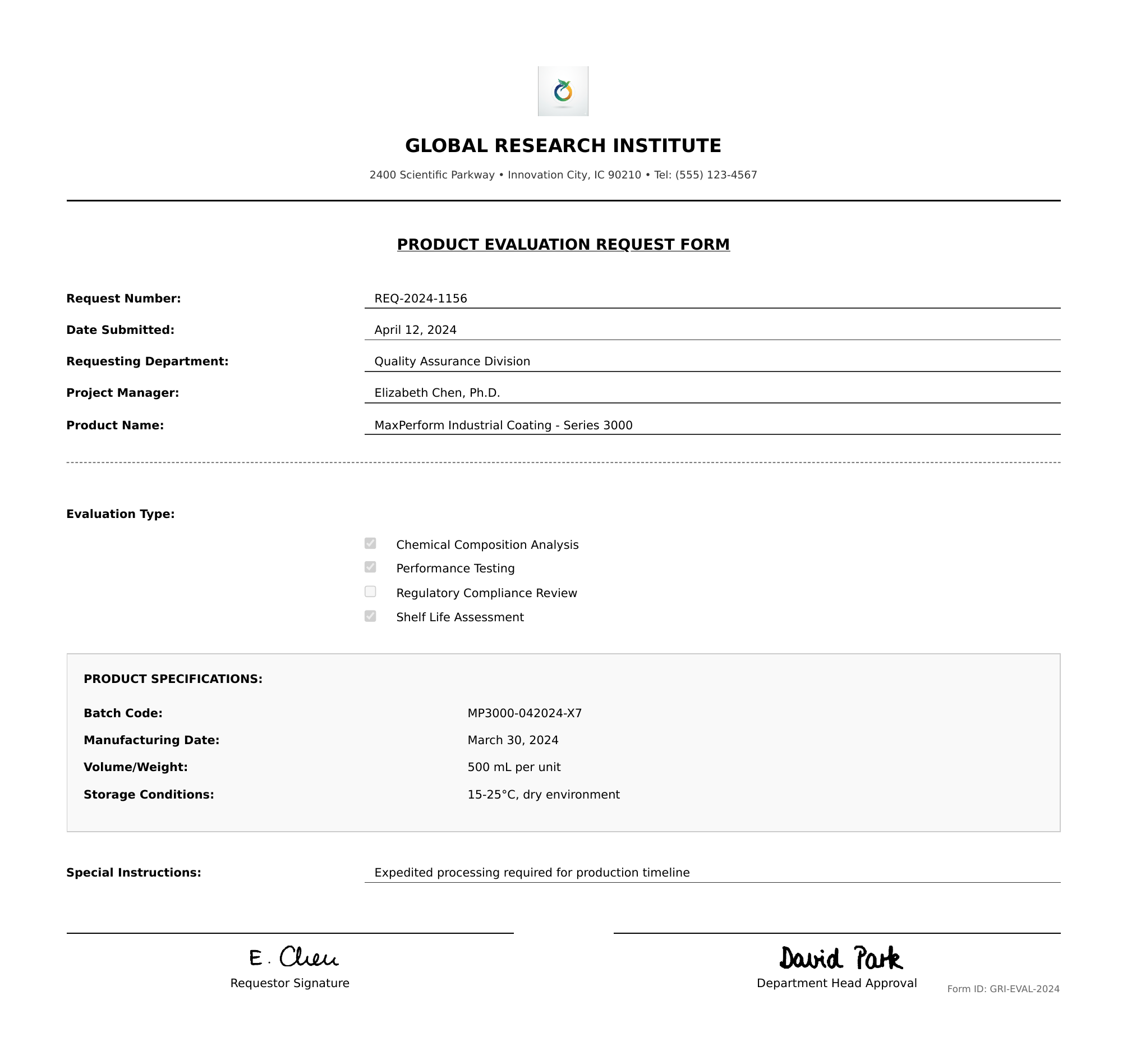}
    \end{subfigure}%
    \hfill
    \begin{subfigure}{\sdsexamplewidth}
        \includegraphics[width=\linewidth,height=\sdsexampleheight,keepaspectratio]{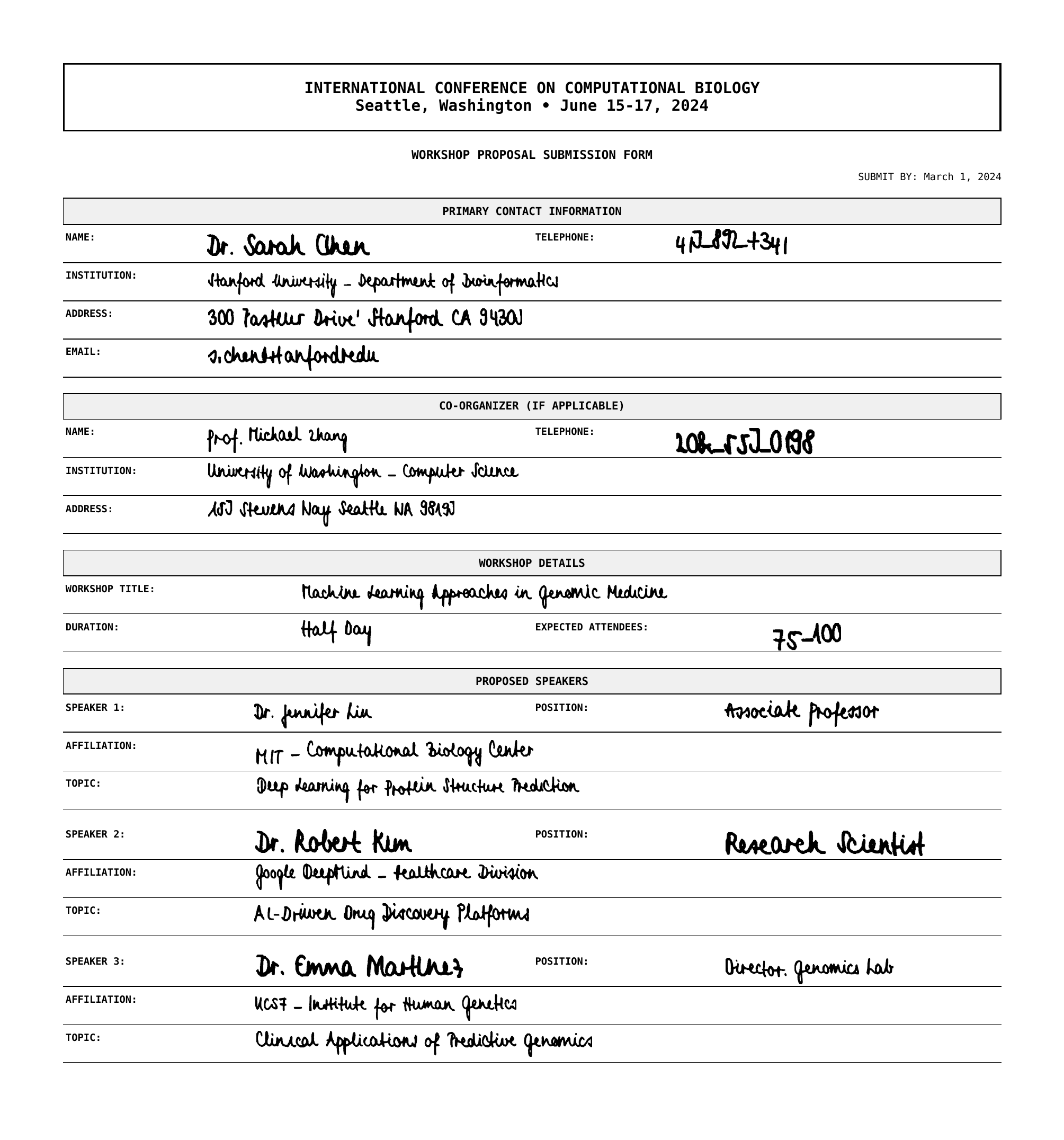}
    \end{subfigure}%
    \hfill
    \begin{subfigure}{\sdsexamplewidth}
        \includegraphics[width=\linewidth,height=\sdsexampleheight,keepaspectratio]{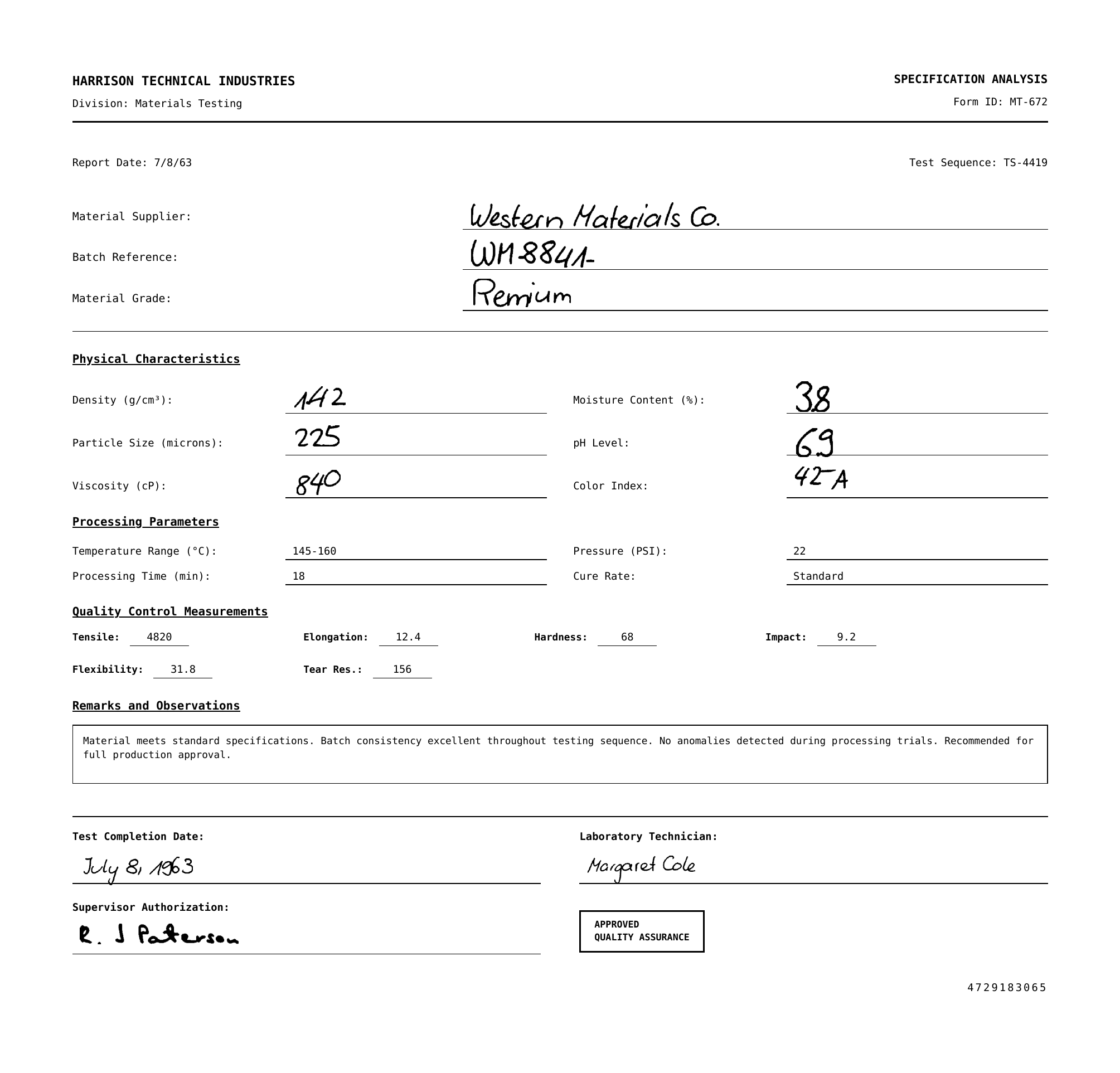}
    \end{subfigure}%
    \caption{\funsd{} synthetic dataset samples for KIE task.}\label{app:sds_ex_funsd}
\end{figure*}
\begin{figure*}[htbp]
    \centering
    \begin{subfigure}{\sdsexamplewidth}
        \includegraphics[width=\linewidth,height=\sdsexampleheight,keepaspectratio]{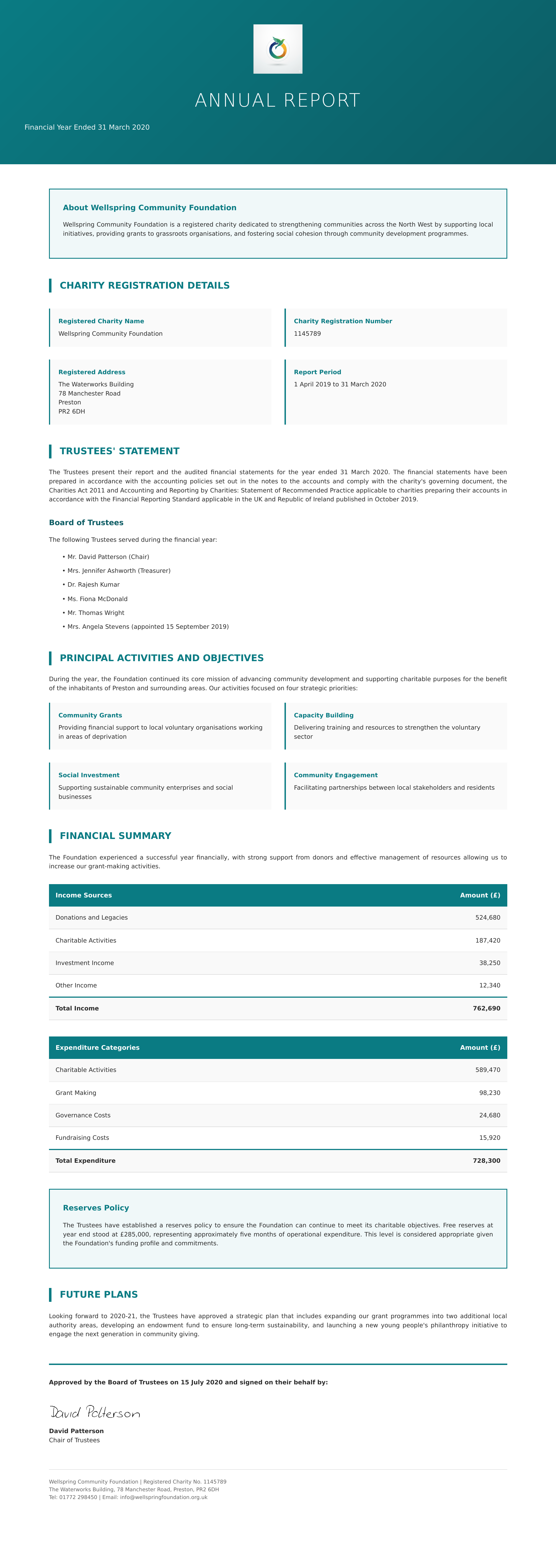}
    \end{subfigure}%
    \hfill
    \begin{subfigure}{\sdsexamplewidth}
        \includegraphics[width=\linewidth,height=\sdsexampleheight,keepaspectratio]{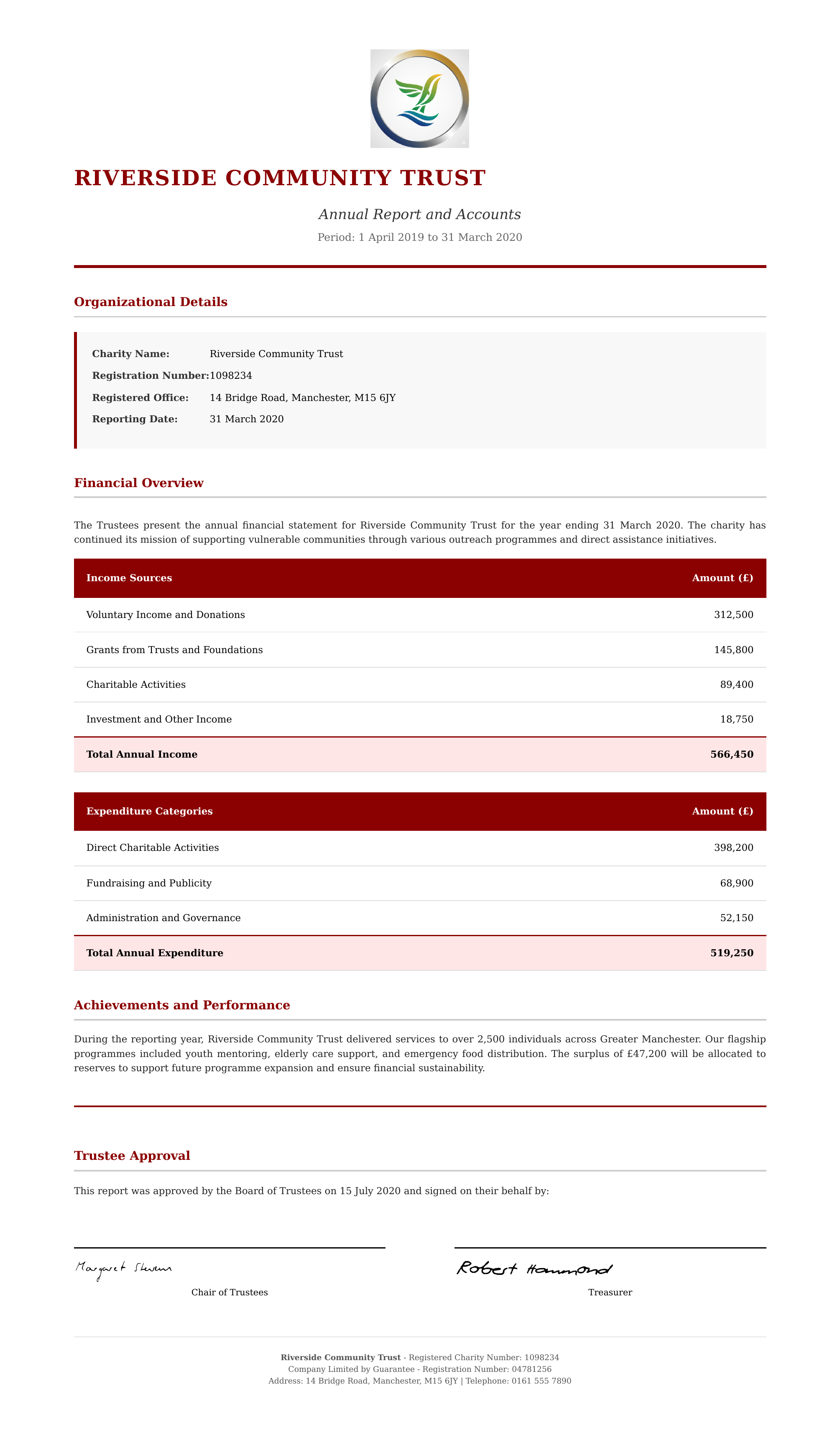}
    \end{subfigure}%
    \hfill
    \begin{subfigure}{\sdsexamplewidth}
        \includegraphics[width=\linewidth,height=\sdsexampleheight,keepaspectratio]{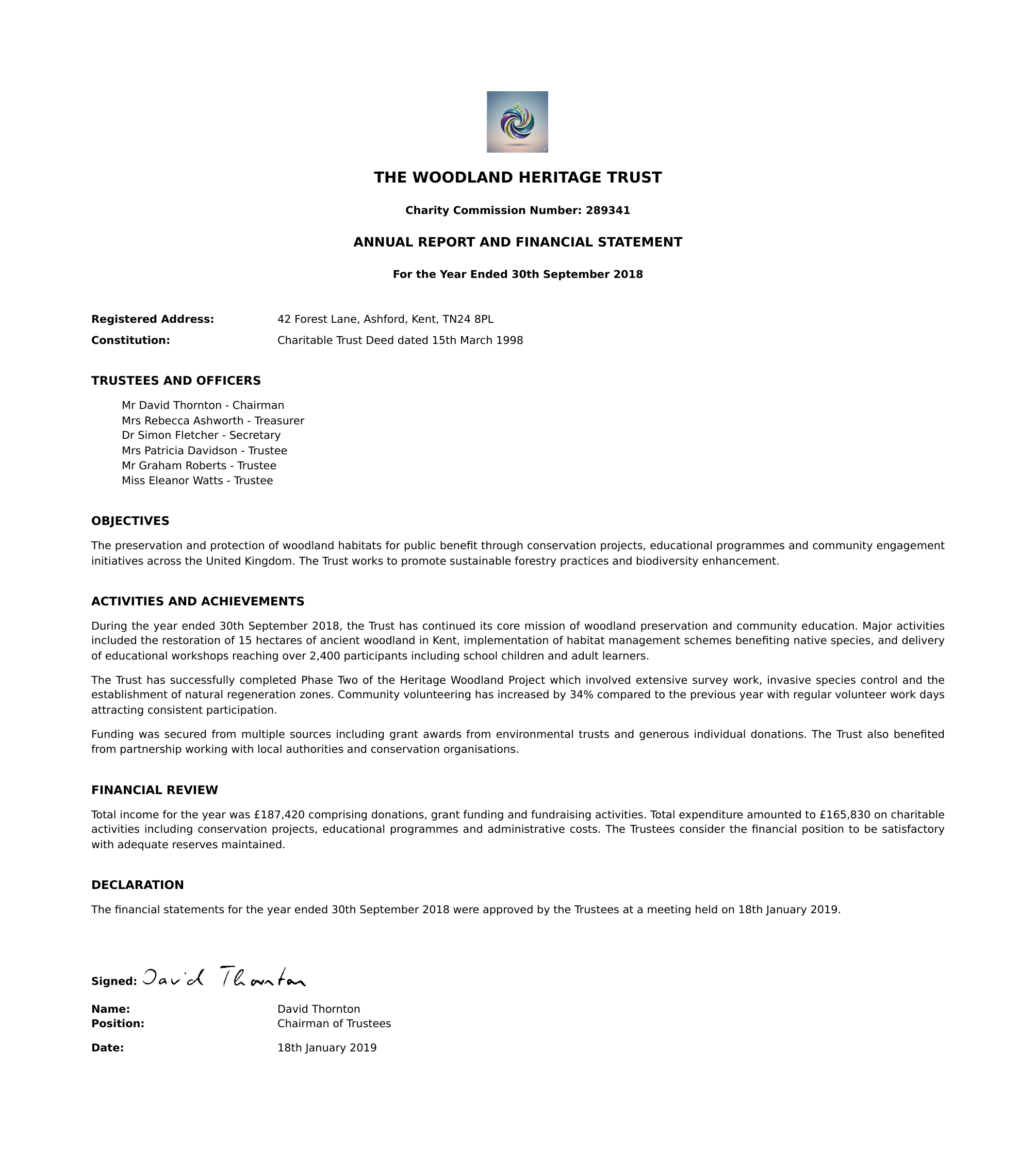}
    \end{subfigure}%
    \hfill
    \begin{subfigure}{\sdsexamplewidth}
        \includegraphics[width=\linewidth,height=\sdsexampleheight,keepaspectratio]{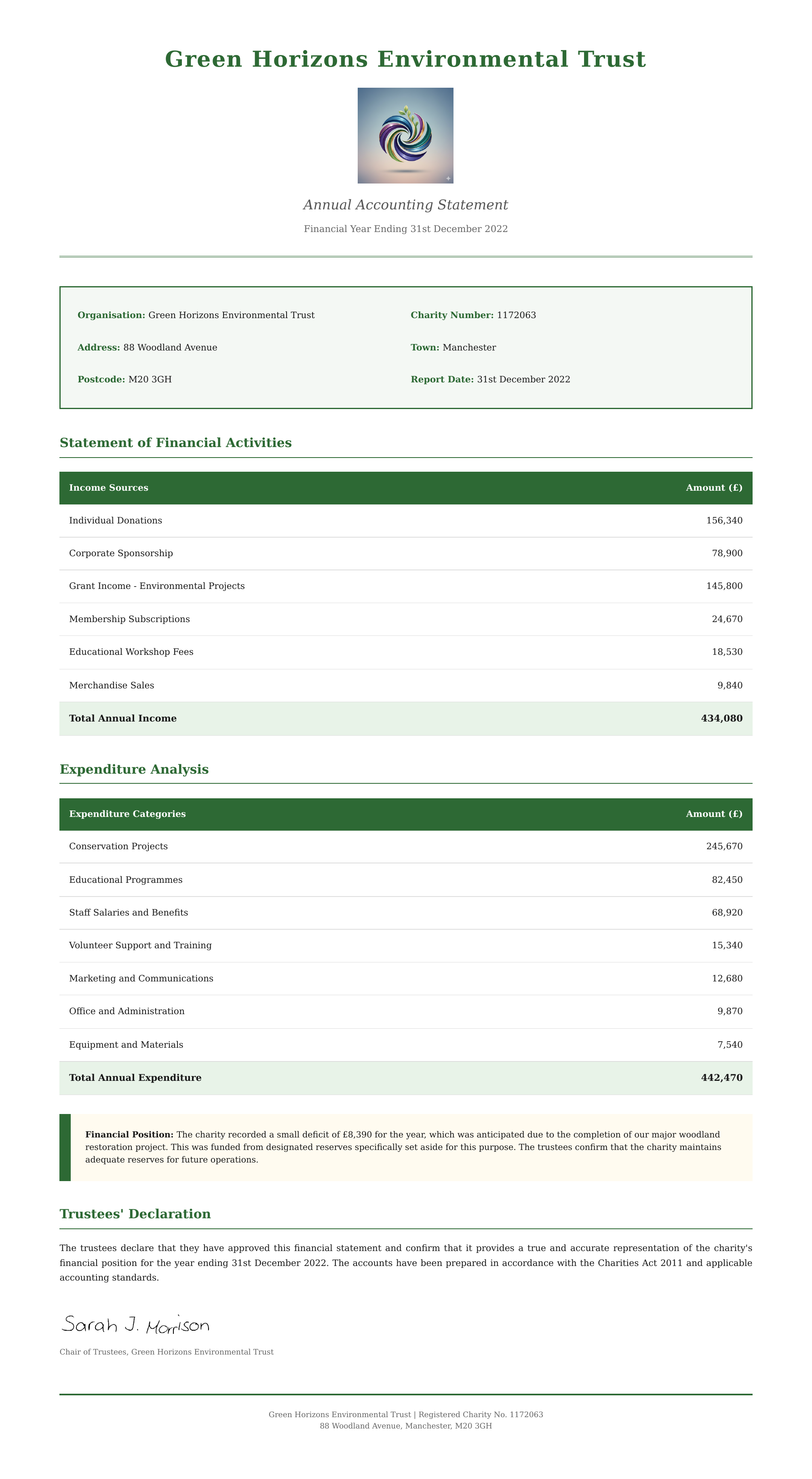}
    \end{subfigure}%
    \hfill
    \begin{subfigure}{\sdsexamplewidth}
        \includegraphics[width=\linewidth,height=\sdsexampleheight,keepaspectratio]{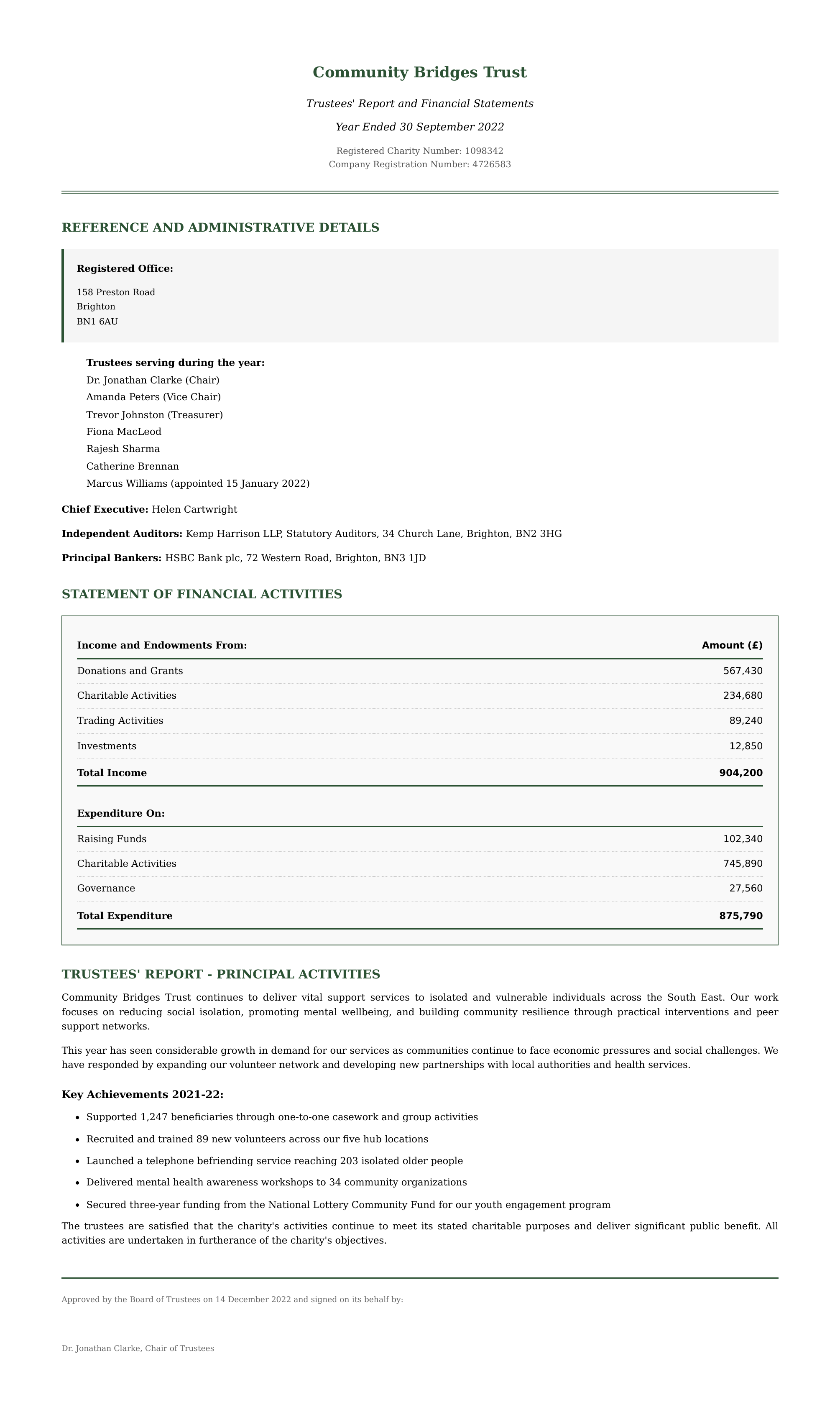}
    \end{subfigure}%
    \caption{\klc{} synthetic dataset samples for VQA task.}\label{app:sds_ex_klc}
\end{figure*}
\begin{figure*}[htbp]
    \centering
    \begin{subfigure}{\sdsexamplewidth}
        \includegraphics[width=\linewidth,height=\sdsexampleheight,keepaspectratio]{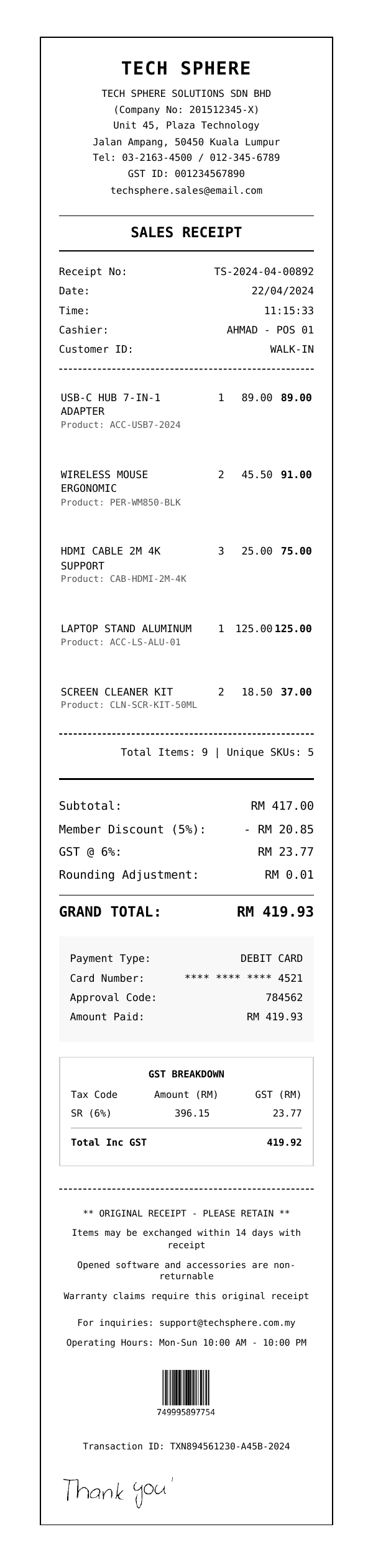}
    \end{subfigure}%
    \hfill
    \begin{subfigure}{\sdsexamplewidth}
        \includegraphics[width=\linewidth,height=\sdsexampleheight,keepaspectratio]{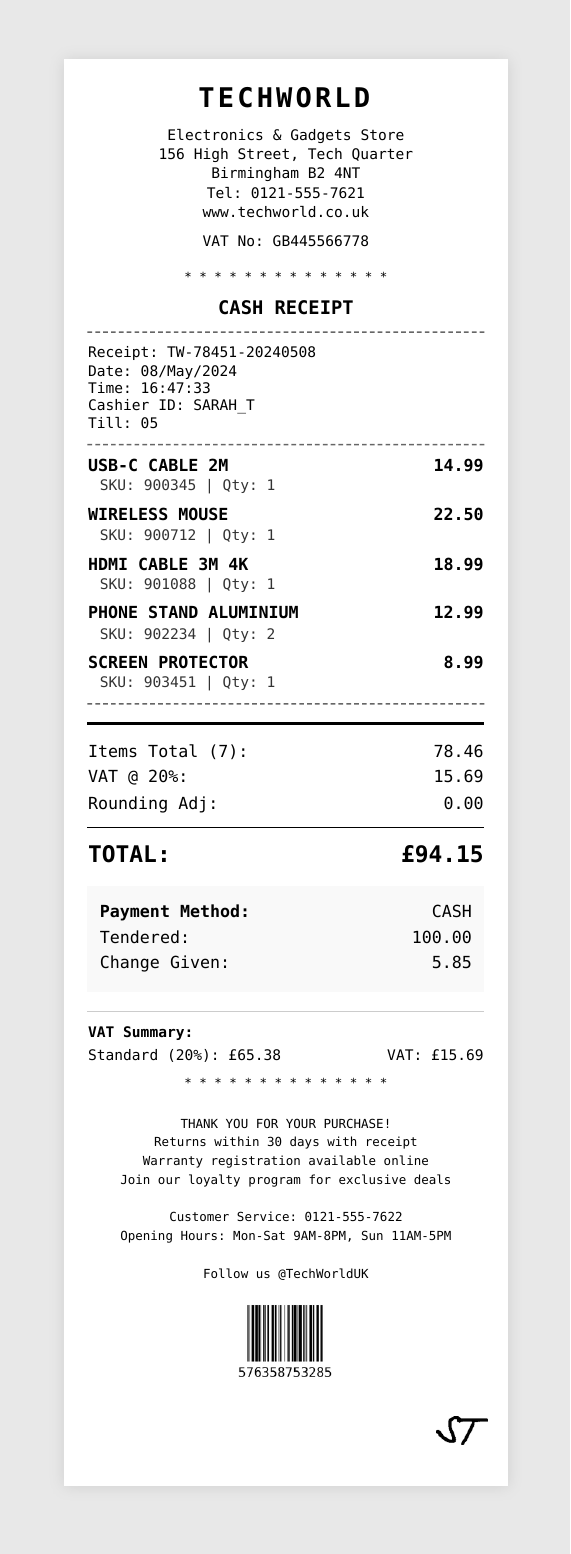}
    \end{subfigure}%
    \hfill
    \begin{subfigure}{\sdsexamplewidth}
        \includegraphics[width=\linewidth,height=\sdsexampleheight,keepaspectratio]{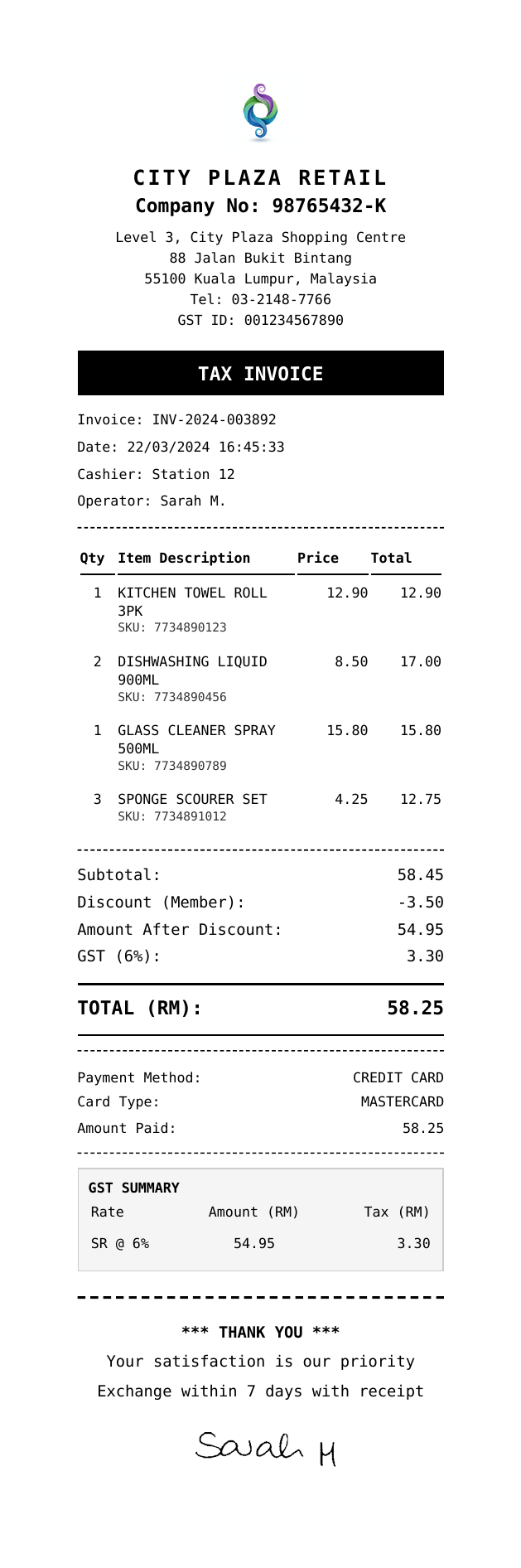}
    \end{subfigure}%
    \hfill
    \begin{subfigure}{\sdsexamplewidth}
        \includegraphics[width=\linewidth,height=\sdsexampleheight,keepaspectratio]{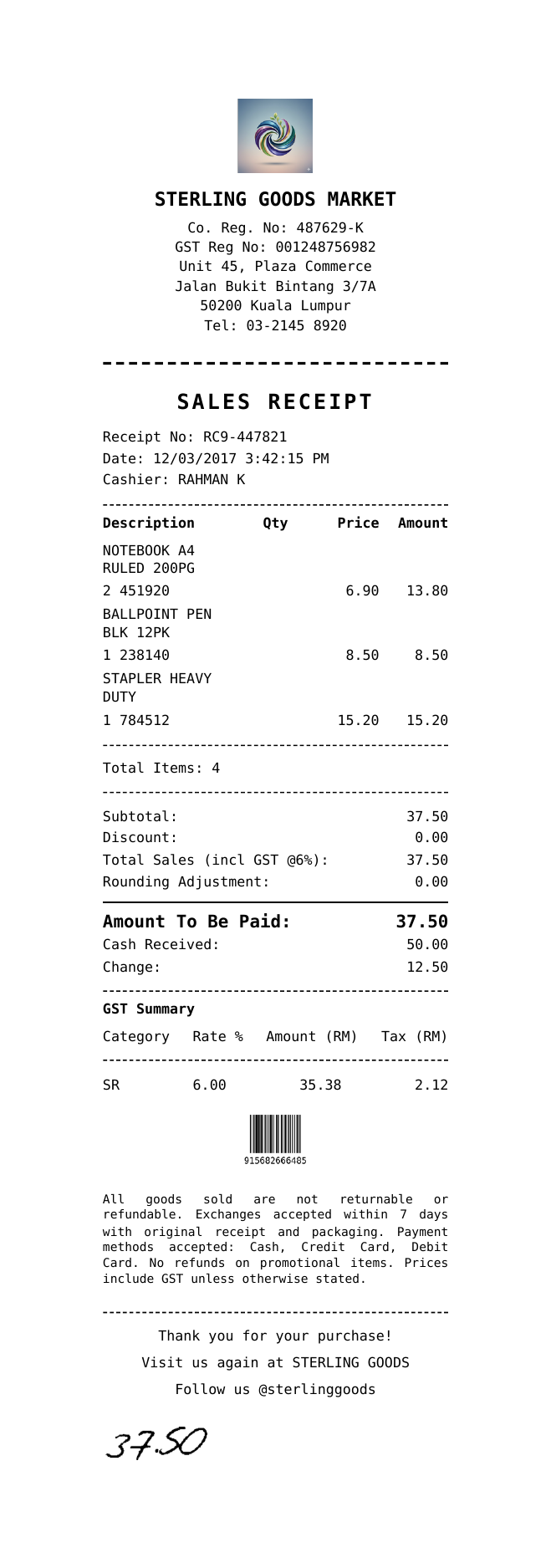}
    \end{subfigure}%
    \hfill
    \begin{subfigure}{\sdsexamplewidth}
        \includegraphics[width=\linewidth,height=\sdsexampleheight,keepaspectratio]{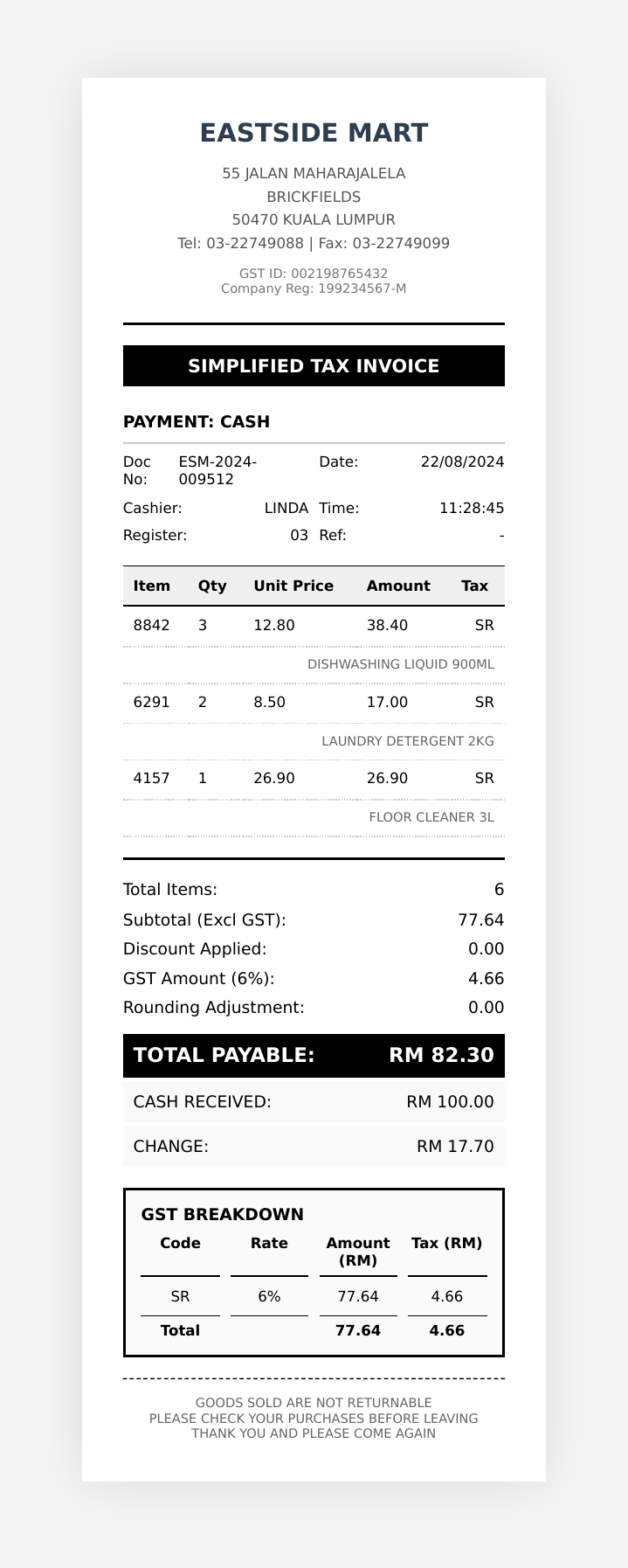}
    \end{subfigure}%
    \caption{\sroie{} syntheitc dataset samples for KIE task.}\label{app:sds_ex_sroie}
\end{figure*}
\begin{figure*}[htbp]
    \centering
    \begin{subfigure}{\sdsexamplewidth}
        \includegraphics[width=\linewidth,height=\sdsexampleheight,keepaspectratio]{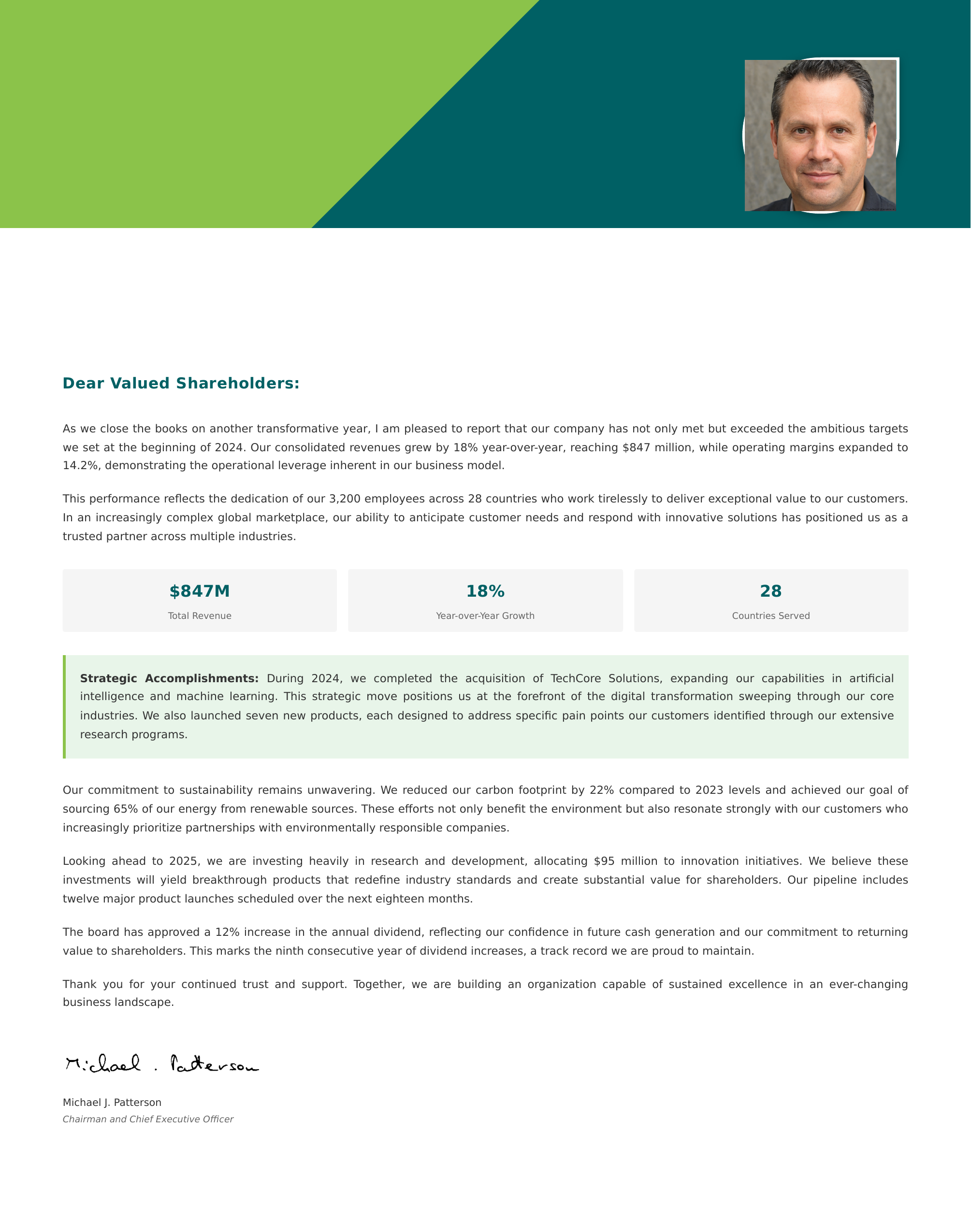}
    \end{subfigure}%
    \hfill
    \begin{subfigure}{\sdsexamplewidth}
        \includegraphics[width=\linewidth,height=\sdsexampleheight,keepaspectratio]{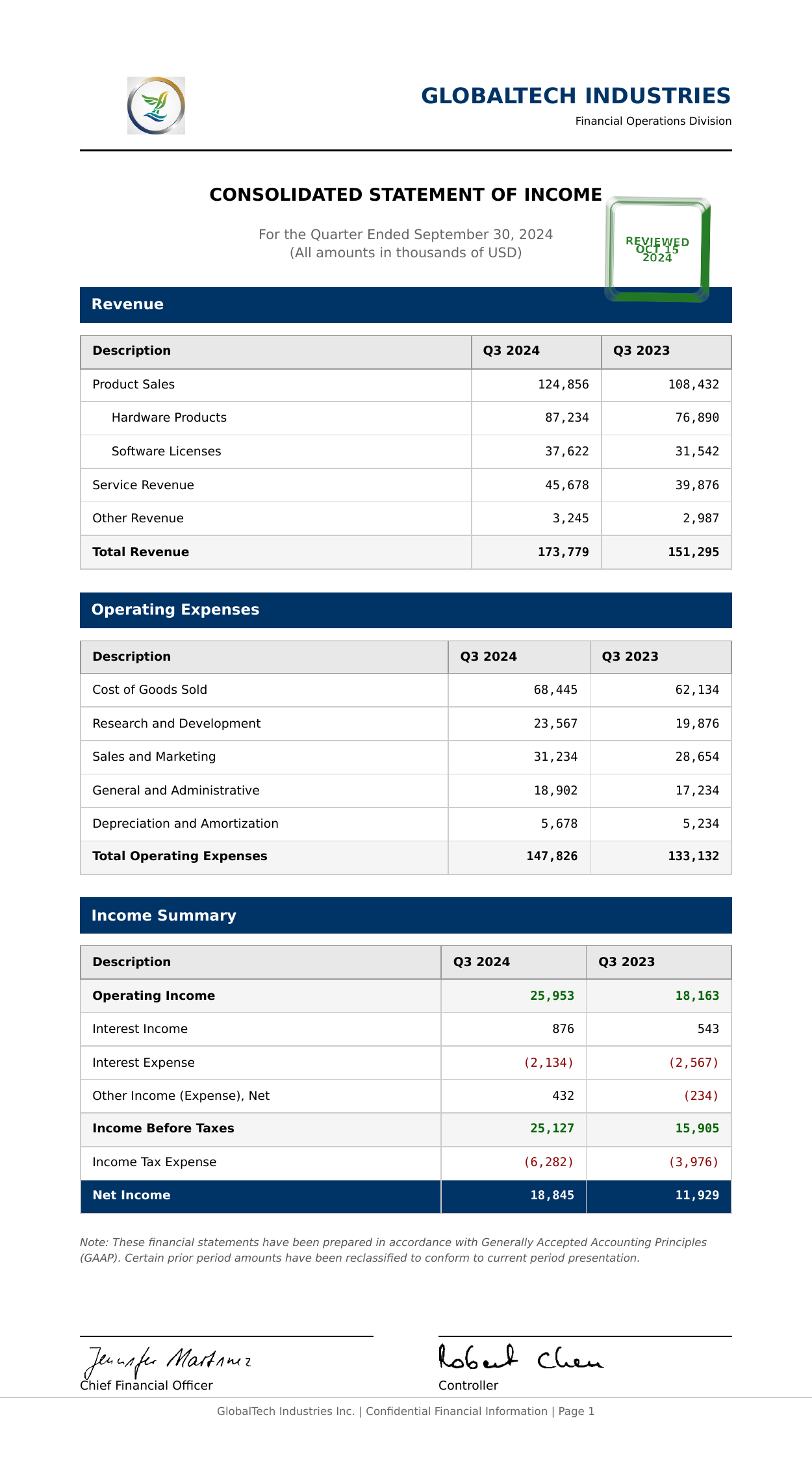}
    \end{subfigure}%
    \hfill
    \begin{subfigure}{\sdsexamplewidth}
        \includegraphics[width=\linewidth,height=\sdsexampleheight,keepaspectratio]{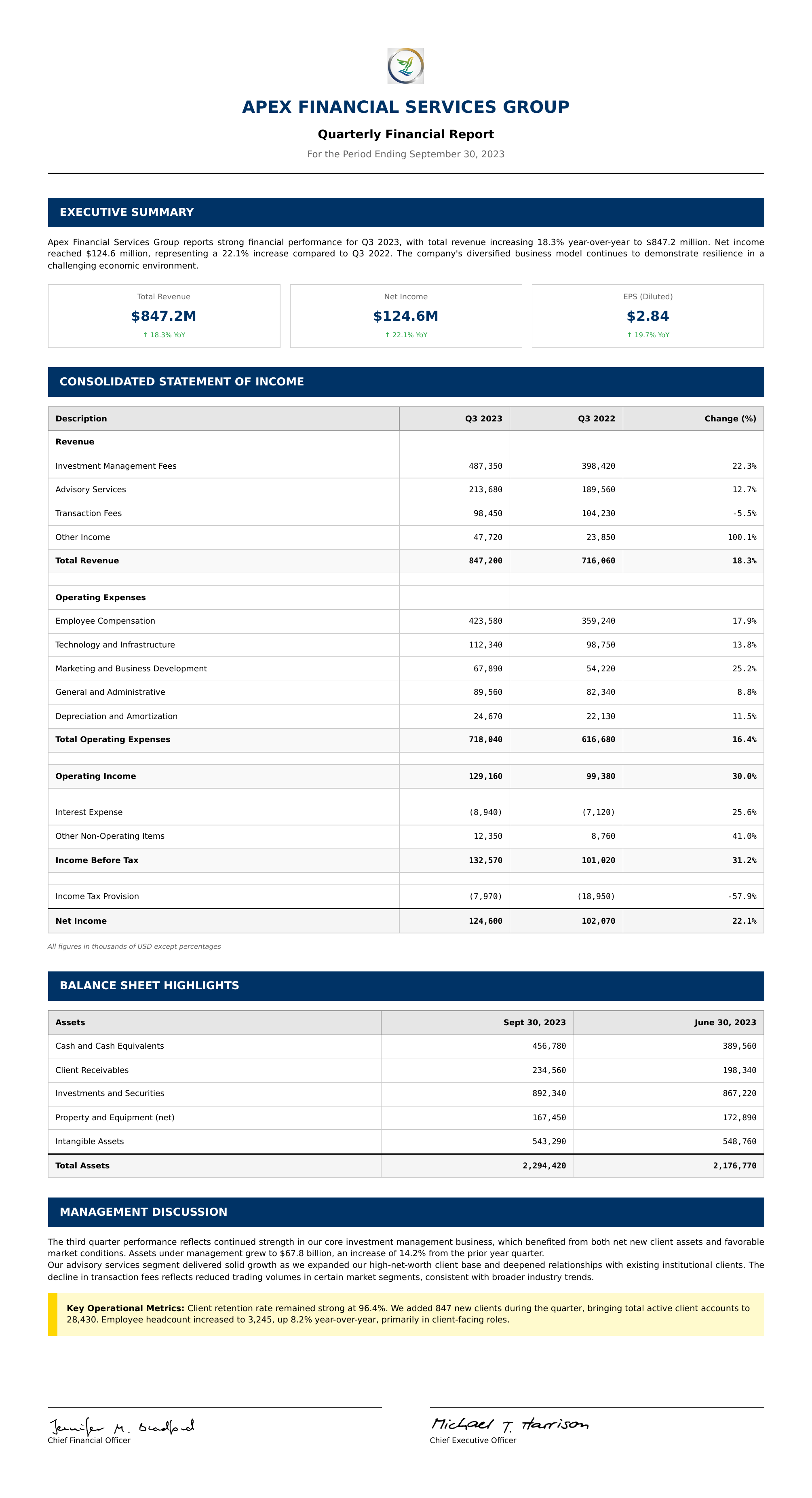}
    \end{subfigure}%
    \hfill
    \begin{subfigure}{\sdsexamplewidth}
        \includegraphics[width=\linewidth,height=\sdsexampleheight,keepaspectratio]{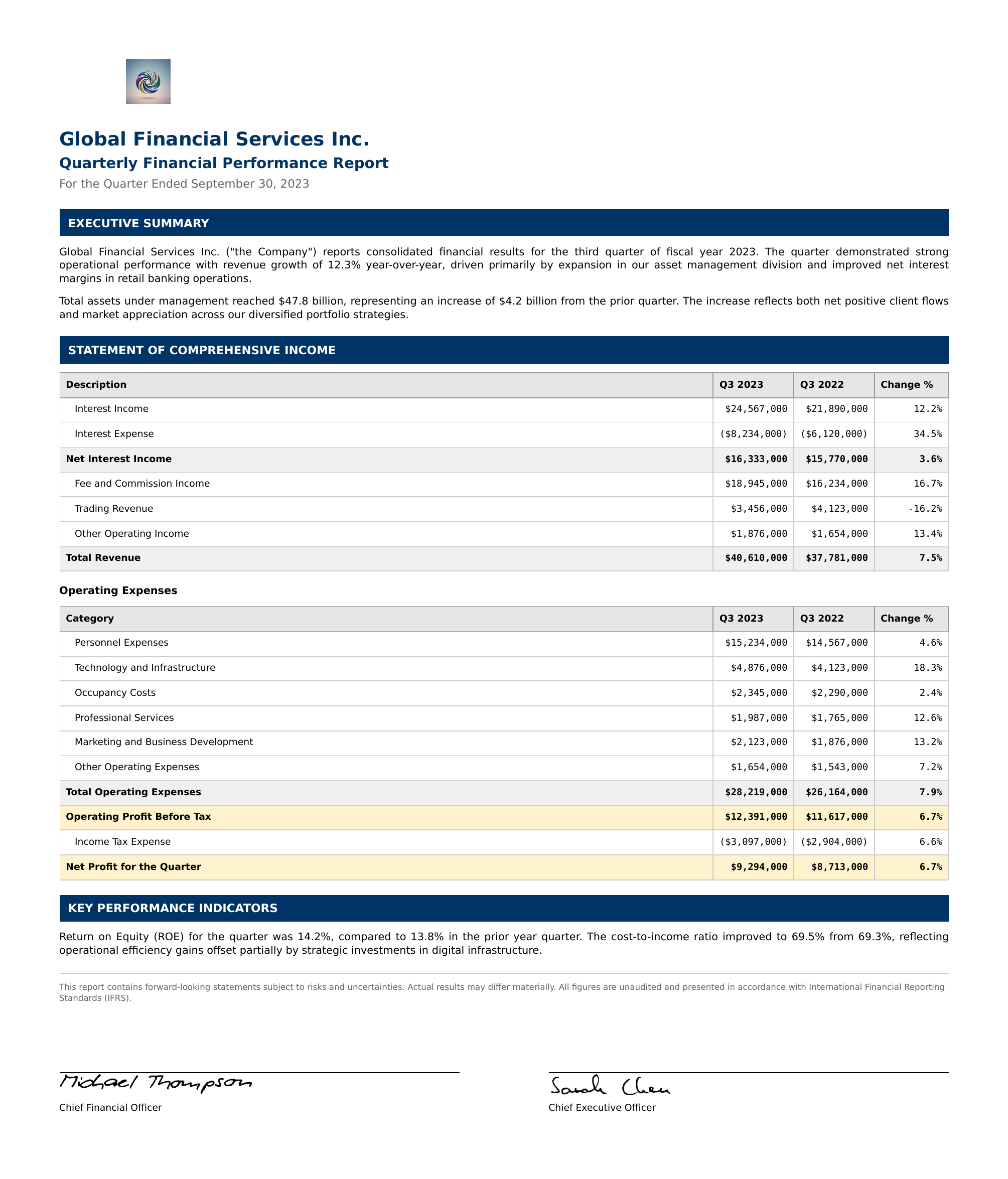}
    \end{subfigure}%
    \hfill
    \begin{subfigure}{\sdsexamplewidth}
        \includegraphics[width=\linewidth,height=\sdsexampleheight,keepaspectratio]{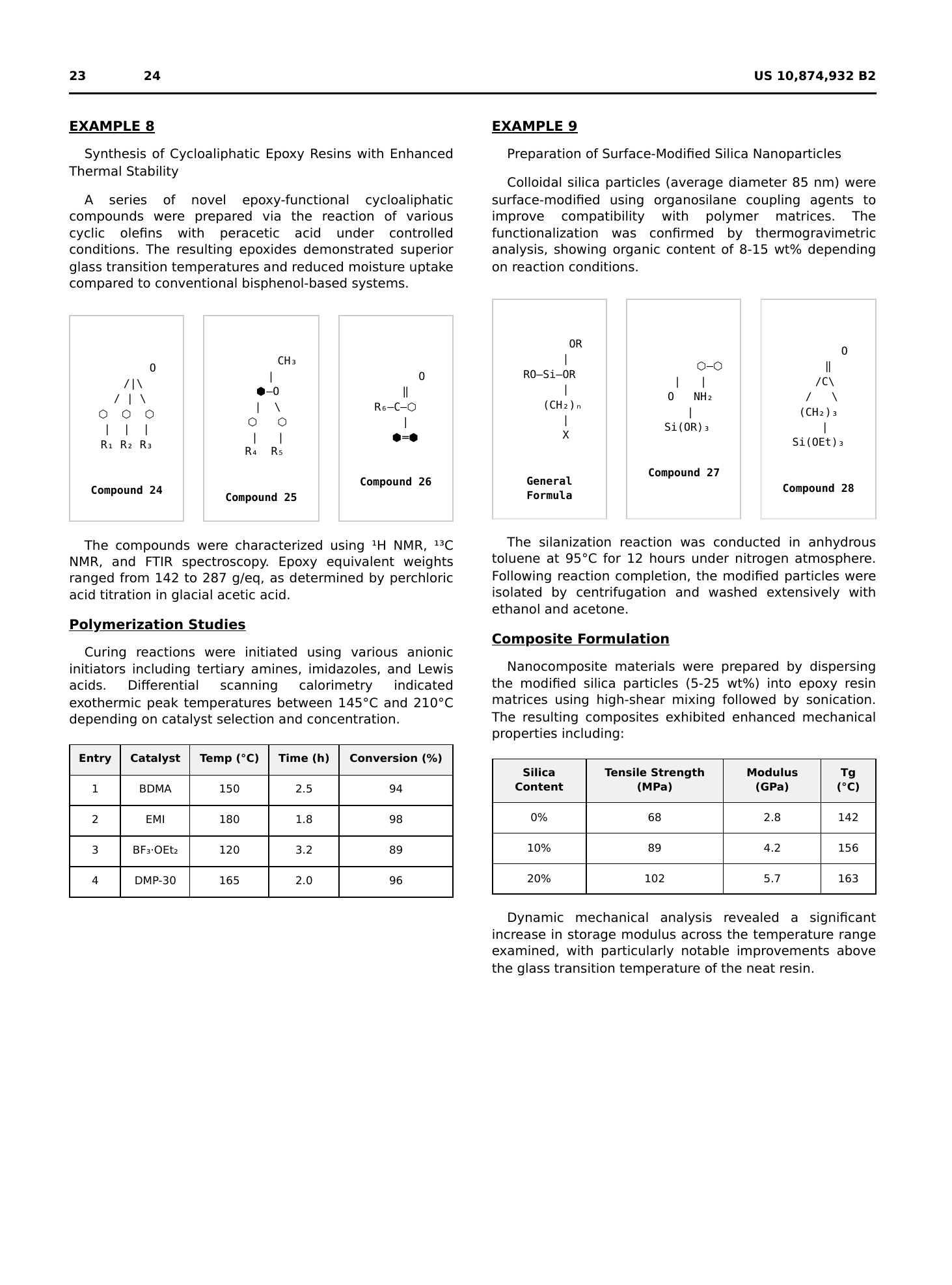}
    \end{subfigure}%
    \caption{\doclaynetcls{} synthetic dataset samples for CLS task.}\label{app:sds_ex_doclaynetcls}
\end{figure*}
\begin{figure*}[htbp]
    \centering
    \begin{subfigure}{\sdsexamplewidth}
        \includegraphics[width=\linewidth,height=\sdsexampleheight,keepaspectratio]{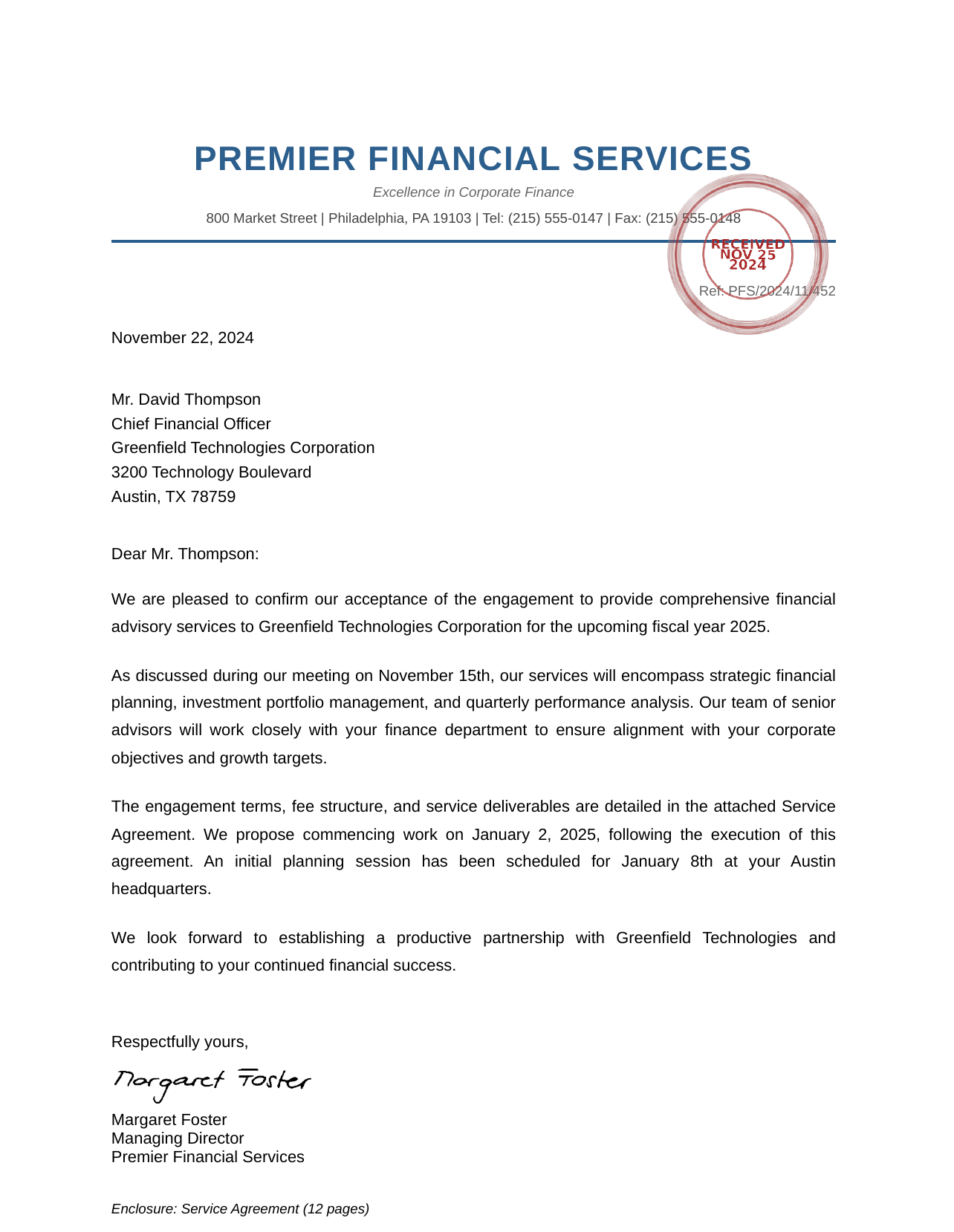}
    \end{subfigure}%
    \hfill
    \begin{subfigure}{\sdsexamplewidth}
        \includegraphics[width=\linewidth,height=\sdsexampleheight,keepaspectratio]{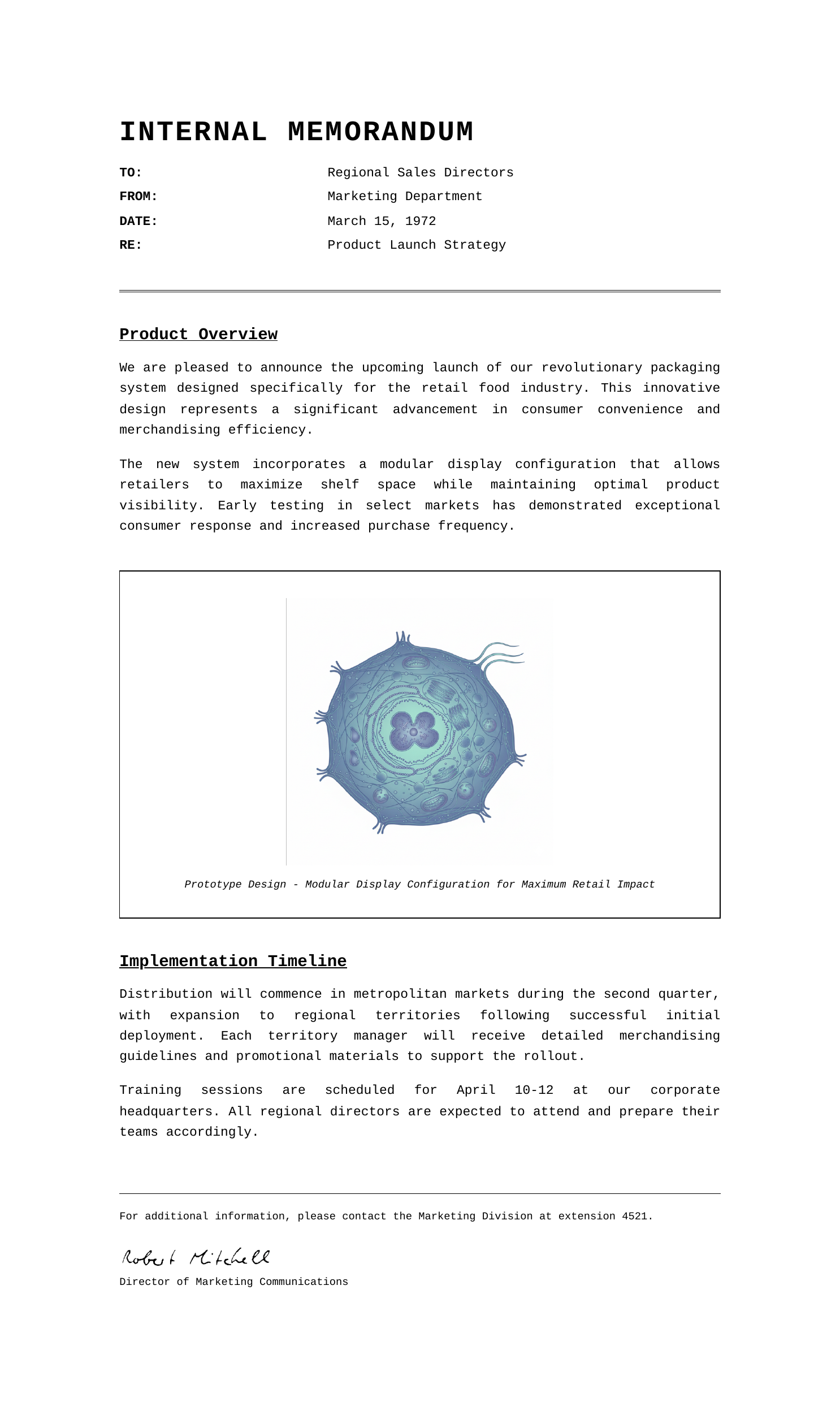}
    \end{subfigure}%
    \hfill
    \begin{subfigure}{\sdsexamplewidth}
        \includegraphics[width=\linewidth,height=\sdsexampleheight,keepaspectratio]{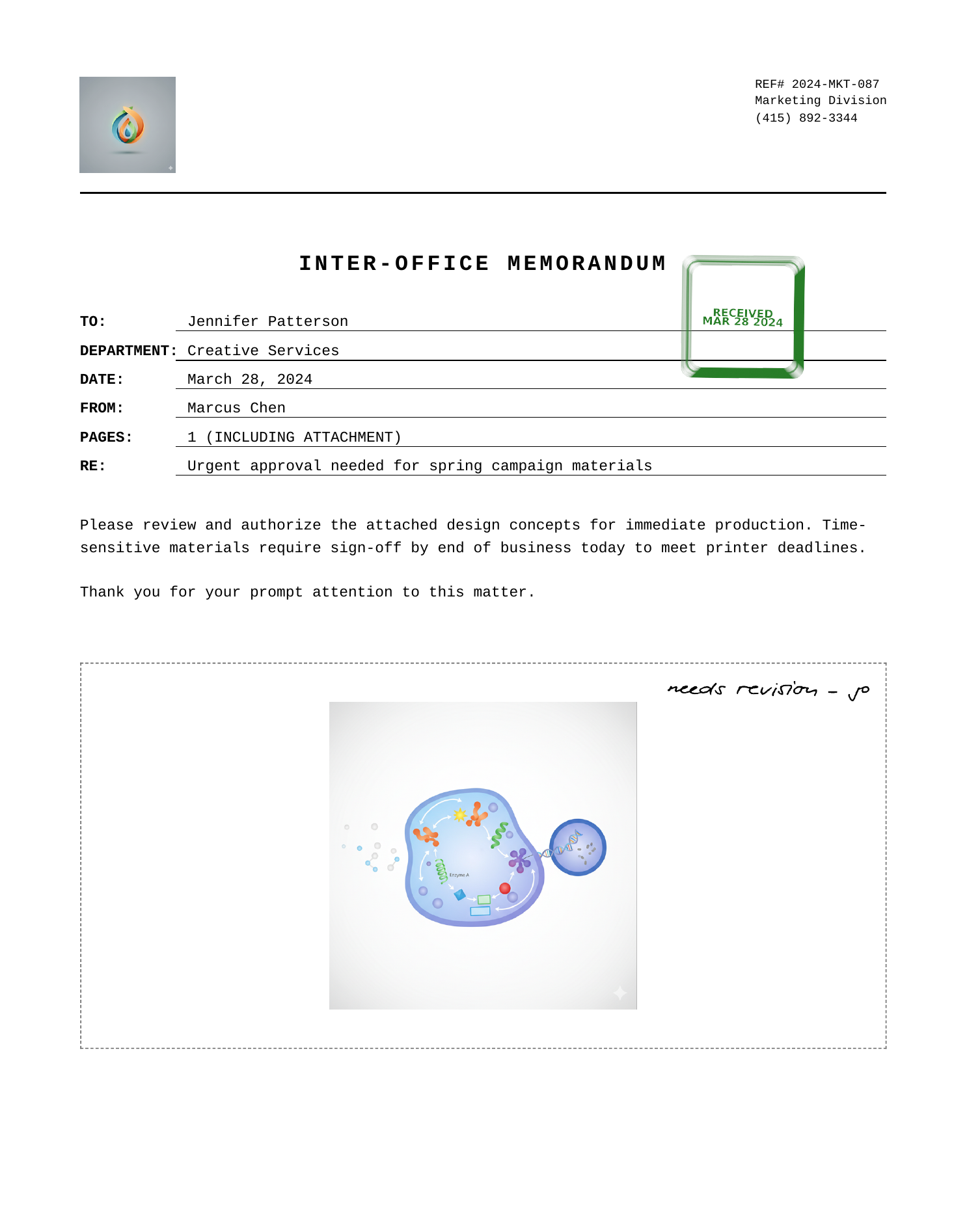}
    \end{subfigure}%
    \hfill
    \begin{subfigure}{\sdsexamplewidth}
        \includegraphics[width=\linewidth,height=\sdsexampleheight,keepaspectratio]{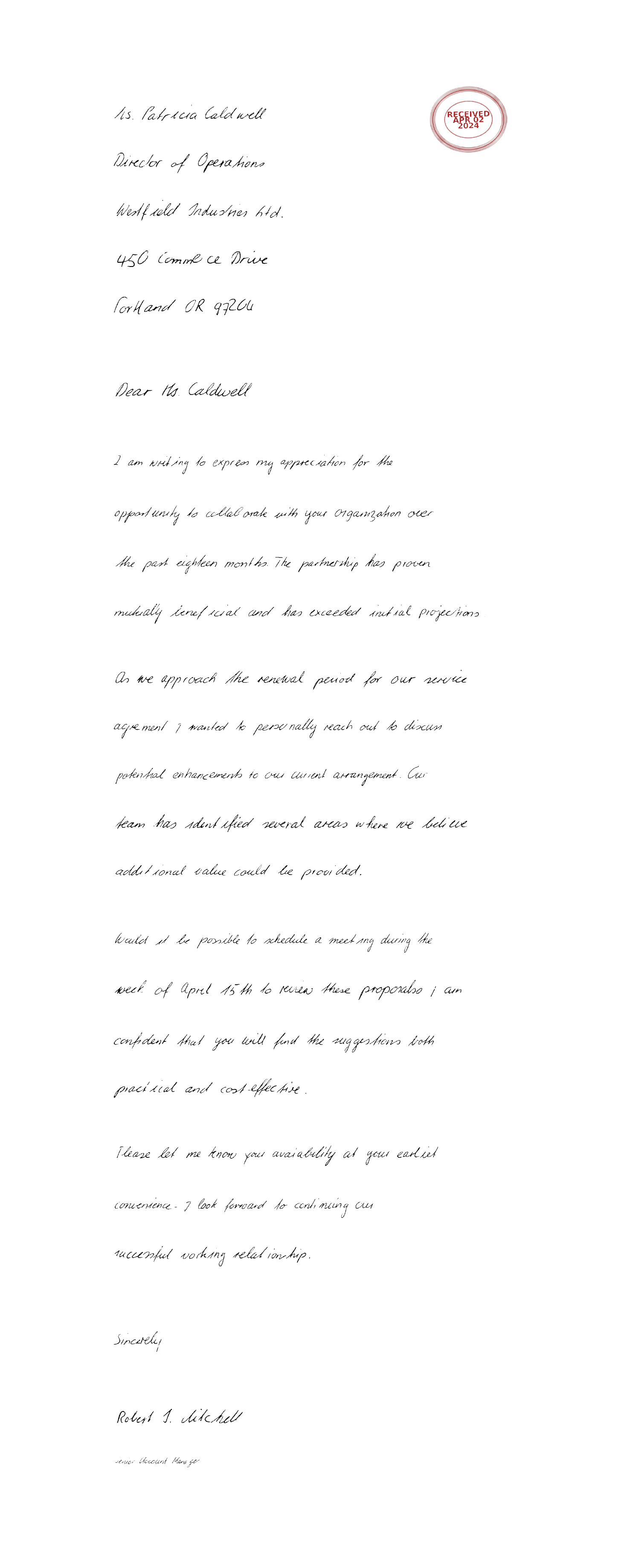}
    \end{subfigure}%
    \hfill
    \begin{subfigure}{\sdsexamplewidth}
        \includegraphics[width=\linewidth,height=\sdsexampleheight,keepaspectratio]{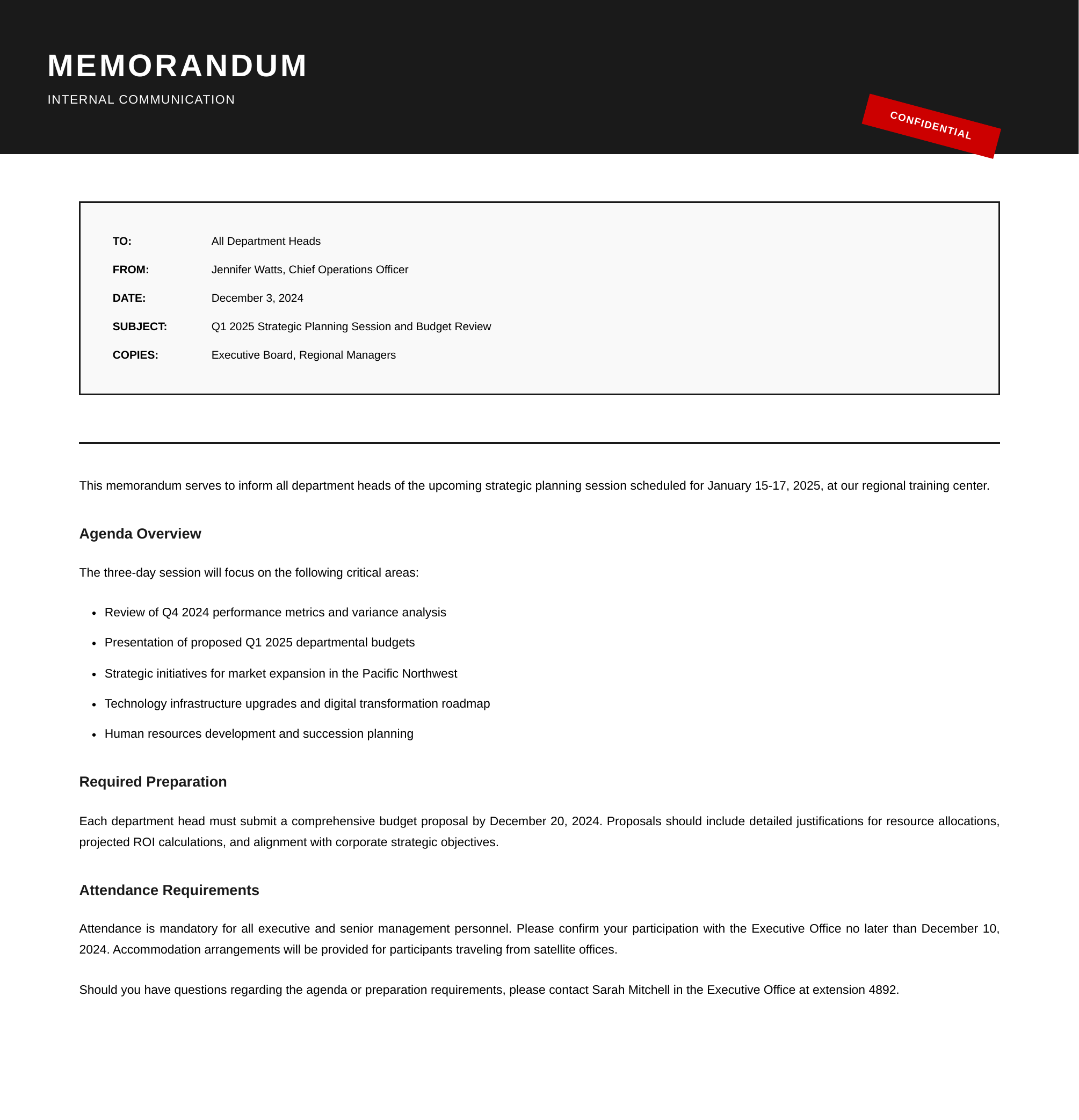}
    \end{subfigure}%
    \caption{\rvlcdip{} synthetic dataset samples for CLS task.}\label{app:sds_ex_rvlcdip}
\end{figure*}
\begin{figure*}[htbp]
    \centering
    \begin{subfigure}{\sdsexamplewidth}
        \includegraphics[width=\linewidth,height=\sdsexampleheight,keepaspectratio]{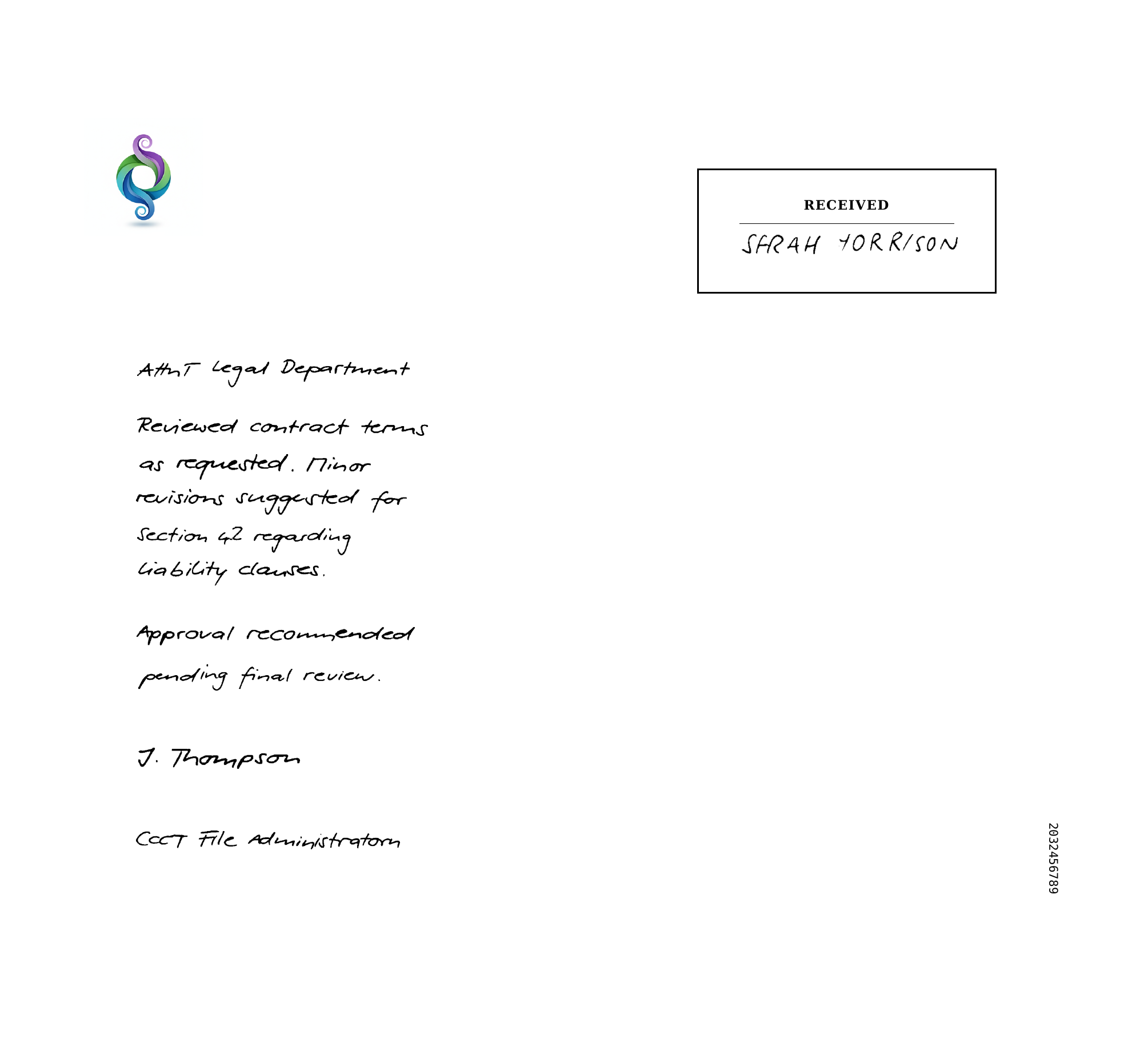}
    \end{subfigure}%
    \hfill
    \begin{subfigure}{\sdsexamplewidth}
        \includegraphics[width=\linewidth,height=\sdsexampleheight,keepaspectratio]{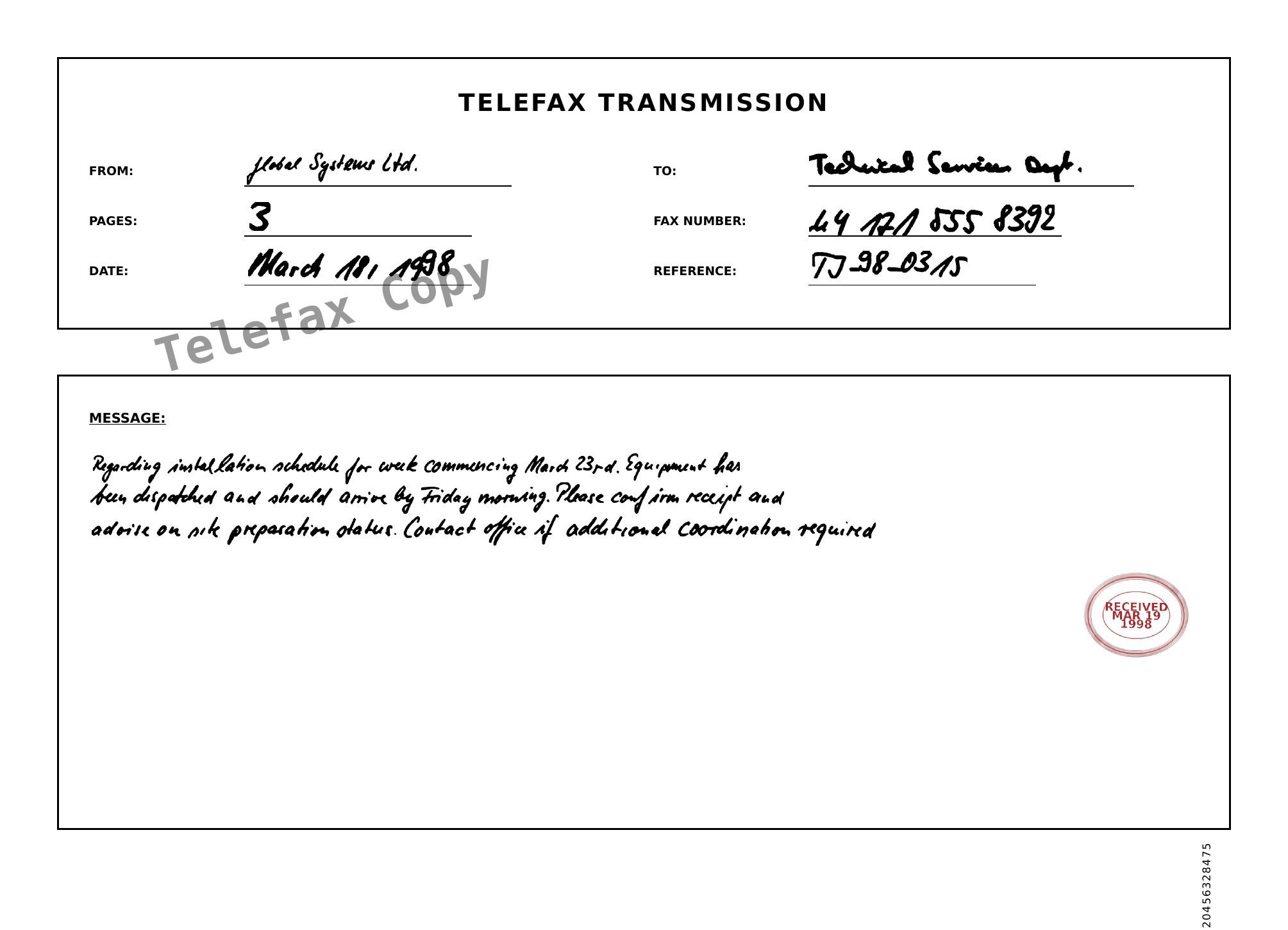}
    \end{subfigure}%
    \hfill
    \begin{subfigure}{\sdsexamplewidth}
        \includegraphics[width=\linewidth,height=\sdsexampleheight,keepaspectratio]{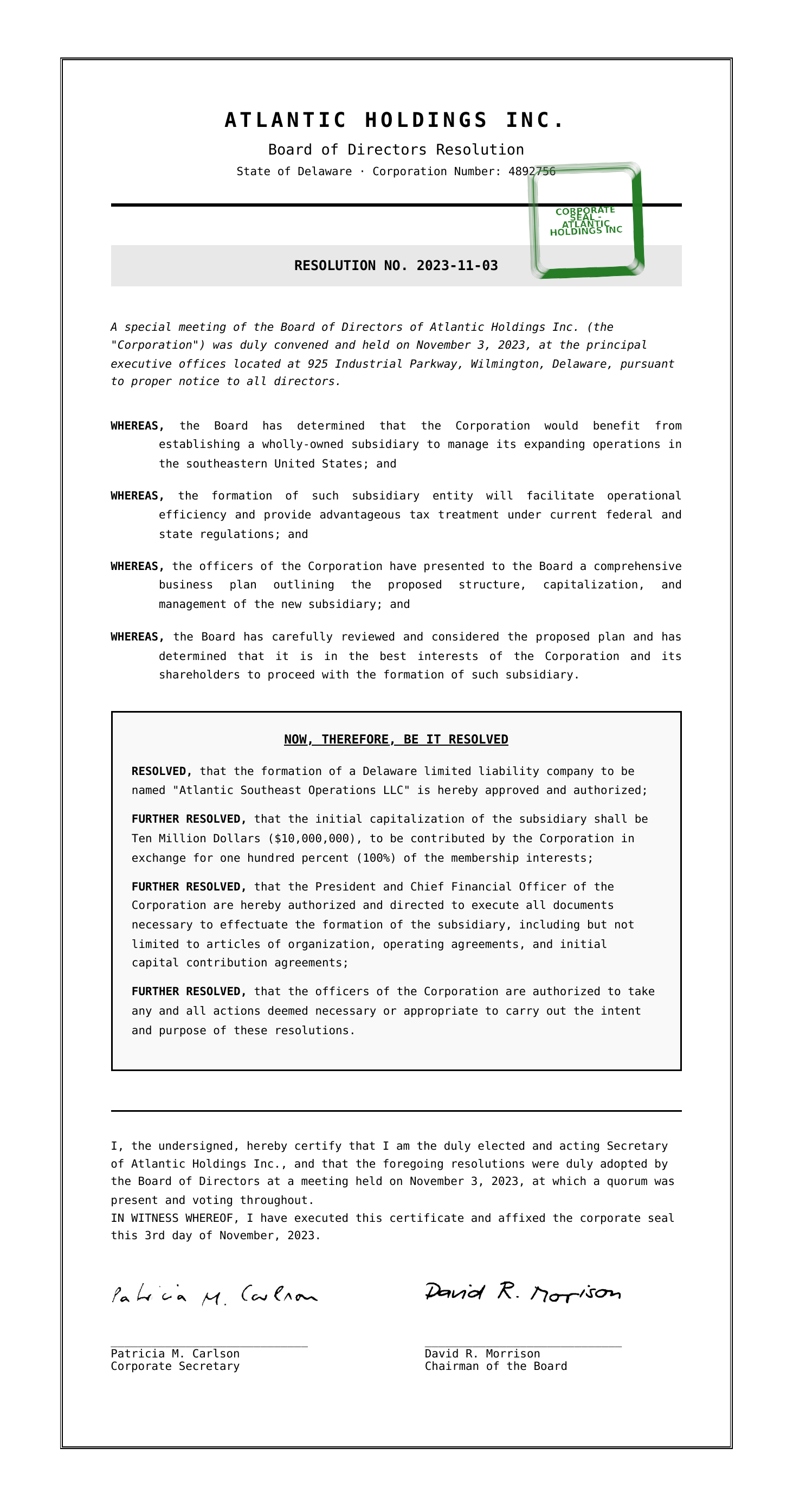}
    \end{subfigure}%
    \hfill
    \begin{subfigure}{\sdsexamplewidth}
        \includegraphics[width=\linewidth,height=\sdsexampleheight,keepaspectratio]{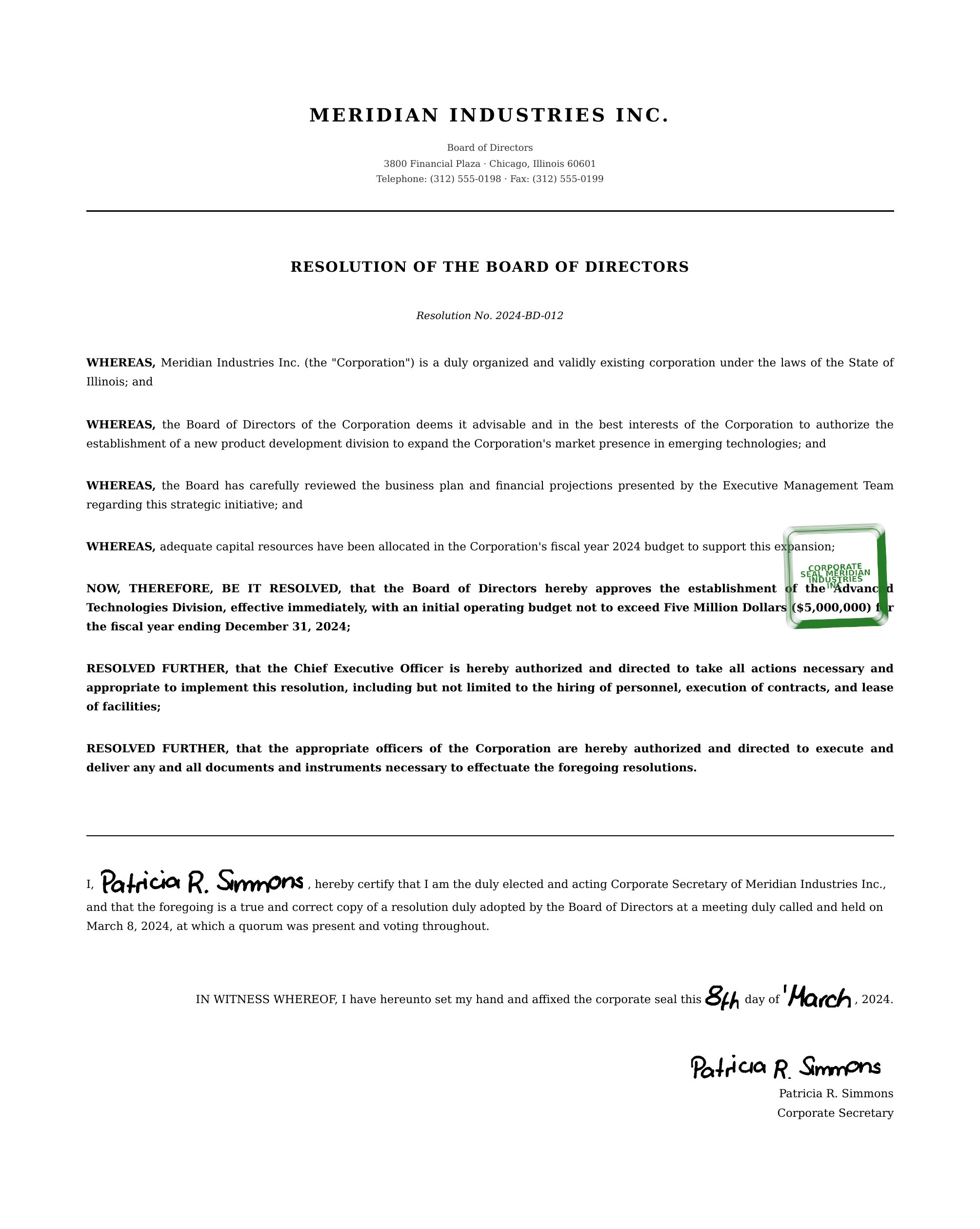}
    \end{subfigure}%
    \hfill
    \begin{subfigure}{\sdsexamplewidth}
        \includegraphics[width=\linewidth,height=\sdsexampleheight,keepaspectratio]{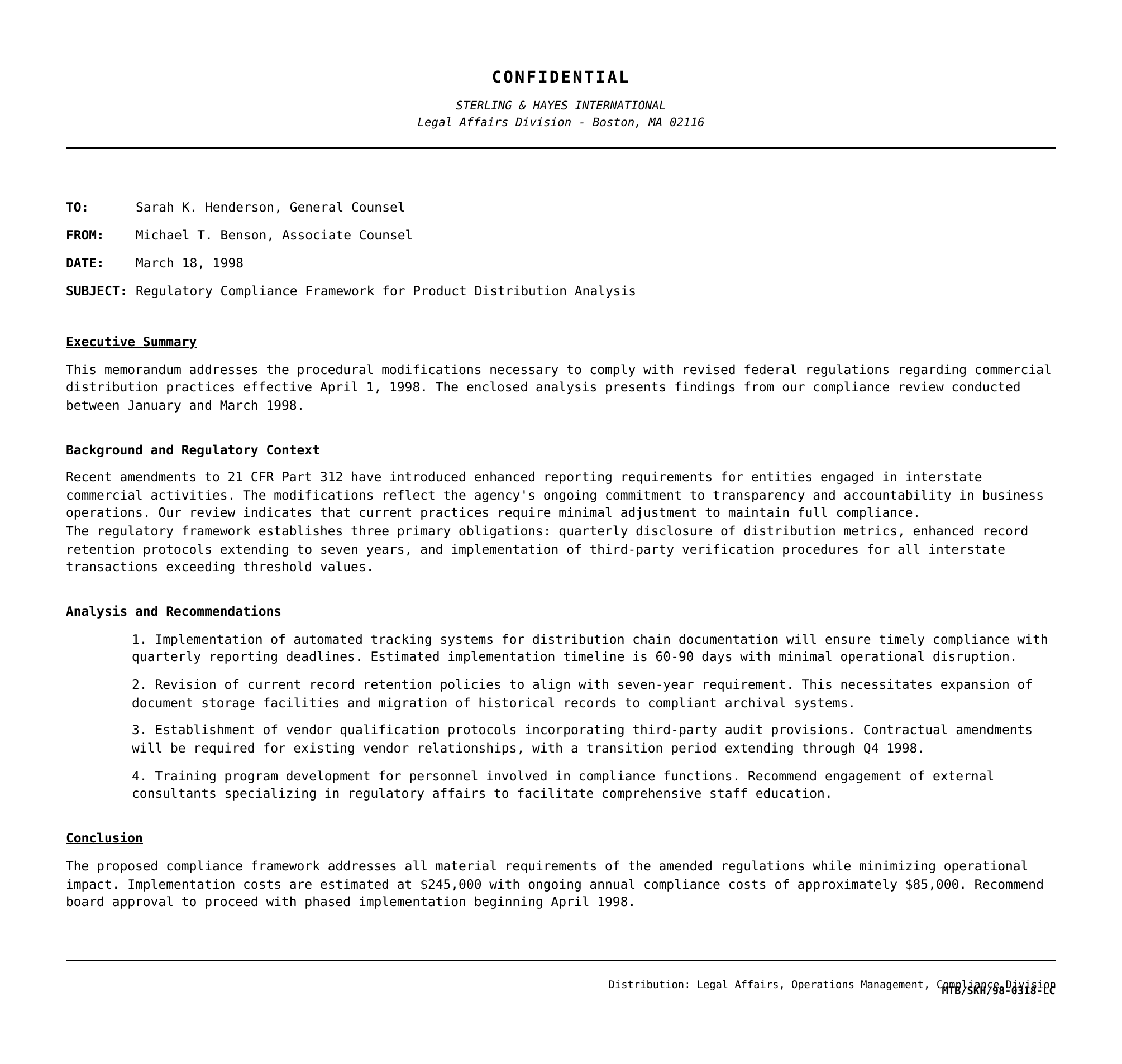}
    \end{subfigure}%
    \caption{\tobacco{} synthetic dataset samples for CLS task.}\label{app:sds_ex_tobacco3482}
\end{figure*}
\begin{figure*}[htbp]
    \centering
    \begin{subfigure}{\sdsexamplewidth}
        \includegraphics[width=\linewidth,height=\sdsexampleheight,keepaspectratio]{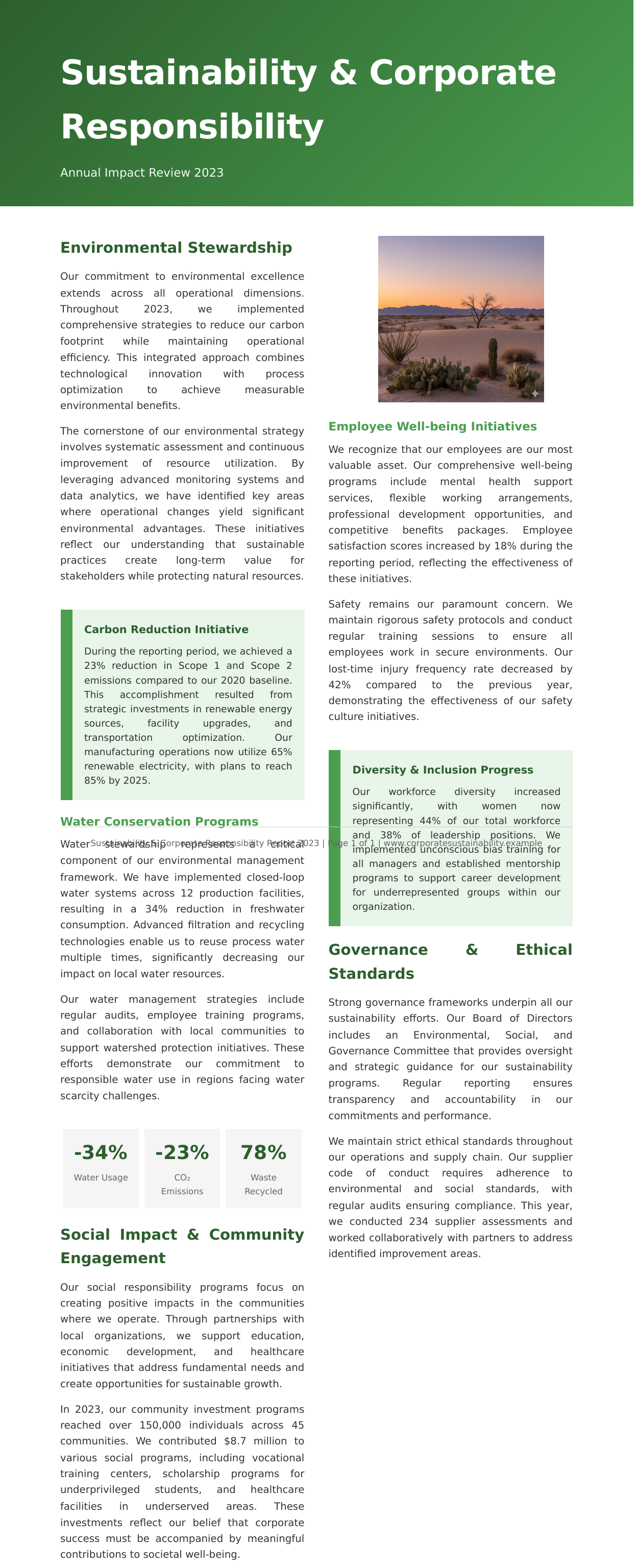}
    \end{subfigure}%
    \hfill
    \begin{subfigure}{\sdsexamplewidth}
        \includegraphics[width=\linewidth,height=\sdsexampleheight,keepaspectratio]{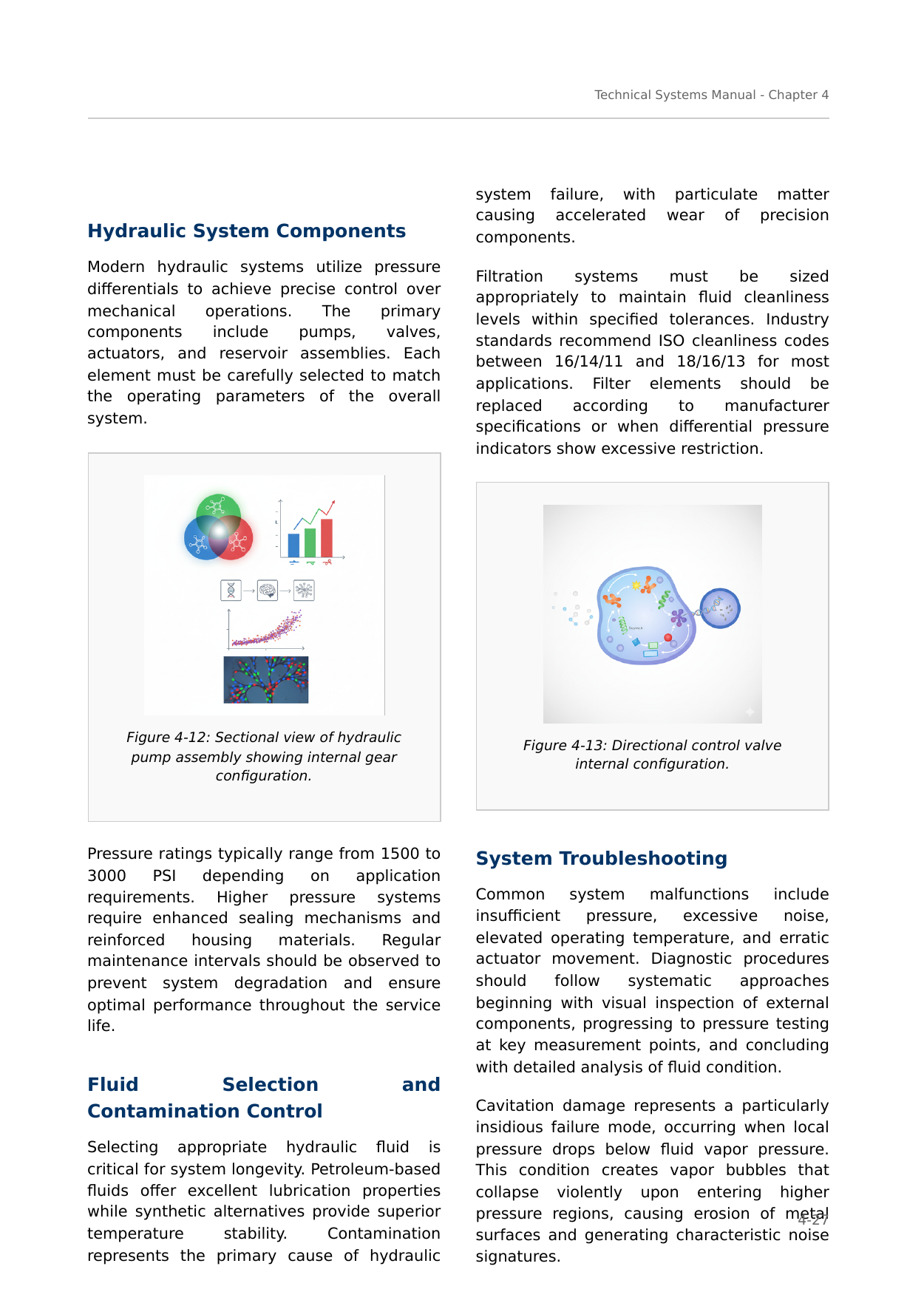}
    \end{subfigure}%
    \hfill
    \begin{subfigure}{\sdsexamplewidth}
        \includegraphics[width=\linewidth,height=\sdsexampleheight,keepaspectratio]{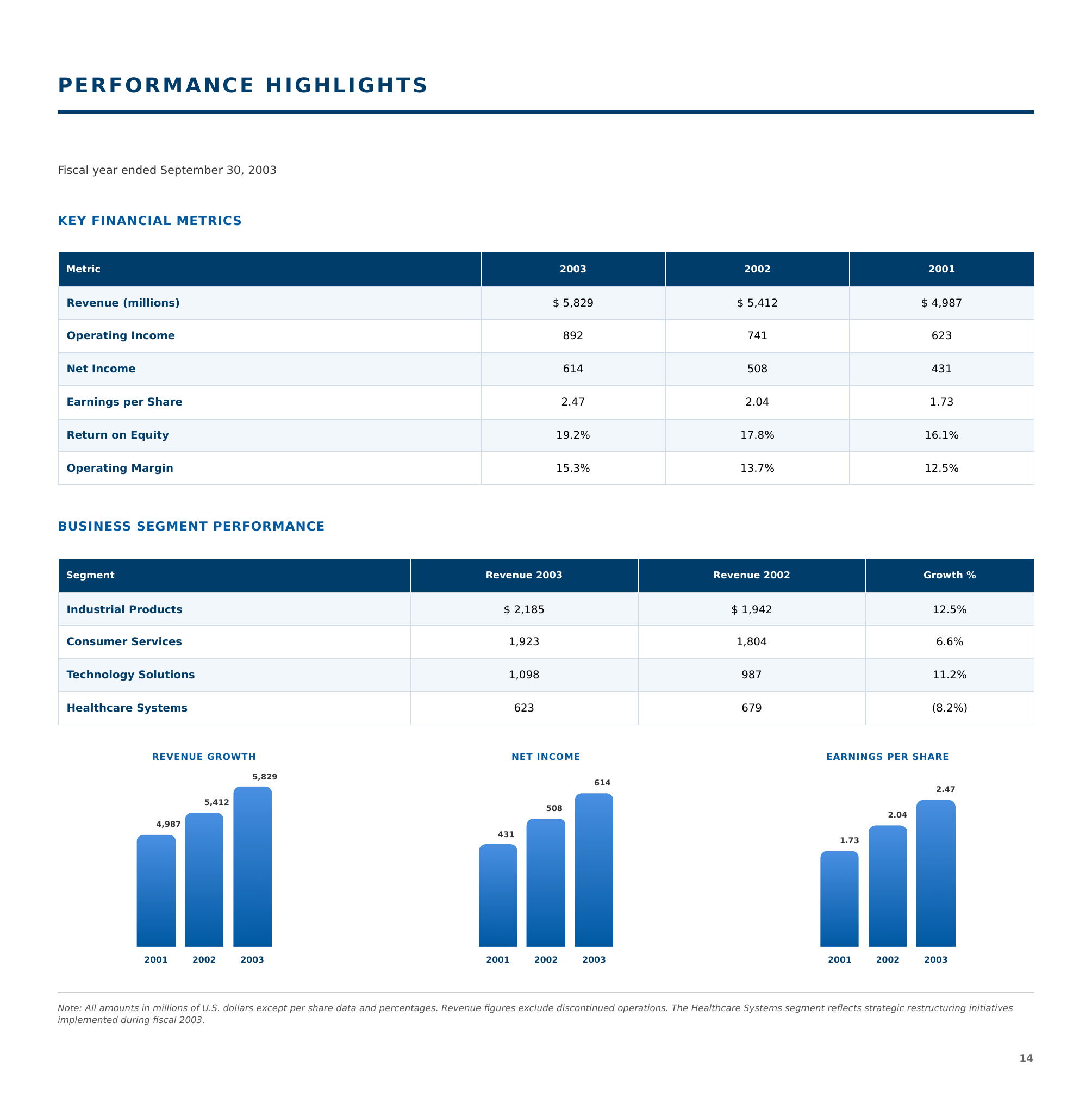}
    \end{subfigure}%
    \hfill
    \begin{subfigure}{\sdsexamplewidth}
        \includegraphics[width=\linewidth,height=\sdsexampleheight,keepaspectratio]{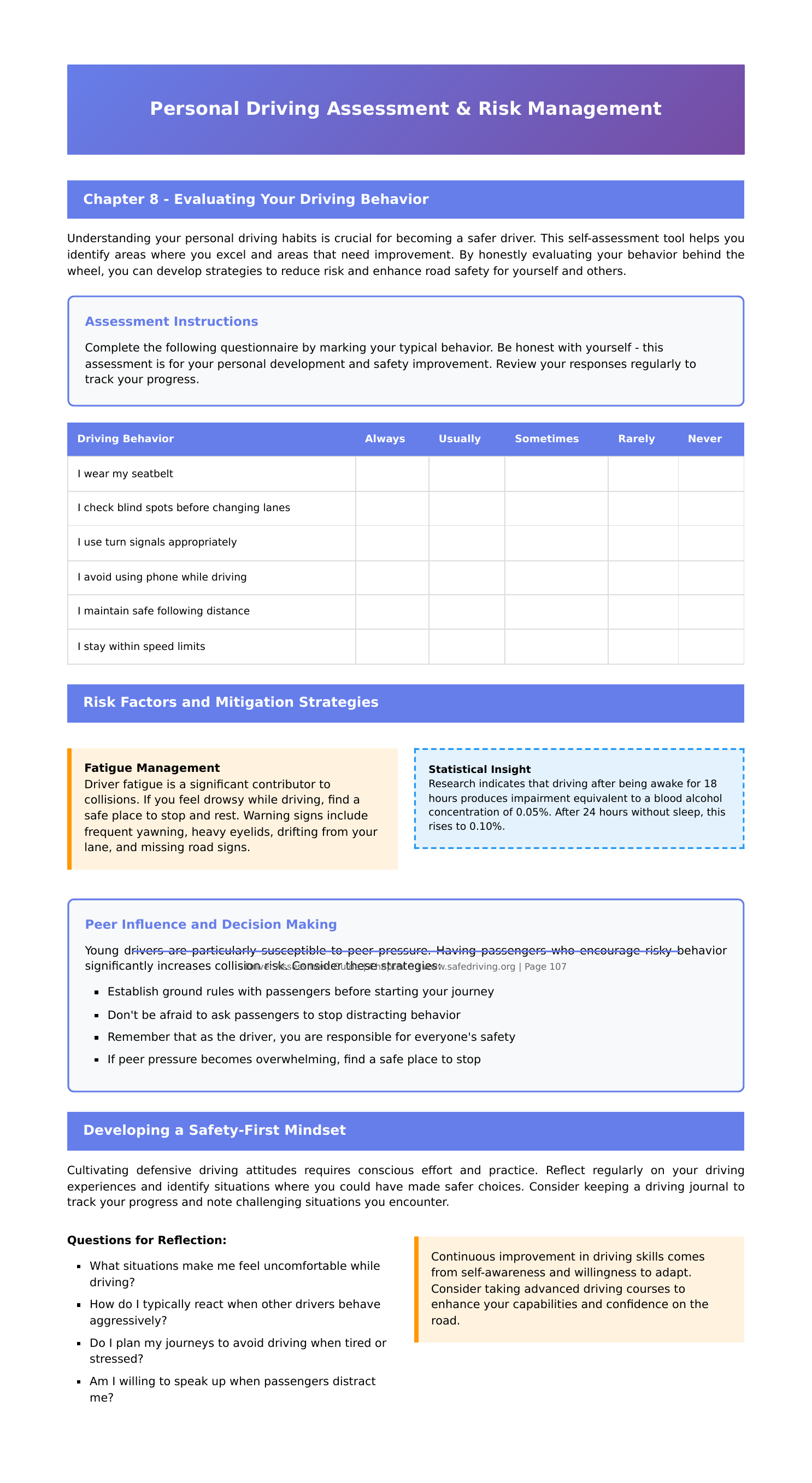}
    \end{subfigure}%
    \hfill
    \begin{subfigure}{\sdsexamplewidth}
        \includegraphics[width=\linewidth,height=\sdsexampleheight,keepaspectratio]{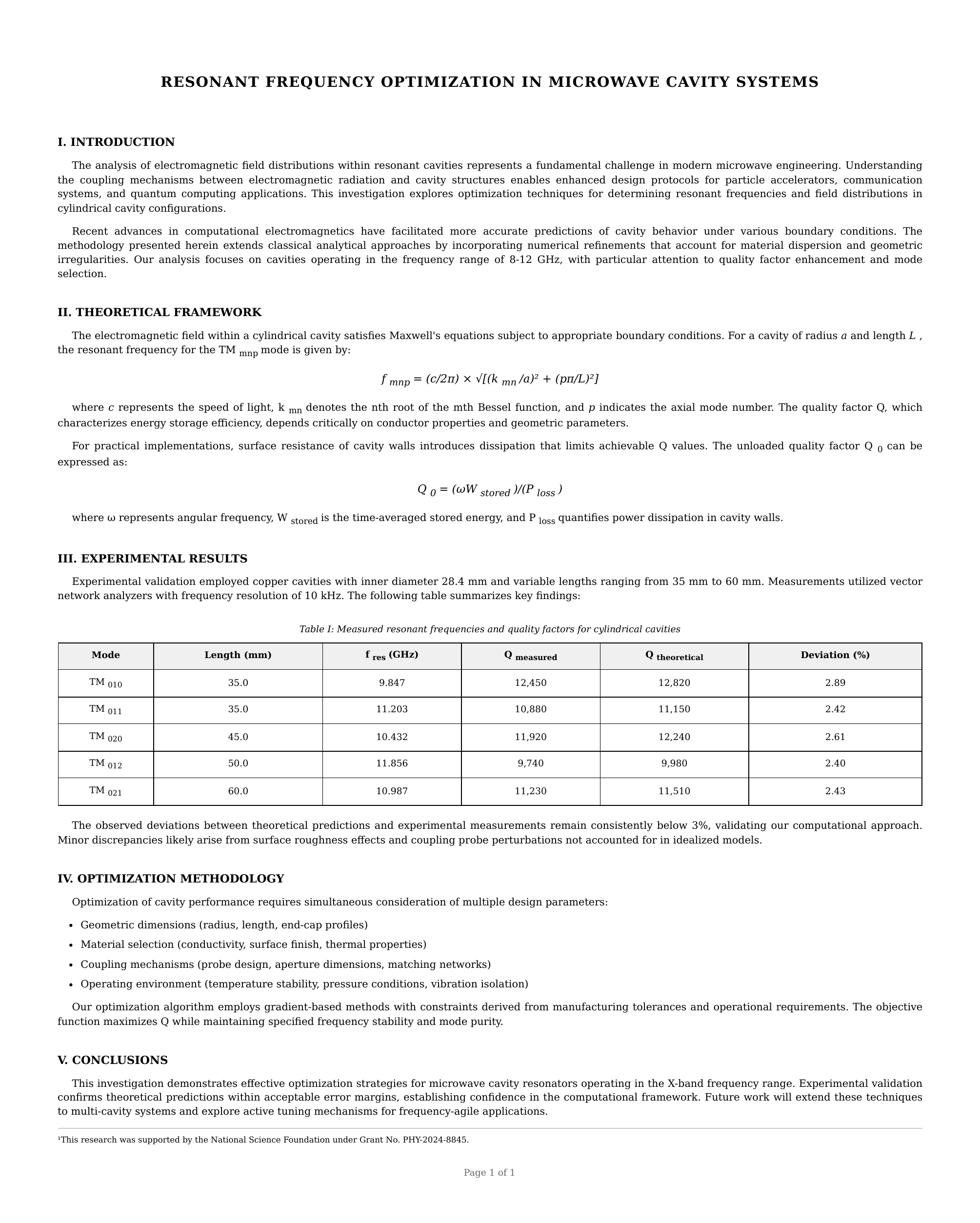}
    \end{subfigure}%
    \caption{\doclaynetdla{} synthetic dataset samples for DLA task.}\label{app:sds_ex_doclaynetdla}
\end{figure*}

\begin{figure*}[htbp]
    \centering
    \begin{subfigure}{\sdsexamplewidth}
        \includegraphics[width=\linewidth,height=\sdsexampleheight,keepaspectratio]{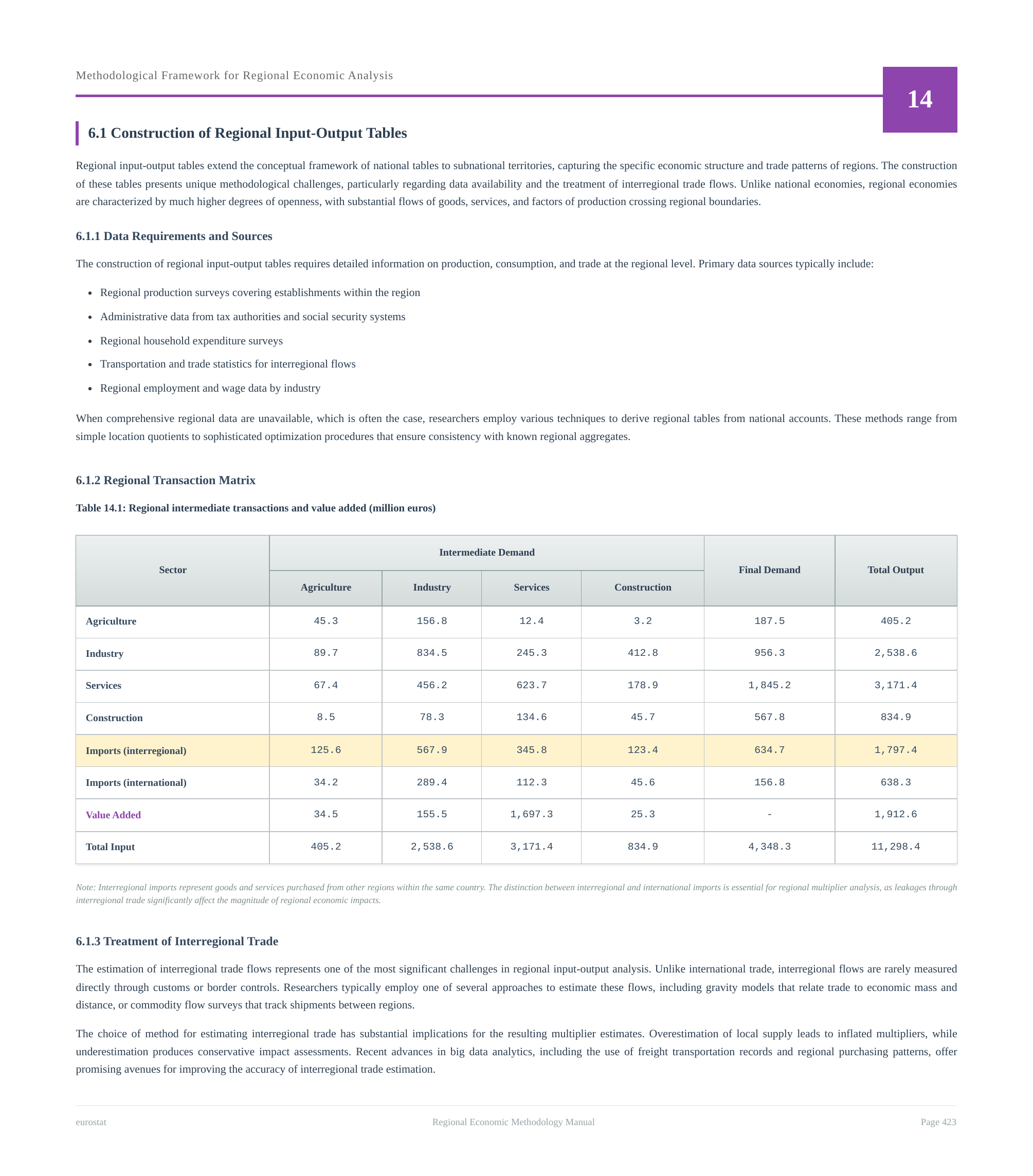}
    \end{subfigure}%
    \hfill
    \begin{subfigure}{\sdsexamplewidth}
        \includegraphics[width=\linewidth,height=\sdsexampleheight,keepaspectratio]{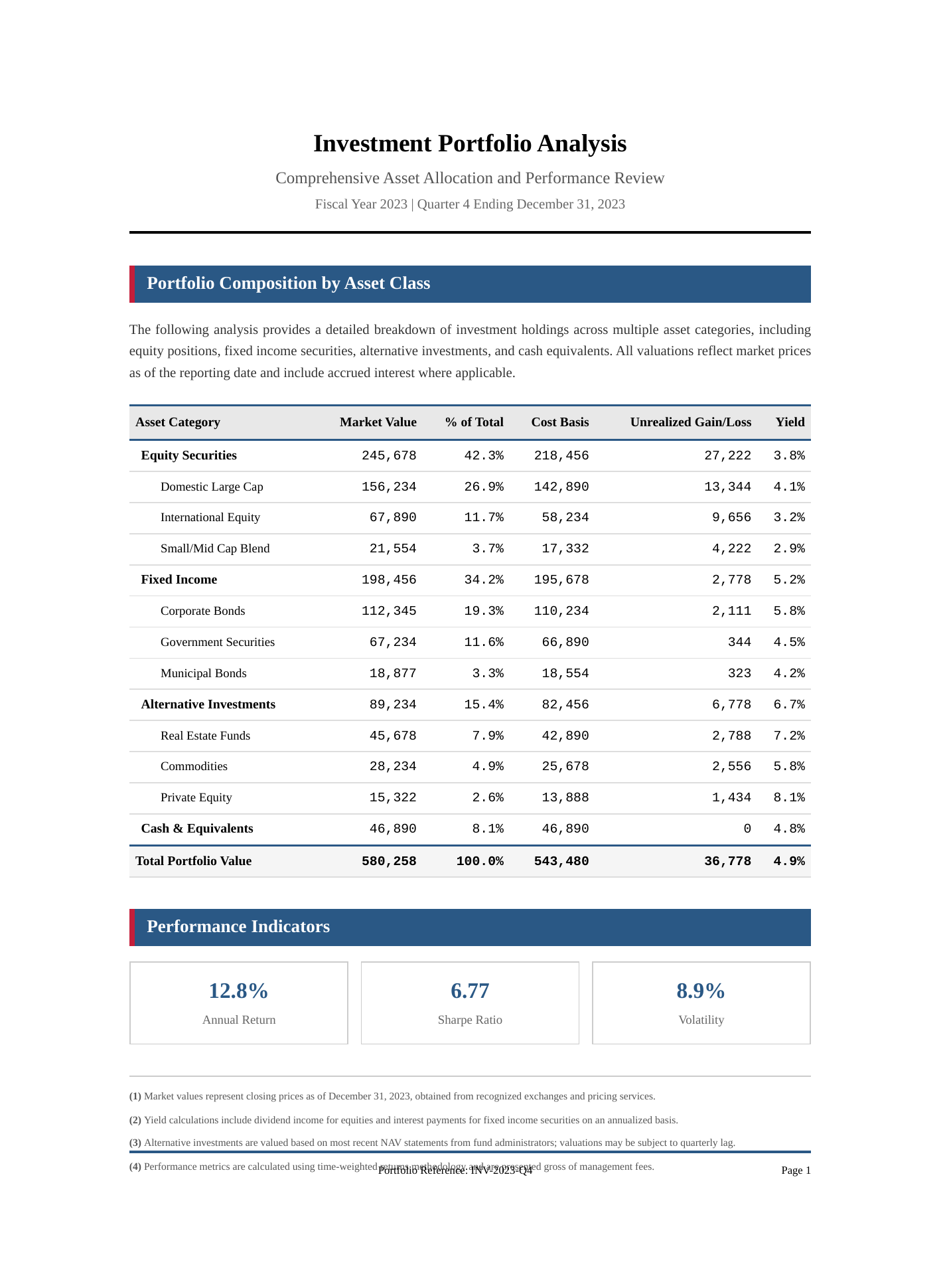}
    \end{subfigure}%
    \hfill
    \begin{subfigure}{\sdsexamplewidth}
        \includegraphics[width=\linewidth,height=\sdsexampleheight,keepaspectratio]{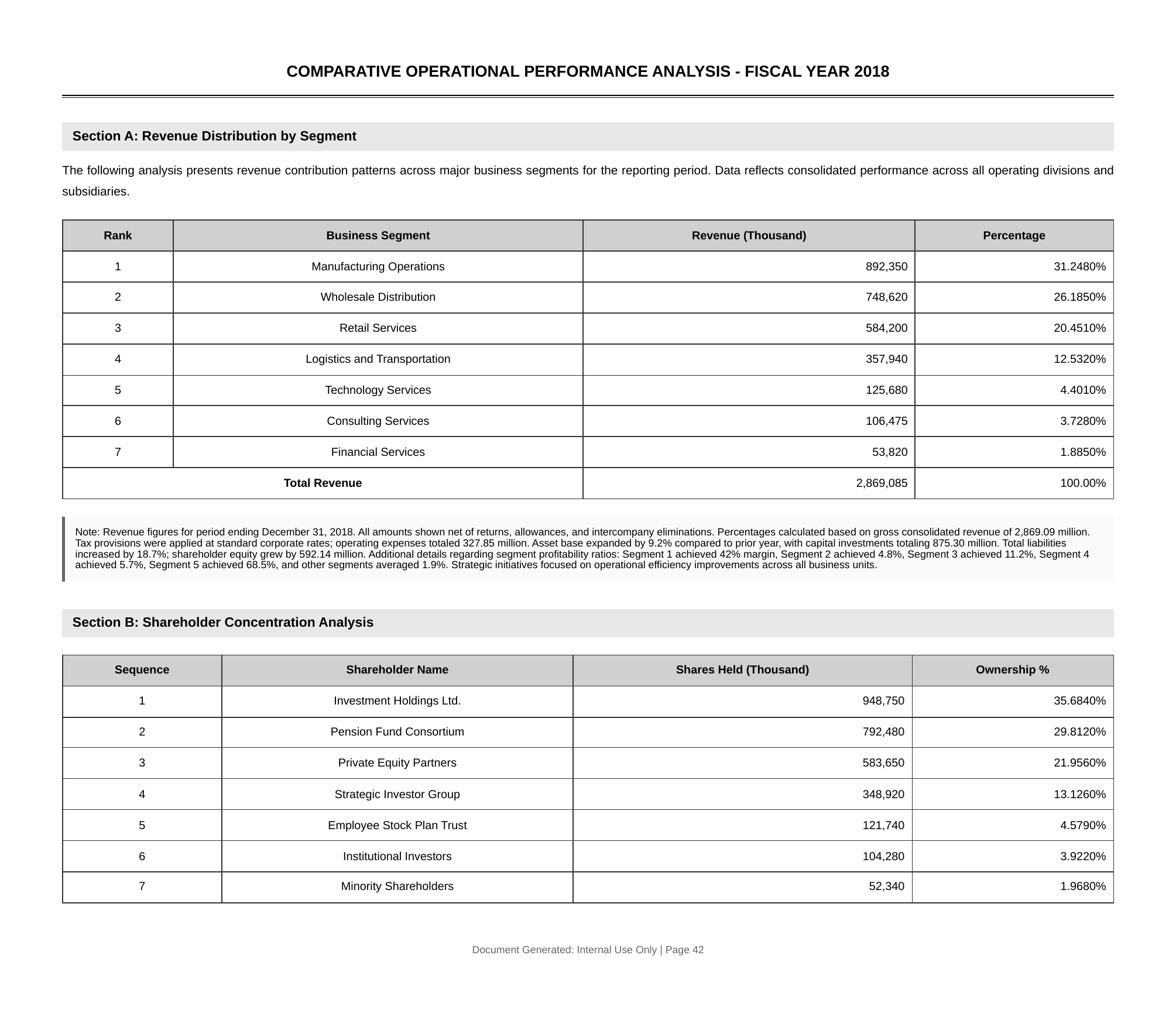}
    \end{subfigure}%
    \hfill
    \begin{subfigure}{\sdsexamplewidth}
        \includegraphics[width=\linewidth,height=\sdsexampleheight,keepaspectratio]{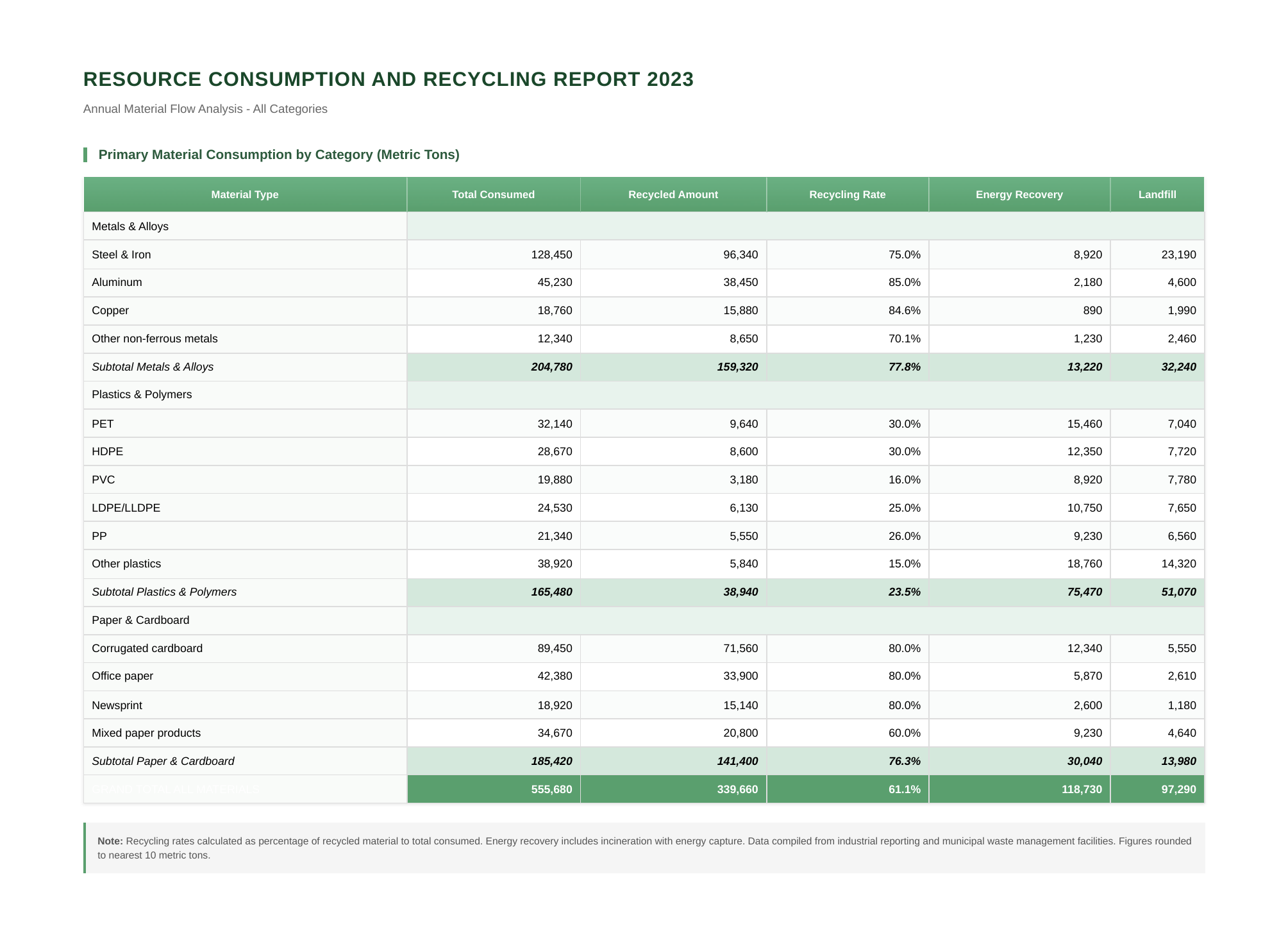}
    \end{subfigure}%
    \hfill
    \begin{subfigure}{\sdsexamplewidth}
        \includegraphics[width=\linewidth,height=\sdsexampleheight,keepaspectratio]{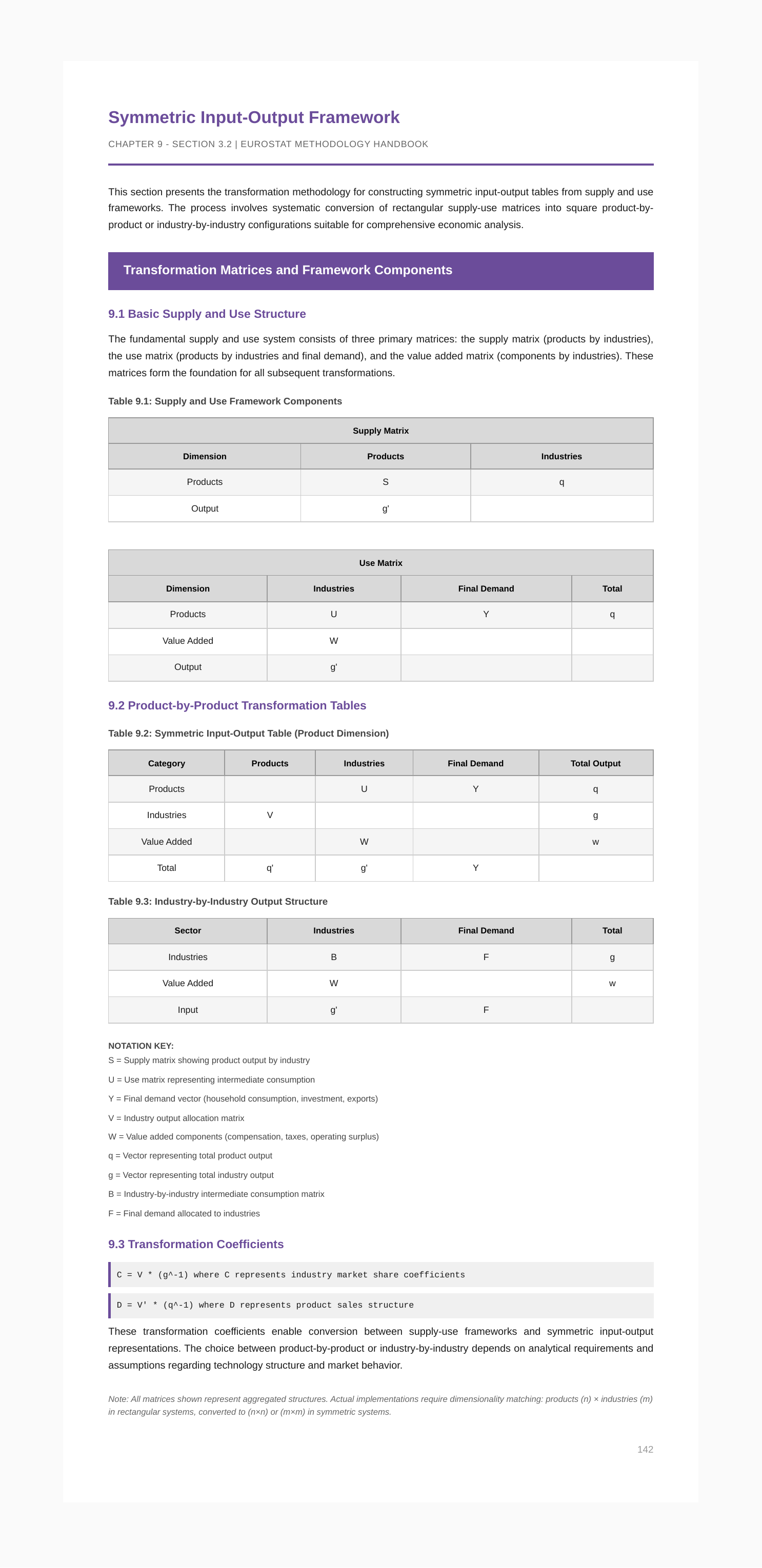}
    \end{subfigure}%
    \caption{\icdarctdar{} synthetic dataset samples for DLA task.}\label{app:sds_ex_icdar2019}
\end{figure*}

\begin{figure*}[htbp]
    \centering
    \begin{subfigure}{\sdsexamplewidth}
        \includegraphics[width=\linewidth,height=\sdsexampleheight,keepaspectratio]{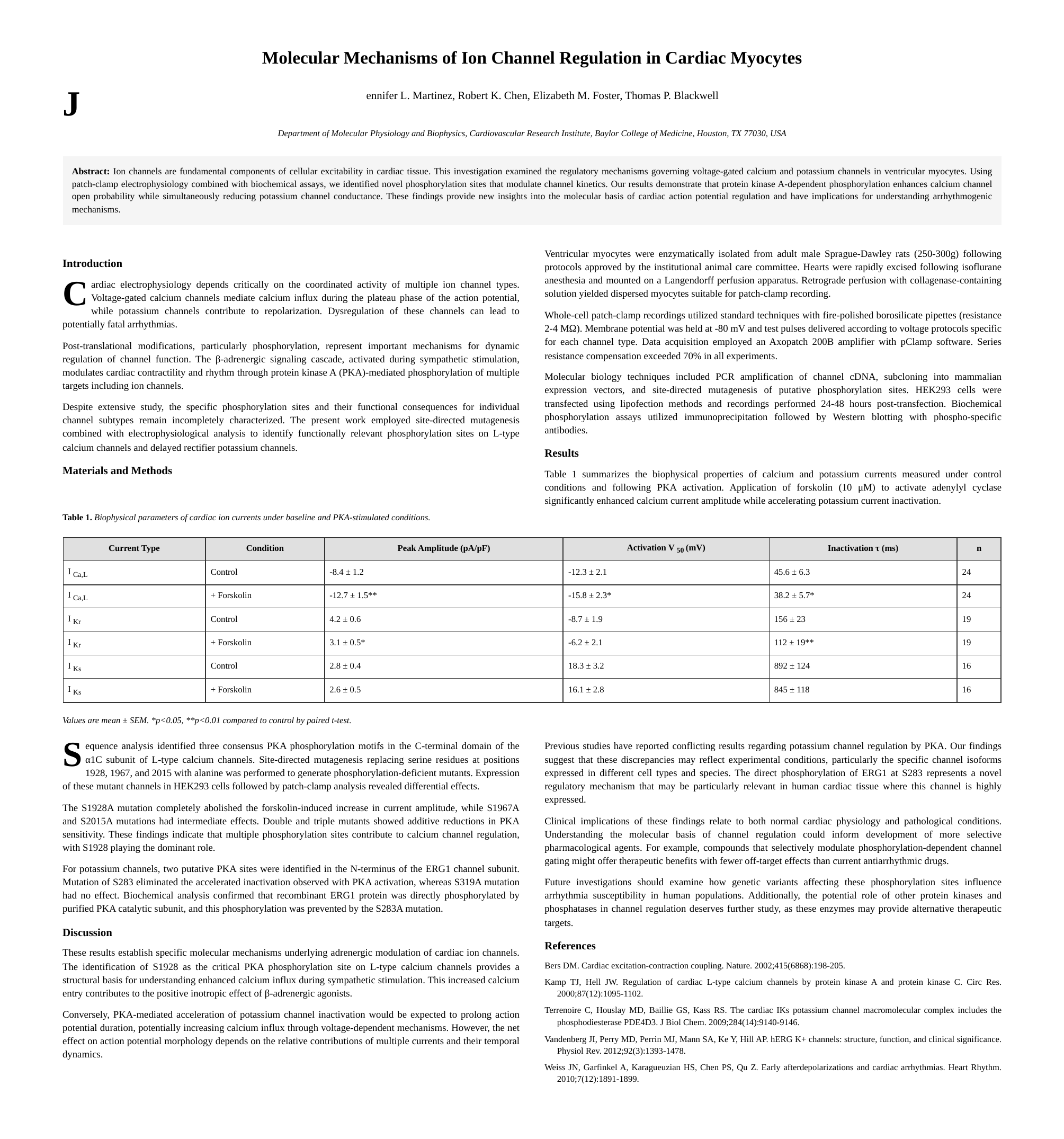}
    \end{subfigure}%
    \hfill
    \begin{subfigure}{\sdsexamplewidth}
        \includegraphics[width=\linewidth,height=\sdsexampleheight,keepaspectratio]{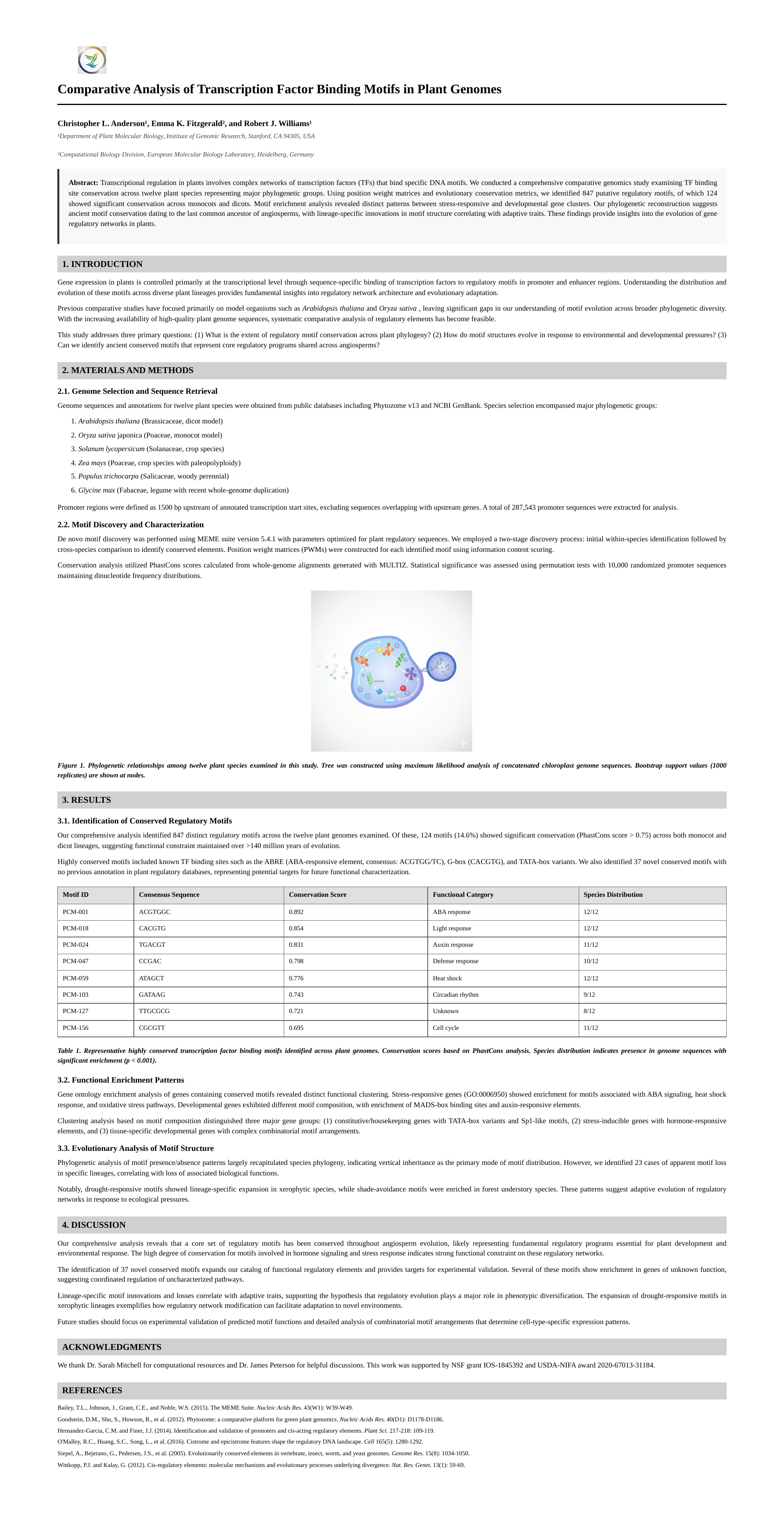}
    \end{subfigure}%
    \hfill
    \begin{subfigure}{\sdsexamplewidth}
        \includegraphics[width=\linewidth,height=\sdsexampleheight,keepaspectratio]{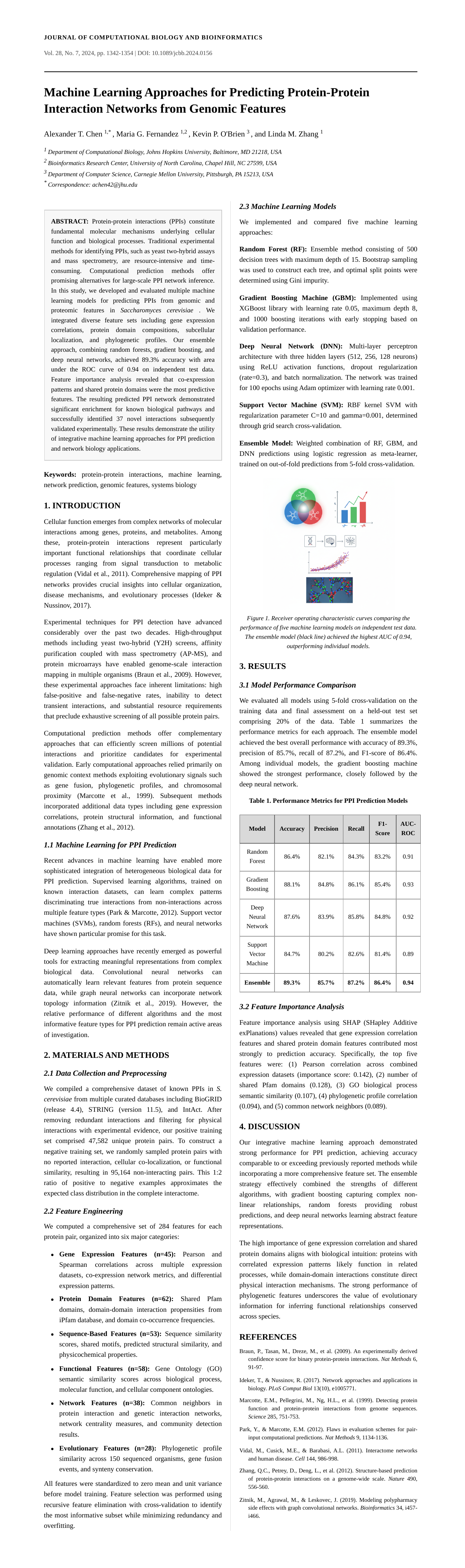}
    \end{subfigure}%
    \hfill
    \begin{subfigure}{\sdsexamplewidth}
        \includegraphics[width=\linewidth,height=\sdsexampleheight,keepaspectratio]{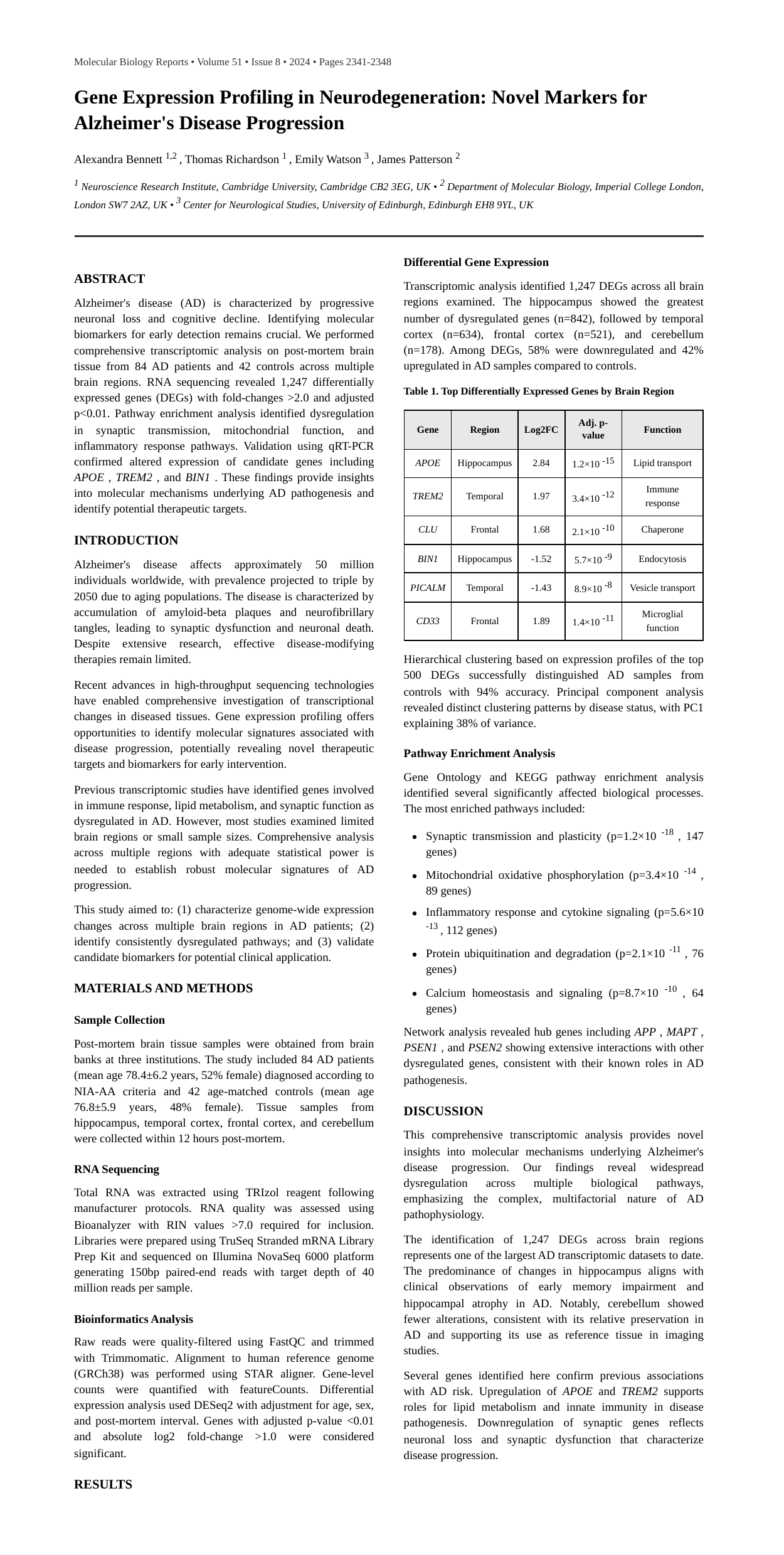}
    \end{subfigure}%
    \hfill
    \begin{subfigure}{\sdsexamplewidth}
        \includegraphics[width=\linewidth,height=\sdsexampleheight,keepaspectratio]{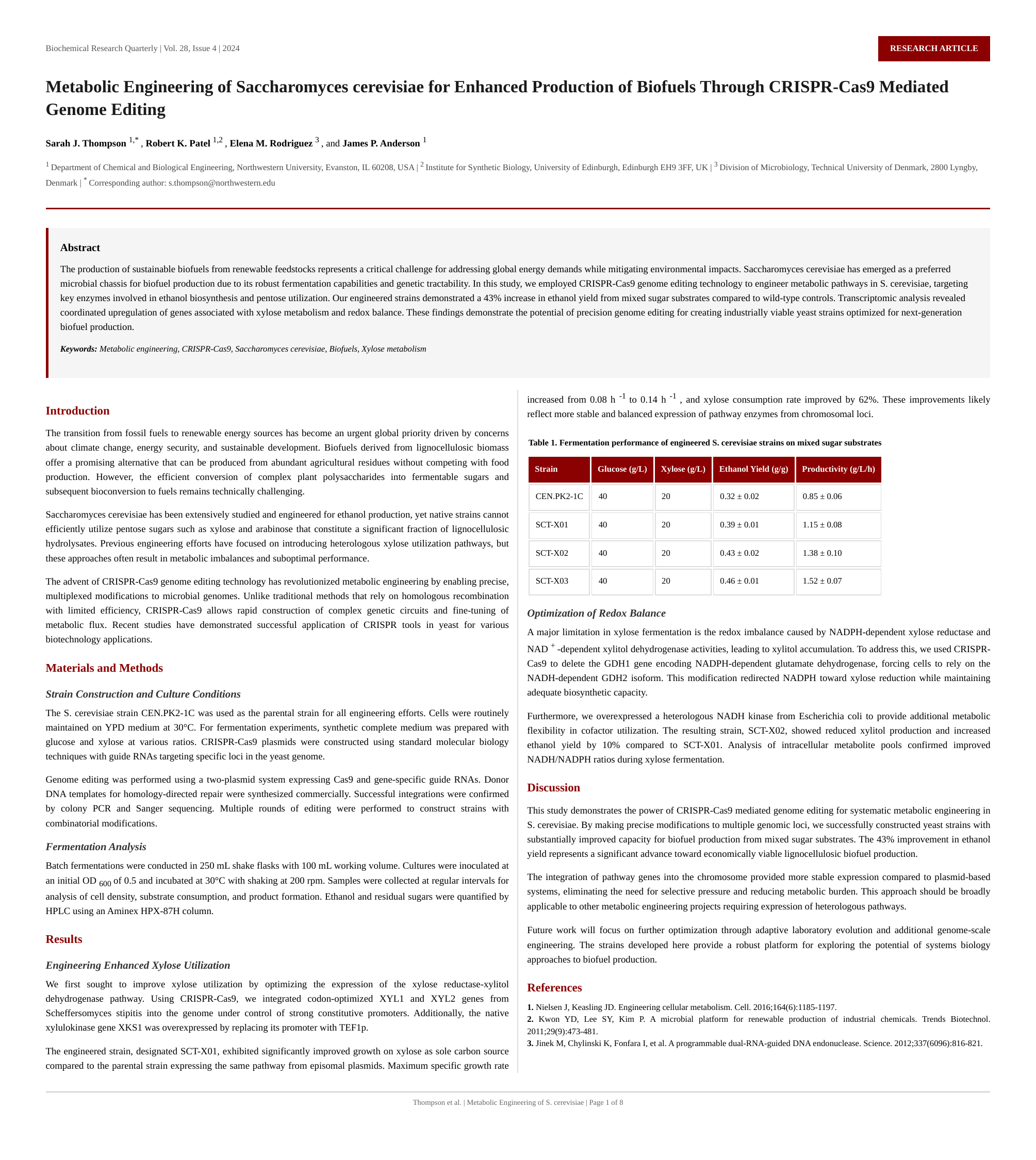}
    \end{subfigure}%
    \caption{\publaynet{} synthetic dataset samples for DLA task.}\label{app:sds_ex_publaynet}
\end{figure*}

\section{Synthetic Dataset Details}\label{app:syn_datasets_details}
\Cref{tab:syn_datasets_details_final,tab:syn_datasets_details_ablation} report 
generation statistics for all synthetic datasets, including sample counts, token usage, estimated API costs, and content composition metrics. The first table shows our final configurations, while the second presents ablation variants across different sampling strategies and strengths.

\begin{table*}[t!] 
\centering
\small
\setlength{\tabcolsep}{3pt}
\renewcommand{\arraystretch}{0.85}
\rowcolors{2}{gray!15}{white} 

\resizebox{\textwidth}{!}{%
\begin{tabular}{l|l|r|r|r|r|r|r|r|r|r|r}
\toprule
 &  & \rotatebox{90}{\textbf{Embedding Type}} & \rotatebox{90}{\textbf{Min. Cluster Size}} & \rotatebox{90}{\textbf{Total Samples}} & \rotatebox{90}{\textbf{Total Valid Samples}} & \rotatebox{90}{\textbf{Input Tokens (M)}} & \rotatebox{90}{\textbf{Output Tokens (M)}} & \rotatebox{90}{\textbf{Cost (USD)}} & \rotatebox{90}{\textbf{Avg. Words}} & \rotatebox{90}{\textbf{Avg. HW Elems}} & \rotatebox{90}{\textbf{Avg. Visual Elems}} \\
{\textbf{Task}} & {\textbf{Dataset Name}} &  &  &  &  &  &  &  &  &  &  \\
\midrule
\multirow[t]{3}{*}{CLASSIFICATION} & \doclaynetcls{} & comb & 10 & 4494 & 3978 & 4.08 & 7.48 & 124 & 407 & 0.052 & 0.334 \\
 & \rvlcdip{} & comb & 10 & 4441 & 3823 & 5.16 & 5.34 & 96 & 206 & 1.466 & 0.371 \\
 & \tobacco{} & comb & 10 & 5292 & 4092 & 6.24 & 6.49 & 116 & 251 & 2.042 & 0.435 \\
\midrule
\multirow[t]{3}{*}{DLA} & \doclaynetdla{} & comb & 10 & 4502 & 3732 & 2.95 & 7.28 & 114 & 372 & 0.061 & 0.301 \\
 & \icdarctdar{} & img & 5 & 1594 & 1515 & 2.01 & 2.89 & 49 & 297 & 0.004 & 0.048 \\
 & \publaynet{} & img & 5 & 4476 & 3835 & 5.9 & 12.14 & 200 & 874 & 0 & 0.542 \\
\midrule
\multirow[t]{4}{*}{KIE} & \cord{} & comb & 10 & 1200 & 1182 & 1.19 & 1.66 & 29 & 67 & 0.007 & 0.158 \\
 & \klc{} & comb & 10 & 4005 & 3441 & 4.92 & 4.91 & 88 & 146 & 0.194 & 0.282 \\
 & \funsd{} & comb & 10 & 291 & 259 & 0.34 & 0.5 & 9 & 128 & 3.629 & 0.205 \\
 & \sroie{} & comb & 10 & 1050 & 1008 & 1.46 & 1.62 & 29 & 113 & 0.089 & 0.124 \\
\midrule
\multirow[t]{2}{*}{QA} & \docvqa{} & comb & 10 & 9990 & 8082 & 10.66 & 14.3 & 246 & 173 & 2.37 & 0.31 \\
 & \wiki{} & comb & 5 & 1604 & 1479 & 1.13 & 3.13 & 50 & 204 & 0.059 & 0.381 \\
\bottomrule
\end{tabular}
}

\caption{Overview of the final synthesized datasets using sampling strategy (v2) and sampling strength ($\alpha=1$) in our experiments. For each dataset, we report the embedding type (image-only or image+text), clustering and sampling parameters, total and valid sample counts, token usage, estimated generation cost, and average text/visual content statistics.}


\label{tab:syn_datasets_details_final}
\end{table*}


\begin{table*}[t!] 
\centering
\small
\setlength{\tabcolsep}{3pt}
\renewcommand{\arraystretch}{0.85}
\rowcolors{2}{gray!15}{white} 

\resizebox{\textwidth}{!}{%
\begin{tabular}{l|l|r|r|r|r|r|r|r|r|r|r|r|r}
\toprule
 &  & \rotatebox{90}{\textbf{Embedding Type}} & \rotatebox{90}{\textbf{Min. Cluster Size}} & \rotatebox{90}{\textbf{Sample Strategy}} & \rotatebox{90}{\textbf{Sample Alpha}} & \rotatebox{90}{\textbf{Total Samples}} & \rotatebox{90}{\textbf{Total Valid Samples}} & \rotatebox{90}{\textbf{Input Tokens (M)}} & \rotatebox{90}{\textbf{Output Tokens (M)}} & \rotatebox{90}{\textbf{Cost (USD)}} & \rotatebox{90}{\textbf{Avg. Words}} & \rotatebox{90}{\textbf{Avg. HW Elems}} & \rotatebox{90}{\textbf{Avg. Visual Elems}} \\
{\textbf{Task}} & {\textbf{Dataset Name}} &  &  &  &  &  &  &  &  &  &  &  &  \\
\midrule
\multirow[t]{6}{*}{CLASSIFICATION} & \rvlcdip{} & comb & 10 & v1 & 1 & 4500 & 3853 & 5.04 & 5.28 & 95 & 211 & 1.305 & 0.397 \\
 & \rvlcdip{} & comb & 10 & v1 & 0.75 & 4306 & 3860 & 4.82 & 5.08 & 91 & 212 & 1.257 & 0.415 \\
 & \rvlcdip{} & comb & 10 & v1 & 0.5 & 4491 & 3997 & 5.03 & 5.32 & 96 & 209 & 1.24 & 0.414 \\
\cmidrule(lr){2-14}
 & \rvlcdip{} & comb & 10 & v2 & 1 & 4441 & 3823 & 5.16 & 5.34 & 96 & 206 & 1.466 & 0.371 \\
 & \rvlcdip{} & comb & 10 & v2 & 0.75 & 4329 & 3891 & 5.04 & 5.25 & 94 & 210 & 1.446 & 0.411 \\
 & \rvlcdip{} & comb & 10 & v2 & 0.5 & 4407 & 3860 & 5.16 & 5.33 & 95 & 206 & 1.31 & 0.411 \\
\midrule
\multirow[t]{6}{*}{DLA} & \publaynet{} & img & 5 & v1 & 1 & 4469 & 3937 & 5.9 & 12.97 & 212 & 939 & 0 & 0.536 \\
 & \publaynet{} & img & 5 & v1 & 0.75 & 4476 & 3929 & 5.9 & 12.99 & 213 & 937 & 0 & 0.551 \\
 & \publaynet{} & img & 5 & v1 & 0.5 & 4481 & 3988 & 5.9 & 13.05 & 213 & 943 & 0 & 0.568 \\
\cmidrule(lr){2-14}
 & \publaynet{} & img & 5 & v2 & 1 & 4476 & 3835 & 5.9 & 12.14 & 200 & 874 & 0 & 0.542 \\
 & \publaynet{} & img & 5 & v2 & 0.75 & 4505 & 3949 & 5.91 & 12.25 & 201 & 882 & 0 & 0.535 \\
 & \publaynet{} & img & 5 & v2 & 0.5 & 4497 & 3930 & 5.9 & 12.41 & 204 & 891 & 0 & 0.55 \\
\midrule
\multirow[t]{6}{*}{KIE} & \cord{} & comb & 10 & v1 & 1 & 1200 & 1187 & 1.18 & 1.67 & 27 & 68 & 0.008 & 0.133 \\
 & \cord{} & comb & 10 & v1 & 0.75 & 1200 & 1193 & 1.18 & 1.68 & 29 & 70 & 0.008 & 0.14 \\
 & \cord{} & comb & 10 & v1 & 0.5 & 1200 & 1191 & 1.17 & 1.66 & 29 & 68 & 0.01 & 0.145 \\
\cmidrule(lr){2-14}
 & \cord{} & comb & 10 & v2 & 1 & 1200 & 1182 & 1.19 & 1.66 & 29 & 67 & 0.007 & 0.158 \\
 & \cord{} & comb & 10 & v2 & 0.75 & 1200 & 1174 & 1.19 & 1.66 & 29 & 67 & 0.009 & 0.15 \\
 & \cord{} & comb & 10 & v2 & 0.5 & 1200 & 1185 & 1.18 & 1.67 & 27 & 68 & 0.007 & 0.163 \\
\midrule
\multirow[t]{6}{*}{QA} & \docvqa{} & comb & 10 & v1 & 1 & 10002 & 8463 & 10.6 & 14.63 & 251 & 167 & 2.659 & 0.318 \\
 & \docvqa{} & comb & 10 & v1 & 0.75 & 10014 & 8463 & 10.55 & 14.71 & 252 & 164 & 2.752 & 0.315 \\
 & \docvqa{} & comb & 10 & v1 & 0.5 & 10013 & 8501 & 10.52 & 14.77 & 253 & 162 & 2.847 & 0.307 \\
\cmidrule(lr){2-14}
 & \docvqa{} & comb & 10 & v2 & 1 & 9990 & 8082 & 10.66 & 14.3 & 246 & 173 & 2.37 & 0.31 \\
 & \docvqa{} & comb & 10 & v2 & 0.75 & 10010 & 8345 & 10.59 & 14.53 & 250 & 168 & 2.364 & 0.328 \\
 & \docvqa{} & comb & 10 & v2 & 0.5 & 9990 & 8088 & 10.55 & 14.6 & 251 & 167 & 2.666 & 0.315 \\
\bottomrule
\end{tabular}
}

\caption{Ablation variants generated by varying the sampling strategy (v1/v2) and sampling strength ($\alpha$). The table reports the same statistics as Table~\ref{tab:syn_datasets_details_final}, enabling comparison of how sampling choices impact dataset size, token usage, and content composition.
}

\label{tab:syn_datasets_details_ablation}
\end{table*}

\section{Synthetic Ground Truth}\label{app:synthdata_gt}
We provide qualitative analysis of the ground truth generated by our VLM-based synthesis pipeline across all tasks. The visualizations validate semantic consistency, spatial coherence, and annotation quality of synthetic data compared to real datasets, while also revealing systematic limitations such as class imbalance in classification tasks.

\subsection{QA}\label{app:synthdata_gt_qa}
\begin{figure*}[p]
    \centering
    \begin{subfigure}[b]{0.48\textwidth}
        \centering
        \includegraphics[width=\textwidth,height=0.35\textheight,keepaspectratio]{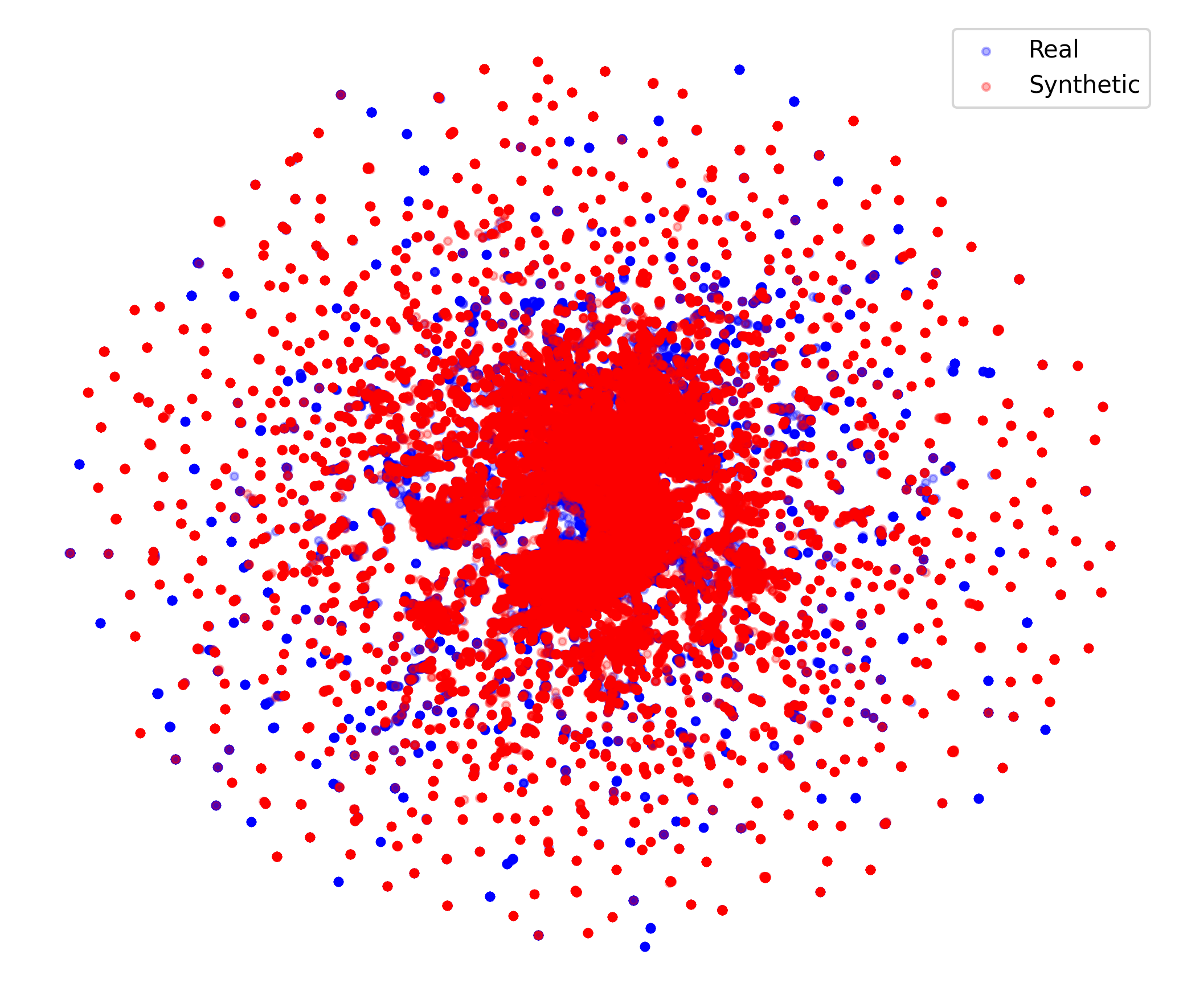}
        \caption{\docvqa{} question embeddings for real and synthetic data overlaid.}
    \end{subfigure}
    \hfill
    \begin{subfigure}[b]{0.48\textwidth}
        \centering
        \includegraphics[width=\textwidth,height=0.35\textheight,keepaspectratio]{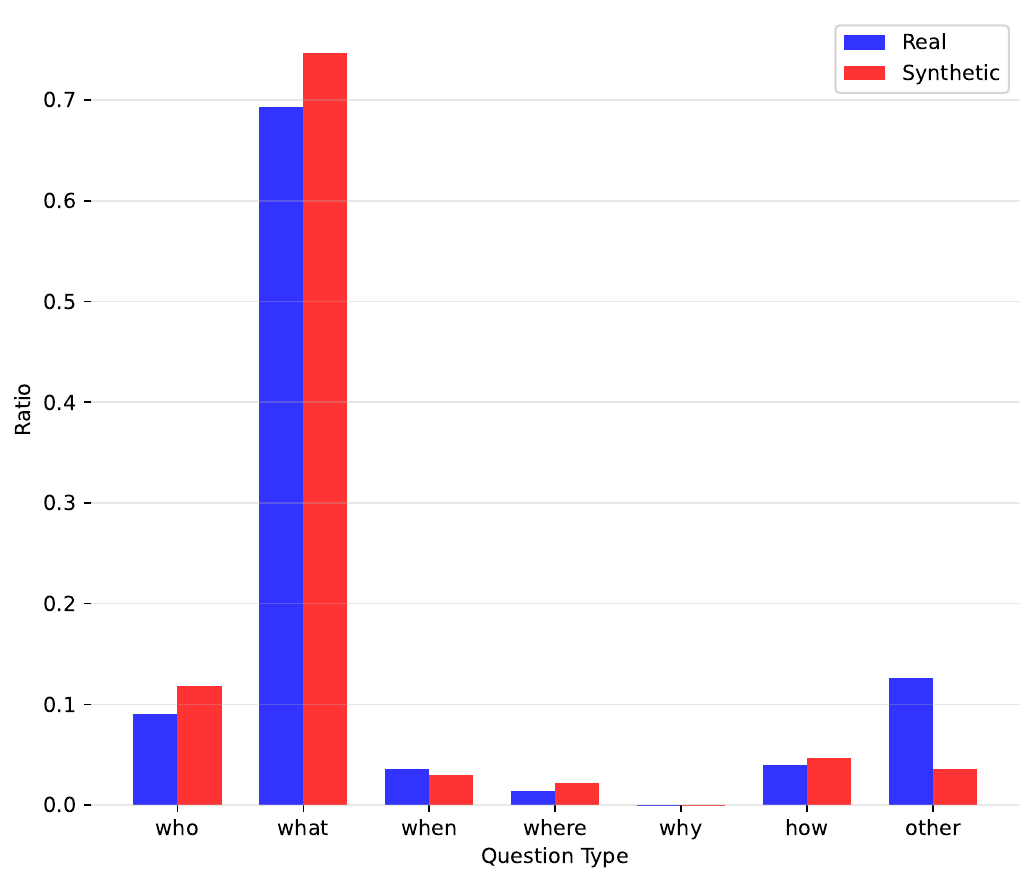}
        \caption{\docvqa{} question type distributions for real and synthetic data.}
    \end{subfigure}
    
    \vspace{1em}
    
    \begin{subfigure}[b]{0.48\textwidth}
        \centering
        \includegraphics[width=\textwidth,height=0.35\textheight,keepaspectratio]{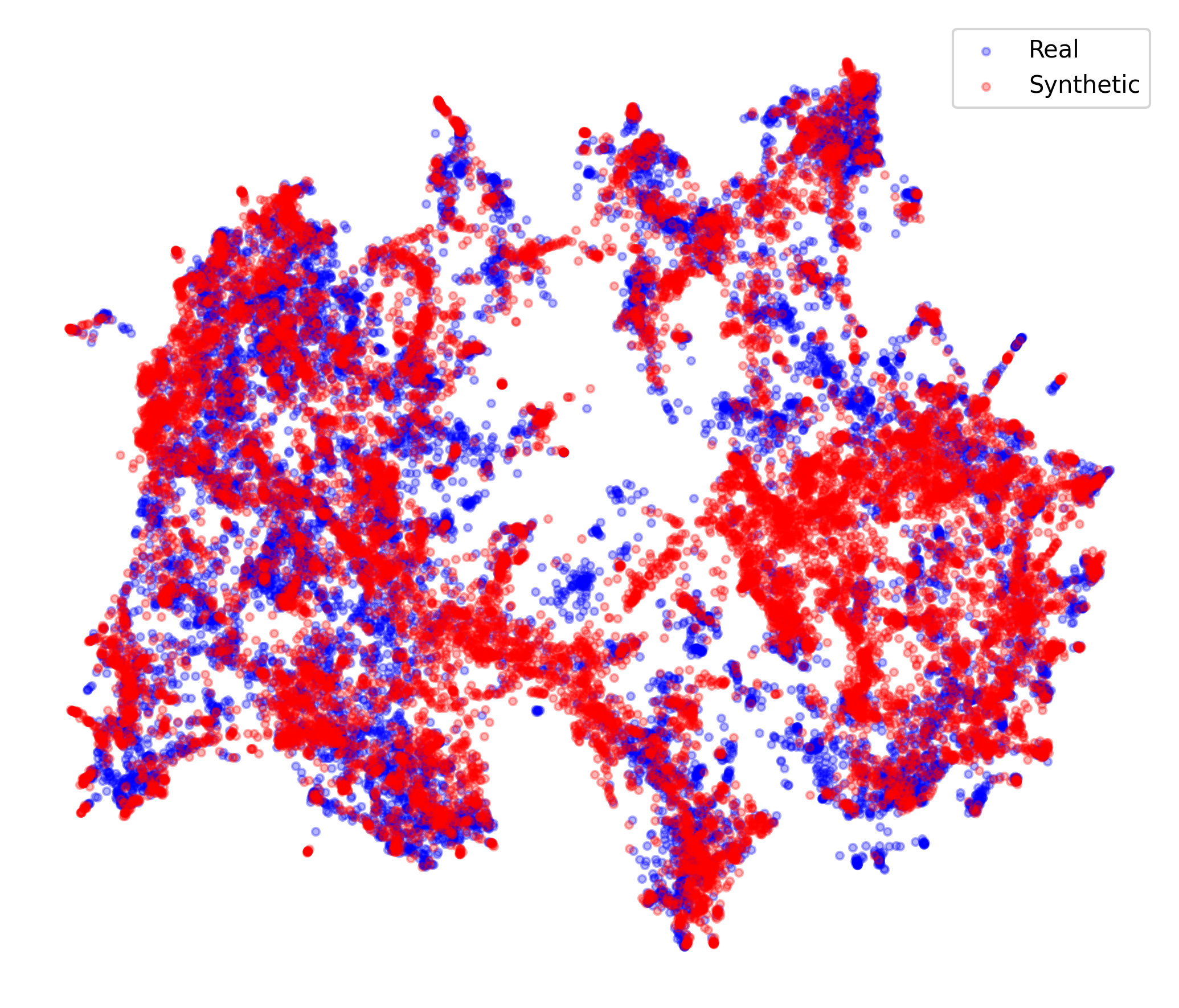}
        \caption{\wiki{} question embeddings for real and synthetic data overlaid.}
    \end{subfigure}
    \hfill
    \begin{subfigure}[b]{0.48\textwidth}
        \centering
        \includegraphics[width=\textwidth,height=0.35\textheight,keepaspectratio]{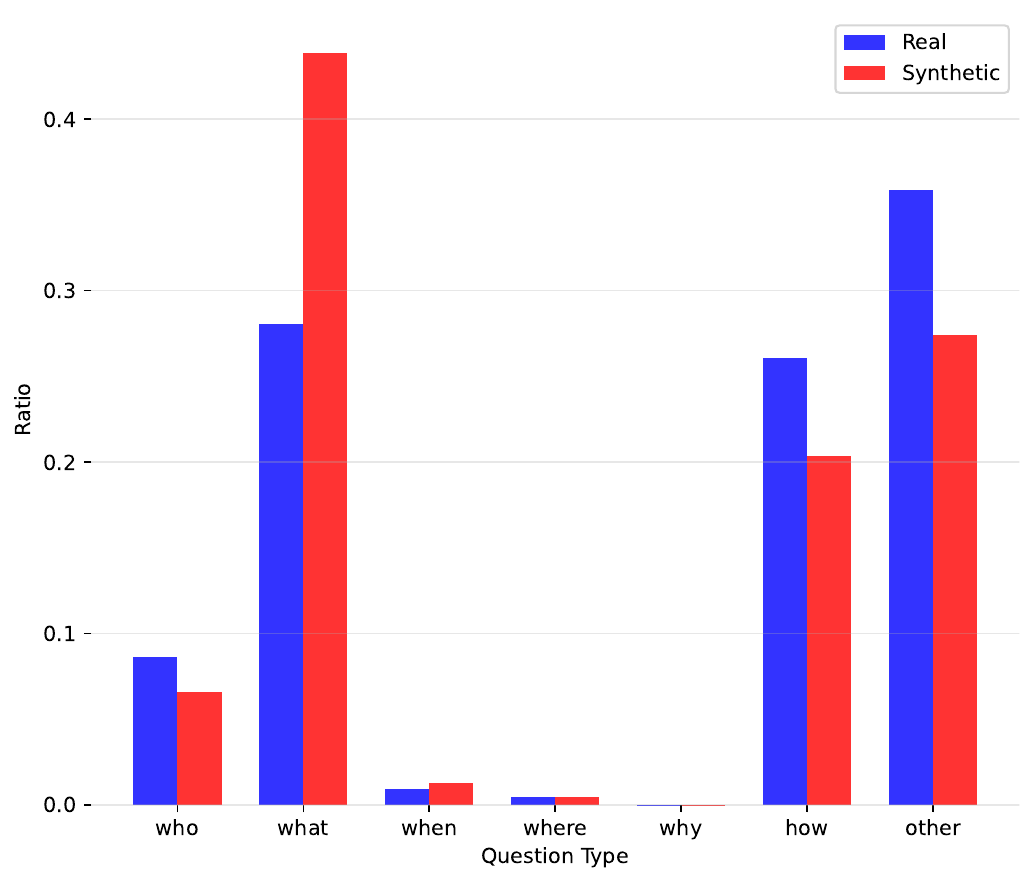}
        \caption{\wiki{} question type distributions for real and synthetic data.}
    \end{subfigure}
    
    \caption{Comparison between real and synthetic data for \docvqa{} (a-b) and \wiki{} (c-d). UMAP embeddings (a, c) show semantic distributions of questions, while bar charts (b, d) compare question type frequencies across datasets.}
\label{fig:qa_analysis}
\end{figure*}

\cref{fig:qa_analysis} shows embeddings of question text and distributions of question types for \docvqa{} and \wiki{}. Question embeddings are retrieved with Sentence Transformers~\cite{RG19} and projected to two dimensions with UMAP~\cite{MHSG18}. Notably, the close alignment between real and synthetic question embeddings demonstrates that our VLM generates semantically similar questions \textit{without ever observing real ground truth annotations}---the synthesis is guided only by seed document images and task-level prompt parameters describing the desired GT structure. Question type distributions are also well-preserved, indicating appropriate task coverage.

\subsection{CLS}\label{app:synthdata_gt_cls}
\begin{figure*}[p]  
    \centering
    \begin{subfigure}[b]{1\textwidth}
        \centering
        \includegraphics[height=0.28\textheight]{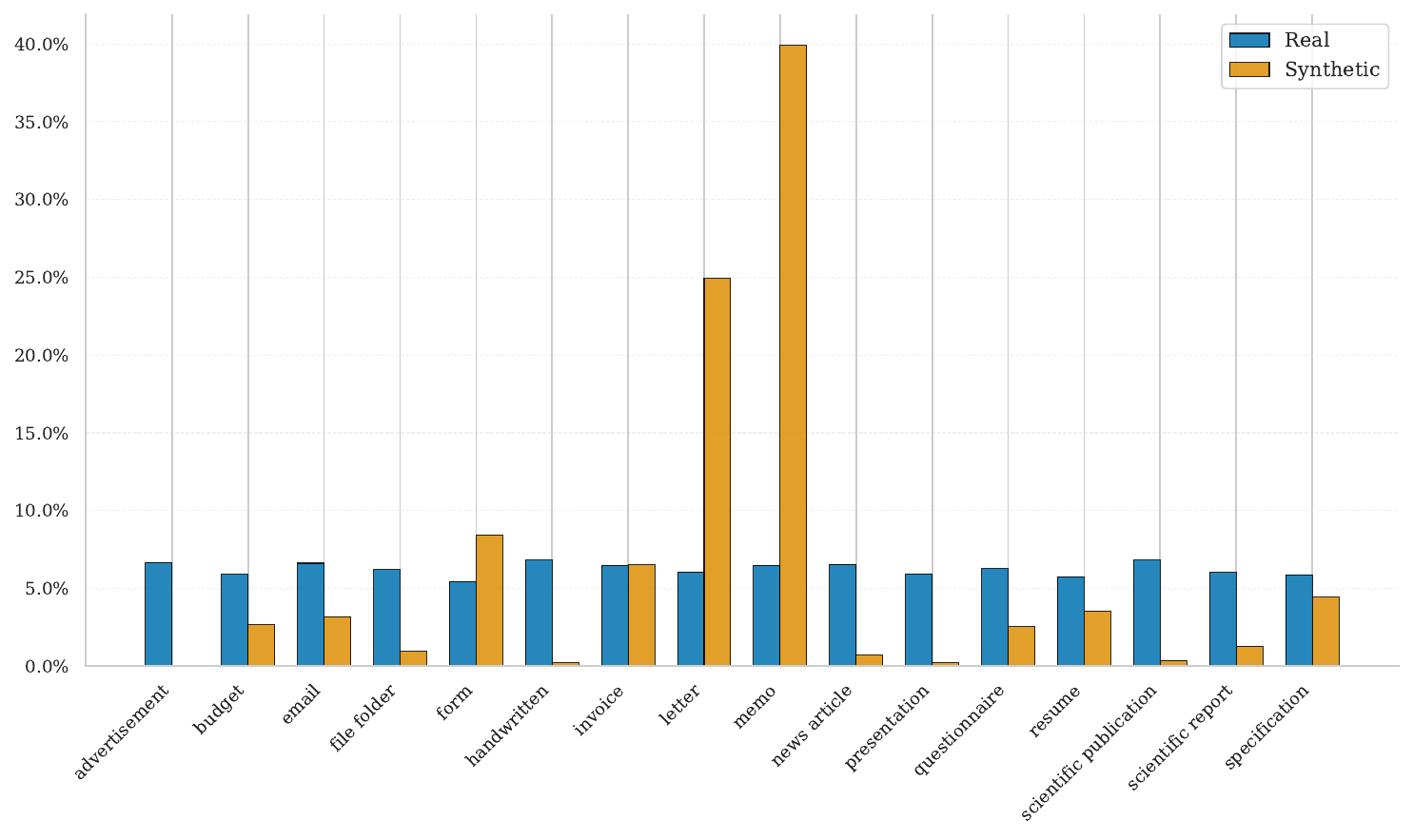}
        \caption{\rvlcdip{}}
    \end{subfigure}
    
    \vspace{2mm}
    
    \begin{subfigure}[b]{1\textwidth}
        \centering
        \includegraphics[height=0.28\textheight]{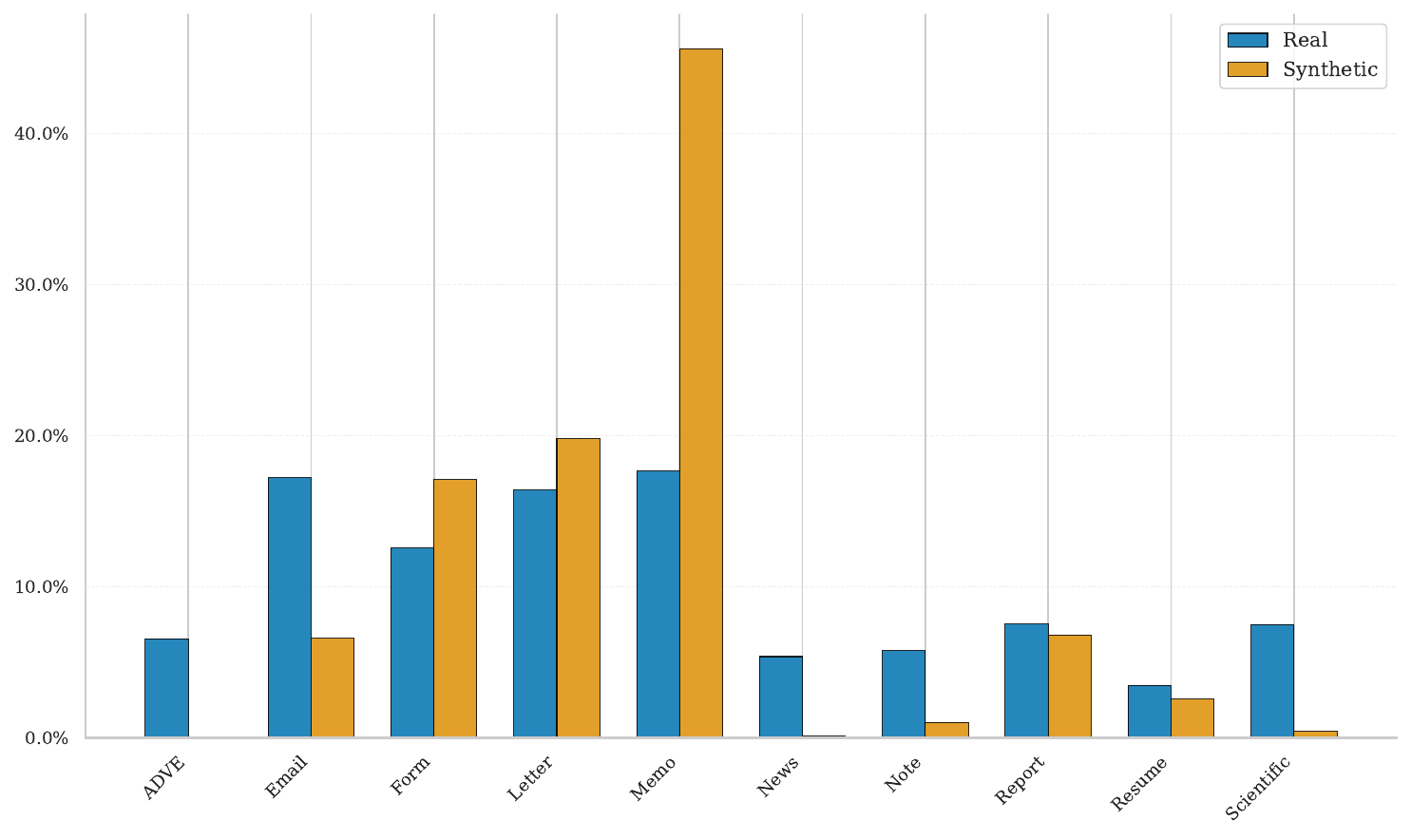}
        \caption{\tobacco{}}
    \end{subfigure}
    
    \vspace{2mm}
    
    \begin{subfigure}[b]{1\textwidth}
        \centering
        \includegraphics[height=0.28\textheight]{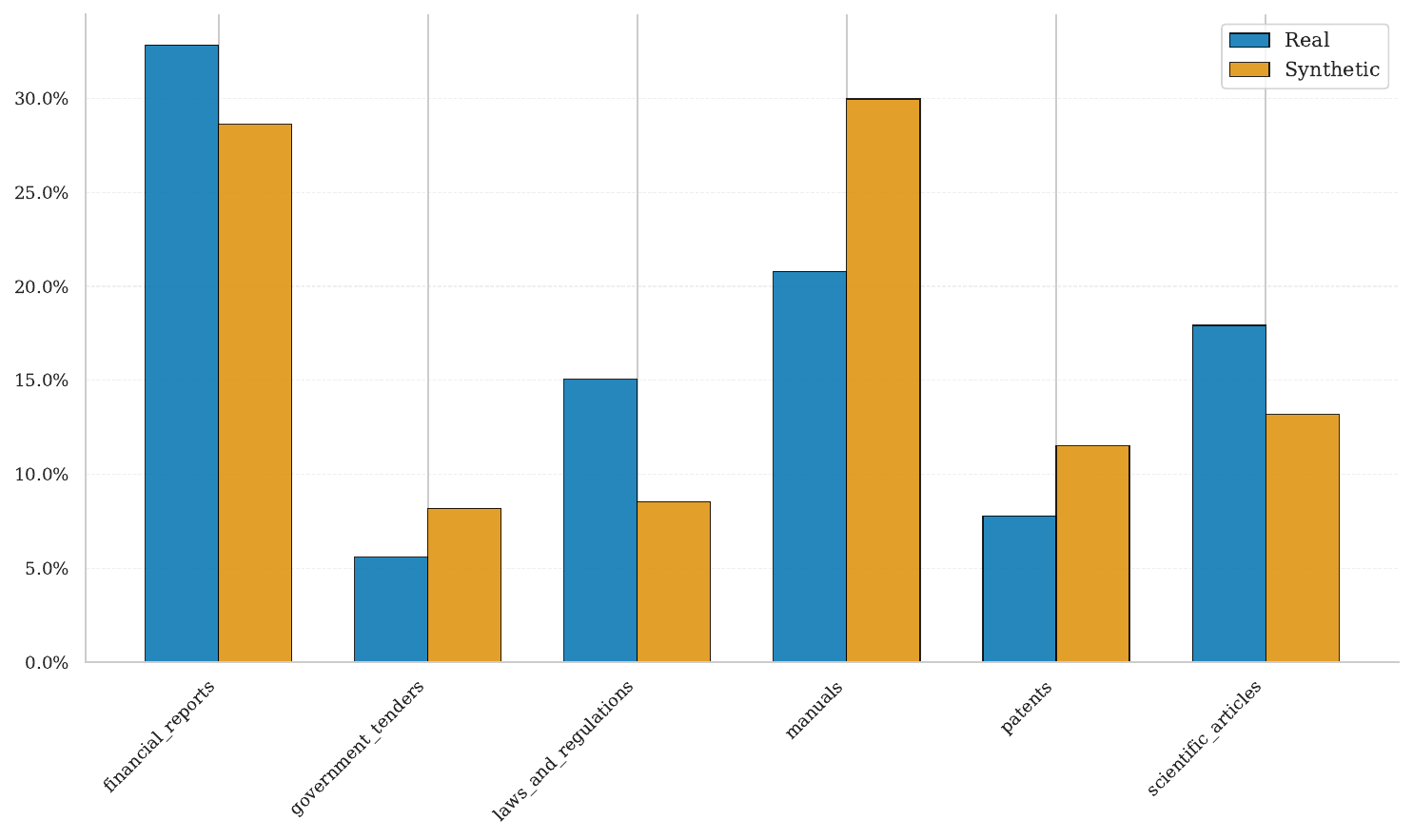}
        \caption{\doclaynetcls{}}
    \end{subfigure}
    
    \caption{Comparisons of class distributions between real and synthetic data for \rvlcdip{} (a), \tobacco{} (b) and \doclaynetcls{} (c).}
    \label{fig:cls_distributions}
\end{figure*}

\cref{fig:cls_distributions} presents class distributions for \rvlcdip{}, \tobacco{}, and \doclaynetcls{} classification datasets. The severe class imbalance in synthetic data is evident, with certain classes (e.g., memo) heavily overrepresented while specialized classes remain undersampled. This imbalance reflects both our intra-cluster sampling strategy with $\alpha=1$ that biases generation toward dominant document patterns, and the VLM's tendency to more readily generate certain document types over others. This imbalance contributes to the performance gap observed in Section~\ref{sec:failure_cases}.

\subsection{KIE}\label{app:synthdata_gt_kie}
\begin{figure*}[p]
    \setlength{\imgheight}{1\textheight}
    \centering

    \includegraphics[height=\imgheight,width=\textwidth,keepaspectratio]{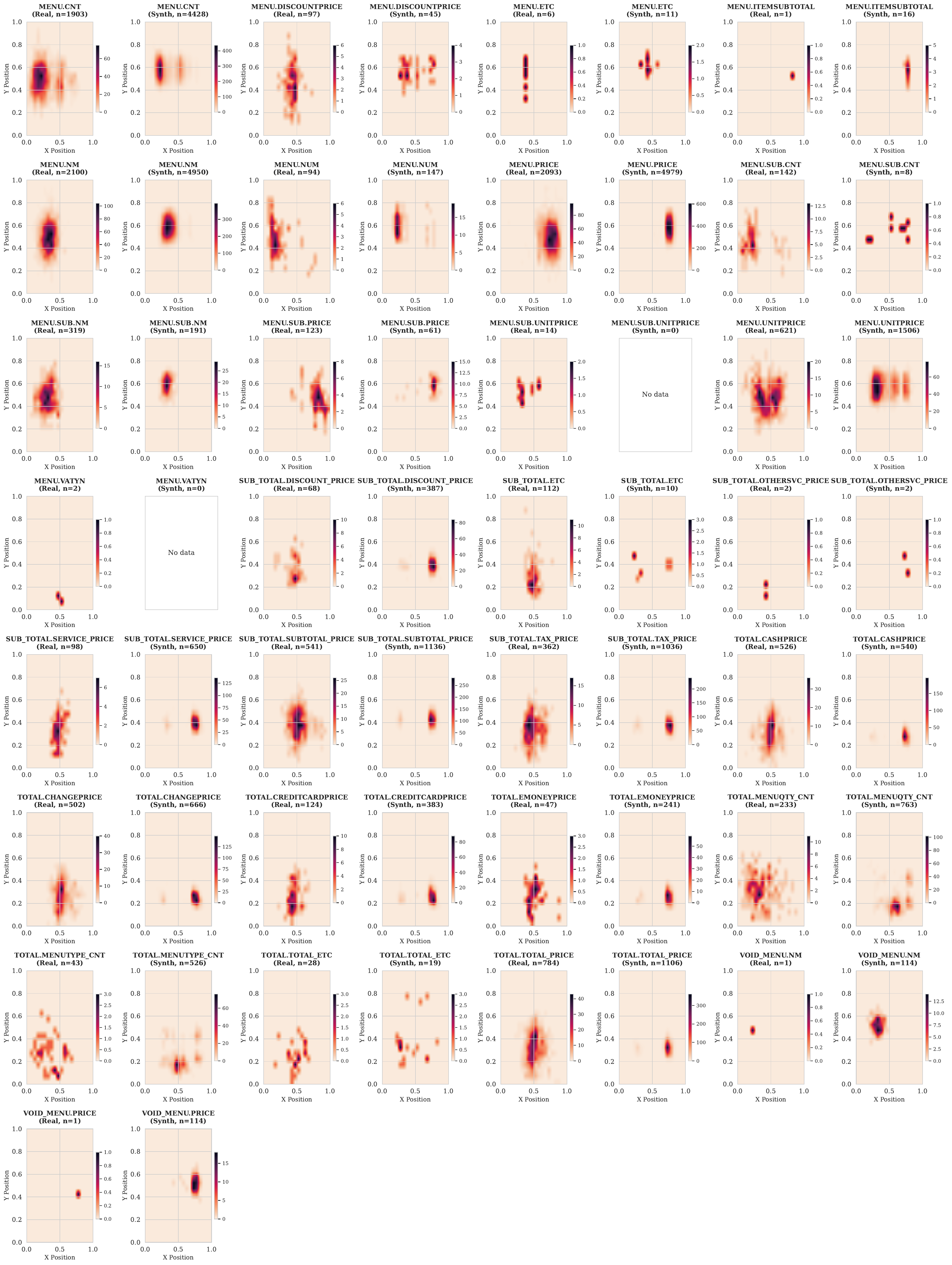}
    \caption{Heatmaps comparing entity placement by type between real and synthetic for \cord{}. 
        \label{fig:heatmap_cord}}
\end{figure*}

\begin{figure*}[h] 
    \centering
\begin{subfigure}[b]{1\textwidth}
        \centering
        \includegraphics[width=1\textwidth]{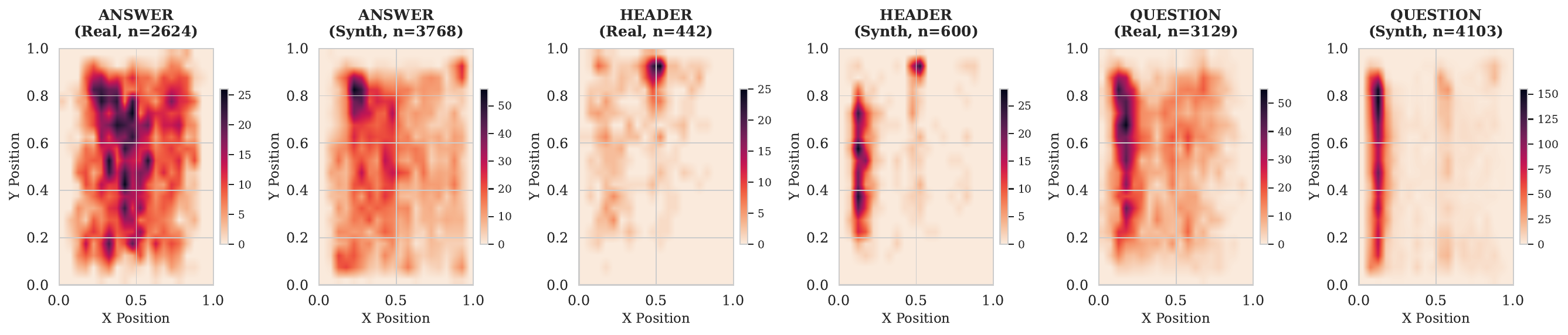}
        \caption{\funsd{}}
    \end{subfigure} 

    \vspace{2mm}
    \begin{subfigure}[b]{1\textwidth}
        \centering
        \includegraphics[width=1\textwidth]{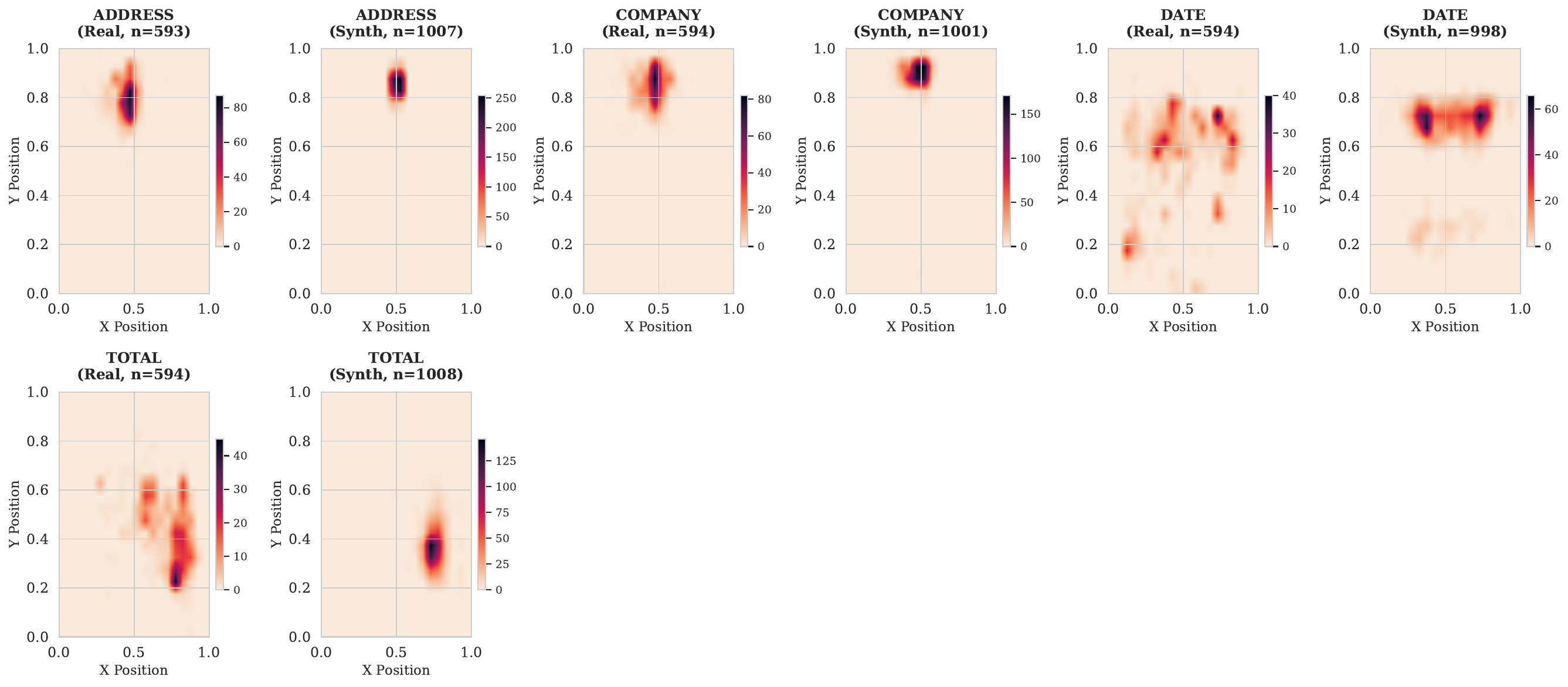}
        \caption{\sroie{}}
    \end{subfigure}

    \caption{Heatmaps comparing entity placement by type between real and synthetic data for \funsd{} (a) and \sroie{} (b).}
    \label{fig:heatmap_funsd_sroie}
\end{figure*}

\cref{fig:heatmap_cord,fig:heatmap_funsd_sroie} show spatial heatmaps comparing entity placement by type between real and synthetic data for \cord{}, \funsd{} and \sroie{}. The heatmaps demonstrate that spatial distributions of key-value entities are well-preserved, validating the quality of element-level annotations generated by our micro template without access to real annotations.

\begin{figure*}[h] 
    \centering
\begin{subfigure}[b]{0.45\textwidth}
        \centering
        \includegraphics[height=0.25\textheight,width=\linewidth,keepaspectratio]{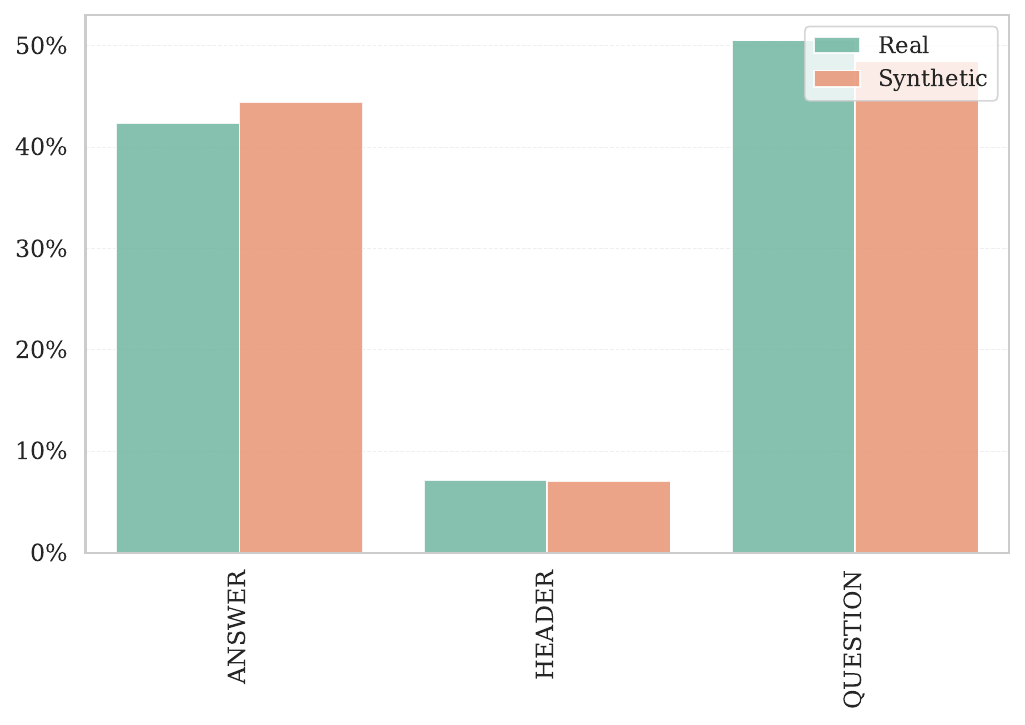}
        \caption{\funsd{}}
    \end{subfigure} \hfill
    \begin{subfigure}[b]{0.45\textwidth}
        \centering
        \includegraphics[height=0.25\textheight,width=\linewidth,keepaspectratio]{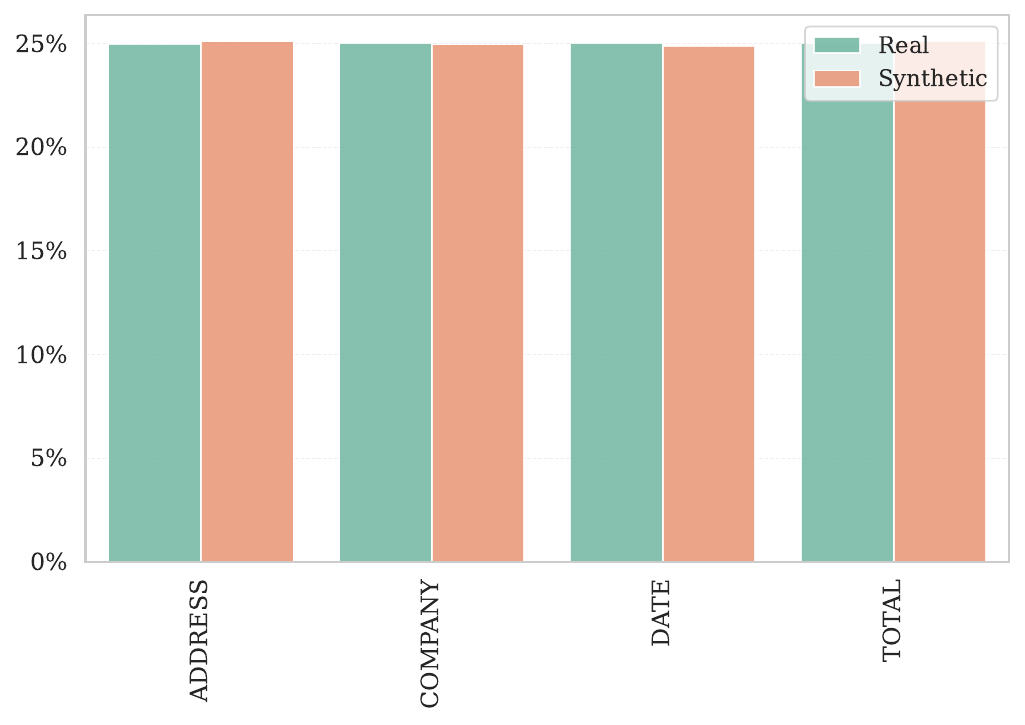}
        \caption{\sroie{}}
    \end{subfigure}

    \vspace{2mm}
    \begin{subfigure}[b]{0.95\textwidth}
        \centering
        \includegraphics[height=0.25\textheight]{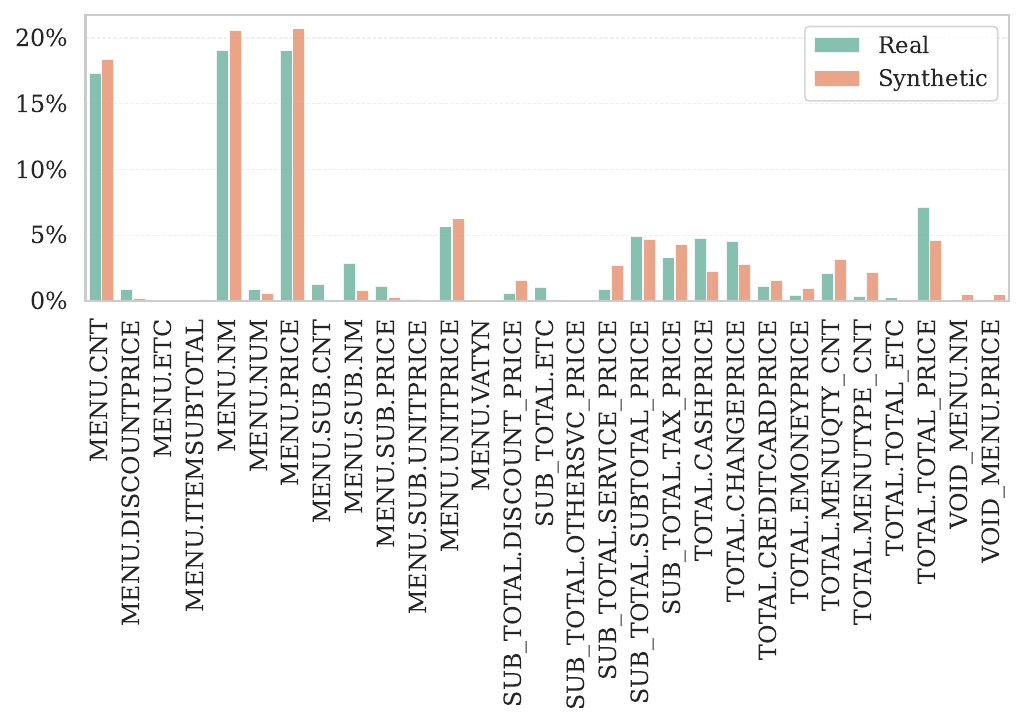}
        \caption{\cord{}}
    \end{subfigure}

    \caption{Comparison of entity type distributions between real and synthetic KIE datasets.}
    \label{fig:synth_gt_kie_distributions}
\end{figure*}

\subsection{DLA}\label{app:synthdata_gt_dla}

\begin{figure*}[htbp]
    \centering
    \begin{subfigure}[b]{0.48\textwidth}
        \centering
        \includegraphics[width=\textwidth,height=0.35\textheight,keepaspectratio]{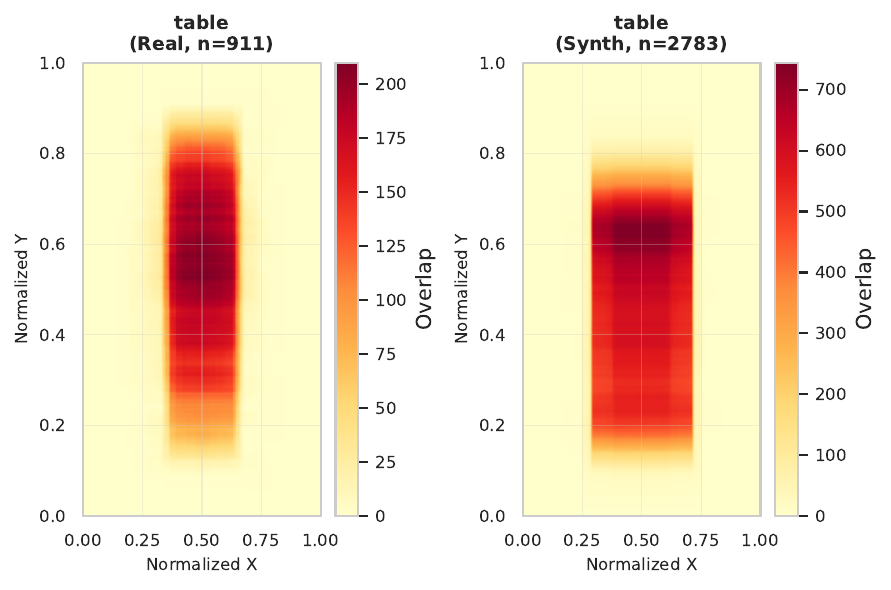}
        \caption{\icdarctdar{} spatial heatmaps.}
    \end{subfigure}
    \hfill
    \begin{subfigure}[b]{0.48\textwidth}
        \centering
        \includegraphics[width=\textwidth,height=0.35\textheight,keepaspectratio]{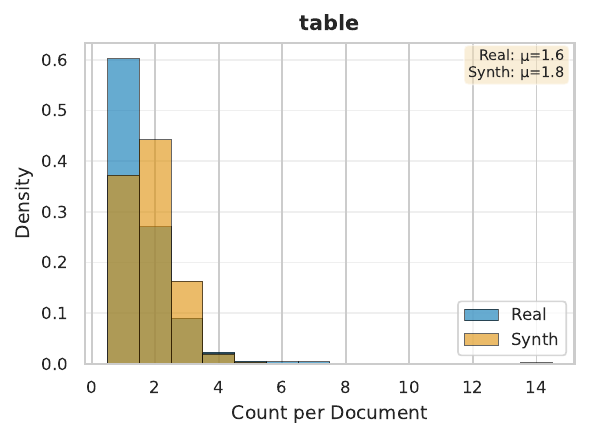}
        \caption{\icdarctdar{} region counts.}
    \end{subfigure}
    
    \vspace{1em}
    
    \begin{subfigure}[b]{0.98\textwidth}
        \centering
        \includegraphics[width=\textwidth,height=0.35\textheight,keepaspectratio]{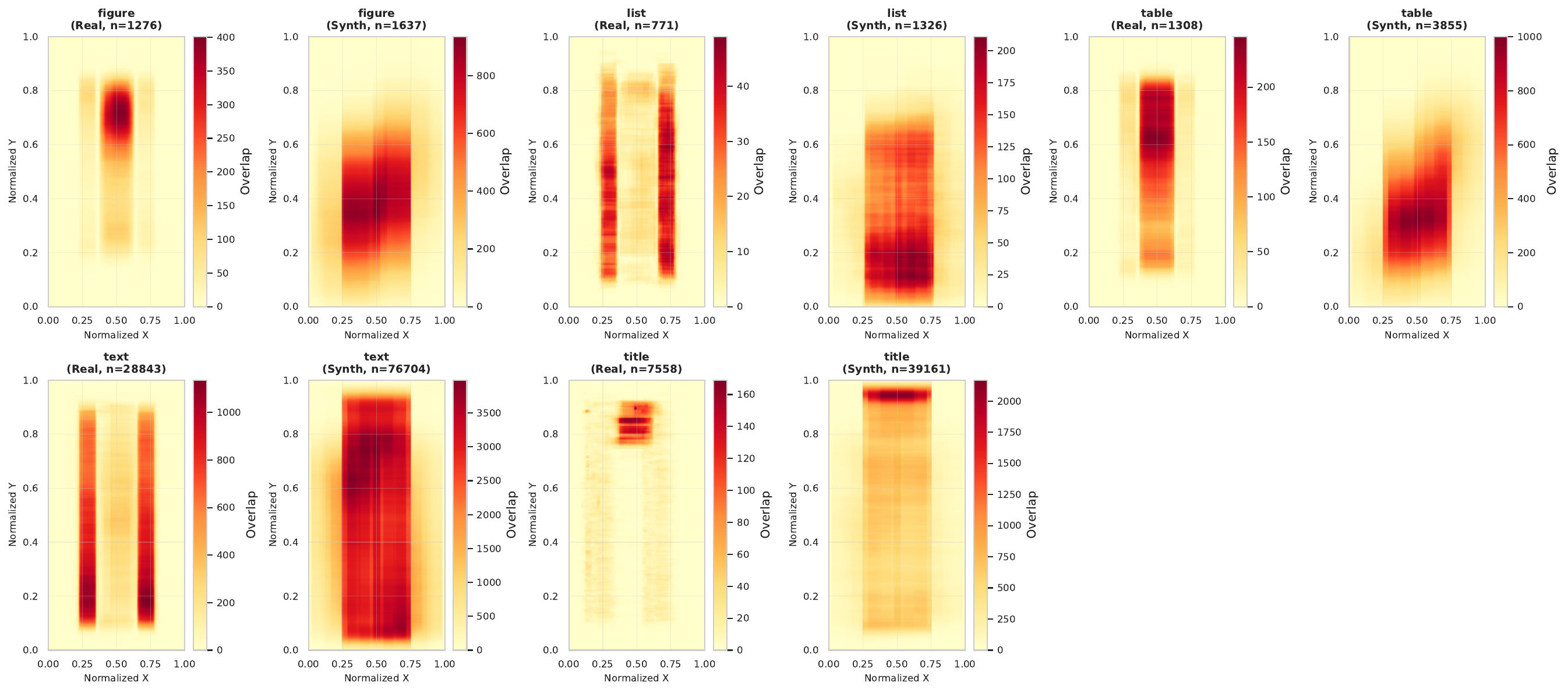}
        \caption{\publaynet{} spatial heatmaps.}
    \end{subfigure}
    \begin{subfigure}[b]{0.98\textwidth}
        \centering
        \includegraphics[width=\textwidth,height=0.35\textheight,keepaspectratio]{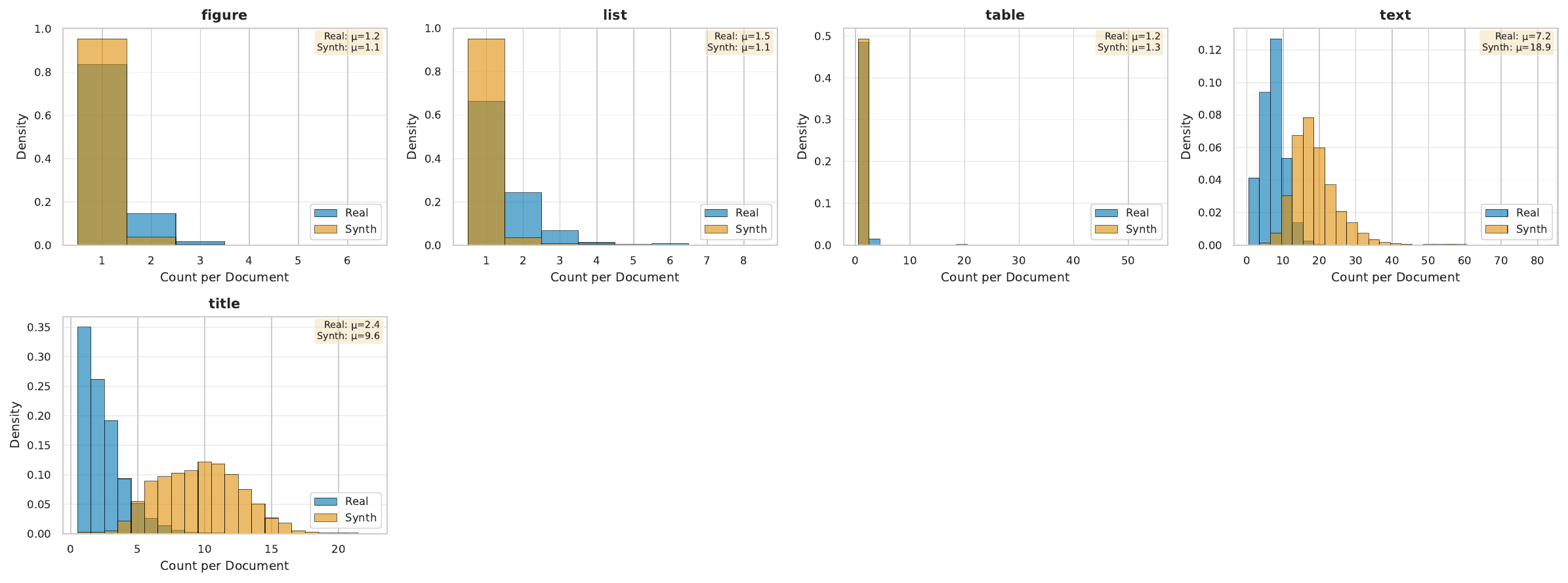}
        \caption{\publaynet{} region counts.}
    \end{subfigure}
    
    \caption{Comparison of annotation spatial heatmaps and class distributions for real and synthetic versions of \icdarctdar{} (a, b) and \publaynet{} (c, d).}
\label{fig:dla_analysis_icdar_publaynet}
\end{figure*}

\begin{figure*}[htbp]
    \centering
    \begin{subfigure}[b]{0.98\textwidth}
        \centering
        \includegraphics[width=\textwidth,height=0.45\textheight,keepaspectratio]{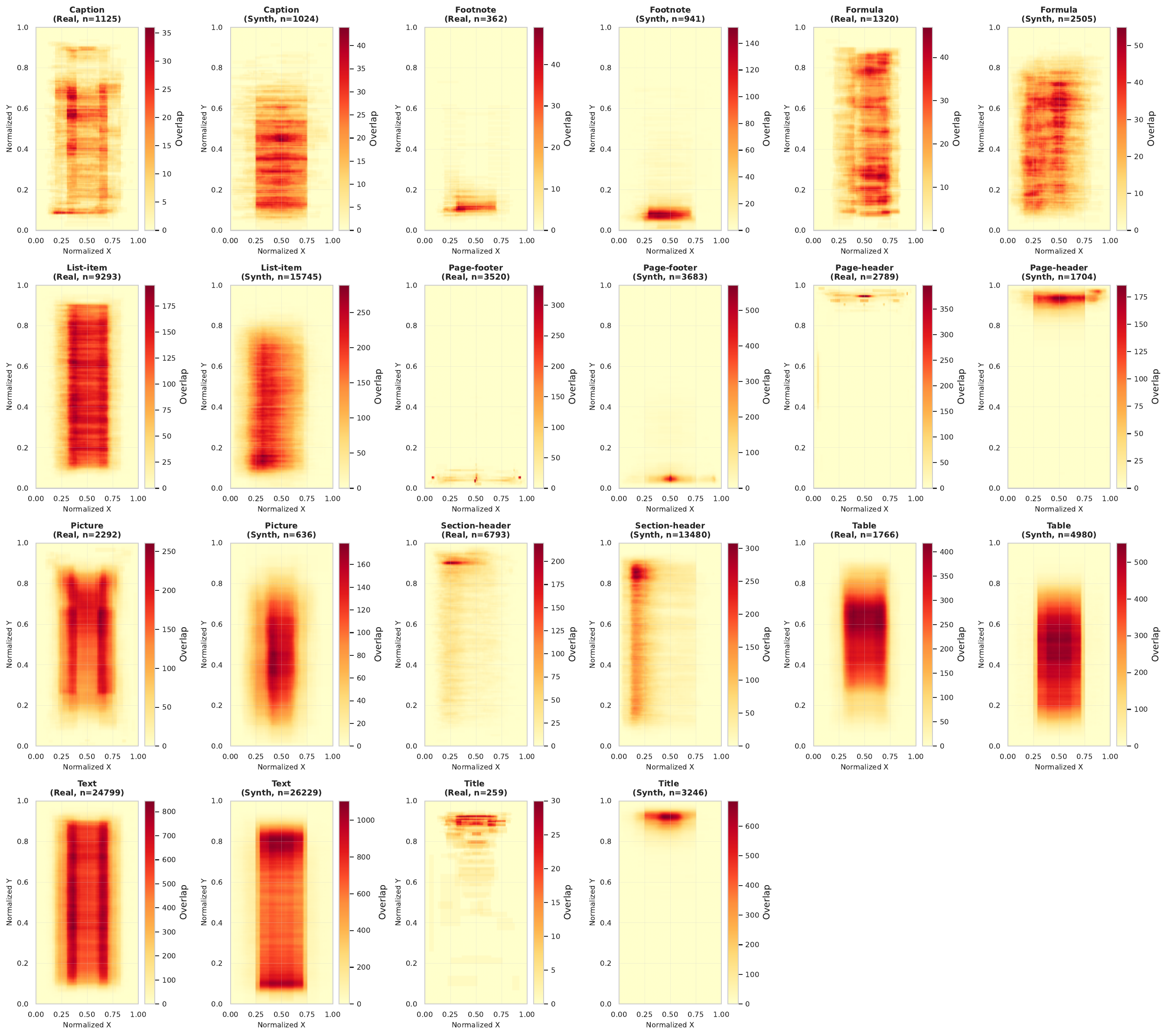}
        \caption{\doclaynetdla{} spatial heatmaps.}
    \end{subfigure}

    \begin{subfigure}[b]{0.98\textwidth}
        \centering
        \includegraphics[width=\textwidth,height=0.45\textheight,keepaspectratio]{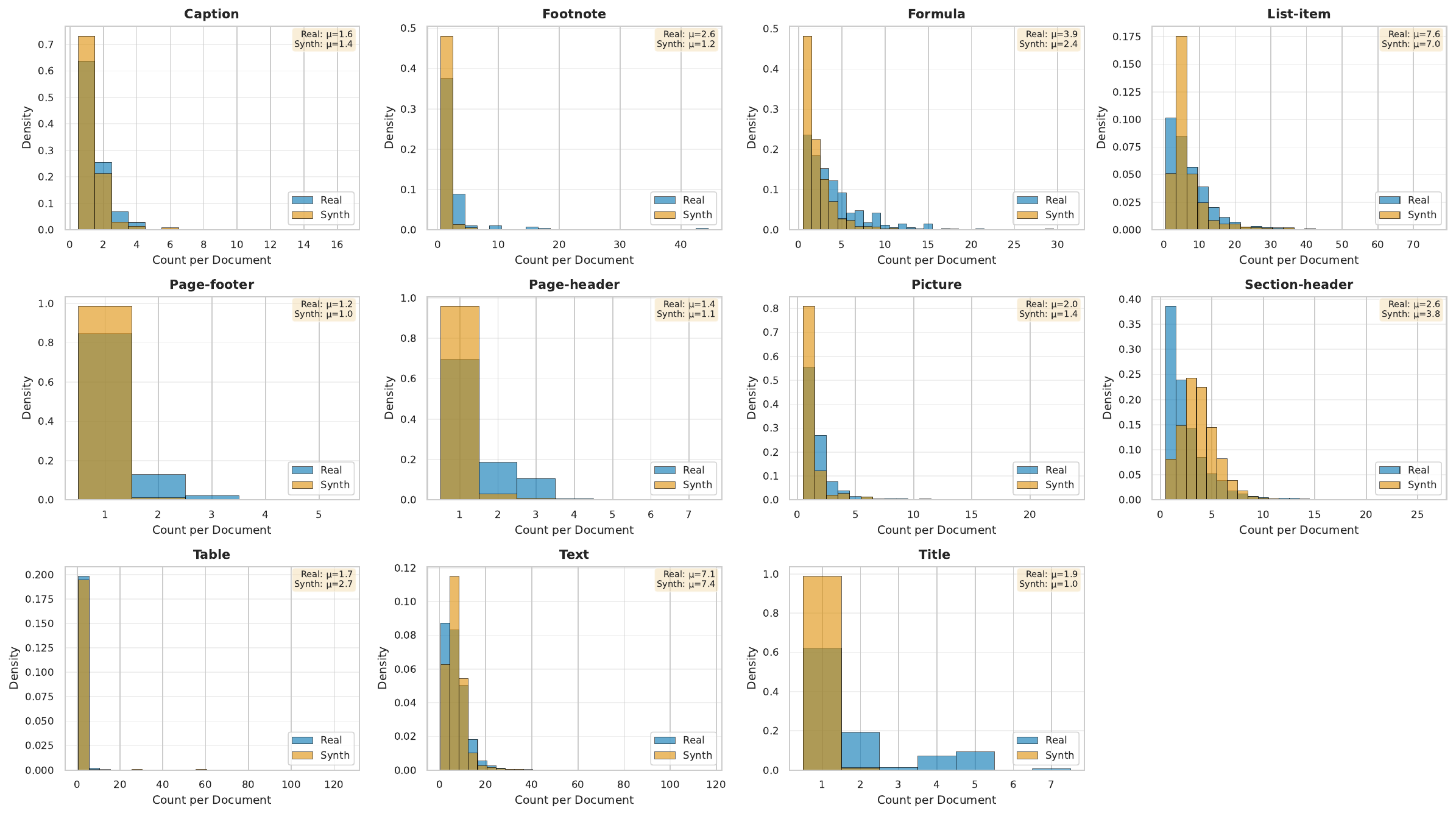}
        \caption{\doclaynetdla{} region counts.}
    \end{subfigure}

    \caption{Comparison of annotation spatial heatmaps (top) and class distributions (bottom) for real and synthetic versions of \doclaynetdla{}.}
\label{fig:dla_analysis_doclaynet}
\end{figure*}

\cref{fig:dla_analysis_icdar_publaynet,fig:dla_analysis_doclaynet} present spatial heatmaps and region count distributions for \icdarctdar{}, \publaynet{} and \doclaynetdla{}. 
While overall spatial layout patterns appear reasonable with region counts and positioning comparable to real data, detailed analysis of \doclaynetdla{} predictions reveals both annotation taxonomy differences and synthesis limitations that explain the low quantitative scores: definitional mismatches in list-item classification (distance-based vs. semantic, see \cref{fig:doclaynet_failure_cases_a}), insufficient variety in visual elements to capture complex images with embedded text (\cref{fig:doclaynet_failure_cases_b}), limited table diversity in size and structure (\cref{fig:doclaynet_failure_cases_c}), systematic labeling differences (e.g., chemistry formulas as "Formula" in synthetic data vs. "Picture" in real data) (\cref{fig:doclaynet_failure_cases_d}), positional biases (top-left elements consistently labeled "Title" in synthetic data, while in real data they are often labeled as "Page Header" or "Section Header"), and near-complete failure on the "Caption" class. 
These findings indicate that performance gaps stem from both annotation inconsistencies and limitations in synthetic document diversity.

\newlength{\doclaynetdlafcwidth}
\setlength{\doclaynetdlafcwidth}{0.49\textwidth}

\newlength{\doclaynetdlafcheight}
\setlength{\doclaynetdlafcheight}{0.39\textheight}

    \begin{figure*}[htbp]
    \centering
        \includegraphics[width=\linewidth,height=\doclaynetdlafcheight,keepaspectratio]{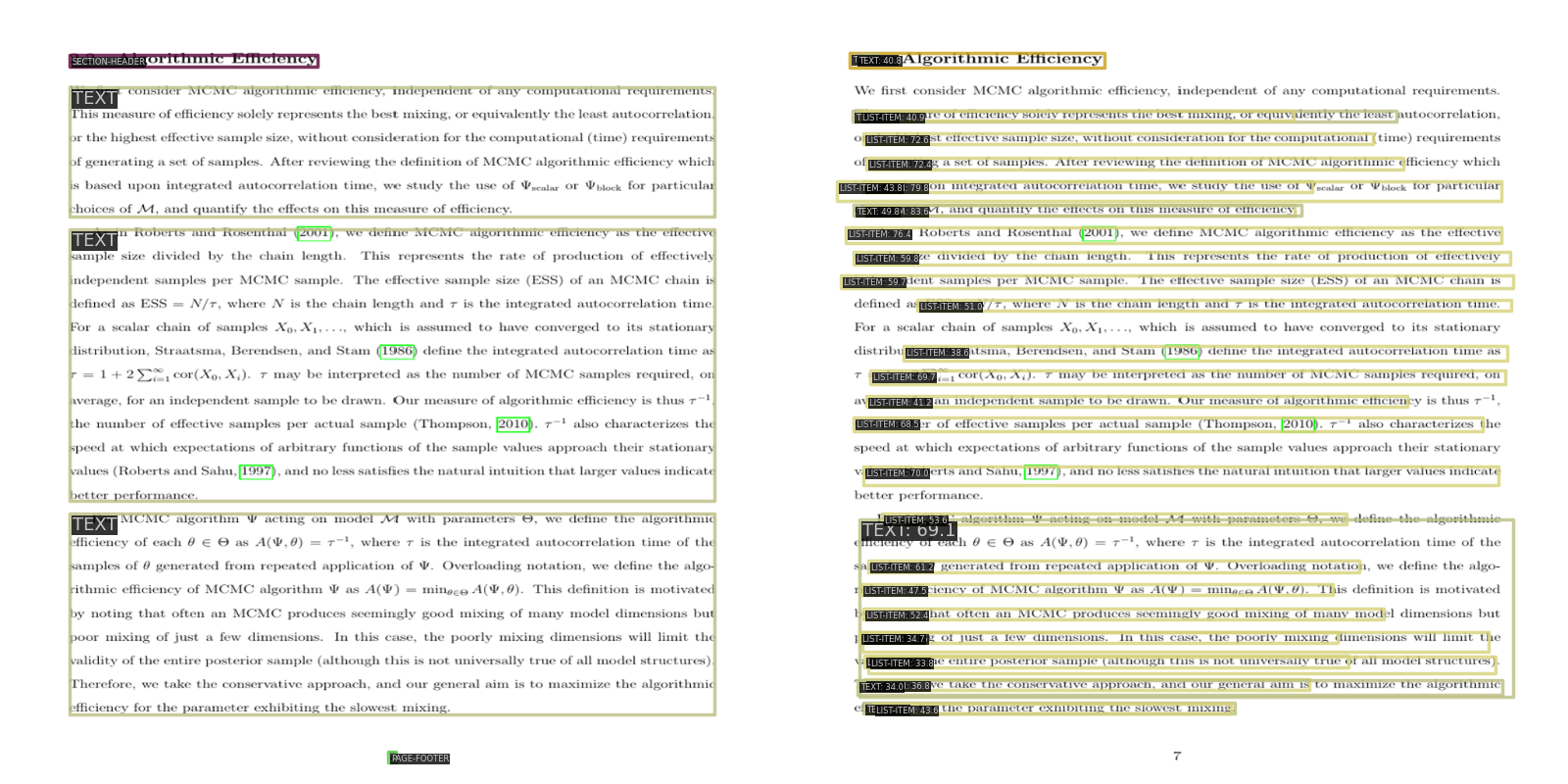}
        \caption{\doclaynetdla{} GT (left) compared to predictions of \fasterrcnn{} trained on purely synthetic data (right). Poor performance likely caused by definitional mismatches in list-item classification (distance-based vs. semantic).}\label{fig:doclaynet_failure_cases_a}
    \end{figure*}%

    \begin{figure*}[htbp]
    \centering
        \includegraphics[width=\linewidth,height=\doclaynetdlafcheight,keepaspectratio]{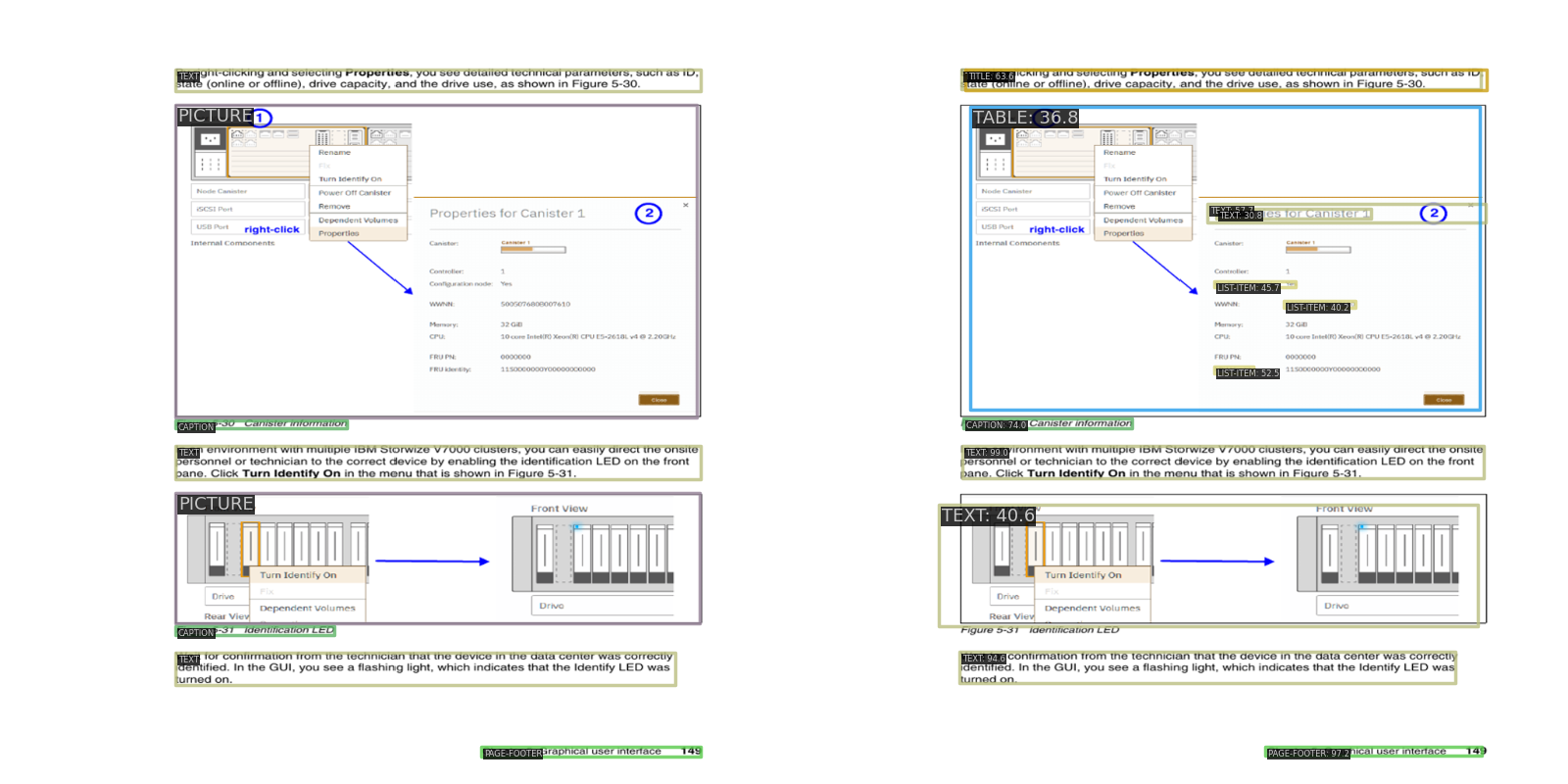}
        \caption{\doclaynetdla{} GT (left) compared to predictions of \fasterrcnn{} trained on purely synthetic data (right). Poor performance likely caused by insufficient variety in visual elements to capture complex images with embedded text.}\label{fig:doclaynet_failure_cases_b}
    \end{figure*}%
   
    \begin{figure*}[htbp]
    \centering
        \includegraphics[width=\linewidth,height=\doclaynetdlafcheight,keepaspectratio]{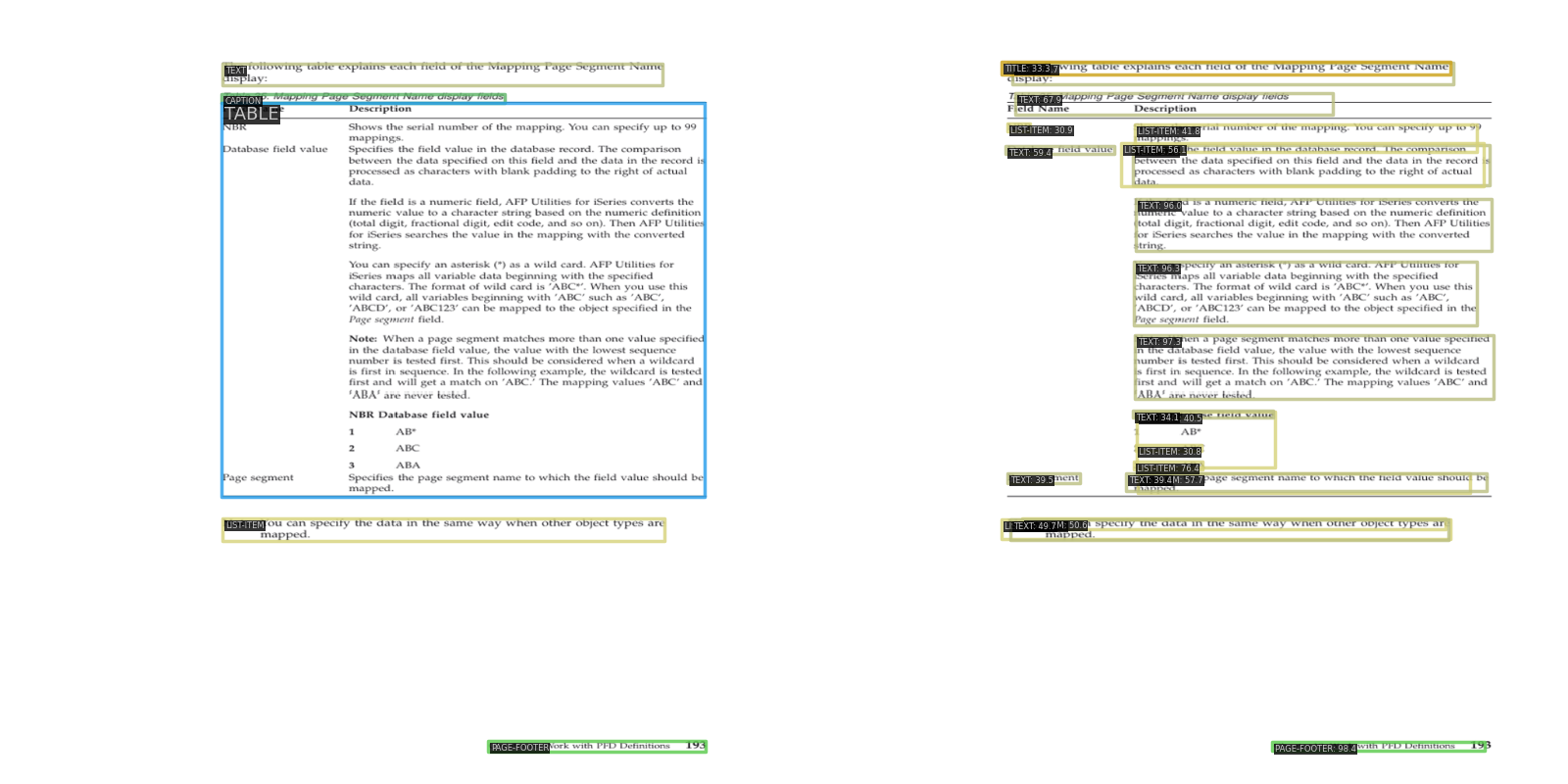}
        \caption{\doclaynetdla{} GT (left) compared to predictions of \fasterrcnn{} trained on purely synthetic data (right). Poor performance likely caused by limited table diversity in size and structure in our synthetic data. Our model fails to annotate the region as a single table.}\label{fig:doclaynet_failure_cases_c}
    \end{figure*}%

    \begin{figure*}[htbp]
    \centering
        \includegraphics[width=\linewidth,height=\doclaynetdlafcheight,keepaspectratio]{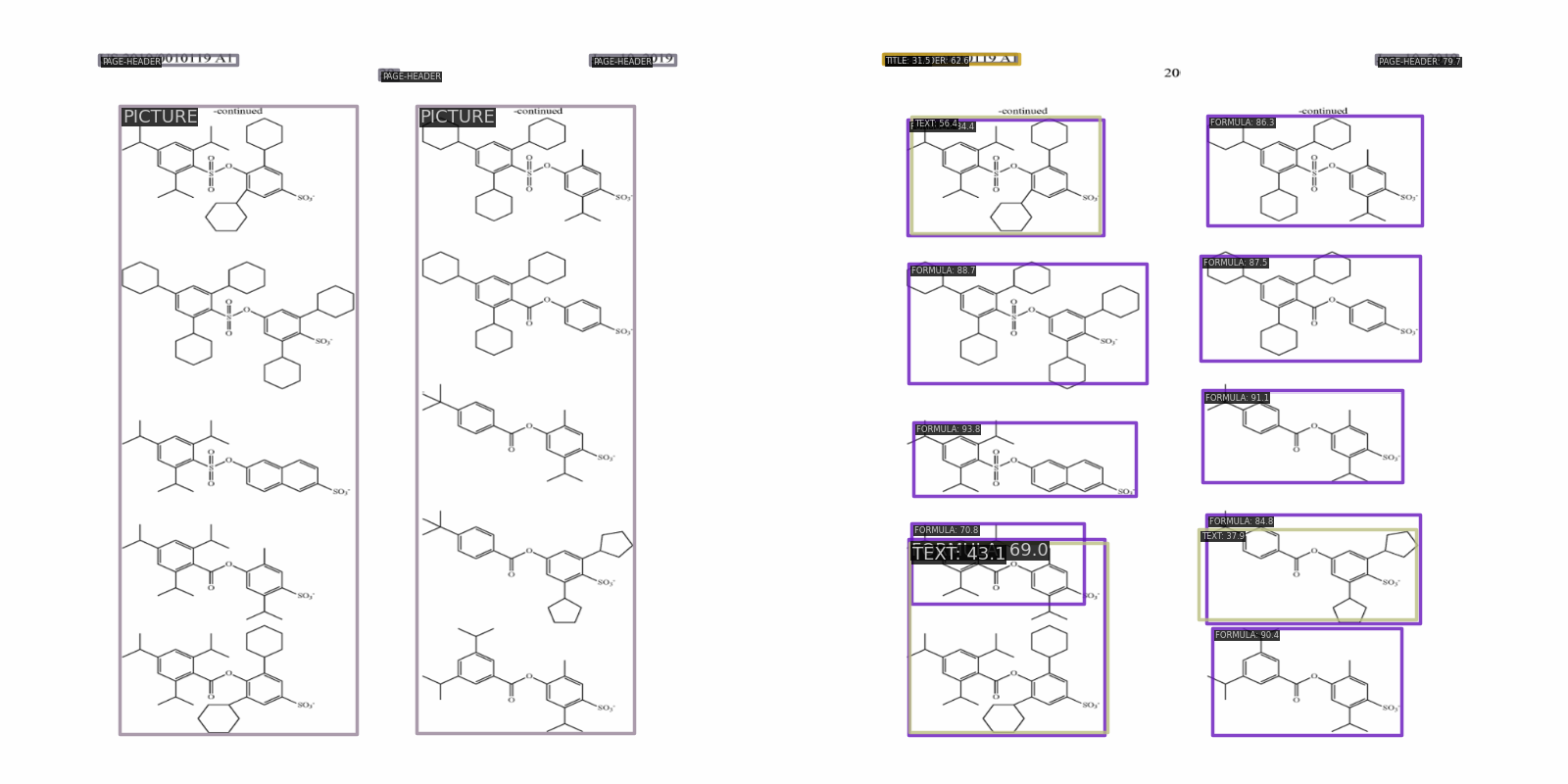}
        \caption{\doclaynetdla{} GT (left) compared to predictions of \fasterrcnn{} trained on purely synthetic data (right). Poor performance likely caused by insufficient detailed GT specification. Real GT annotates the formulas as "Picture", while our model annotates them as "Formula".}\label{fig:doclaynet_failure_cases_d}
    \end{figure*}%

\section{Evaluation Setup}

\label{app:training_hyperparameters} 
\cref{tab:hyperparams_all} summarizes the default training configurations for all models and tasks used in our experiments. For each task (CLS, KIE, VQA, DLA) and model combination, we report learning rate, batch size, number of epochs, optimizer settings, regularization parameters, and other training details. These hyperparameters were selected based on preliminary experiments and follow common practices for document understanding tasks. 

\begin{sloppypar}
For training experiments in tasks CLS, KIE, and VQA, we fine-tune the \bert{}, \lilt{}, and \layoutlm{} models using their pretrained checkpoints:
\texttt{bert-base-uncased},
\texttt{SCUT-DLVCLab/\-lilt-roberta-en-base}, and
\texttt{micro\-soft/\-layoutlmv3-base}\footnotemark[1]\footnotetext[1]{We use the pretrained checkpoints available at \url{https://huggingface.co/} for these models.}, respectively. For the DLA task, we use the MMDetection\footnotemark[2]\footnotetext[2]{\url{https://mmdetection.readthedocs.io/}} library to fine-tune the models using their available pretrained checkpoints.
\end{sloppypar}

To reduce training time and computational overhead, we adopt early stopping\footnotemark[3]\footnotetext[3]{Mahsereci, Maren, et al. "Early stopping without a validation set." arXiv preprint arXiv:1703.09580 (2017).} with a patience of 10 wherever specified. Specifically, we evaluate the model on the validation set after every training epoch, and if the target metric does not exceed its best value for 10 consecutive epochs, we terminate training early.
However, for training configurations Few\textsubscript{A} (R) and Few\textsubscript{B} (R), since the dataset sizes are extremely small (100, 300, or 1000 samples), we disable early stopping and train for the full number of epochs for a fair comparison. Across all experiments, we use the validation set to select the best checkpoint and report test performance using the checkpoint that achieves the highest validation score.

\begin{table*}[!t]
\centering

\resizebox{\textwidth}{!}{%
\rowcolors{2}{gray!15}{white} 

\begin{tabular}{l l l c c c c c c c c c c c c c c c}
\textbf{Task} & \textbf{Model} & \textbf{Modality} & \rotatebox{90}{\textbf{Learning Rate}} & \rotatebox{90}{\textbf{Batch Size}} & \rotatebox{90}{\textbf{Epochs}} & \rotatebox{90}{\textbf{Optimizer}} & \rotatebox{90}{\textbf{Weight Decay}} & \rotatebox{90}{\textbf{Momentum}} & \rotatebox{90}{\textbf{Warmup Ratio}} & \rotatebox{90}{\textbf{Dropout}} & \rotatebox{90}{\textbf{Segment-Level Layout}} & \rotatebox{90}{\textbf{Early Stopping}} & \rotatebox{90}{\textbf{Mixed Precision}} & \rotatebox{90}{\textbf{Image Size}} \\
\midrule
 \cellcolor{white}& \bert{} & T & 1.00E-05 & 32 & 50 & Adam & 0.01 & N/A & 0.1 & 0.1 & \xmark & \cmark & \xmark & N/A \\
                     \cellcolor{white}& \lilt{} & T+L & 1.00E-05 & 32 & 50 & Adam & 0.01 & N/A & 0.1 & 0.1 & \xmark & \cmark & \xmark & N/A \\
                     \cellcolor{white}\multirow{-3}{*}{CLS} & \layoutlm{} & T+L+I & 1.00E-05 & 32 & 50 & Adam & 0.01 & N/A & 0.1 & 0.1 & \cmark & \cmark & \xmark & 224×224 \\
\midrule
 \cellcolor{white}& \bert{} & T & 2.00E-05 & 16 & 100 & AdamW & 0.01 & N/A & 0.1 & 0.1 & \xmark & \xmark & \xmark & N/A \\
                     \cellcolor{white}& \lilt{} & T+L & 2.00E-05 & 16 & 100 & AdamW & 0.01 & N/A & 0.1 & 0.1 & \xmark & \xmark & \xmark & N/A \\
                     \cellcolor{white}\multirow{-3}{*}{KIE}& \layoutlm{} & T+L+I & 2.00E-05 & 16 & 100 & AdamW & 0.01 & N/A & 0.1 & 0.1 & \cmark & \xmark & \xmark & 224×224 \\
\midrule
 & \bert{} & T & 5.00E-05 & 32 & 50 & Adam & 0.01 & N/A & 0.02 & 0.1 & \xmark & \cmark & \xmark & N/A \\
                     \cellcolor{white}& \lilt{} & T+L & 5.00E-05 & 32 & 50 & Adam & 0.01 & N/A & 0.02 & 0.1 & \xmark & \cmark & \xmark & N/A \\
                     \cellcolor{white}\multirow{-3}{*}{VQA}& \layoutlm{} & T+L+I & 5.00E-05 & 16 & 50 & Adam & 0.01 & N/A & 0.02 & 0.1 & \cmark & \cmark & \xmark & 224×224 \\
\midrule
 \cellcolor{white}& \fasterrcnn{} & I & 2.00E-02 & 16 & 40 & SGD & 0.0001 & 0.9 & 0.05 & 0.1 & \xmark & \cmark & \cmark & 480–800×1333\tnote{*} \\
                     \multirow{-2}{*}{DLA}& \cascadercnn{} & I & 2.00E-02 & 16 & 40 & SGD & 0.0001 & 0.9 & 0.05 & 0.1 & \xmark & \cmark & \cmark & 480–800×1333\tnote{*} \\
\end{tabular}%

}
\caption{Default training hyperparameters for all models and tasks in our experiments. For each task (CLS, KIE, VQA, DLA) and model combination, we report learning rate, batch size, number of epochs, optimizer settings, and other training parameters. These settings were chosen based on preliminary experiments and common practices for document understanding tasks.}
\label{tab:hyperparams_all}
\end{table*}

\section{Data \& Code Availability}
\label{app:code_release}
Complete source code is provided in this supplementary material, including the VLM-based synthesis pipeline, clustering and sampling procedures, model training scripts, and evaluation tools. Due to size constraints, we include representative samples of the synthetic datasets and the complete \docvqahw{} subset in the supplementary material. The full synthetic datasets (140K+ samples across eleven benchmarks) will be released upon publication. All real-world benchmark datasets are publicly available from their original sources as cited in the paper.

\end{document}